\documentclass{article}

\usepackage{microtype}
\usepackage{graphicx}
\usepackage{subfigure}
\usepackage{booktabs}
\usepackage{hyperref}

\usepackage[accepted]{icml2025}

\usepackage{amsmath}
\usepackage{amssymb}
\usepackage{mathtools}
\usepackage{amsthm}

\newcommand{\secref}[1]{Section~\ref{#1}}
\newcommand{\figref}[1]{Figure~\ref{#1}}
\newcommand{\tabref}[1]{Table~\ref{#1}}
\newcommand{\algref}[1]{Algorithm~\ref{#1}}
\newcommand{\defref}[1]{Definition~\ref{#1}}
\newcommand{\asmref}[1]{Assumption~\ref{#1}}

\newcommand{\figsref}[2]{Figures~\ref{#1} and~\ref{#2}}
\newcommand{\secsref}[2]{Sections~\ref{#1} and~\ref{#2}}

\theoremstyle{plain}
\newtheorem{theorem}{Theorem}[section]
\theoremstyle{definition}
\newtheorem{definition}[theorem]{Definition}
\newtheorem{assumption}[theorem]{Assumption}

\newcommand{\argmax}{\operatornamewithlimits{\arg \max}}

\renewcommand{\algorithmicrequire}{\textbf{Input:}}
\renewcommand{\algorithmicensure}{\textbf{Output:}}

\newcommand{\bbE}{\mathbb{E}}

\newcommand{\bbR}{\mathbb{R}}

\newcommand{\bbI}{\mathbb{I}}

\newcommand{\bc}{\mathbf{c}}

\newcommand{\bx}{\mathbf{x}}

\newcommand{\bz}{\mathbf{z}}

\newcommand{\bw}{\mathbf{w}}
\newcommand{\bl}{\mathbf{l}}
\newcommand{\bu}{\mathbf{u}}

\newcommand{\bA}{\mathbf{A}}
\newcommand{\bC}{\mathbf{C}}
\newcommand{\bD}{\mathbf{D}}

\newcommand{\bP}{\mathbf{P}}

\newcommand{\bX}{\mathbf{X}}

\newcommand{\bW}{\mathbf{W}}

\newcommand{\bI}{\mathbf{I}}

\newcommand{\bsSigma}{\boldsymbol \Sigma}

\newcommand{\calC}{\mathcal{C}}
\newcommand{\calD}{\mathcal{D}}

\newcommand{\calN}{\mathcal{N}}

\newcommand{\calS}{\mathcal{S}}

\newcommand{\calX}{\mathcal{X}}

\newcommand{\ours}{DRE-BO-SSL}
\graphicspath{{./figures/}}

\icmltitlerunning{Density Ratio Estimation-based Bayesian Optimization with Semi-Supervised Learning}

\begin{document}

\twocolumn[
\icmltitle{Density Ratio Estimation-based Bayesian Optimization\\with Semi-Supervised Learning}

\icmlsetsymbol{equal}{*}

\begin{icmlauthorlist}
\icmlauthor{Jungtaek Kim}{xxx}
\end{icmlauthorlist}

\icmlaffiliation{xxx}{University of Wisconsin--Madison, Madison, WI 53706, USA}

\icmlcorrespondingauthor{Jungtaek Kim}{jungtaek.kim@wisc.edu}

\icmlkeywords{Bayesian Optimization, Density Ratio Estimation-based Bayesian Optimization, Bayesian Optimization with Semi-Supervised Learning}

\vskip 0.3in
]

\printAffiliationsAndNotice{}

\begin{abstract}
Bayesian optimization has attracted huge attention from diverse research areas in science and engineering, since it is capable of efficiently finding a global optimum of an expensive-to-evaluate black-box function. In general, a probabilistic regression model is widely used as a surrogate function to model an explicit distribution over function evaluations given an input to estimate and a training dataset. Beyond the probabilistic regression-based methods, density ratio estimation-based Bayesian optimization has been suggested in order to estimate a density ratio of the groups relatively close and relatively far to a global optimum. Developing this line of research further, supervised classifiers are employed to estimate a class probability for the two groups instead of a density ratio. However, the supervised classifiers used in this strategy are prone to be overconfident for known knowledge on global solution candidates. Supposing that we have access to unlabeled points, e.g., predefined fixed-size pools, we propose density ratio estimation-based Bayesian optimization with semi-supervised learning to solve this challenge. Finally, we show the empirical results of our methods and several baseline methods in two distinct scenarios with unlabeled point sampling and a fixed-size pool, and analyze the validity of our methods in diverse experiments.
\end{abstract}

\section{Introduction}
\label{sec:introduction}

Bayesian optimization~\citep{BrochuE2010arxiv,GarnettR2023book} has attracted immense attention
from various research areas
such as hyperparameter optimization~\citep{BergstraJ2011neurips},
battery lifetime optimization~\citep{AttiaPM2020nature},
chemical reaction optimization~\citep{ShieldsBJ2021nature},
nanophotonic structure optimization~\citep{KimJ2024dd},
and language model fine-tuning~\citep{JangC2024neurips},
since it is capable of efficiently finding a global optimum
of an expensive-to-evaluate black-box function.
As studied in previous literature on Bayesian
optimization~\citep{SnoekJ2012neurips,MartinezCantinR2018aistats,SpringenbergJT2016neurips,HutterF2011lion},
a probabilistic regression model, which can estimate a distribution of function evaluations
over inputs, is widely used as a surrogate
function;
A Gaussian process~\citep{RasmussenCE2006book}
is a predominant choice for the surrogate function.
An analogy between probabilistic regression models in Bayesian optimization
is that they rely on an explicit function over function evaluations
$p(y \mid \bx, \calD)$
given an input to estimate, denoted as $\bx$,
and a training dataset $\calD$.

Beyond the probabilistic regression-based Bayesian optimization,
density ratio estimation (DRE)-based Bayesian optimization has been studied
recently~\citep{BergstraJ2011neurips,TiaoLC2021icml}.
Furthermore,
likelihood-free Bayesian optimization,
which is equivalent to DRE-based Bayesian optimization with a particular utility function, has been proposed by~\citet{SongJ2022icml}.
\citet{BergstraJ2011neurips} attempt to model two densities $p(\bx \mid y \leq y^\dagger, \calD)$ and $p(\bx \mid y > y^\dagger, \calD)$,
where $y^\dagger$ is a threshold for dividing inputs to two groups that are relatively close and relatively far to a global solution,
in order to estimate $\zeta$-relative density ratio~\citep{YamadaM2011neurips}.
On the other hand,
instead of modeling two densities separately,
\citet{TiaoLC2021icml,SongJ2022icml} estimate a density ratio
using class-probability estimation~\citep{QinJ1998biometrika}.
As discussed in the previous work,
this line of research provides a new understanding of Bayesian optimization,
which allows us to solve Bayesian optimization
using binary classification.
Moreover,
it can reduce the amount of computation required
for building surrogate functions.

\begin{figure*}[t]
    \centering
    \subfigure[MLP, BORE, Iterations 1 to 5]{
        \centering
        \includegraphics[width=0.19\textwidth]{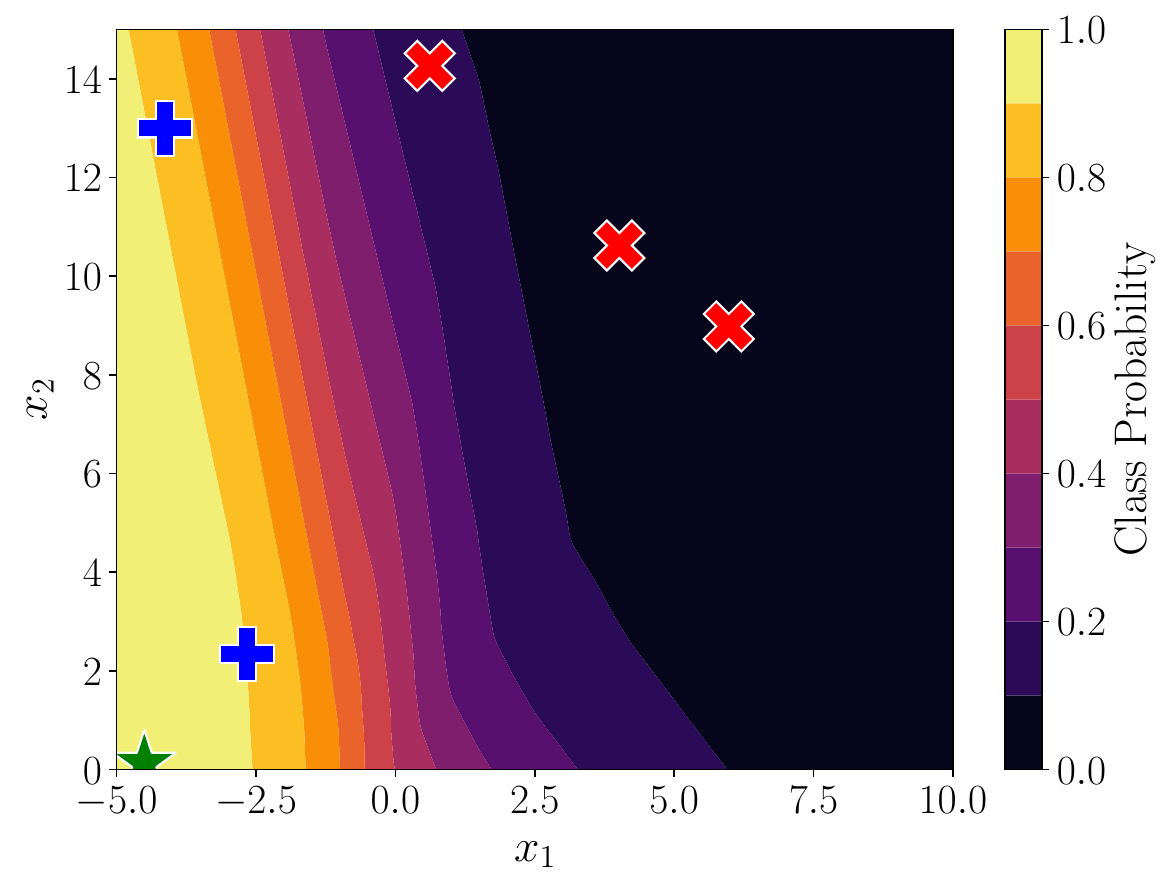}
        \includegraphics[width=0.19\textwidth]{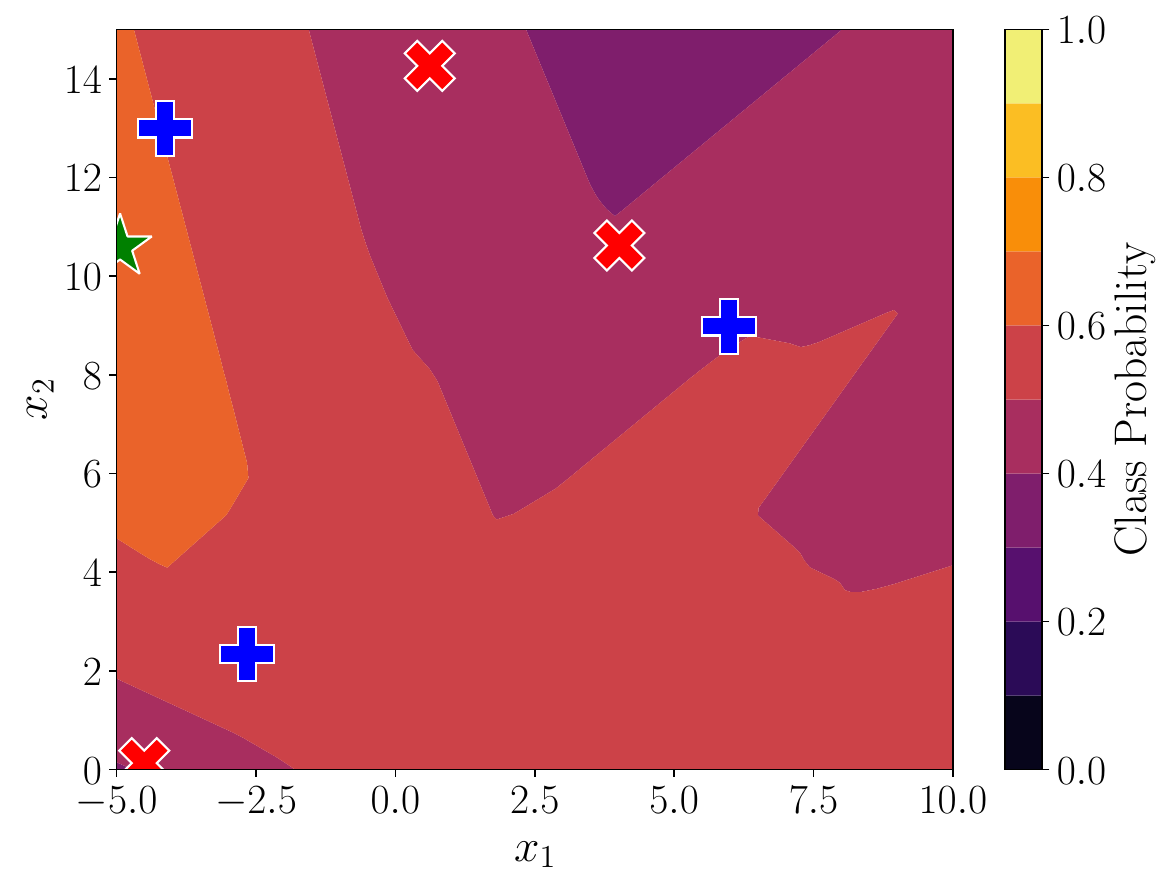}
        \includegraphics[width=0.19\textwidth]{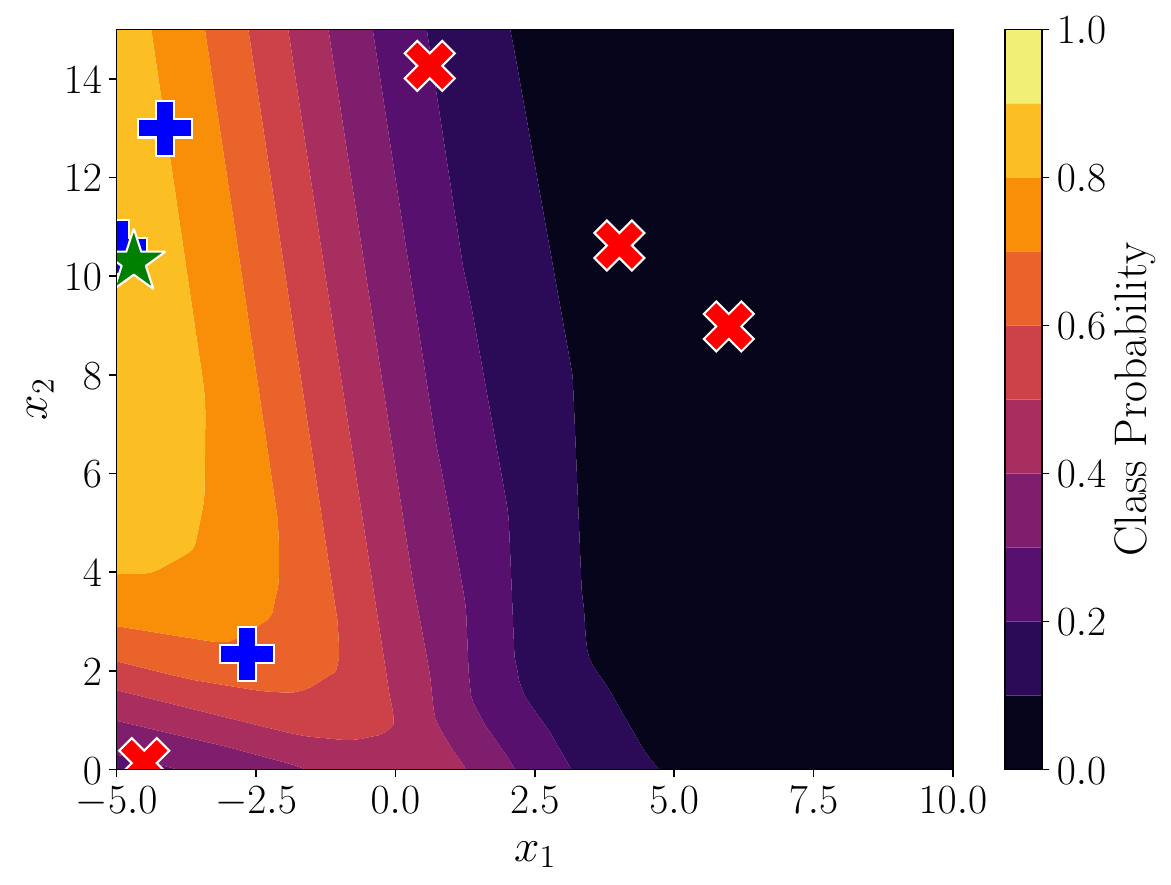}
        \includegraphics[width=0.19\textwidth]{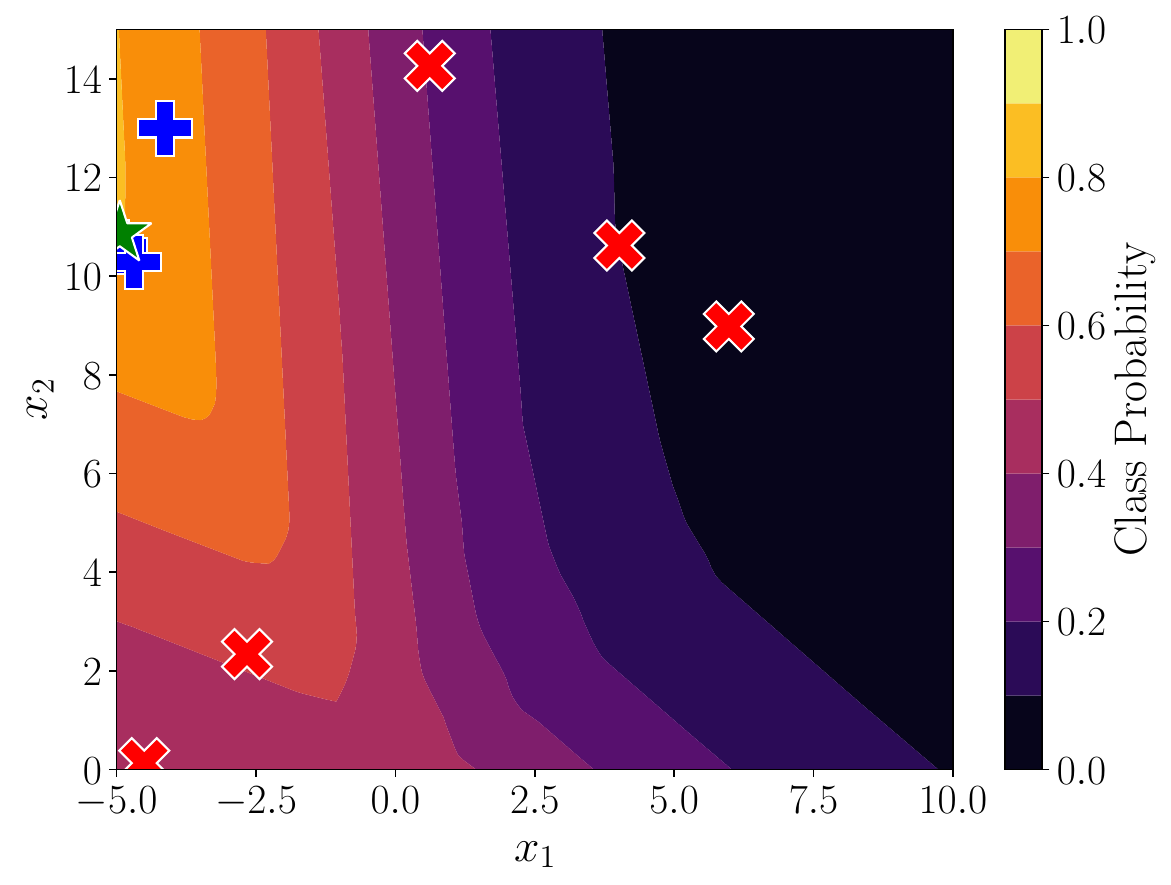}
        \includegraphics[width=0.19\textwidth]{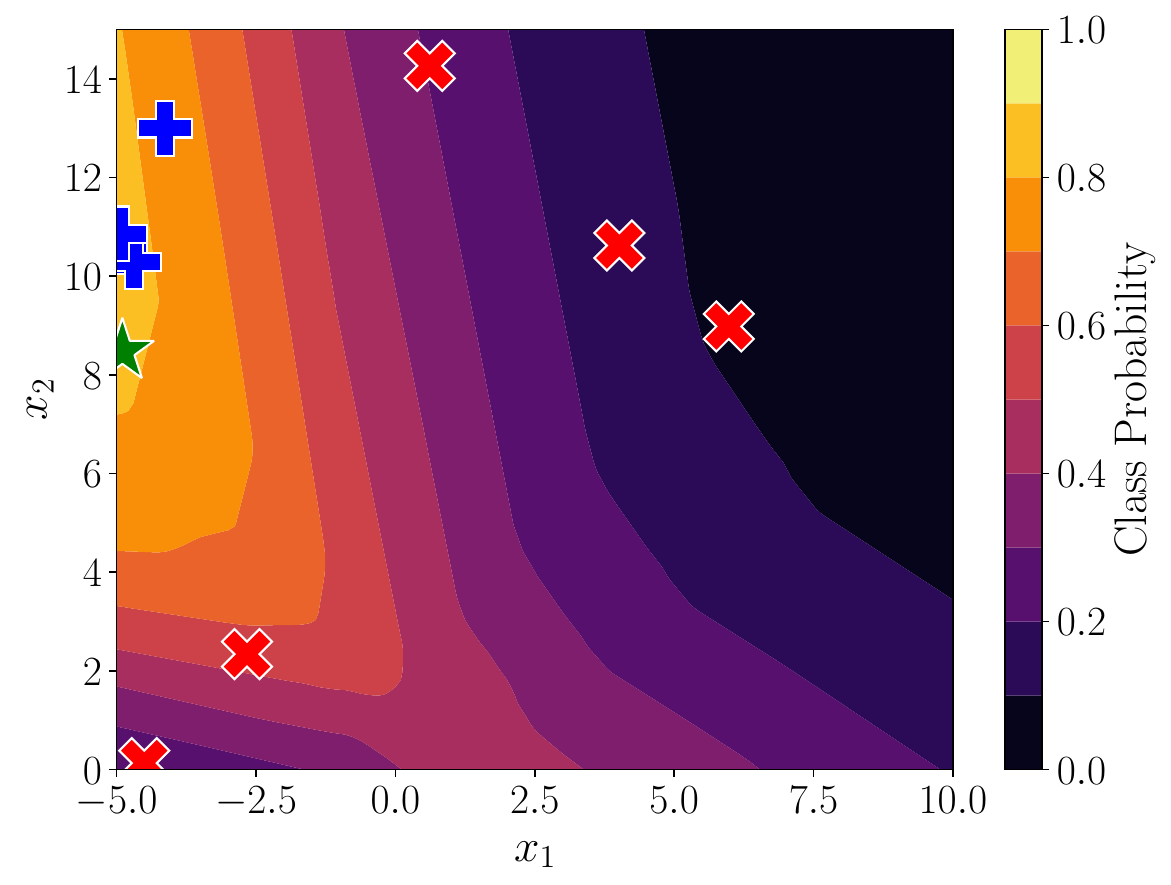}
    }
    \subfigure[MLP, LFBO, Iterations 1 to 5]{
        \centering
        \includegraphics[width=0.19\textwidth]{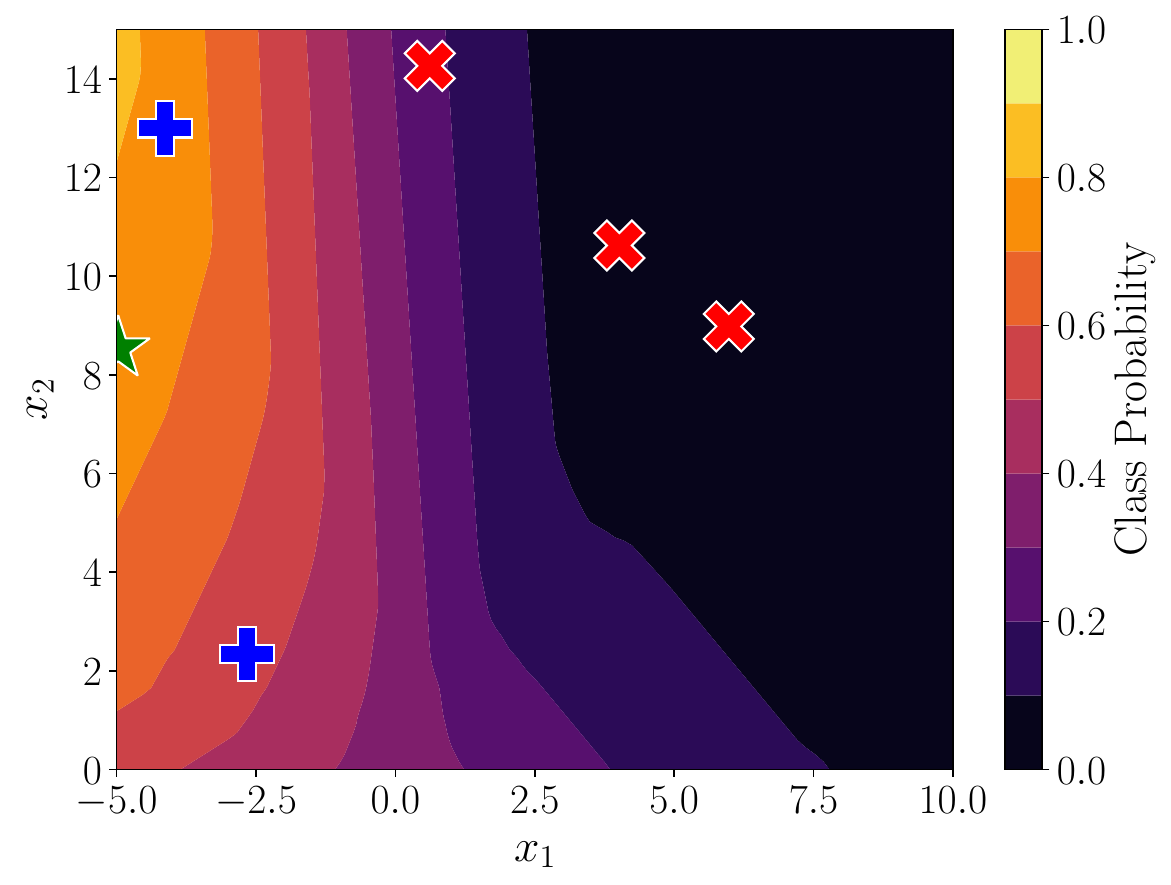}
        \includegraphics[width=0.19\textwidth]{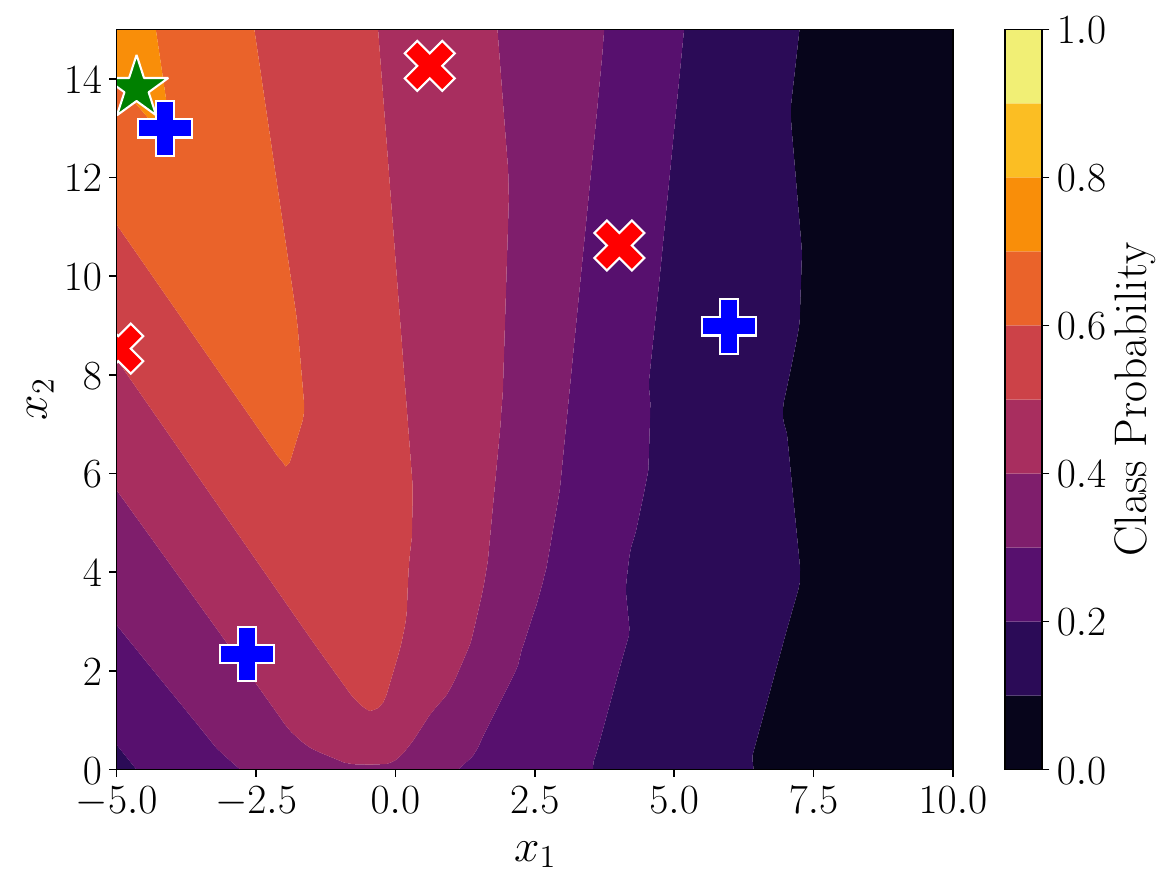}
        \includegraphics[width=0.19\textwidth]{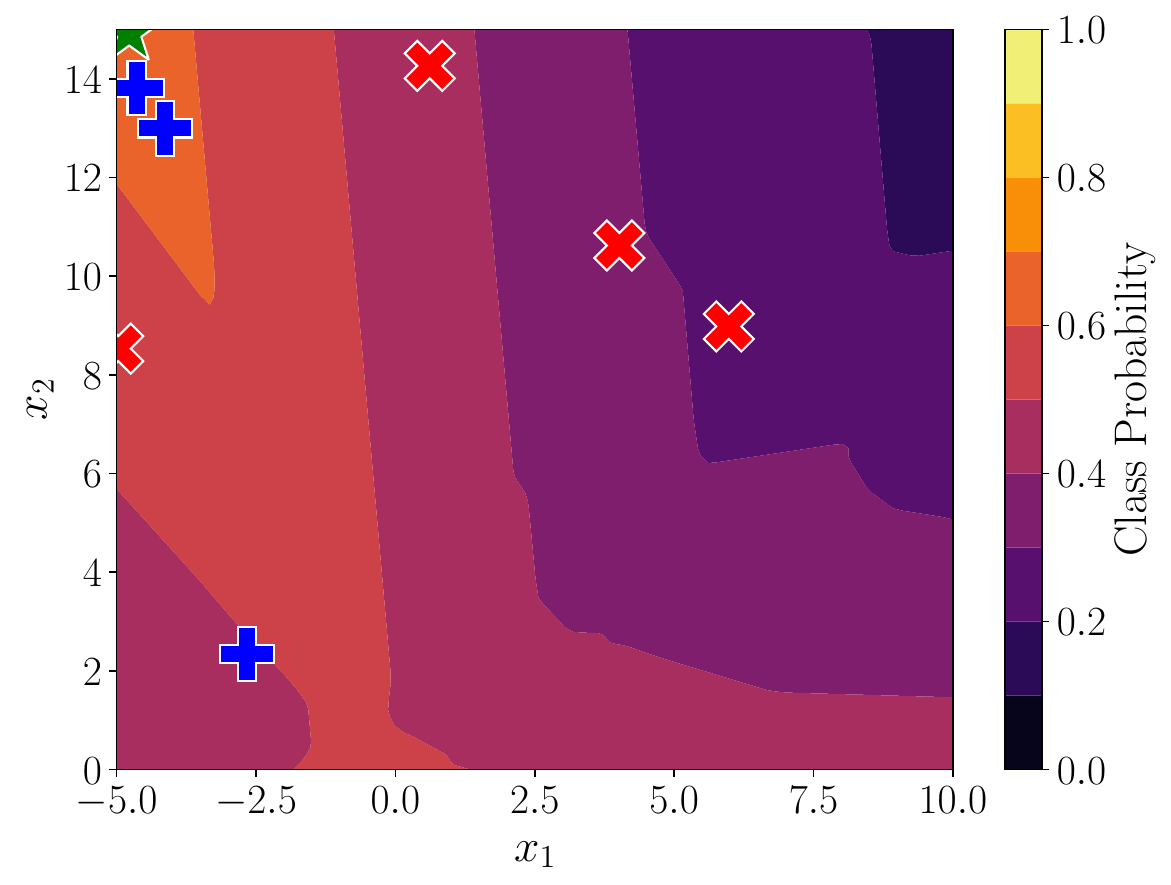}
        \includegraphics[width=0.19\textwidth]{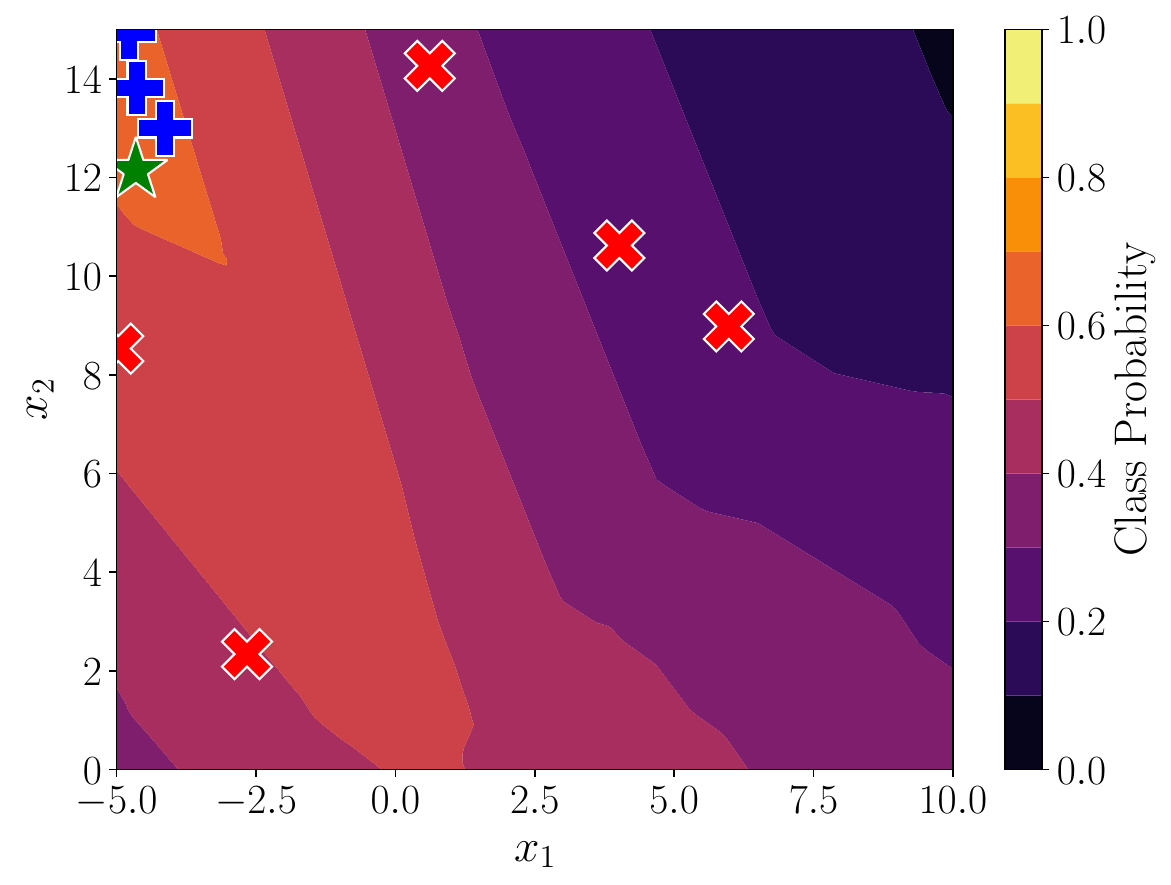}
        \includegraphics[width=0.19\textwidth]{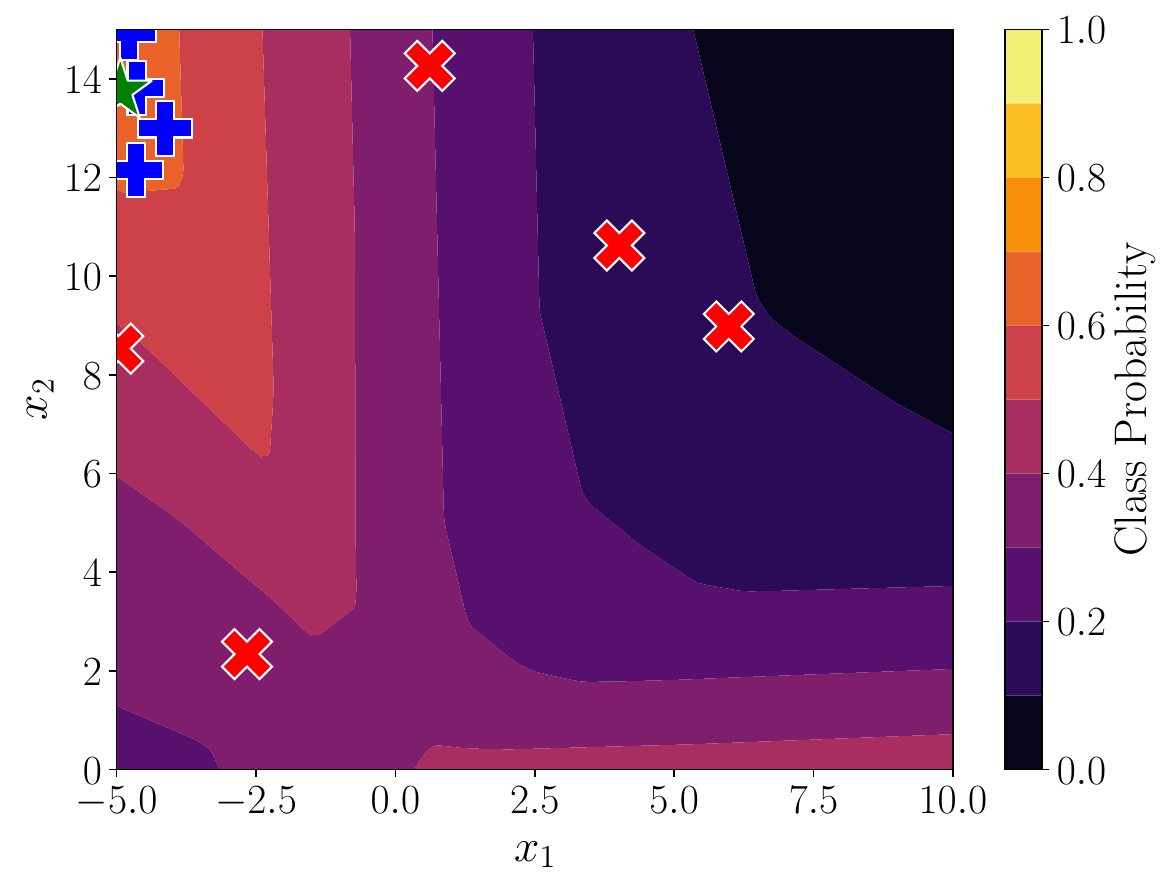}
    }
    \subfigure[\ours, Label propagation, Iterations 1 to 5]{
        \centering
        \includegraphics[width=0.19\textwidth]{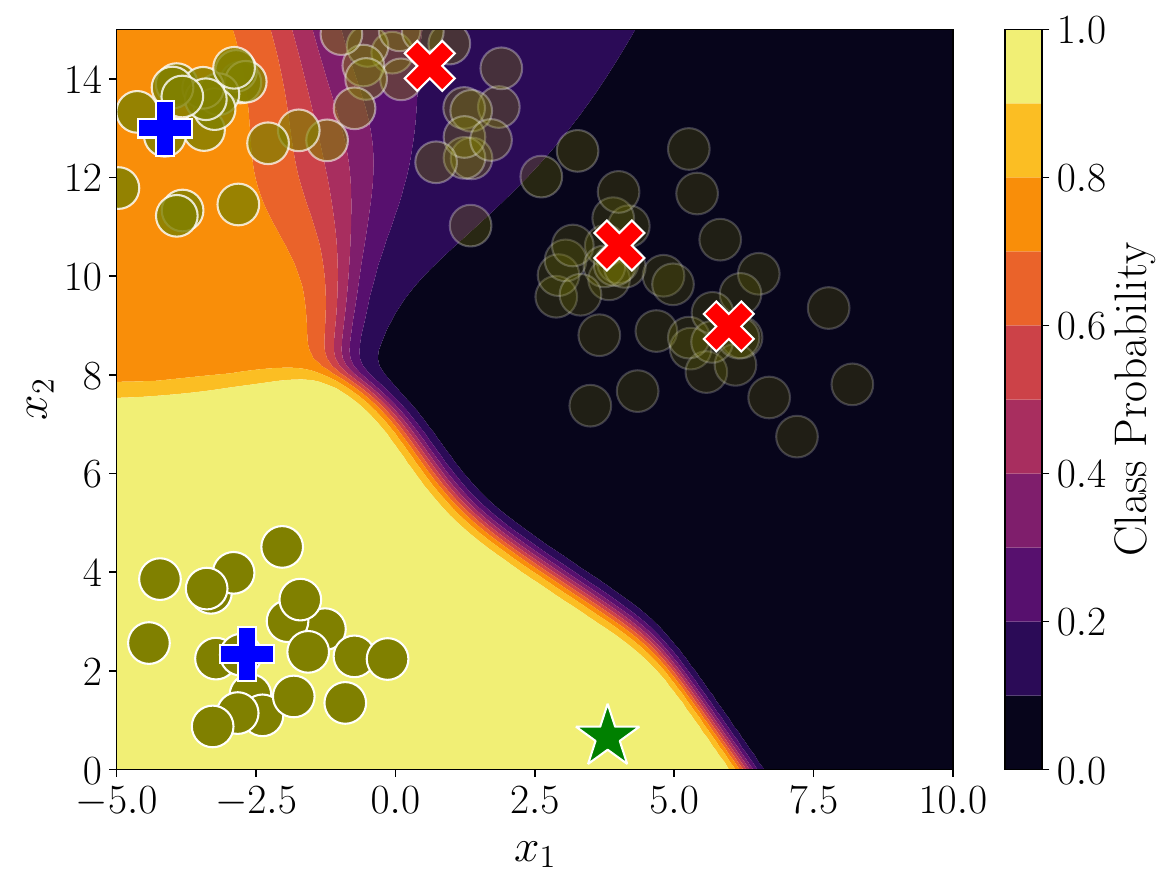}
        \includegraphics[width=0.19\textwidth]{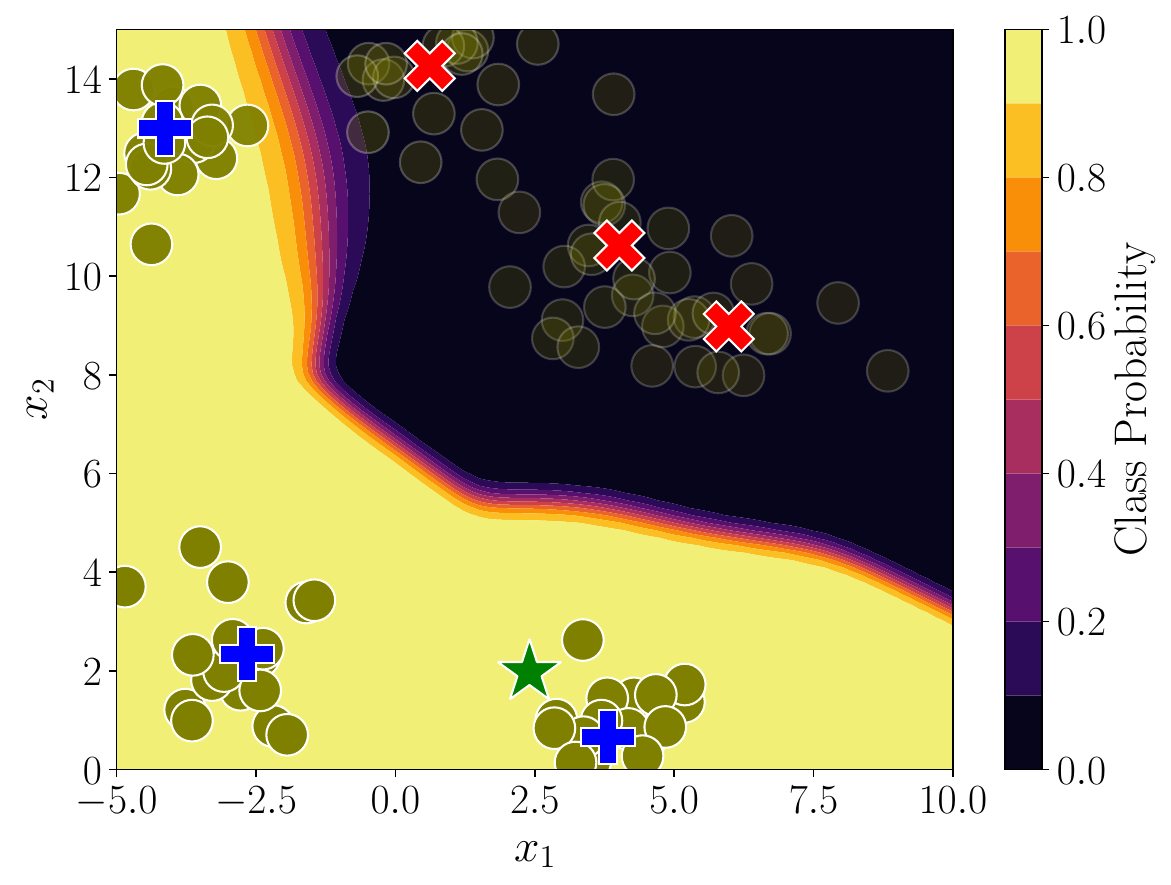}
        \includegraphics[width=0.19\textwidth]{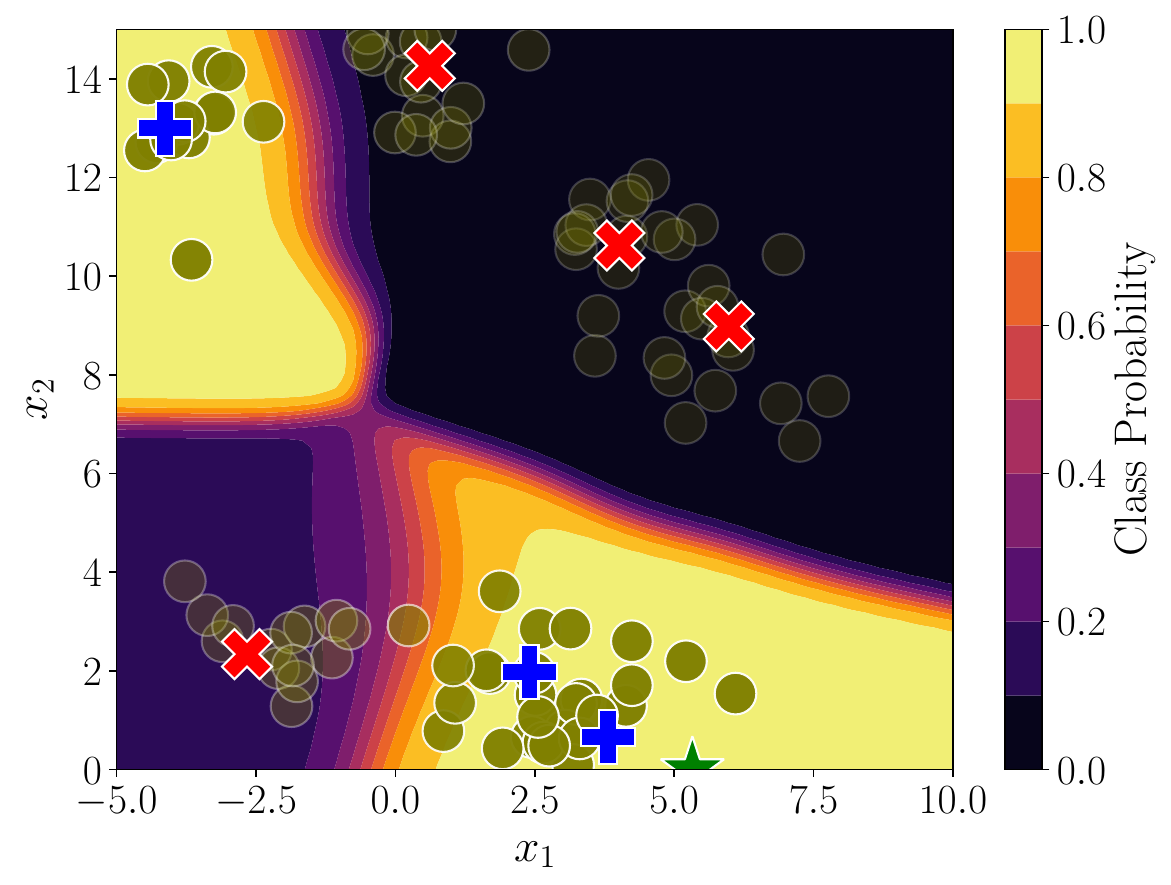}
        \includegraphics[width=0.19\textwidth]{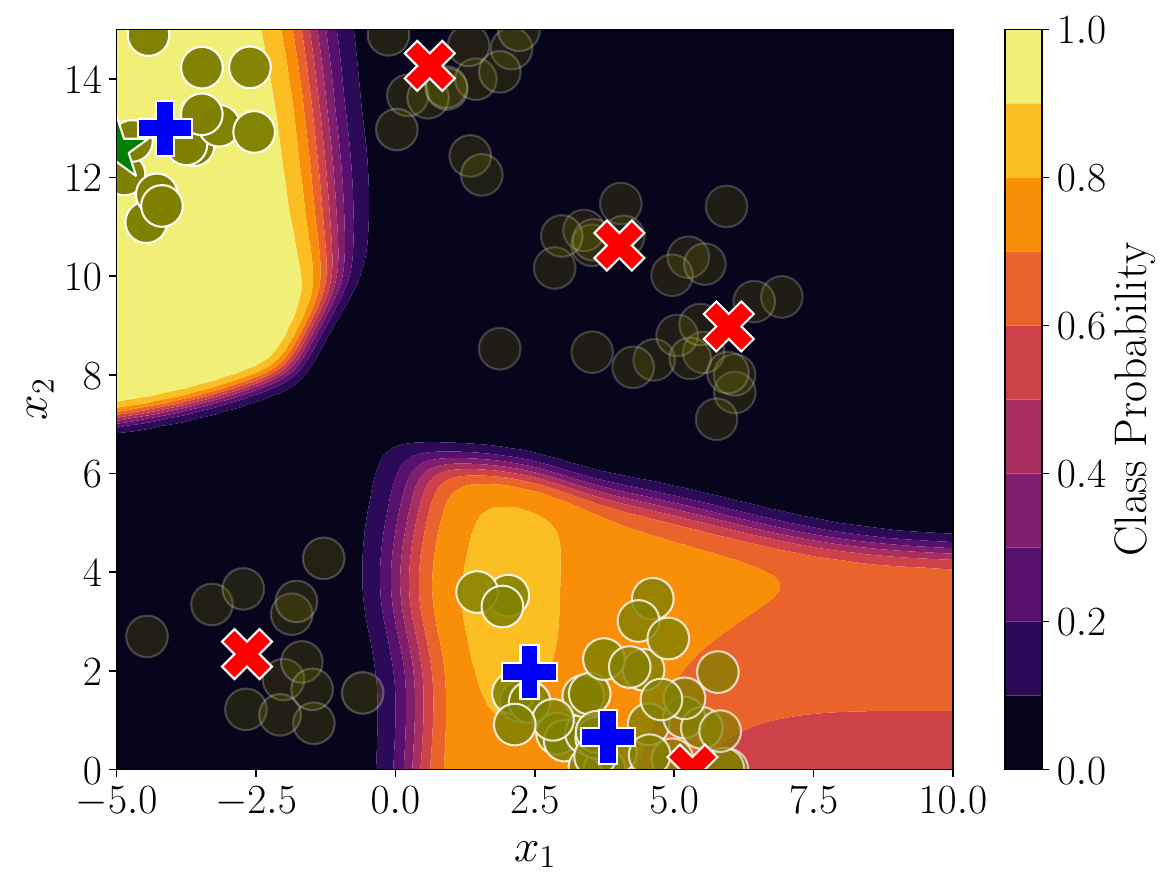}
        \includegraphics[width=0.19\textwidth]{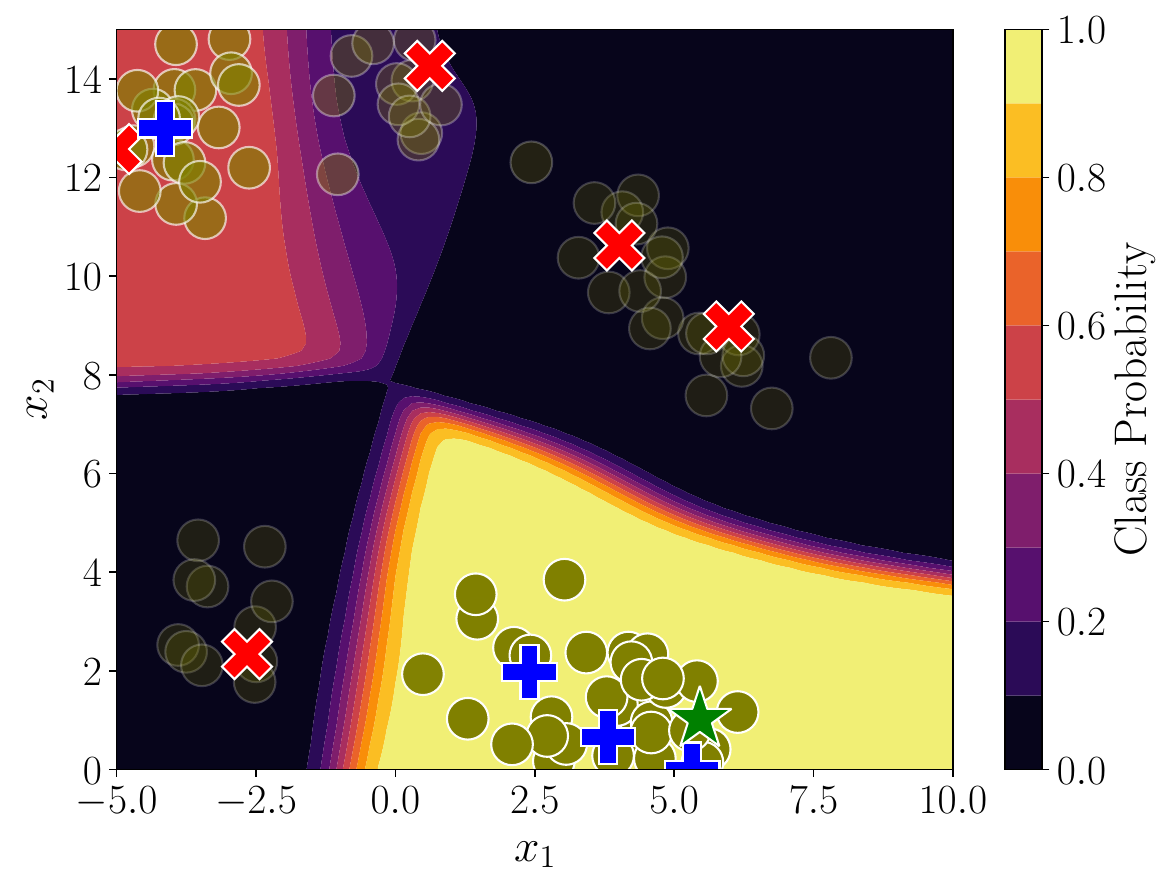}
    }
    \subfigure[\ours, Label spreading, Iterations 1 to 5]{
        \centering
        \includegraphics[width=0.19\textwidth]{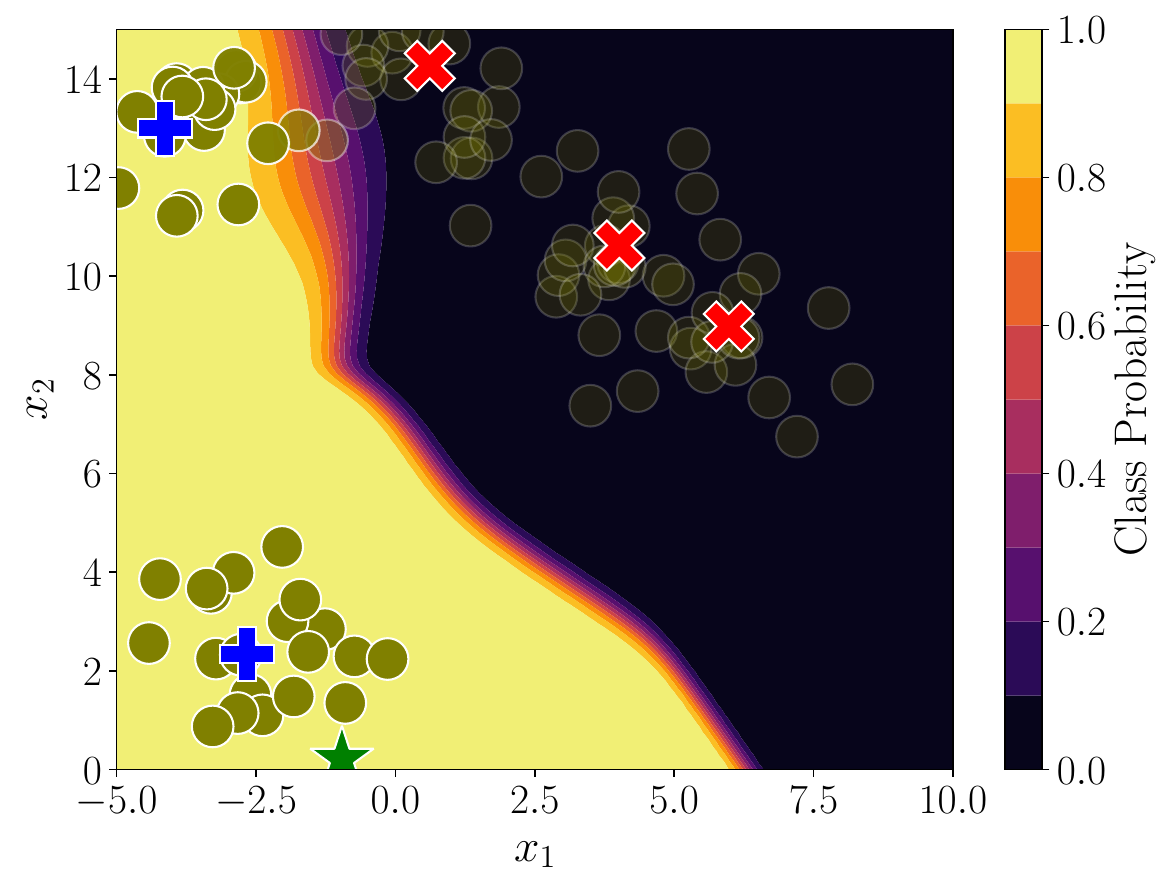}
        \includegraphics[width=0.19\textwidth]{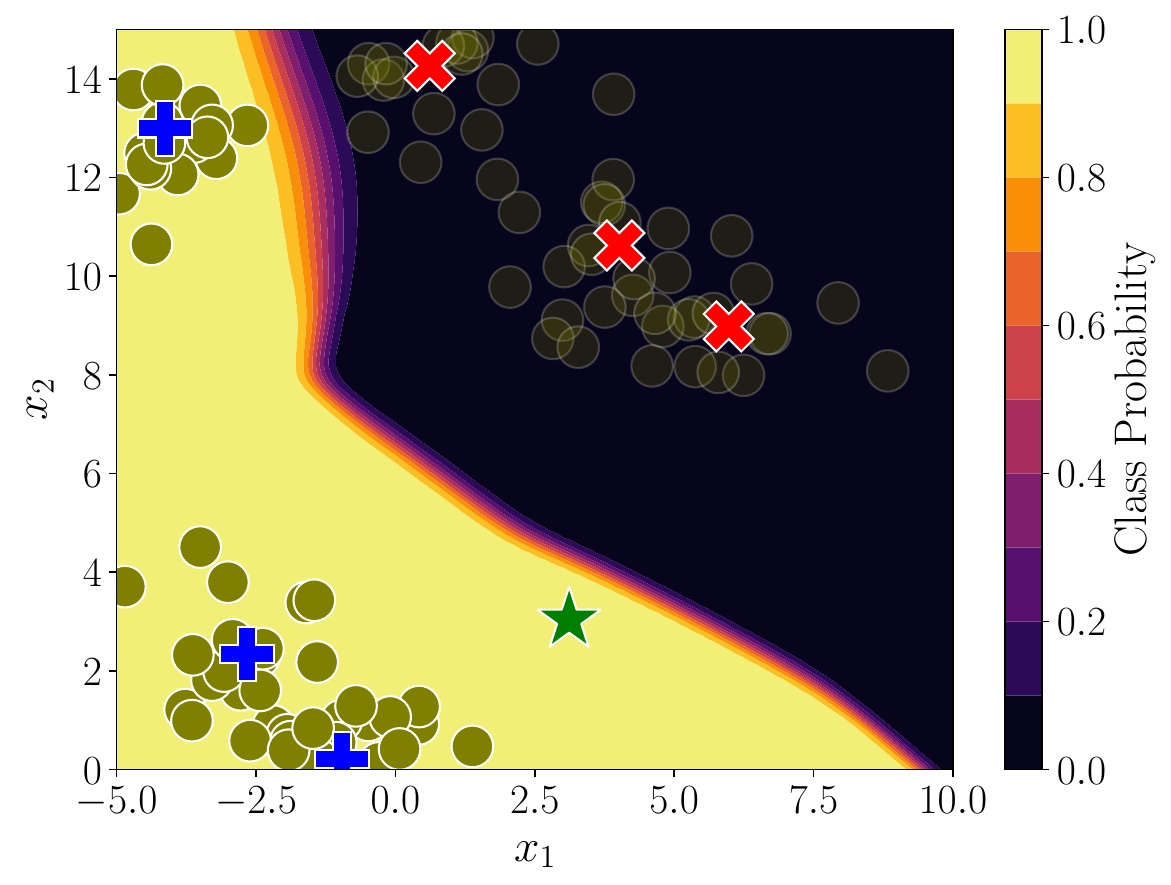}
        \includegraphics[width=0.19\textwidth]{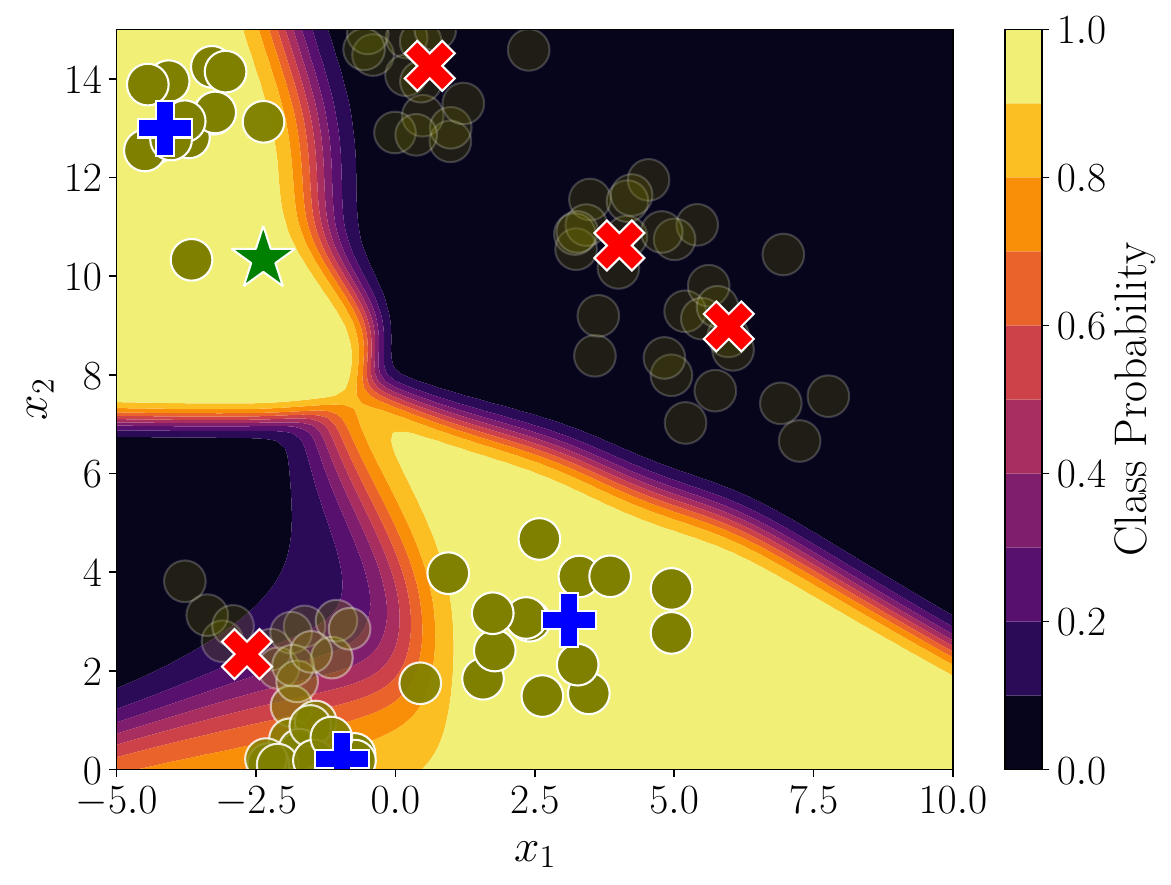}
        \includegraphics[width=0.19\textwidth]{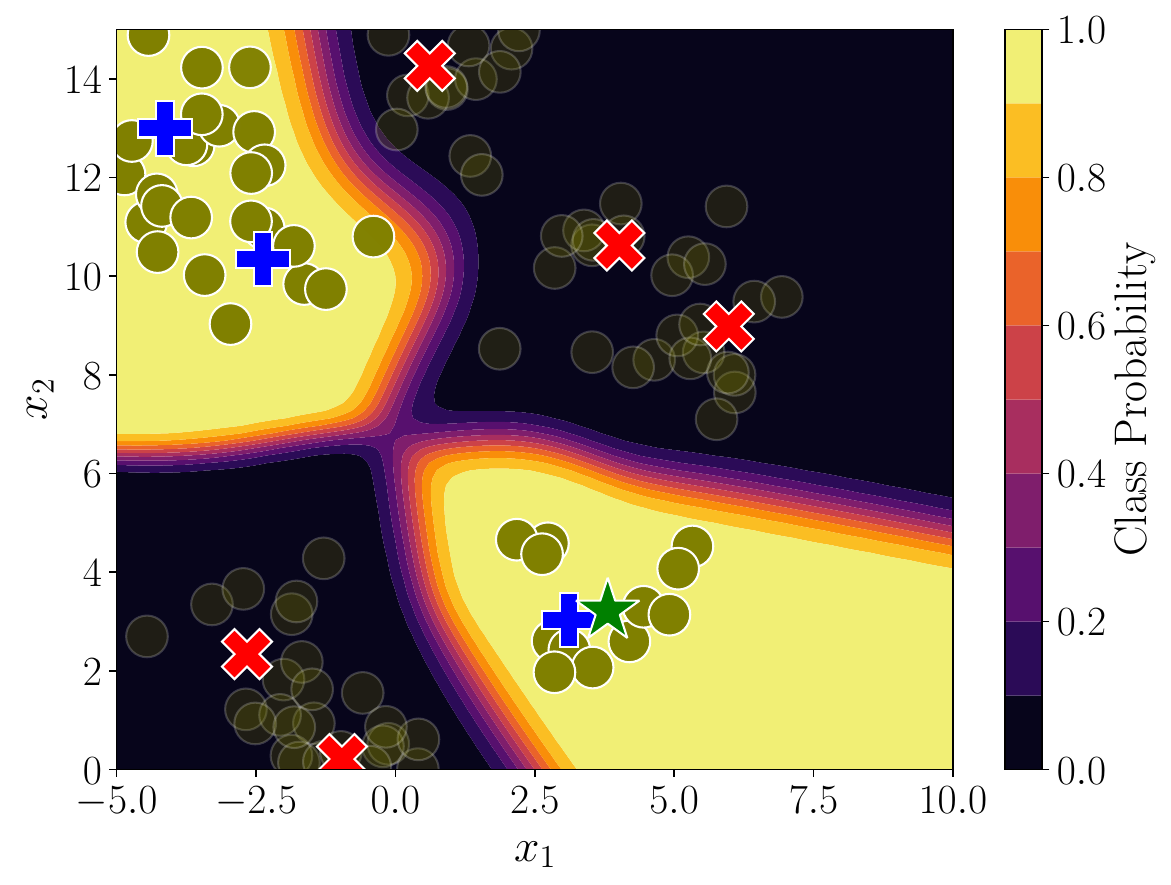}
        \includegraphics[width=0.19\textwidth]{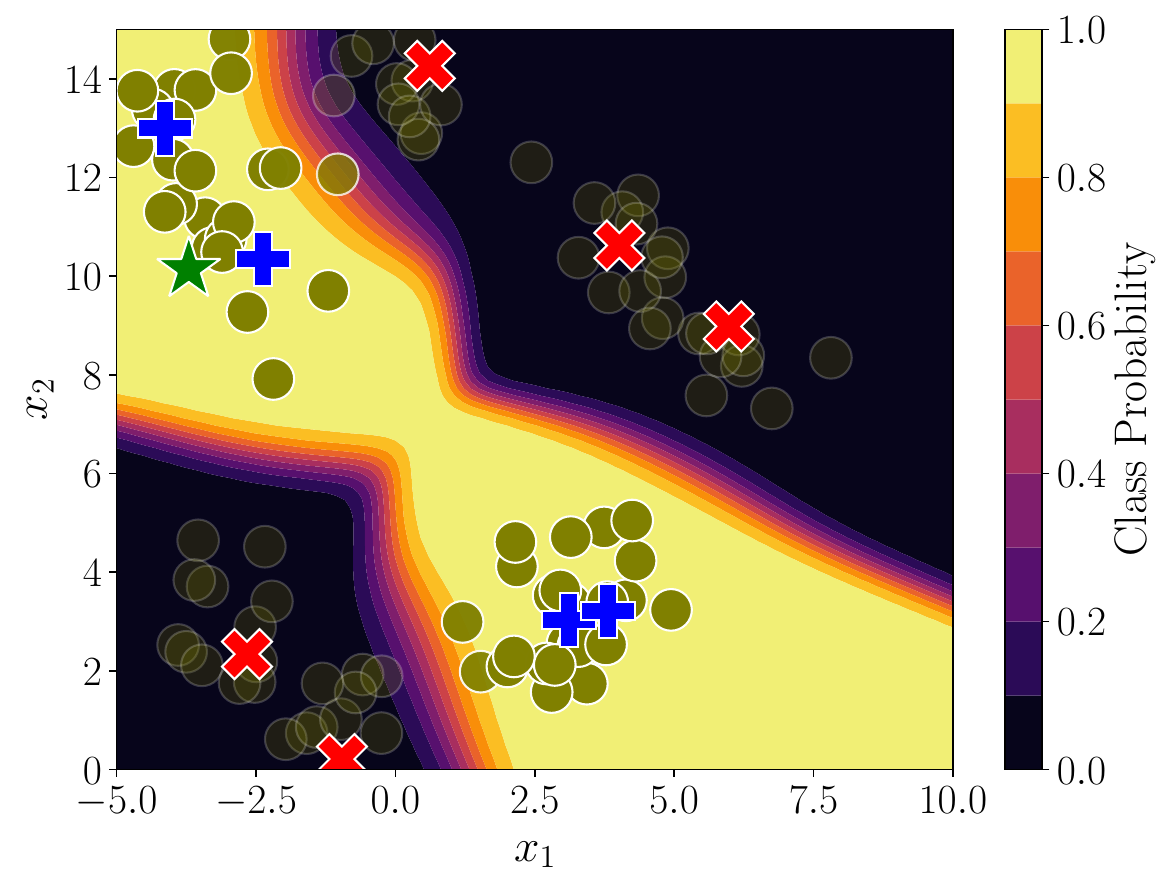}
    }
    \caption{Comparisons of BORE and LFBO by multi-layer perceptrons, and \ours~with label propagation and label spreading for the Branin function. Each row shows Iterations 1 to 5 with five initial points. \texttt{+} (blue), \texttt{x} (red), and $\star$ (green) indicate data points of $y \leq y^\dagger$, data points of $y > y^\dagger$, and query points, respectively. Moreover, \texttt{o} (olive) stands for unlabeled points and its transparency represents the class probability predicted by a semi-supervised classifier. A query point is chosen by maximizing the output of the classifier. More results are shown in~\figref{fig:prob_spaces_random_forest_gradient_boosting_xgboost}.}
    \label{fig:prob_spaces}
\end{figure*}

However, the supervised classifiers utilized in the DRE-based Bayesian optimization are prone to be overconfident for known knowledge on global solution candidates or a region that have been already exploited.
In this paper,
supposing that we have access to unlabeled points,
we propose a novel DRE-based method with semi-supervised learning,
which is named~\ours, to solve this over-exploitation problem.
Although our direct competitors,
i.e., BORE~\citep{TiaoLC2021icml} and LFBO~\citep{SongJ2022icml},
show their strengths through theoretical and empirical analyses,
our algorithm betters an ability to consider a wider region that satisfies $p(\bx \mid y \leq y^\dagger, \calD) \geq p(\bx \mid y > y^\dagger, \calD)$,
than the competitors,
as shown in~\figref{fig:prob_spaces}.
By this intuitive example in~\figref{fig:prob_spaces},
we presume that~\ours~appropriately balances exploration and exploitation
rather than the existing methods.
Compared to a supervised classifier,
e.g., random forests~\citep{BreimanL2001ml},
gradient boosting~\citep{FriedmanJH2001aos},
and multi-layer perceptrons,
our semi-supervised classifiers,
i.e., label propagation~\citep{ZhuX2002tr}
and label spreading~\citep{ZhouD2003neurips_b},
are less confident in terms of the regions of global solution candidates
using unlabeled data points;
see~\figsref{fig:prob_spaces}{fig:prob_spaces_random_forest_gradient_boosting_xgboost}
and \secsref{subsec:overexploitation}{sec:dre_based} for detailed examples and analyses.

To make use of semi-supervised classifiers,
we take into account \emph{two distinct scenarios
with unlabeled point sampling
and with a predefined fixed-size pool}.
For the first scenario,
we randomly sample unlabeled data points
from the truncated multivariate normal distribution using a minimax tilting method~\citep{BotevZI2017jrssb},
to allow for the possibility of adopting a cluster assumption~\citep{SeegerM2000tr}.
Finally, we demonstrate that our method shows superior performance
compared to the exiting methods in diverse experiments
including synthetic benchmarks,
Tabular Benchmarks~\citep{KleinA2019arxiv},
NATS-Bench~\citep{DongX2021ieeetpami},
and 64D minimum multi-digit MNIST search.
Note that Tabular Benchmarks, NATS-Bench,
and 64D minimum multi-digit MNIST search have access to fixed-size pools.
To validate our methods,
we provide thorough analyses on the components of \ours~in~\secref{sec:discussion} and the appendices.

To sum up,
our contributions are itemized as follows:
\begin{itemize}
    \item we identify the over-exploitation problem of supervised classifiers in DRE-based Bayesian optimization;
    \item we propose DRE-based Bayesian optimization with semi-supervised learning, named \ours~for two distinct scenarios with unlabeled point sampling and a predefined fixed-size pool;
    \item we demonstrate the effectiveness of our method in various experiments including NATS-Bench and 64D minimum multi-digit MNIST search.
\end{itemize}

\subsection{Over-Exploitation Problem}
\label{subsec:overexploitation}

As illustrated in~\figsref{fig:prob_spaces}{fig:prob_spaces_random_forest_gradient_boosting_xgboost},
the supervised classifiers used in DRE-based Bayesian optimization
suffer from the over-exploitation problem.
Interestingly,
many deep learning models also share a similar problem, which is called an overconfidence problem~\citep{GuoC2017icml,MullerR2019neurips},
due to various reasons, but primarily due to overparameterization.
It is noteworthy that the definition of the over-exploitation problem in DRE-based Bayesian optimization
is different from the overconfidence problem in general classification.
The definition in DRE-based Bayesian optimization does not imply that a single data point has a high probability for a particular class,
but it implies that a few data points have high probabilities for a class of interest.
More precisely,
our definition indicates the problem of overconfidence over known knowledge on global solution candidates.
For example,
the definition in general classification includes
a case that 50\% of data points are biased to one class
and the remainder is biased to another class.
On the contrary,
this definition does not include such a case
and it only includes a case that a small number of data points or the small region of a search space,
which is colored in yellow in~\figsref{fig:prob_spaces}{fig:prob_spaces_random_forest_gradient_boosting_xgboost},
are biased to some particular class,
i.e., Class 1 in DRE-based Bayesian optimization.

By the aforementioned definition,
at early iterations of Bayesian optimization,
a supervised classifier tends to
overfit to a small size of $\calD_t$
due to a relatively large model capacity.
This consequence makes a Bayesian optimization algorithm
highly focus on exploitation.
Similar to our observation,
the imbalance of exploration and exploitation in the DRE-based approaches is also discussed in the recent work by~\citet{OliveiraR2022neurips,PanJ2024icml}.
Moreover,
the consequence mentioned above is different from the characteristics of probabilistic regression-based Bayesian optimization
because the regression-based methods are capable of exploring unseen regions by dealing with uncertainties.
Besides, even though a threshold ratio $\zeta$ might be able to mitigate this problem,
an overconfident supervised classifier is likely to keep getting stuck in a local optimum
as $y^\dagger$ cannot change dramatically;
more detailed analysis on $\zeta$ is provided in~\secsref{sec:discussion}{sec:discussion_threshold_ratios}.

\section{Background and Related Work}
\label{sec:related_work}

\paragraph{Bayesian Optimization.}

It is a principled and efficient approach
to finding a global solution of a challenging objective,
e.g., expensive-to-evaluate black-box functions~\citep{BrochuE2010arxiv,GarnettR2023book}.
To focus on a probabilistic regression model as a surrogate function,
we omit the details of Bayesian optimization here;
see the references by~\citet{BrochuE2010arxiv,GarnettR2023book}
for details.
In Bayesian optimization,
Gaussian process regression~\citep{RasmussenCE2006book}
is widely used as a surrogate function~\citep{SrinivasN2010icml,SnoekJ2012neurips}
because of its flexibility with minimum assumptions on model and smoothness.
While a Gaussian process is a probable choice,
Bayesian optimization with diverse surrogates
such as Student-$t$ process regression~\citep{MartinezCantinR2018aistats},
Bayesian neural networks~\citep{SpringenbergJT2016neurips},
and tree-based models~\citep{HutterF2011lion,KimJ2022aistats}
has been proposed.
An analogy between such models is
that they model $p(y \mid \bx, \calD)$ explicitly,
so that it can be used to define an acquisition function
with the statistics of $p(y \mid \bx, \calD)$.
Note that we solve the problem of minimizing the objective functions in this work.

\paragraph{Density-Ratio Estimation.}

Whereas knowing a data distribution $p(\bx)$ is important,
it is difficult to directly estimate $p(\bx)$~\citep{SugiyamaM2012book}.
For specific machine learning problems such as importance sampling~\citep{KloekT1978econm}
and mutual information estimation~\citep{BishopCM2006book},
we can bypass direct density estimation
and estimate a density ratio.
More recently,
\citet{RhodesB2020neurips} tackle a density-chasm problem using telescopic density-ratio estimation,
which replaces an original problem with a set of logistic regression problems.
Since the work by~\citet{RhodesB2020neurips} suffers from the issue on distribution shift, the recent work by~\citet{SrivastavaA2023tmlr} proposes a density ratio estimation method with multinomial logistic regression, which is capable of mitigating the distribution shift using multi-class classification.
In Bayesian optimization,
\citet{BergstraJ2011neurips} have proposed a strategy with tree-structured Parzen estimators
to estimate a density ratio
as an alternative to probabilistic regression-based acquisition functions.
In addition,
the existing work by~\citet{TiaoLC2021icml,SongJ2022icml}
suggests methods with binary classifiers
in order to estimate class probabilities as a density ratio;
see~\secref{sec:dre_based} for the details of this literature.

\paragraph{Semi-Supervised Learning.}

It is a learning scheme with both labeled and unlabeled data~\citep{ZhuX2005tr,ChapelleO2006book,BengioY2006sslbook}.
To cooperate with labeled and unlabeled data,
this strategy generally utilizes geometry of data points
or connectivity between points,
and assigns pseudo-labels to unlabeled data points,
which referred to as transductive learning~\citep{GammermanA1998uai}.
As a semi-supervised learning method on a similarity graph,
\citet{ZhuX2002tr} propose a label propagation algorithm
which iteratively propagates the labels of unlabeled data points using labeled data.
\citet{ZhouD2003neurips_b} compute
the labels of labeled and unlabeled data points
by a weighted iterative algorithm with initial labels.
\citet{BelkinM2002neurips} predict pseudo-labels
by finding a linear combination of a few smallest eigenvectors of the graph Laplacian.

\section{DRE-based Bayesian Optimization}
\label{sec:dre_based}

Unlike probabilistic regression-based Bayesian optimization,
DRE-based Bayesian optimization employs
a density ratio-based acquisition function,
defined with a density $p(\bx \mid y \leq y^\dagger, \calD_t)$,
where $\bx$ is a $d$-dimensional vector,
$y$ is its function evaluation,
$y^\dagger$ is a threshold,
and $\calD_t = \{(\bx_i, y_i)\}_{i = 0}^t$ is a dataset of $t+1$ pairs of data point and evaluation.
In particular,
the work by~\citet{BergstraJ2011neurips}
defines an acquisition function
based on $\zeta$-relative density ratio~\citep{YamadaM2011neurips}:
\begin{align}
    &A(\bx \mid \zeta, \calD_t) \nonumber\\
    &= \frac{p(\bx \mid y \leq y^\dagger, \calD_t)}{\zeta p(\bx \mid y \leq y^\dagger, \calD_t) + (1 - \zeta) p(\bx \mid y > y^\dagger, \calD_t)},
    \label{eqn:density_ratio}
\end{align}
where $\zeta = p(y \leq y^\dagger) \in [0, 1)$ is a threshold ratio.
We need to find a maximizer of~\eqref{eqn:density_ratio},
by optimizing the following composite function:
$h(p(\bx \mid y \leq y^\dagger, \calD_t) / p(\bx \mid y > y^\dagger, \calD_t))$,
where $h(x) = (\zeta + (1 - \zeta) x^{-1})^{-1}$.
Since $h$ is a strictly increasing function,
we can directly maximize $p(\bx \mid y \leq y^\dagger, \calD_t) / p(\bx \mid y > y^\dagger, \calD_t)$.
In the previous work by~\citet{BergstraJ2011neurips},
two tree-structured Parzen estimators are used to estimate
the respective densities,
$p(\bx \mid y \leq y^\dagger, \calD_t)$
and $p(\bx \mid y > y^\dagger, \calD_t)$.

While the work by~\citet{BergstraJ2011neurips} utilizes two distinct tree-structured Parzen estimators,
\citet{TiaoLC2021icml} propose
a method to directly estimate \eqref{eqn:density_ratio}
using class-probability estimation~\citep{QinJ1998biometrika,SugiyamaM2012book},
which is called BORE.
Since it can be formulated as a problem of binary classification in which Class 1 is a group of the top $\zeta$ of $\calD_t$ and Class 0 is a group of the bottom $1 - \zeta$ of $\calD_t$ in terms of function evaluations,
the acquisition function defined in \eqref{eqn:density_ratio} induces the following:
\begin{equation}
    A(\bx \mid \zeta, \calD_t) = \frac{p(\bx \mid z = 1)}{\zeta p(\bx \mid z = 1) + (1 - \zeta) p(\bx \mid z = 0)}.
    \label{eqn:density_ratio_class}
\end{equation}
By the Bayes' theorem, \eqref{eqn:density_ratio_class}
becomes the following:
\begin{equation}
    A(\bx \mid \zeta, \calD_t) = \zeta^{-1} \frac{p(z = 1 \mid \bx)}{p(z = 1 \mid \bx) + p(z = 0 \mid \bx)}.
\end{equation}
Therefore, a class probability over $\bx$ for Class 1
is considered as an acquisition function;
it is simply derived by~\citet{TiaoLC2021icml} as the following:
\begin{equation}
    A(\bx \mid \zeta, \calD_t) = \zeta^{-1} \pi(\bx).
\end{equation}
Eventually, the class probability is estimated by
various off-the-shelf classifiers such as random forests and multi-layer perceptrons.

\citet{SongJ2022icml} have suggested a general framework of likelihood-free Bayesian optimization, called LFBO:
\begin{align}
    &A(\bx \mid \zeta, \calD_t) \nonumber\\
    &= \argmax_{S: \calX \to \bbR} \bbE_{\calD_t} [u(y; y^\dagger) f'(S(\bx)) - f^*(f'(S(\bx)))],
\end{align}
which is versatile for any non-negative utility function $u(y; y^\dagger)$,
where $f$ is a strictly convex function,
$f'$ is the derivative of $f$,
and $f^*$ is the convex conjugate of $f$.
By the properties of LFBO,
it is equivalent to an expected utility-based acquisition function.
Along with the general framework,
\citet{SongJ2022icml} prove that BORE is equivalent to the probability of improvement~\citep{KushnerHJ1964jbe}
and LFBO with $u(y; y^\dagger) = \bbI(y \leq y^\dagger)$
is also equivalent to the probability of improvement,
where $\bbI$ is an indicator function.
Moreover, they show that
their method with $u(y; y^\dagger) = \max(y^\dagger - y, 0)$,
which can be implemented as weighted binary classification,
is equivalent to the expected improvement~\citep{JonesDR1998jgo}.

\section{DRE-based Bayesian Optimization with Semi-Supervised Learning}
\label{sec:proposed_method}

We introduce DRE-based Bayesian optimization with semi-supervised learning, named~\ours,
by following the previous studies in DRE-based and likelihood-free Bayesian optimization~\citep{BergstraJ2011neurips,TiaoLC2021icml,SongJ2022icml},
which were discussed in~\secref{sec:dre_based}.

\begin{algorithm}[t]
    \caption{\ours}
    \label{alg:ours}
    \begin{algorithmic}[1]
        \REQUIRE Iteration budget $T$, a search space $\calX$, a black-box objective $f$, a threshold ratio $\zeta$, a semi-supervised classifier $\pi_{\bC}$, and unlabeled data points $\bX_u$ if available and the number of unlabeled data points $n_u$, otherwise.
        \ENSURE Best candidate $\bx^+$.
        \STATE Initialize $\calD_0 = \{ (\bx_0, y_0) \}$ by randomly selecting $\bx_0$ from $\calX$ and evaluating $\bx_0$ by $f$.
		\FOR{$t = 0$ {\bfseries to} $T-1$}
            \STATE Calculate a threshold $y_t^\dagger$ using $\zeta$.
            \STATE Assign labels $\bC_t$ of data points in $\calD_t$ with $y_t^\dagger$.
            \IF{$\bX_u$ are not available}
                \STATE Sample $n_u$ unlabeled data points $\bX_u$ from $\calX$.
            \ENDIF
            \STATE Estimate pseudo-labels $\widehat{\bC}_t$ by following the procedure in~\algref{alg:labeling}.
            \STATE Choose the next query point $\bx_{t+1}$ by maximizing $\pi_{\widehat{\bC}_t}(\bx; \zeta, \calD_t, \bX_u)$ for $\bx \in \calX$.
            \STATE Evaluate $\bx_{t+1}$ by $f$ and update $\calD_{t + 1}$.
		\ENDFOR
        \STATE Determine the best candidate $\bx^+$, considering $y_{0:T}$.
	\end{algorithmic}
\end{algorithm}

Similar to the standard Bayesian optimization
and existing DRE-based Bayesian optimization,
\ours~iterates the undermentioned steps as presented in~\algref{alg:ours}.
Firstly, a threshold $y_t^\dagger$ is calculated by considering $y_{1:t}$ with $\zeta$.
Secondly, labels $\bC_t$ of data points in $\calD_t$ are assigned to one of two classes;
a group of $y \leq y_t^\dagger$ is assigned to Class 1
and a group of $y > y_t^\dagger$ is assigned to Class 0.
If we are given unlabeled data points $\bX_u$,
the corresponding points $\bX_u$ are used,
but if not available it samples $n_u$ unlabeled data points $\bX_u$ from $\calX$ by utilizing a strategy described in~\secref{subsec:unlabeled_sampling}.
Then, it estimates pseudo-labels $\widehat{\bC}_t$ of a semi-supervised learning model
by following the procedure presented in~\secref{subsec:semi_supervised} and \algref{alg:labeling}.
Using $\widehat{\bC}_t$,
it chooses the next query point $\bx_{t+1}$:
\begin{equation}
    \bx_{t+1} = \argmax_{\bx \in \calX} \pi_{\widehat{\bC}_t}(\bx; \zeta, \calD_t, \bX_u),
    \label{eqn:argmax_pi}
\end{equation}
where $\pi_{\widehat{\bC}_t}(\bx; \zeta, \calD_t, \bX_u)$ predicts a class probability over $\bx$ for Class 1;
see \eqref{eqn:semi_inductive}.

We adopt a multi-started local optimization technique, e.g., L-BFGS-B~\citep{ByrdRH1995siamjsc},
to solve \eqref{eqn:argmax_pi}.
However,
a flat landscape of $\pi_{\widehat{\bC}_t}(\bx; \zeta, \calD_t, \bX_u)$ over $\bx$ may occur,
so that optimization performance can be degraded.
To tackle this issue,
a simple heuristic of randomly selecting a query point among points with identical highest class probabilities complements our method.
Since the multi-started technique is utilized
and the output of $\pi_{\widehat{\bC}_t}$ is bounded in $[0, 1]$,
a flat landscape is easily recognized
by comparing the outcomes of the multi-started strategy.

\subsection{Label Propagation and Label Spreading}
\label{subsec:semi_supervised}

Here we describe semi-supervised learning techniques~\citep{ZhuX2005tr,ChapelleO2006book,BengioY2006sslbook} to build~\ours.
We cover a transductive setting~\citep{GammermanA1998uai},
which is to label unlabeled data by utilizing given labeled data,
and then an inductive setting, which is to predict any point
using pseudo-labels of unlabeled and labeled data.

Suppose that each data point is defined
on a $d$-dimensional compact space $\calX \subset \bbR^{d}$.
We consider $n_l$ labeled points $\bX_l \in \bbR^{n_l \times d}$,
their corresponding labels $\bC_l \in \bbR^{n_l \times c}$,
and $n_u$ unlabeled points $\bX_u \in \bbR^{n_u \times d}$,
where $c$ is the number of classes.
$\bX_l$ and $\bC_l$ are query points that have already been evaluated and their class labels;
we define
$\bX_l = [\bx_0, \ldots, \bx_{n_l - 1}]^\top$ for $\calD_t = \{ (\bx_i, y_i) \}_{i = 0}^{n_l - 1}$.
For the sake of brevity,
the concatenated data points of $\bX_l$ and $\bX_u$ are defined as
$\bX = [\bX_l; \bX_u] \in \bbR^{(n_l + n_u) \times d}$.
Note that in our problem $c = 2$,
because we address the problem with two classes.

As shown in~\algref{alg:labeling},
we first initialize labels to propagate $\widehat{\bC} \in \bbR^{(n_l + n_u) \times 2}$;
it is initialized as the following:
\begin{equation}
    \widehat{\bC} = [\bc_0, \bc_1, \ldots, \bc_{n_l - 1}, \bc_{n_l}, \bc_{n_l + 1}, \ldots, \bc_{n_l + n_u - 1}]^\top,
\end{equation}
where $\bc_0, \bc_1, \ldots, \bc_{n_l - 1}$ are one-hot representations
$[\bC_l]_{1:}, [\bC_l]_{2:}, \ldots, [\bC_l]_{n_l:}$,
and $\bc_{n_l}, \bc_{n_l + 1}, \ldots, \bc_{n_l + n_u -1}$ are zero vectors.
Denote that $[\bC]_{i:}$ is $i$th row of $\bC$.
Then, we compute
a similarity between two data points $\bx_i, \bx_j \in \calX$,
so that we compare all pairs in $\bX$.
For example, one of popular similarity functions, i.e., a radial basis function kernel, can be used:
\begin{equation}
    w_{ij} = \exp\left( -\beta \| \bx_i - \bx_j \|_2^2 \right),
    \label{eqn:w_ij}
\end{equation}
where $\beta$ is a free parameter given.
As discussed in the work by~\citet{ZhuX2002tr},
we can learn $\beta$ in~\eqref{eqn:w_ij}
by minimizing an entropy for propagated labels $\widehat{\bC}$:
\begin{equation}
    H(\widehat{\bC}) = -\sum_{i = 1}^{n_l + n_u} [\widehat{\bC}]_{i:}^\top \log [\widehat{\bC}]_{i:}.
    \label{eqn:entropy}
\end{equation}
See \figref{fig:gammas} for the results of learning $\beta$
and \secref{sec:discussion_free_parameters} for analysis on the effects of $\beta$.
By \eqref{eqn:w_ij},
we compute a transition probability from $\bx_j$ to $\bx_i$ by
$p_{ij} = w_{ij} / \sum_{k = 1}^{n_l + n_u} w_{kj}$.
Note that similarities $\bW \in \bbR^{(n_l + n_u) \times (n_l + n_u)}$ and transition probabilities $\bP \in \bbR^{(n_l + n_u) \times (n_l + n_u)}$ are defined, where $[\bW]_{ij} = w_{ij}$ and $[\bP]_{ij} = p_{ij}$.
Moreover,
by the definition of $p_{ij}$,
$\bP = \bD^{-1} \bW$,
where $\bD$ is a degree matrix
whose diagonal entry is defined as the following:
\begin{equation}
    [\bD]_{ii} = \sum_{j = 1}^{n_l + n_u} [\bW]_{ij}.
\end{equation}

With initial $\widehat{\bC}$ and $\bP$,
we repeat the following steps:
(i) computing the next $\widehat{\bC}$;
(ii) normalizing $\widehat{\bC}$ row-wise,
until a change of $\widehat{\bC}$ converges to a tolerance $\varepsilon$
or the number of iterations propagated reaches to maximum iterations $\tau$.
In particular, label propagation~\citep{ZhuX2002tr}
updates $\widehat{\bC}$ and constrains the labels of labeled data to $\bC_l$:
\begin{align}
    \widehat{\bC}_{t + 1} &\leftarrow \bP^\top \widehat{\bC}_t, \\
    [\widehat{\bC}_{t + 1}]_{i:} &\leftarrow [\bC_l]_{i:},
\end{align}
for $i \in [n_l]$ at Iteration $t$.
On the other hand,
label spreading~\citep{ZhouD2003neurips_b}
propagates $\widehat{\bC}$ by allowing a change of the labels of labeled data with a clamping factor $\alpha \in (0, 1)$:
\begin{equation}
    \widehat{\bC}_{t + 1} \leftarrow \alpha \bD^{-1/2} \bW \bD^{-1/2} \widehat{\bC}_t + (1 - \alpha) \widehat{\bC}_0,
\end{equation}
where $\widehat{\bC}_0$ is initial propagated labels,
which are defined in Line 1 of~\algref{alg:labeling}.
Note that $\bD^{-1/2} \bW \bD^{-1/2}$ can be pre-computed before the loop defined from Lines 4 to 7 of~\algref{alg:labeling}.

By the nature of transductive setting,
it only predicts the labels of particular data using the known data~\citep{GammermanA1998uai},
which implies that it cannot estimate a categorical distribution of unseen data.
To enable it,
given unseen $\bx$,
we define an inductive model with $\widehat{\bC}$:
\begin{equation}
    \widehat{c}_i = \frac{\bw^\top [\widehat{\bC}]_{:i}}{\sum_{j = 1}^{c} \bw^\top [\widehat{\bC}]_{:j}},
    \label{eqn:semi_inductive}
\end{equation}
for $i \in [2]$,
where $\bw \in \bbR^{n_l + n_u}$ is similarities of $\bx$ and $\bX$
by~\eqref{eqn:w_ij}.
$[\widehat{\bC}]_{:i}$ denotes $i$th column of $\widehat{\bC}$.
This inductive model is better than
or equivalent to other classifiers without unlabeled data in certain circumstances;
its analysis is provided in~\secref{sec:analysis}.

\subsection{Unlabeled Point Sampling}
\label{subsec:unlabeled_sampling}

As described above,
if unlabeled points are not available,
we need to generate unlabeled points
under a transductive learning scheme.
However, it is a challenging problem
unless we know a landscape of $\pi_{\bC}$ adequately.
Many studies by~\citet{SeegerM2000tr,RigolletP2007jmlr,SinghA2008neurips,BenDavidS2008colt,CarmonY2019neurips,WeiC2020iclr,ZhangS2022iclr}
investigate how unlabeled data can affect a model
and whether unlabeled data helps improve the model or not.

In order to make use of the theoretical findings of previous literature,
we define a cluster assumption:
\begin{assumption}[Cluster assumption in the work by~\citet{SeegerM2000tr}]
    \label{asm:cluster}
    Two points $\bx_i, \bx_j \in \calX$ should belong to the same label if there is a path between $\bx_i$ and $\bx_j$ which passes only through regions of relatively high $P(X)$,
    where $P$ is a distribution over a random variable $X \in \calX$.
\end{assumption}

\begin{figure*}[t]
    \centering
    \subfigure[Beale]{
        \centering
        \includegraphics[width=0.23\textwidth]{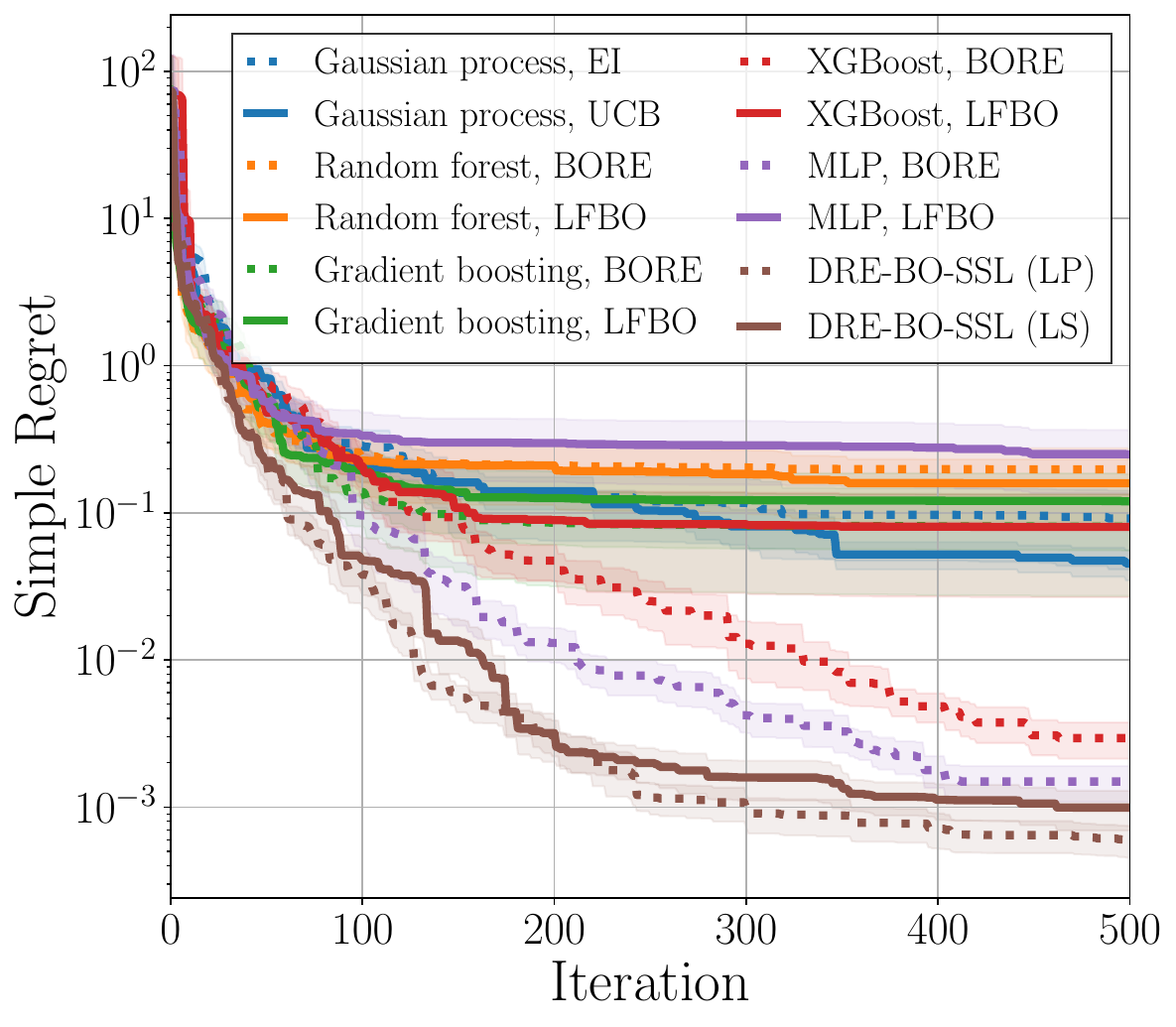}
    }
    \subfigure[Branin]{
        \centering
        \includegraphics[width=0.23\textwidth]{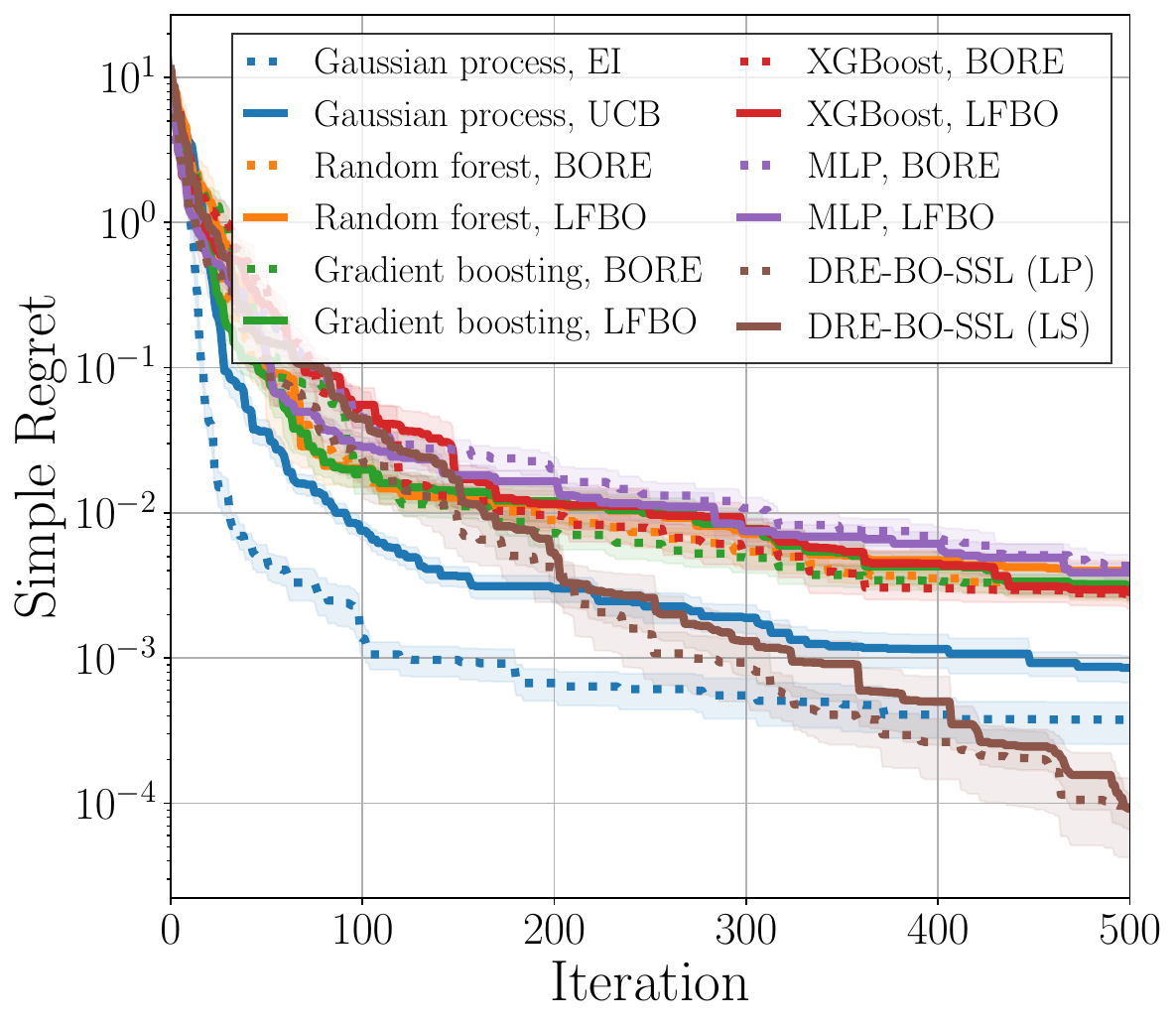}
    }
    \subfigure[Bukin6]{
        \centering
        \includegraphics[width=0.23\textwidth]{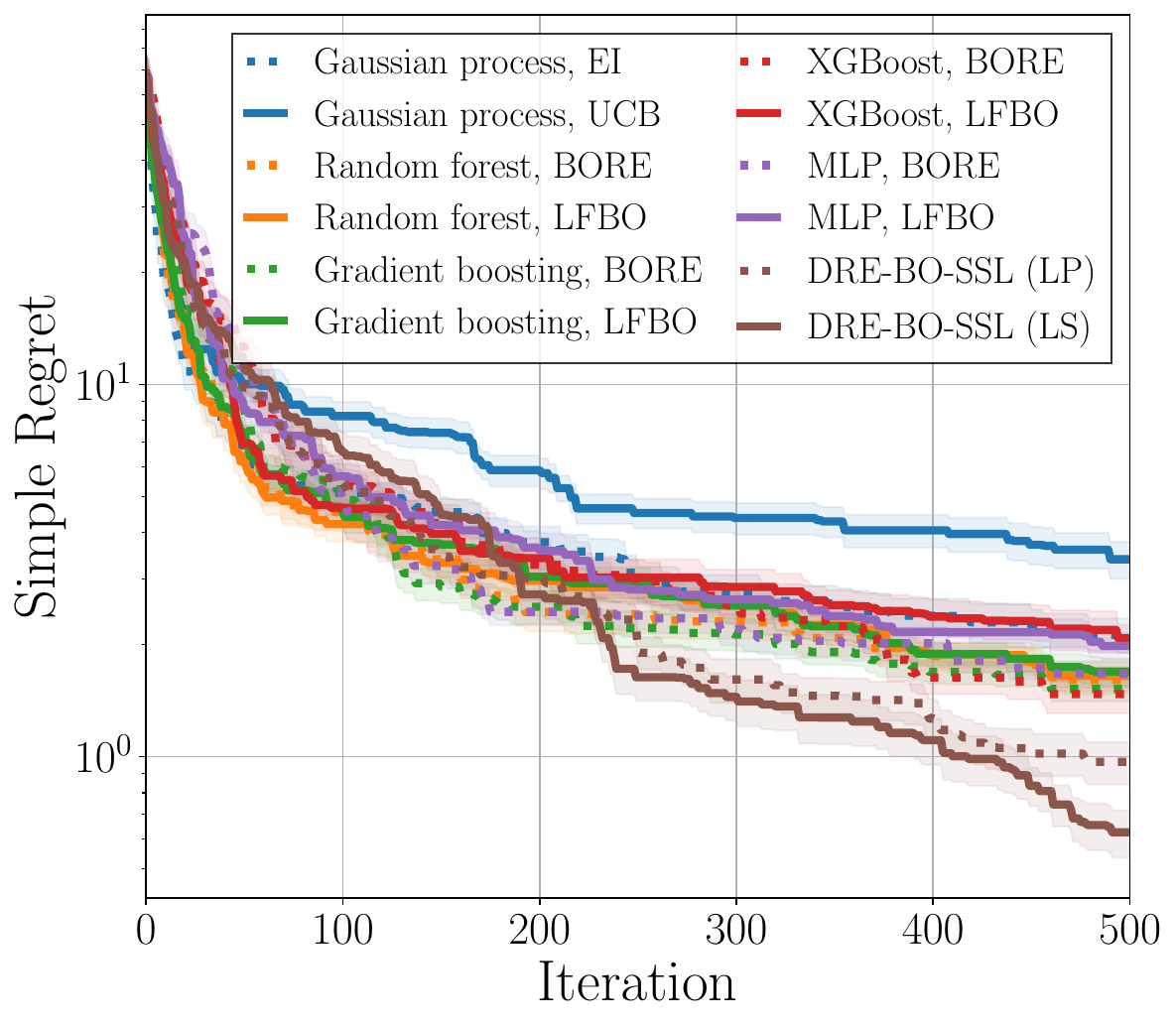}
    }
    \subfigure[Six-hump camel]{
        \centering
        \includegraphics[width=0.23\textwidth]{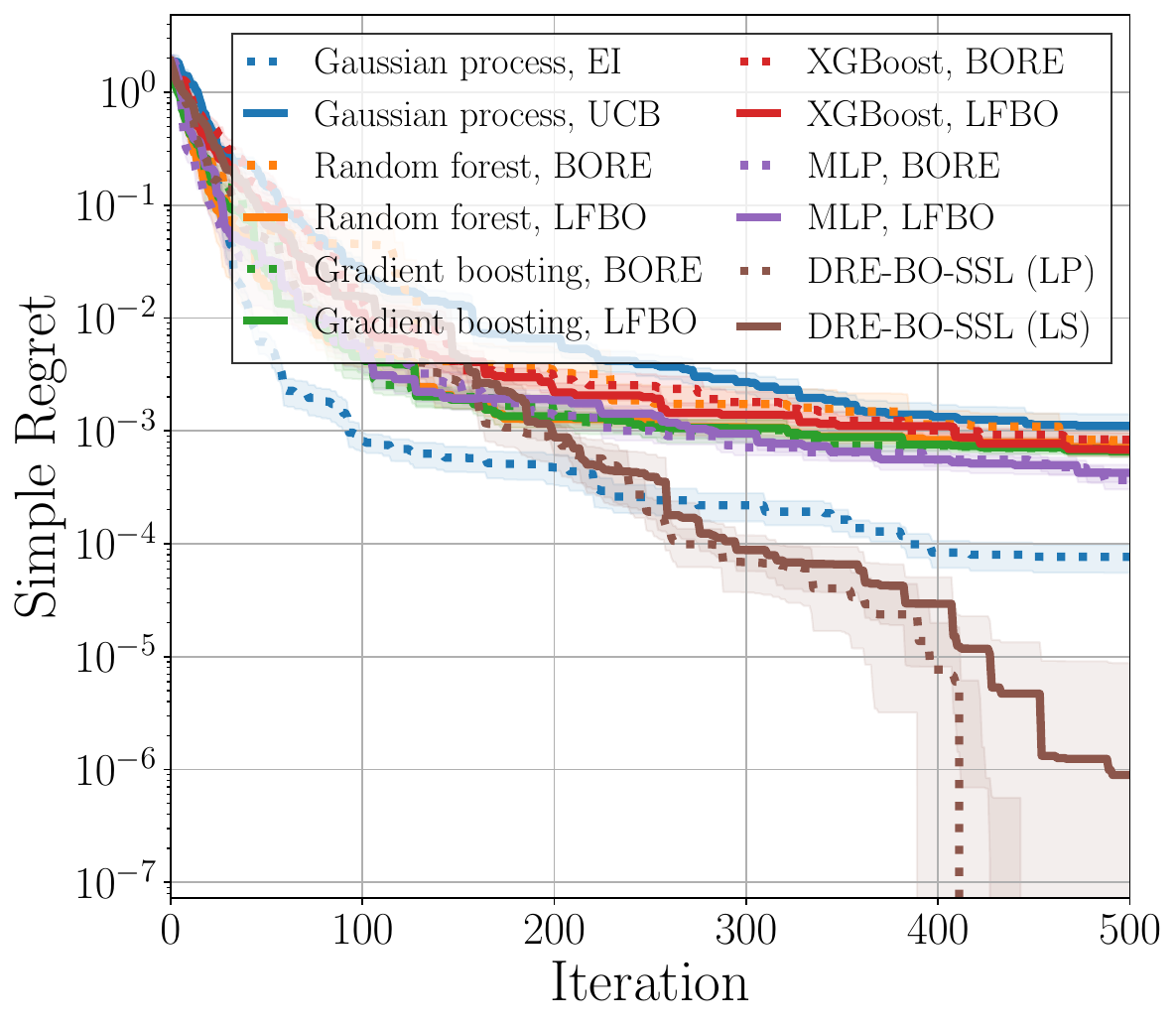}
    }
    \caption{Results with 20 repeated experiments on synthetic benchmark functions for a scenario with unlabeled point sampling. LP and LS stand for label propagation and label spreading, respectively.}
    \label{fig:synthetic_sampling}
\end{figure*}

\begin{figure*}[t]
    \centering
    \subfigure[Beale]{
        \centering
        \includegraphics[width=0.23\textwidth]{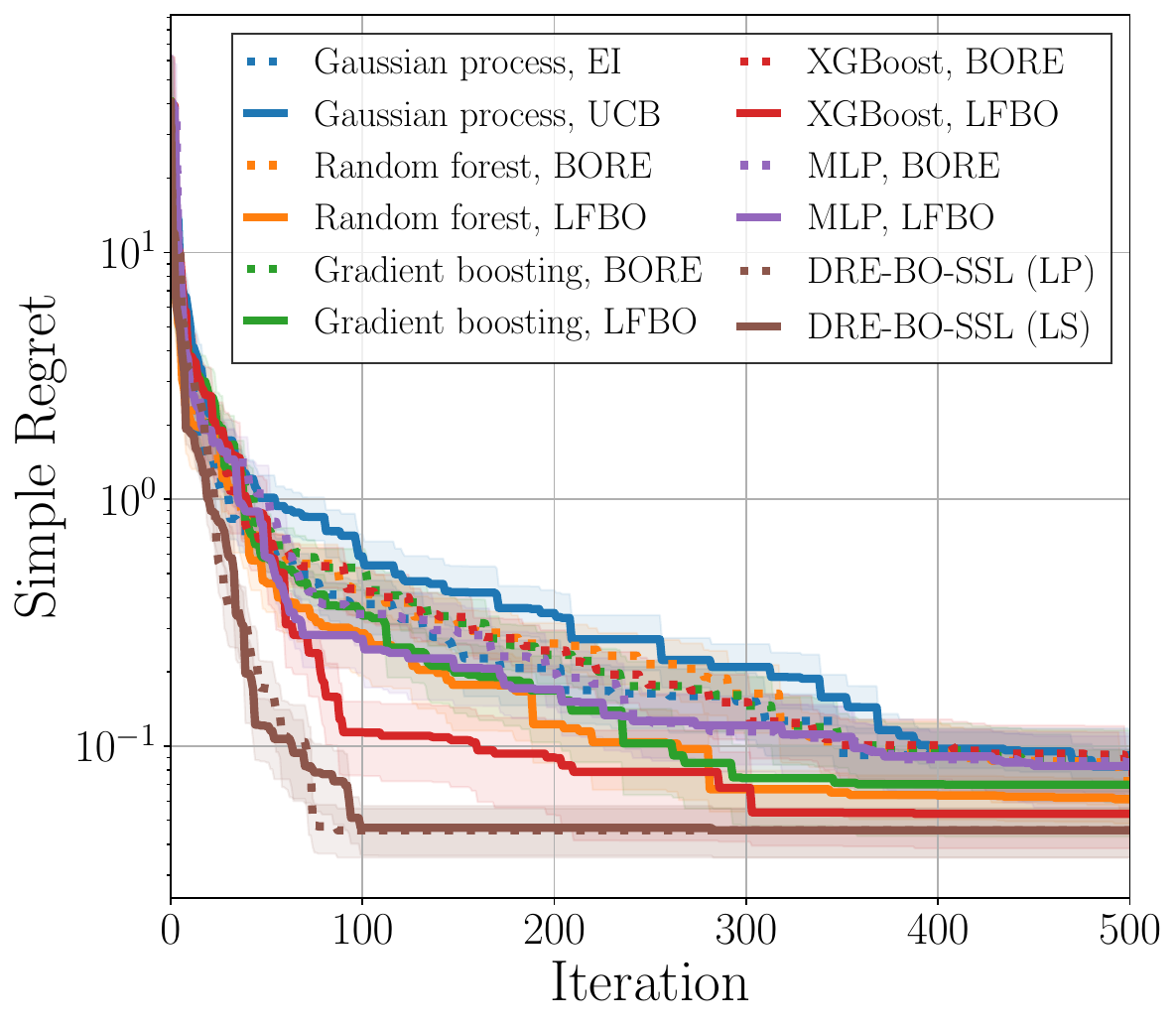}
    }
    \subfigure[Branin]{
        \centering
        \includegraphics[width=0.23\textwidth]{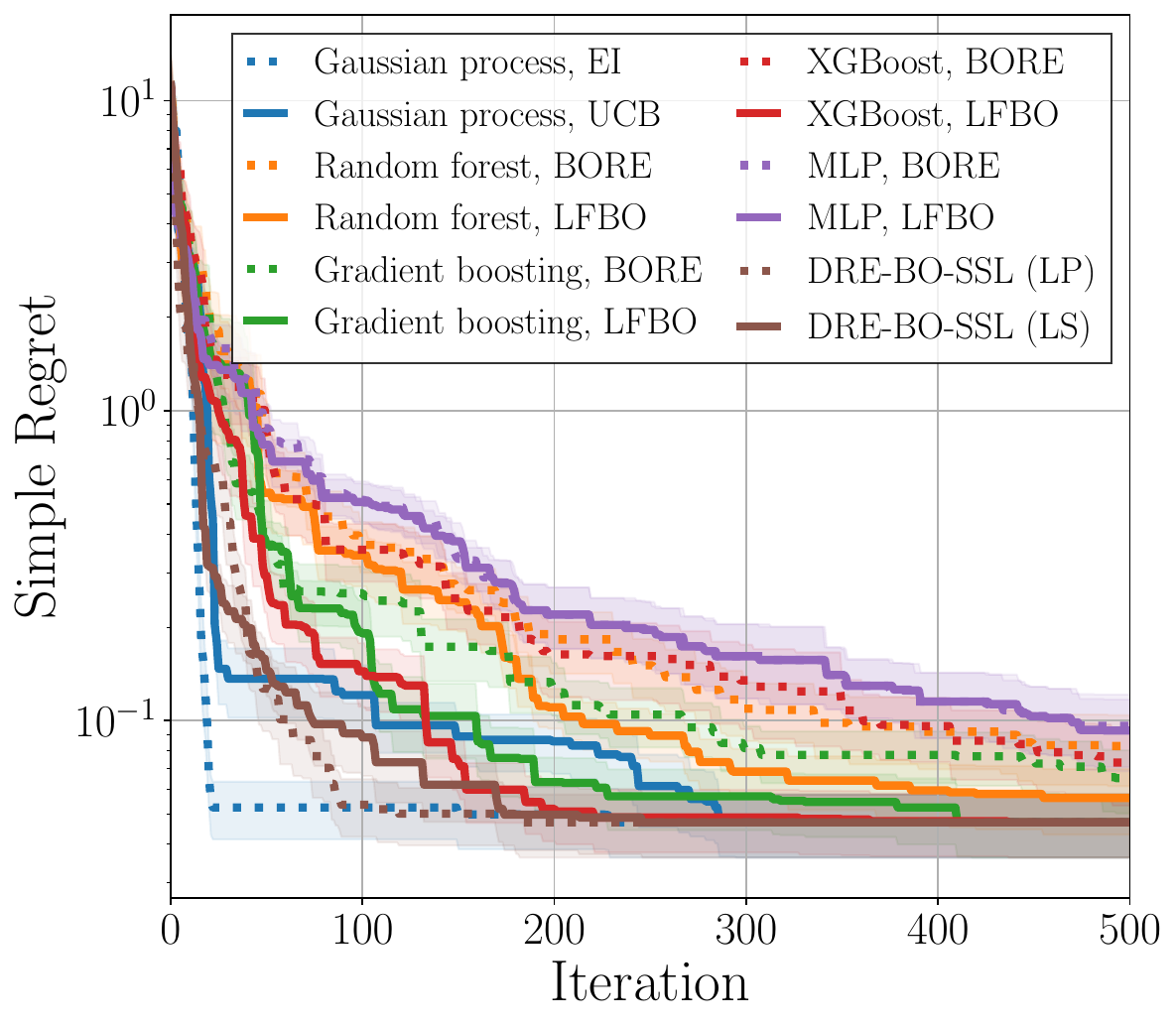}
    }
    \subfigure[Bukin6]{
        \centering
        \includegraphics[width=0.23\textwidth]{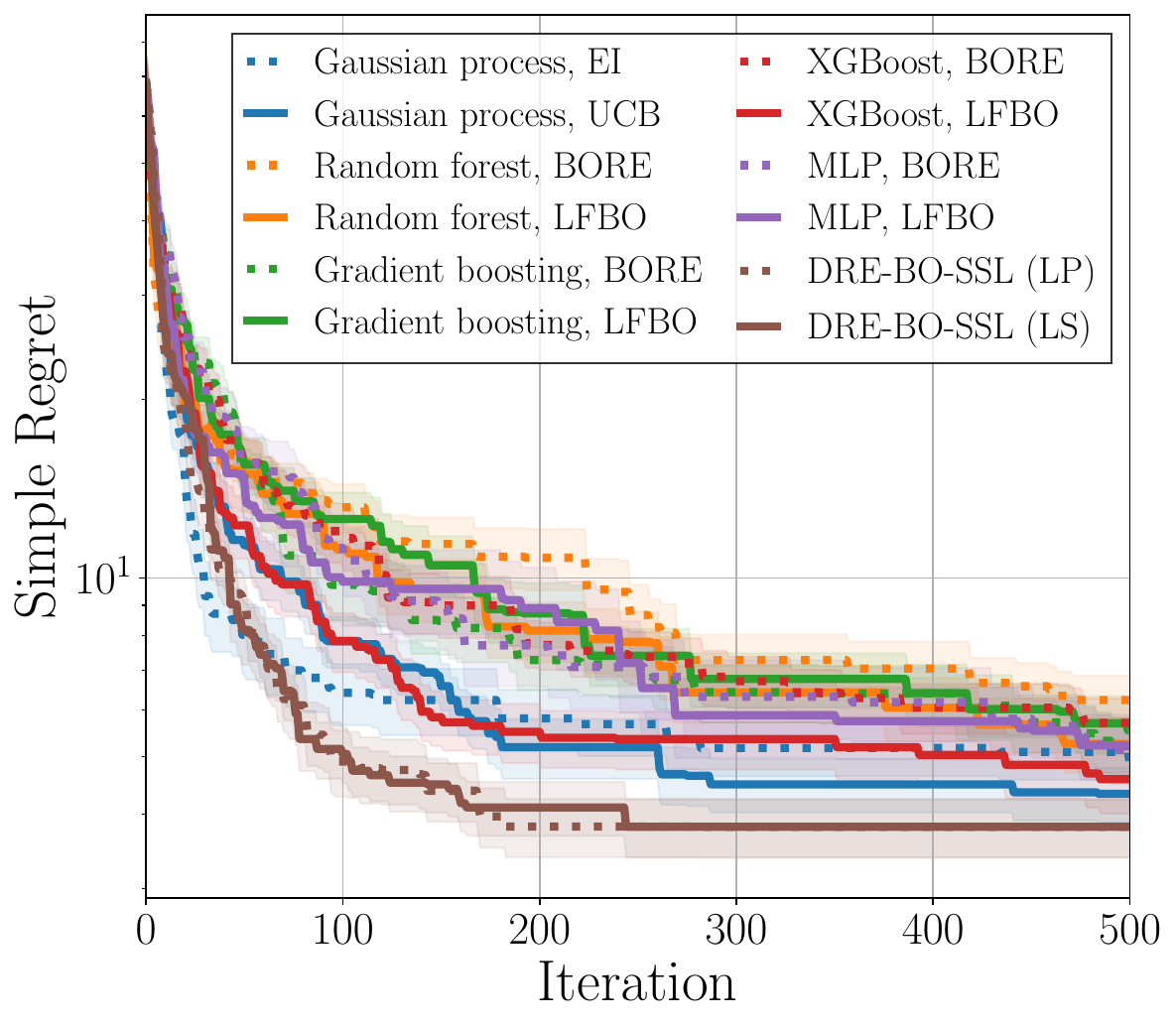}
    }
    \subfigure[Six-hump camel]{
        \centering
        \includegraphics[width=0.23\textwidth]{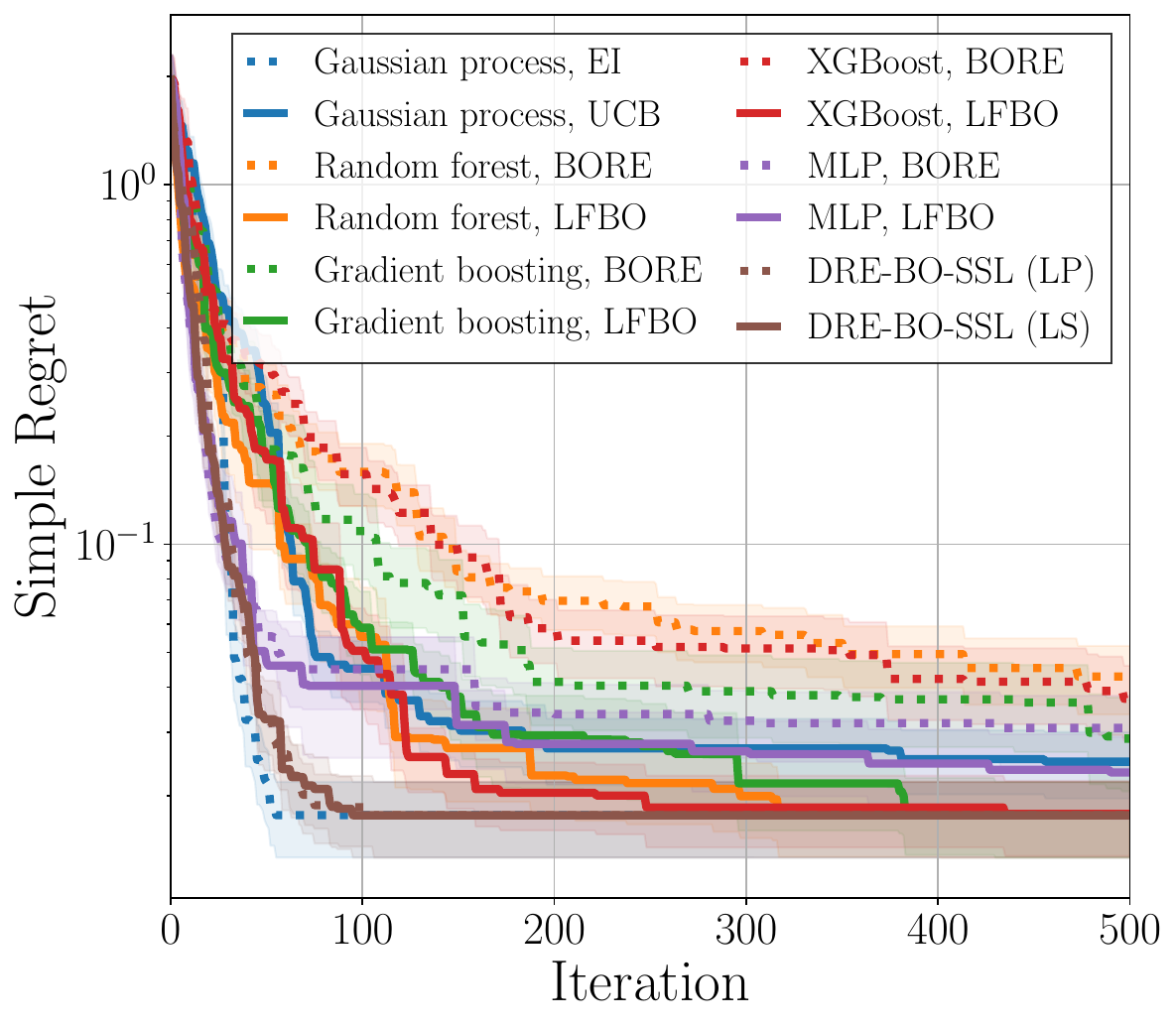}
    }
    \caption{Results with 20 repeated experiments on synthetic benchmark functions for a scenario with fixed-size pools. LP and LS stand for label propagation and label spreading, respectively.}
    \label{fig:synthetic_pools}
\end{figure*}

By~\asmref{asm:cluster},
the idea of clustering on the Euclidean space
or spectral clustering on a graph
can be naturally applied in semi-supervised learning~\citep{SeegerM2000tr,JoachimsT2003icml},
which is not the scope of this work.

To build~\ours~associated with~\asmref{asm:cluster},
we sample unlabeled data points
from the truncated multivariate normal distribution
so that each sample is in a compact $\calX$:
\begin{equation}
    f(\bz) = \frac{\exp(-\frac{1}{2} \bz^\top \bz) \bbI(\bl \leq \bA \bz \leq \bu)}{P(\bl \leq \bA Z \leq \bu)},
    \label{eqn:truncated_mvn}
\end{equation}
where $\bl, \bu \in \bbR^d$ are lower and upper bounds,
$\bsSigma = \bA \bA^\top$ is a covariance matrix,
$\bbI$ is an indicator function,
and $Z \sim \calN(\boldsymbol 0, \bI_d)$ is a random variable.
It is challenging to calculate a denominator of~\eqref{eqn:truncated_mvn},
$P(\bl \leq \bA^\top Z \leq \bu)$,
and simulate from $f(\bz)$
because an integration of the denominator and an accept-reject sampling strategy from $f(\bz)$ are cumbersome in this multi-dimensional case.
To effectively sample from the truncated multivariate normal distribution,
we adopt the minimax tilting method~\citep{BotevZI2017jrssb}.
Compared to the method by~\citet{GenzA1992jcgs},
it yields a high acceptance rate and accurate sampling.
In this paper $\bsSigma$ is set as an identity matrix, and $\bl$ and $\bu$ are determined by a search space.
We will provide more detailed discussion on point sampling in~\secsref{sec:discussion}{sec:discussion_sampling}.

\section{Experiments}
\label{sec:experiments}

\begin{figure*}[t!]
    \centering
    \subfigure[Sampling, LP]{
        \centering
        \includegraphics[width=0.23\textwidth]{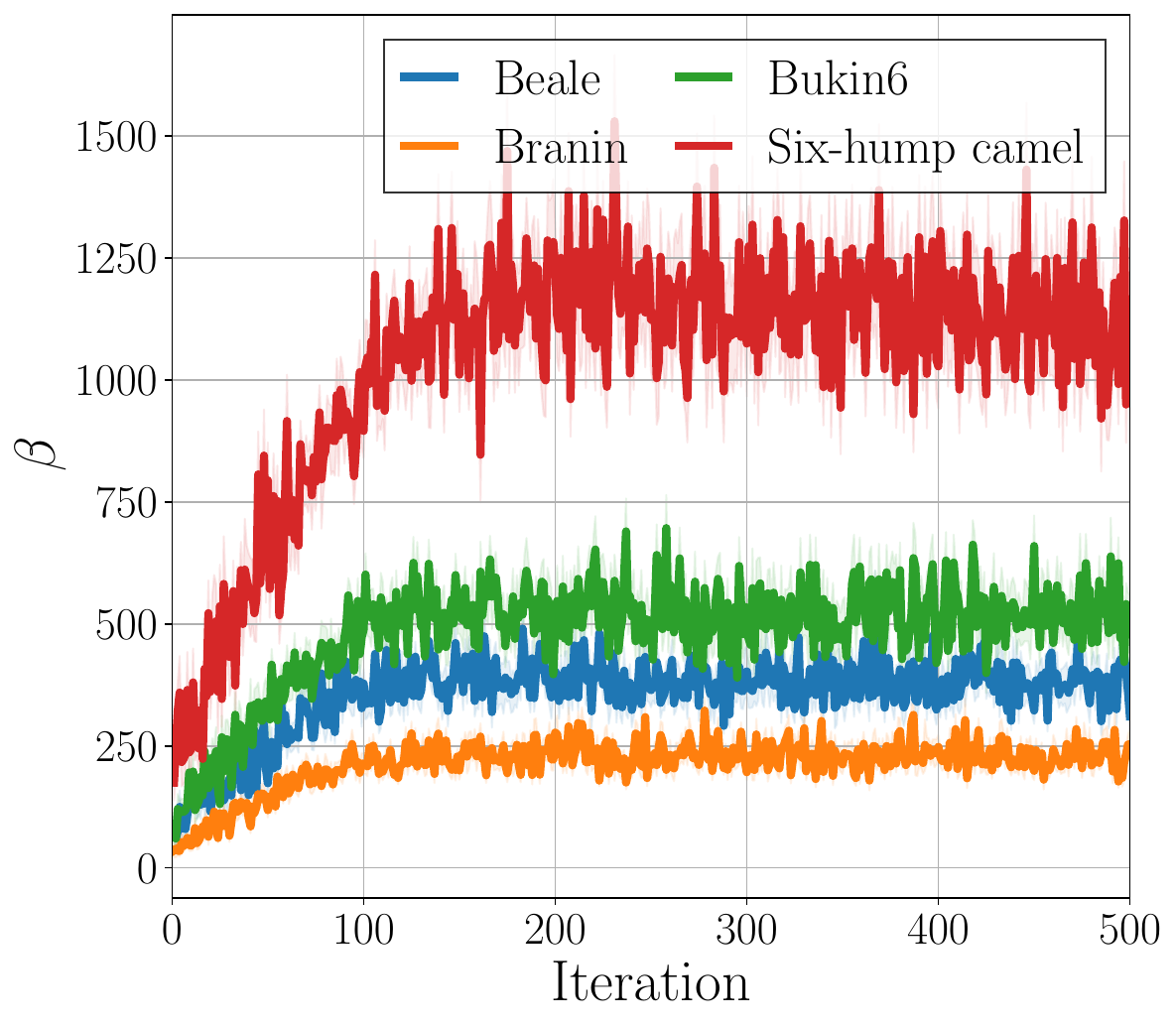}
    }
    \subfigure[Sampling, LS]{
        \centering
        \includegraphics[width=0.23\textwidth]{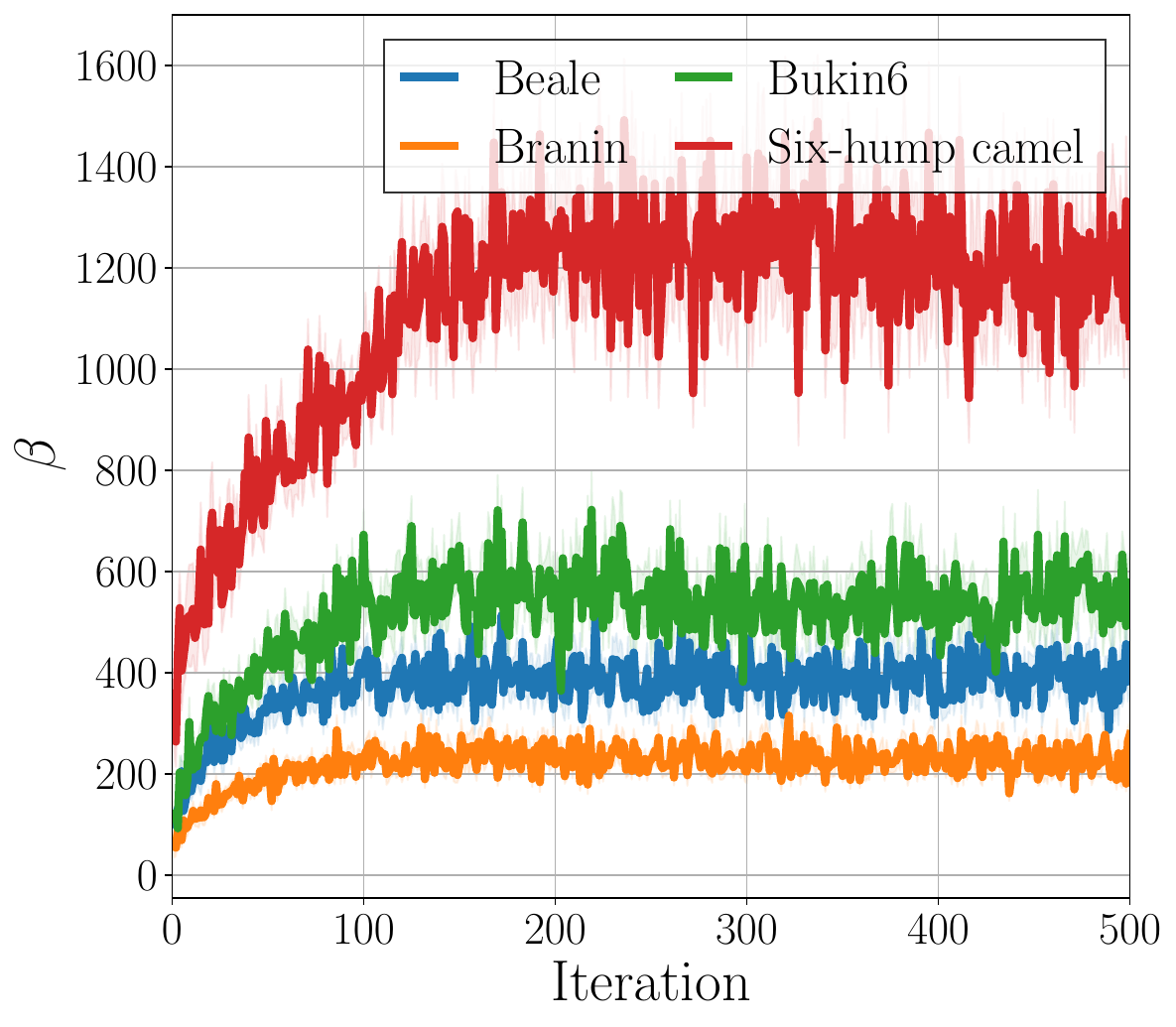}
    }
    \subfigure[Pool, LP]{
        \centering
        \includegraphics[width=0.23\textwidth]{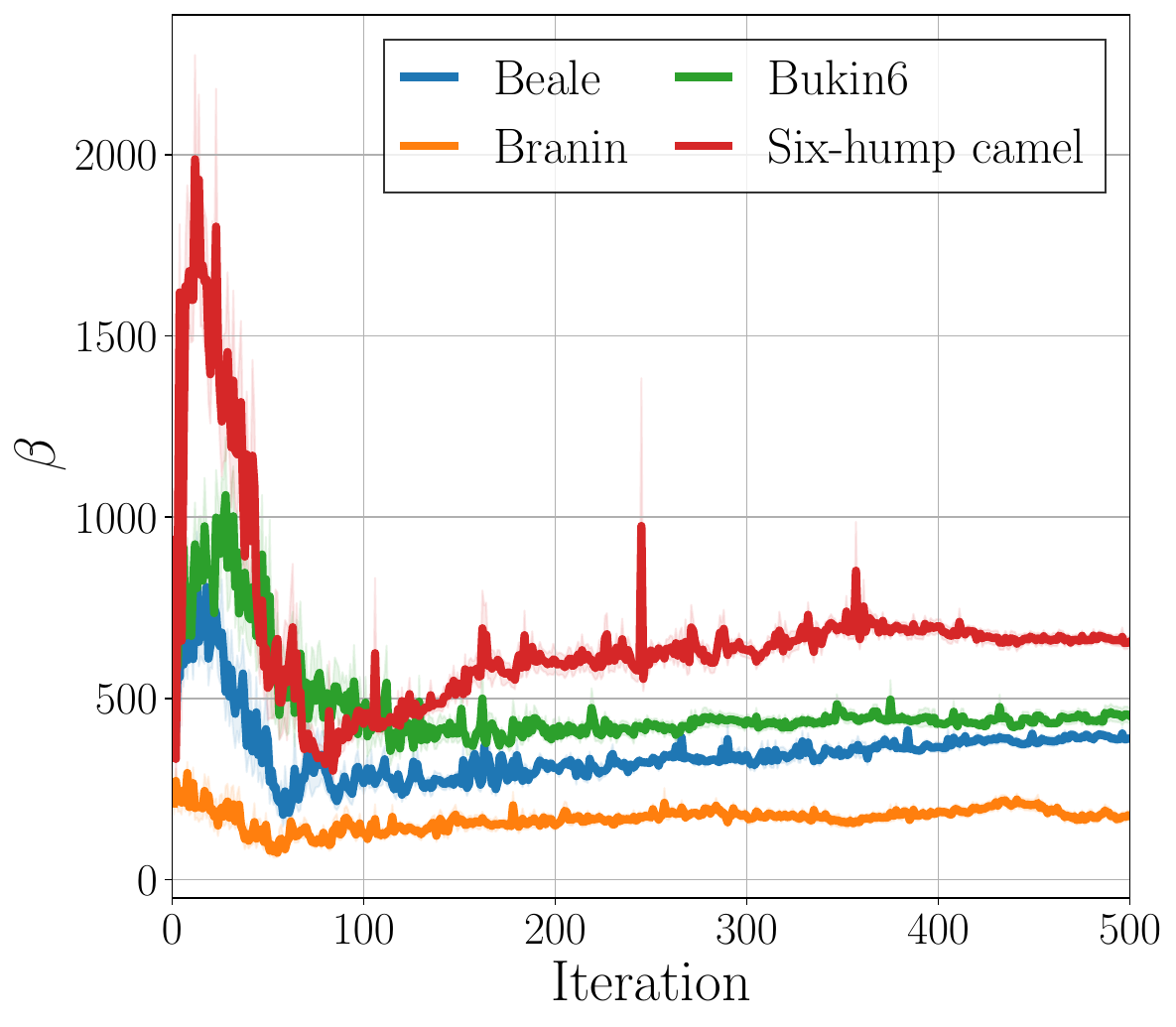}
    }
    \subfigure[Pool, LS]{
        \centering
        \includegraphics[width=0.23\textwidth]{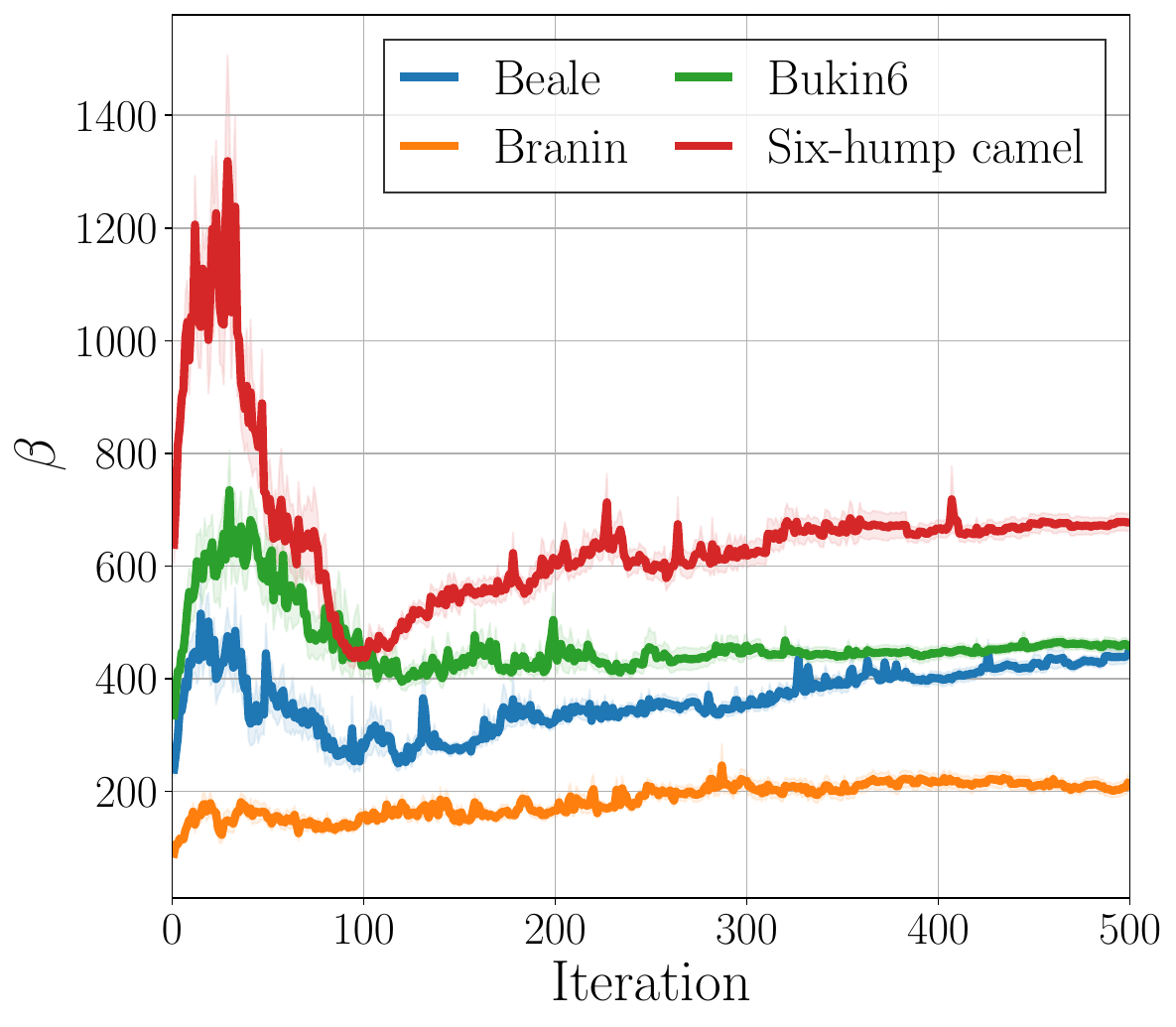}
    }
    \caption{Results with 20 repeated experiments on learning $\beta$ for label propagation, which is denoted as LP, and label spreading, which is denoted as LS. Sampling and pool indicate the experiments in \secsref{subsec:scenario_sampling}{subsec:scenarios_pools}, respectively.}
    \label{fig:gammas}
\end{figure*}

We compare baseline methods with \ours~in the following optimization problems:
synthetic benchmarks for a scenario with unlabeled point sampling,
and synthetic benchmarks,
Tabular Benchmarks~\citep{KleinA2019arxiv},
NATS-Bench~\citep{DongX2021ieeetpami},
and minimum multi-digit MNIST search
for a scenario with a fixed-size pool.
Note that Tabular Benchmarks, NATS-Bench, and minimum multi-digit MNIST search are defined with a fixed number of possible solution candidates, which implies that they are considered as combinatorial optimization problems.
By following the previous work by~\citet{TiaoLC2021icml,SongJ2022icml},
we set a threshold ratio as $\zeta = 0.33$
for all experiments;
the effects of a threshold ratio $\zeta$ are analyzed in~\secsref{sec:discussion}{sec:discussion_threshold_ratios}.
To solve \eqref{eqn:argmax_pi},
we use L-BFGS-B~\citep{ByrdRH1995siamjsc} with 1,000 different initializations.
All experiments are repeated 20 times with 20 fixed random seeds,
where 5 initial points are given to each experiment.
The sample mean and the standard error of the sample mean are reported.
Other missing details including the details of the competitors of our methods are presented in~\secref{sec:exp_details}.

As the competitors of our method,
we test the following baseline methods:
\begin{itemize}
    \item Gaussian process, EI and UCB: It is a Bayesian optimization strategy, which is defined with Gaussian process regression with the Mat\'ern 5/2 kernel~\citep{RasmussenCE2006book}, where expected improvement~\citep{JonesDR1998jgo} or Gaussian process upper confidence bound~\citep{SrinivasN2010icml} is used as an acquisition function;
    \item Random forest, BORE and LFBO: These are BORE and LFBO that employ random forests~\citep{BreimanL2001ml} with 1,000 decision trees, where minimum samples to split are set to 2 for these baselines;
    \item Gradient boosting, BORE and LFBO: These methods are BORE and LFBO with gradient boosting classifiers~\citep{FriedmanJH2001aos} with 100 decision trees, where a learning rate for the classifier is set to 0.3;
    \item XGBoost, BORE and LFBO: Similar to gradient boosting, BORE and LFBO with XGBoost~\citep{ChenT2016kdd} have 100 decision trees as base learners with a learning rate of 0.3;
    \item MLP, BORE and LFBO: These methods are built with two-layer fully-connected networks; the detail of the multi-layer perceptron is described as follows.
\end{itemize}

The architecture of the multi-layer perceptron is set as the following:
(i) first layer: fully-connected, input dimensionality $d$, output dimensionality 32, ReLU;
(ii) second layer: fully-connected, input dimensionality 32, output dimensionality 1, Logistic,
where $d$ is the dimensionality of the problem we solve.

Note that most configurations for the baselines
follow the configurations described in the work by~\citet{SongJ2022icml}.

\begin{figure*}[t]
    \centering
    \subfigure[Naval]{
        \centering
        \includegraphics[width=0.23\textwidth]{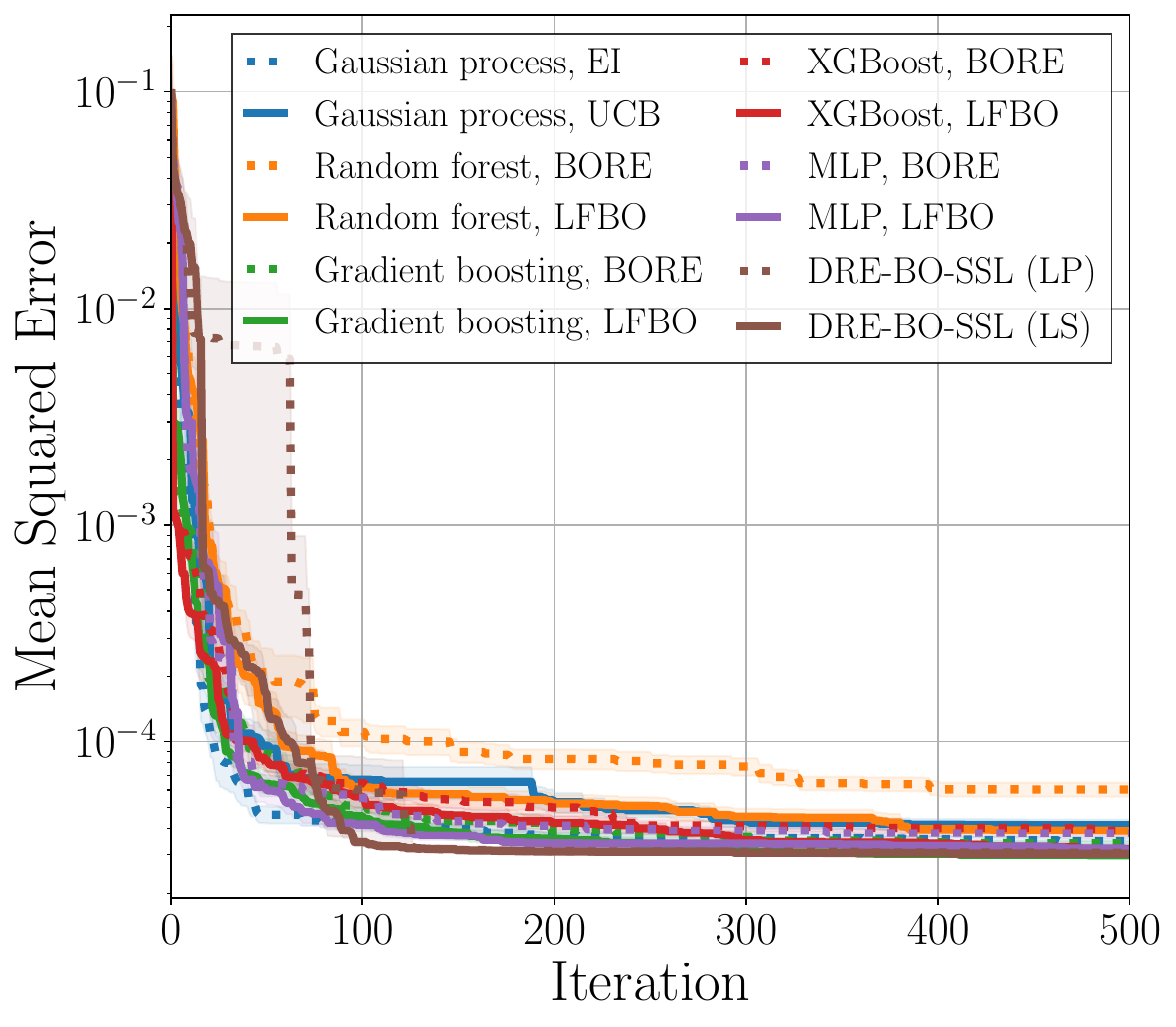}
    }
    \subfigure[Protein]{
        \centering
        \includegraphics[width=0.23\textwidth]{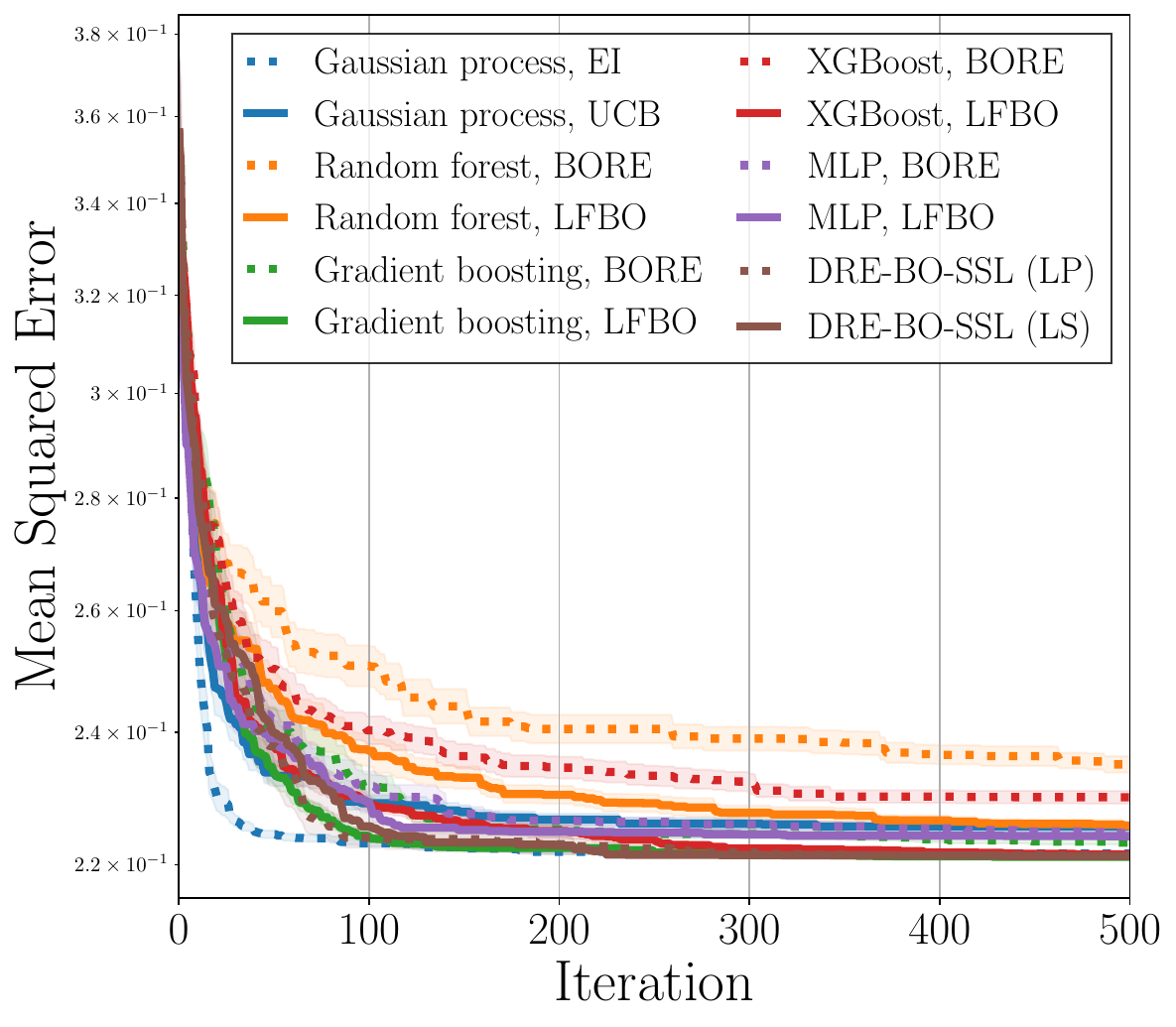}
    }
    \subfigure[Parkinson's]{
        \centering
        \includegraphics[width=0.23\textwidth]{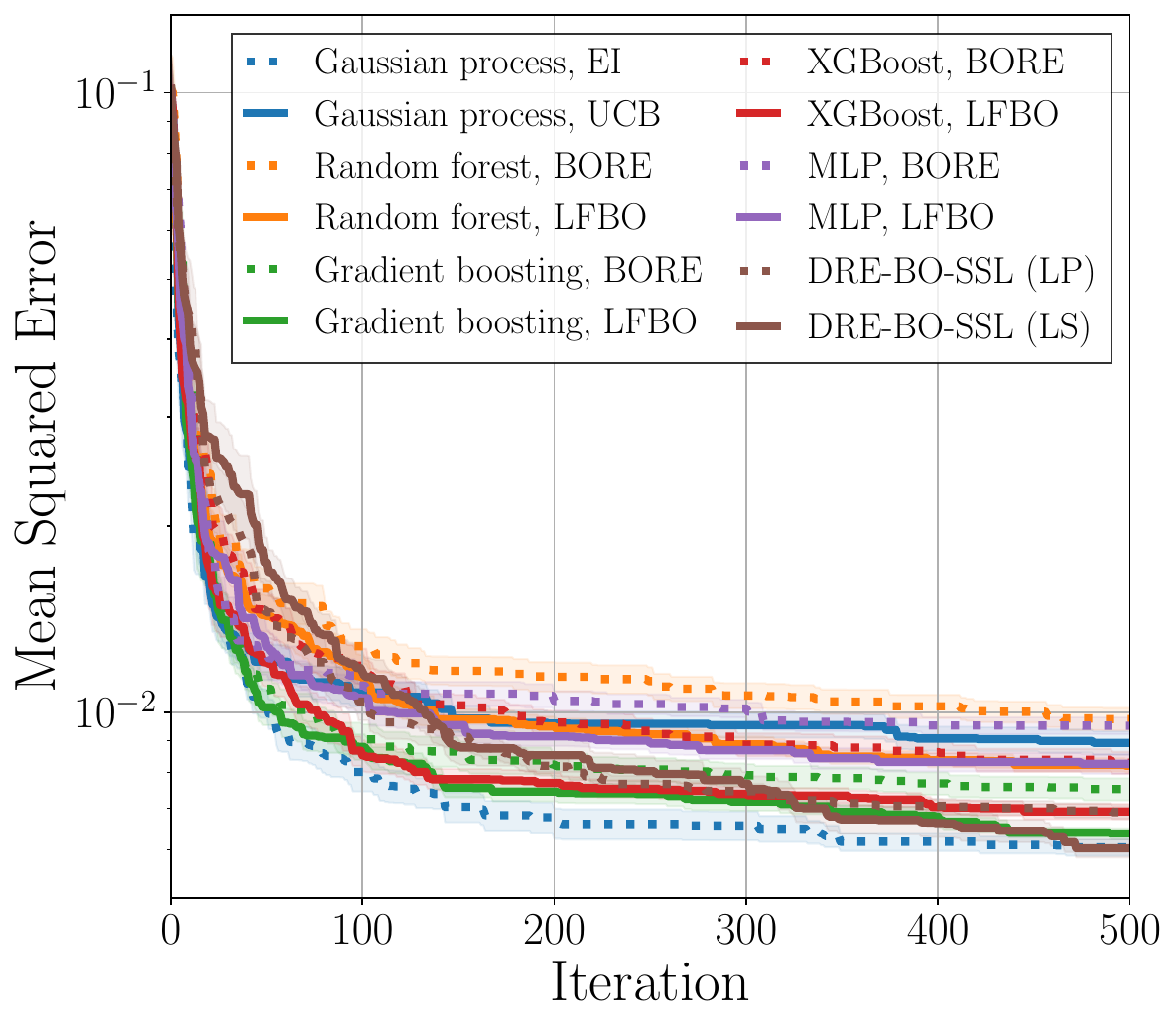}
    }
    \subfigure[Slice]{
        \centering
        \includegraphics[width=0.23\textwidth]{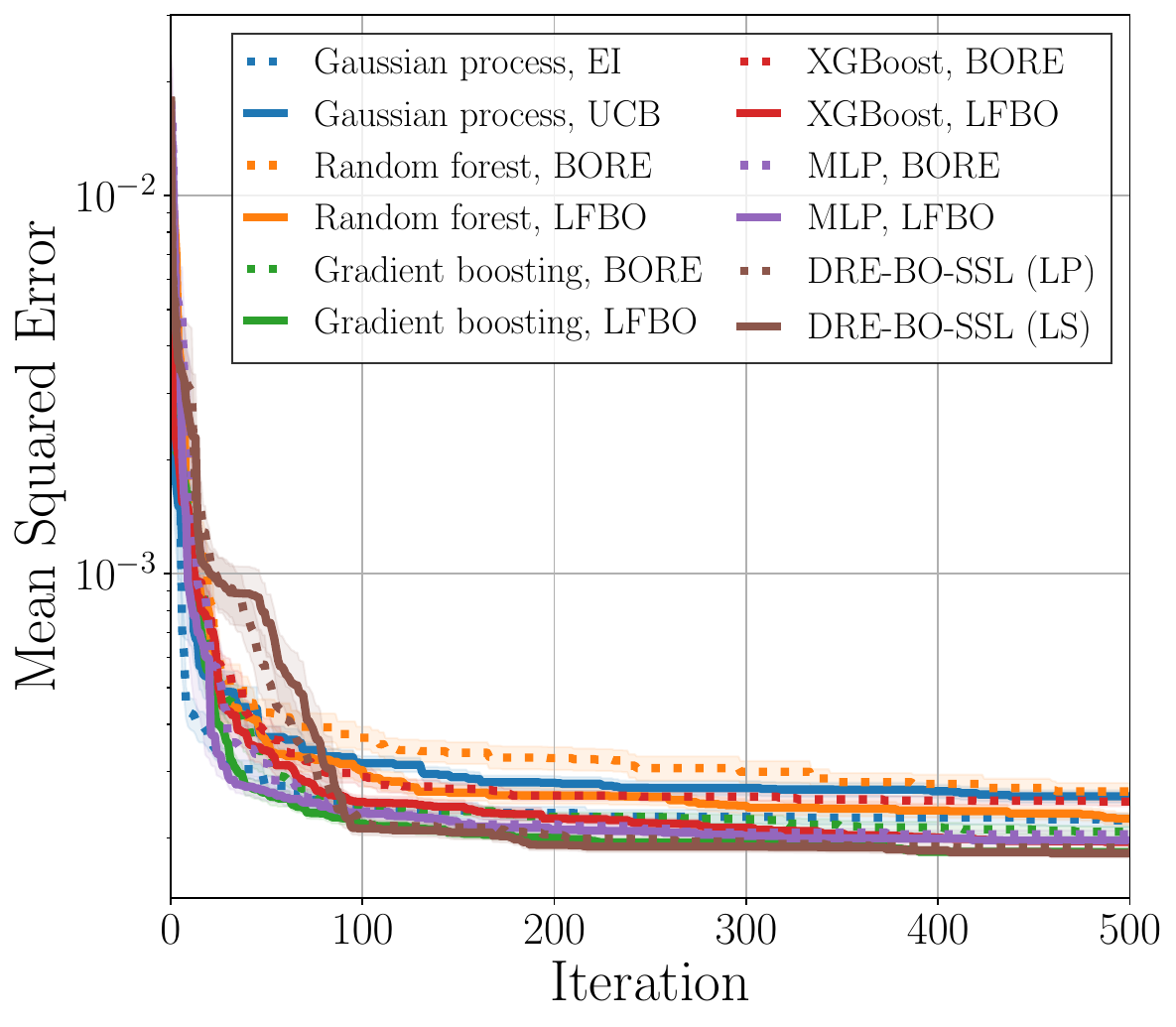}
    }
    \caption{Results with 20 repeated experiments on Tabular Benchmarks for a scenario with fixed-size pools. LP and LS stand for label propagation and label spreading, respectively.}
    \label{fig:tabular}
\end{figure*}

\begin{figure*}[t]
    \centering
    \subfigure[CIFAR-10]{
        \includegraphics[width=0.315\textwidth]{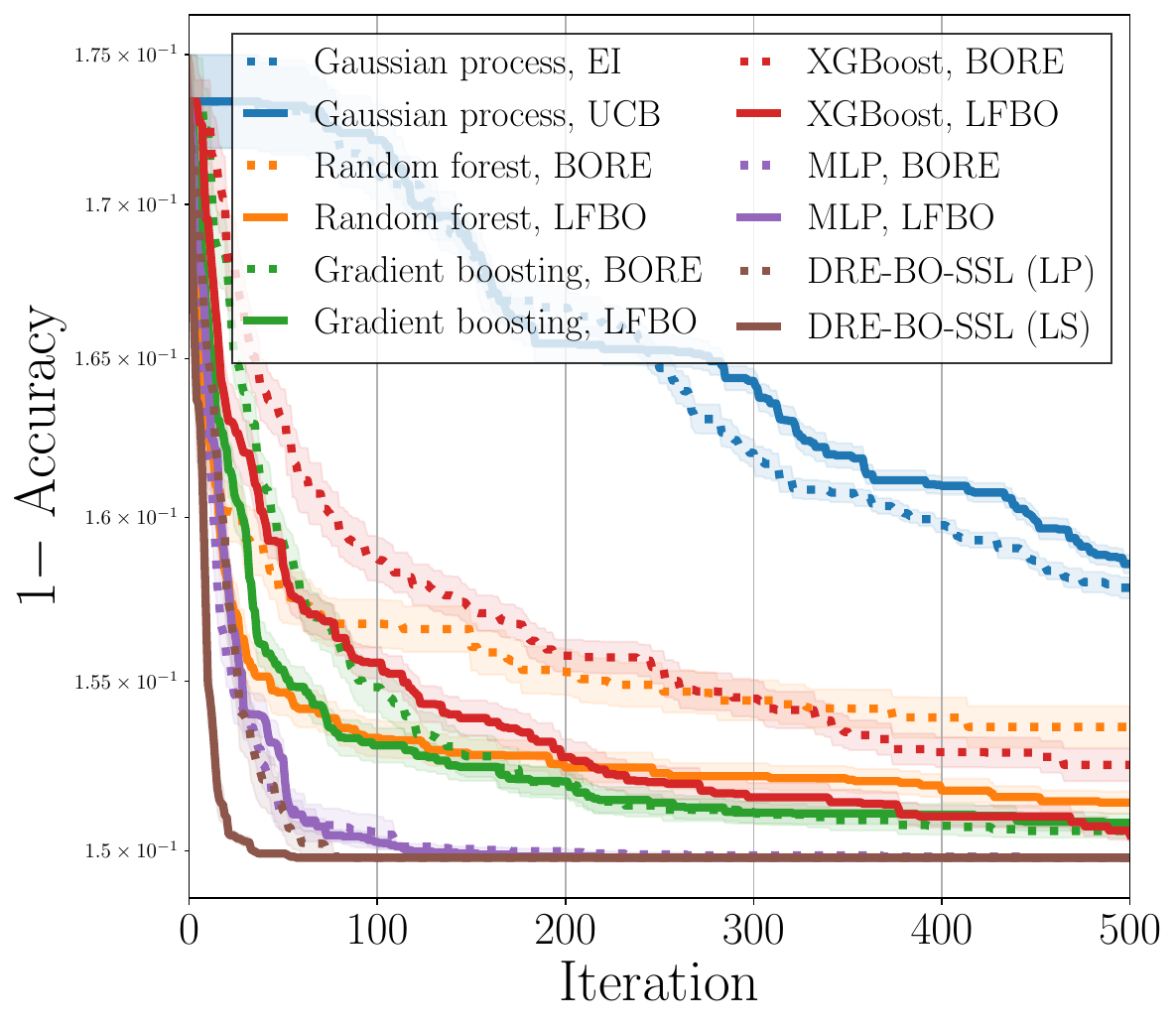}
    }
    \subfigure[CIFAR-100]{
        \includegraphics[width=0.315\textwidth]{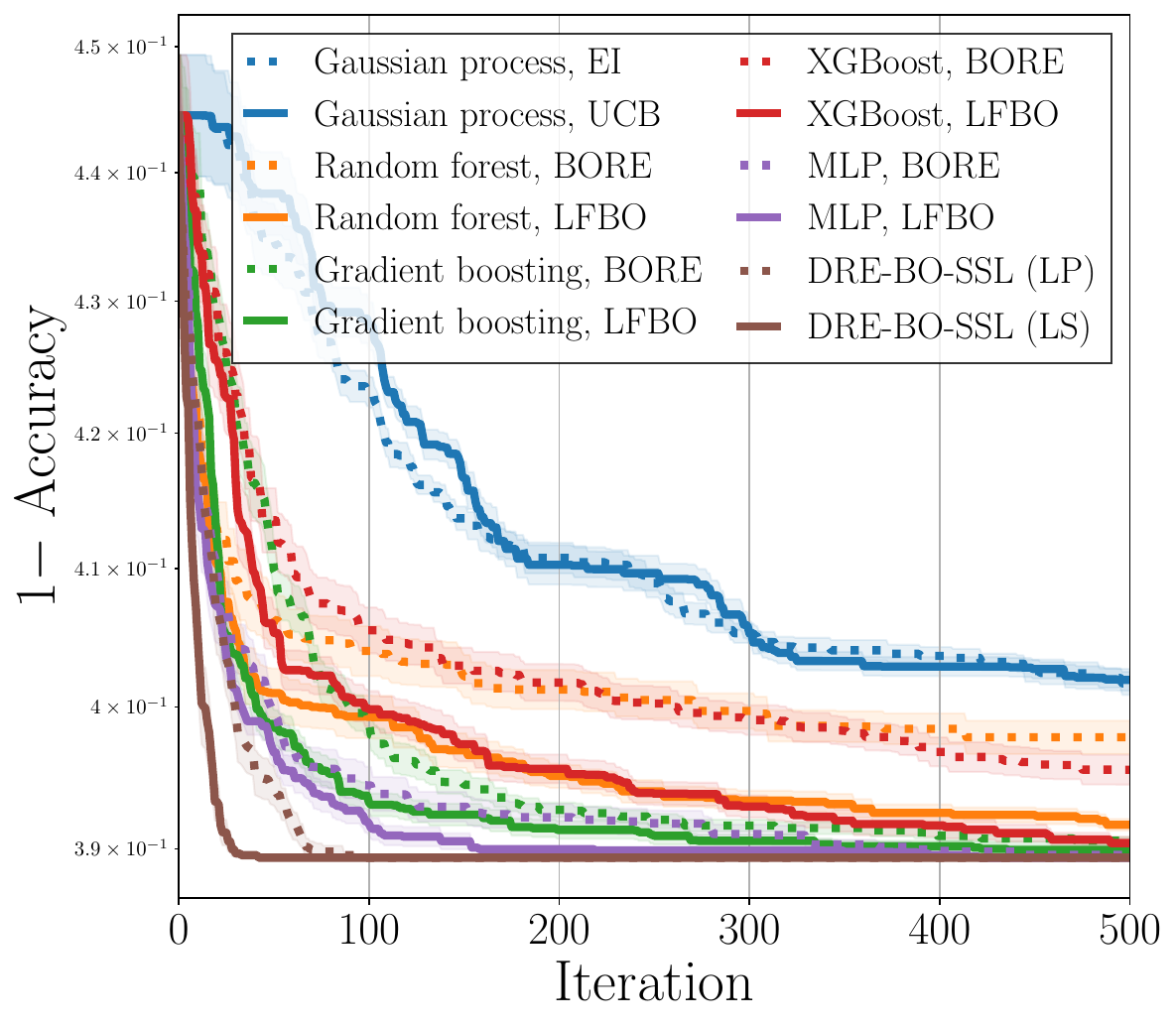}
    }
    \subfigure[ImageNet-16-120]{
        \includegraphics[width=0.315\textwidth]{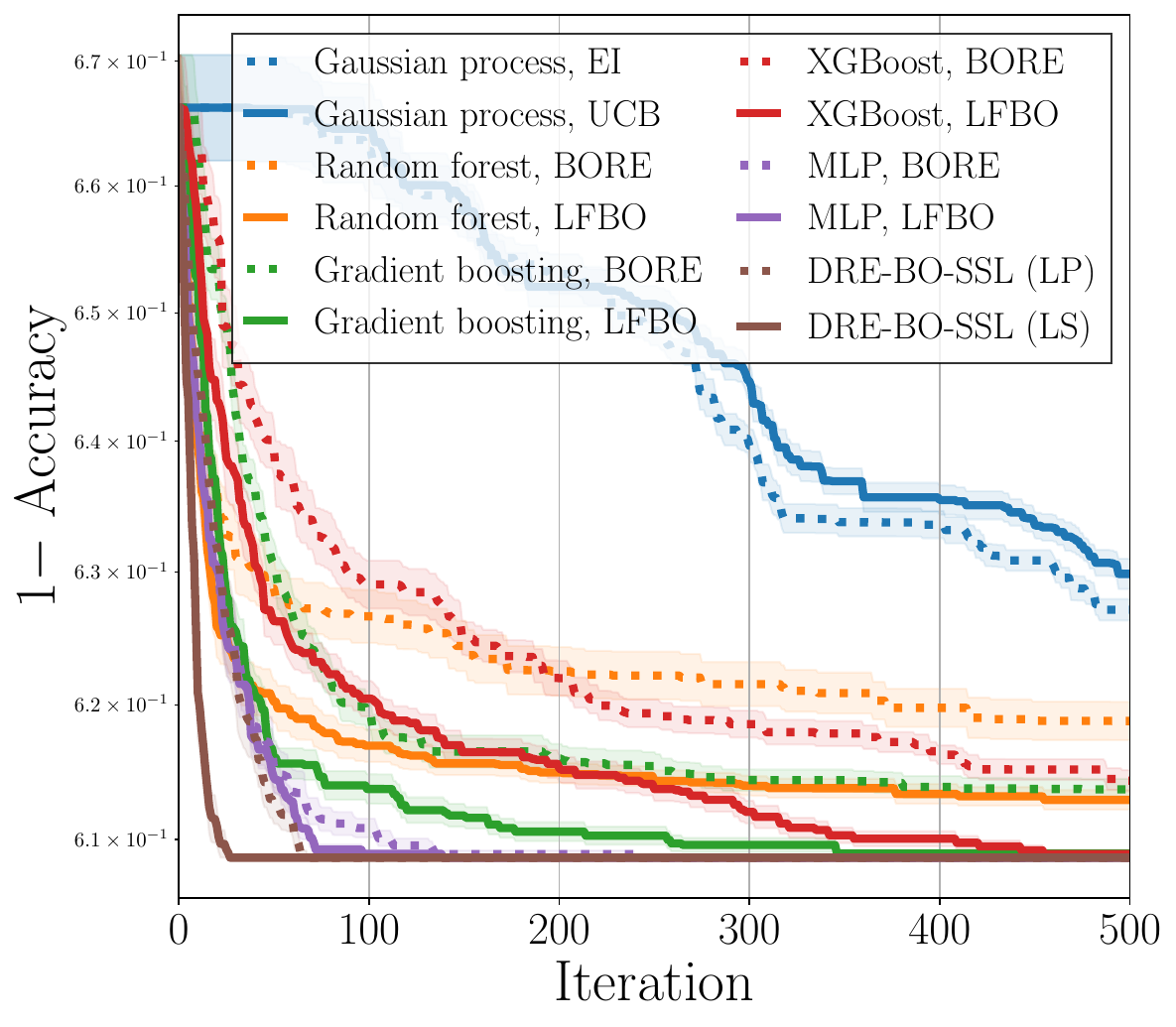}
    }
	\caption{Results with 20 repeated experiments on NATS-Bench for a scenario with fixed-size pools.}
	\label{fig:natsbench}
\end{figure*}

\subsection{A Scenario with Unlabeled Point Sampling}
\label{subsec:scenario_sampling}

\paragraph{Synthetic Benchmarks.}

We run several synthetic functions
for our methods and the baseline methods.
For unlabeled point sampling,
we sample 100 points from $\calN(\bx_1, \bI_d), \calN(\bx_2, \bI_d), \ldots, \calN(\bx_{n_l}, \bI_d)$,
where $\lfloor n_u / n_l \rfloor$ or $\lfloor n_u / n_l \rfloor + 1$ points are sampled from each truncated distribution, so that $n_u$ points are sampled in total.
As shown in~\figref{fig:synthetic_sampling},
our methods outperform the baseline methods
incorporating labeled data with unlabeled points.
Interestingly,
out methods beat Gaussian process-based Bayesian optimization.
It implies that ours can fairly balance exploration and exploitation.
Furthermore,
we present the results of learning $\beta$ in \figref{fig:gammas},
where $\beta$ is adaptively selected by~\eqref{eqn:entropy} every iteration.

\subsection{Scenarios with Fixed-Size Pools}
\label{subsec:scenarios_pools}

\paragraph{Synthetic Benchmarks.}

Several synthetic benchmark functions are tested
for our methods and the baseline methods.
To generate a fixed-size pool for each benchmark, we uniformly sample 1000 points from a bounded search space before an optimization round is started.
As presented in~\figref{fig:synthetic_pools},
our methods perform better than the baseline methods.
It implies that the use of unlabeled data helps improve optimization performance.
Also, the results of learning $\beta$ are demonstrated in~\figref{fig:gammas}.
Learned $\beta$ is likely to converge to some value
as iterations proceed according to the results.

\paragraph{Tabular Benchmarks.}

Comparisons of our methods and the existing methods are carried out in these hyperparameter optimization benchmarks~\citep{KleinA2019arxiv},
as in~\figref{fig:tabular}.
We can benchmark a variety of machine learning models, which are defined with specific hyperparameters and trained on one of four datasets: naval propulsion,
protein structure,
Parkinson's telemonitoring,
and slice localization.
There exist 62,208 models, which are used as a pool in this paper, for each dataset.
Our algorithms show superior performance compared to other approaches.
Similar to the synthetic functions,
we argue that the adoption of a predefined pool leverages its performance.
In some cases, the Gaussian process-based strategy is better than our methods.

\paragraph{NATS-Bench.}

NATS-Bench~\citep{DongX2021ieeetpami},
which is the up-to-date version of NAS-Bench-201~\citep{DongX2019iclr},
is used to test our methods and the baseline methods.
NATS-Bench is a neural architecture search benchmark
with three popular datasets: CIFAR-10, CIFAR-100, and ImageNet-16-120,
and it has 32,768 architectures,
i.e., a fixed-size pool in this paper,
for each dataset.
Similar to the experiments mentioned above,
our methods work well in three datasets,
compared to the existing methods;
see \figref{fig:natsbench} for the results.

\begin{figure}[t]
    \centering
    \includegraphics[width=0.315\textwidth]{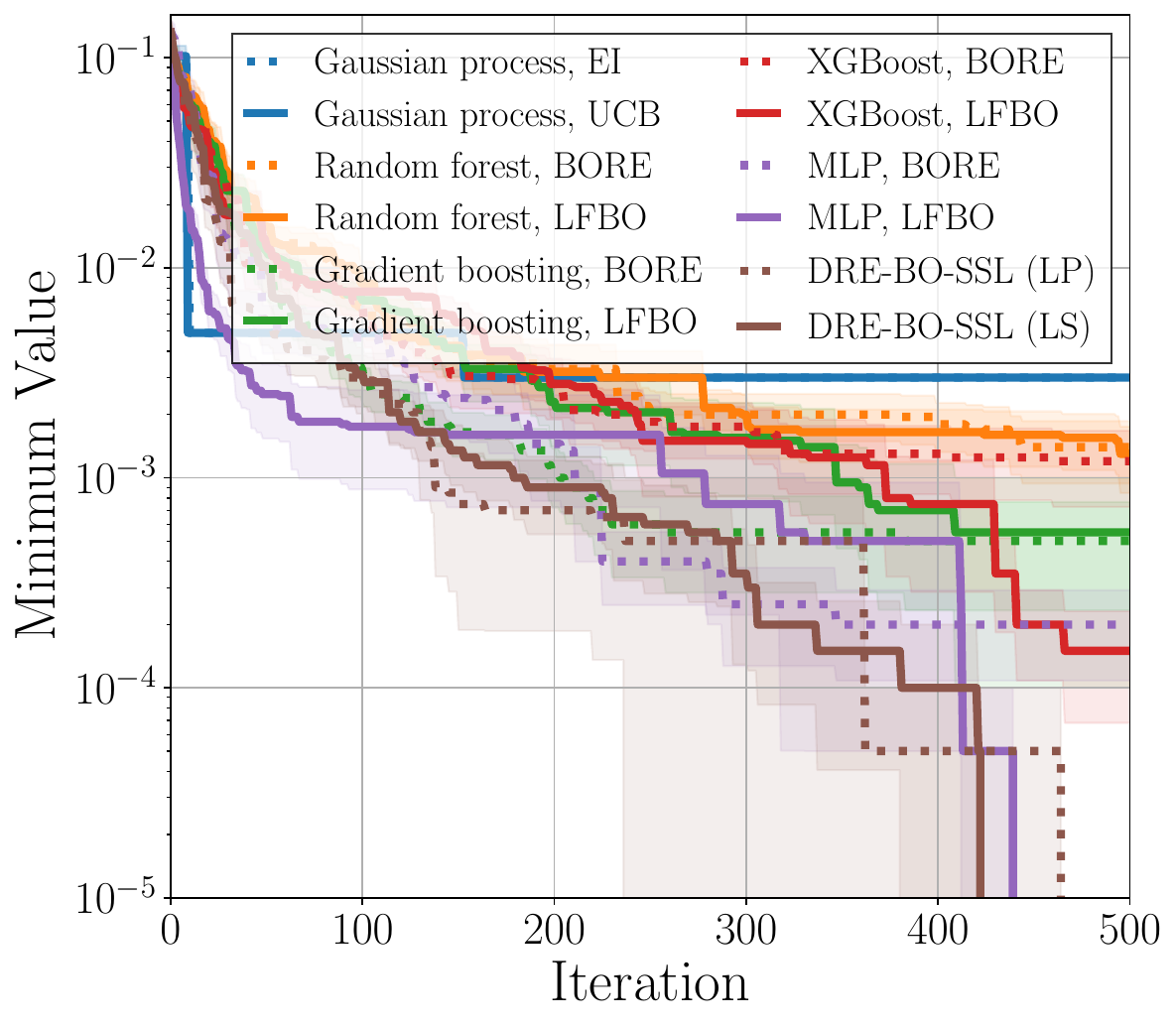}
    \caption{Results with 20 repeated experiments on 64D minimum multi-digit MNIST search.}
    \label{fig:multi_digit_mnist_results}
\end{figure}

\paragraph{64D Minimum Multi-Digit MNIST Search.}

This task, which is proposed in this work, is to find a minimum multi-digit number,
where a fixed number of multi-digit MNIST images are given.
As visualized in \figref{fig:multi_digit_mnist},
three random images in the MNIST dataset~\citep{LeCunY1998mnist}
are concatenated.
Eventually, ``000'' and ``999'' are global minimum and global maximum, respectively.
Since inputs are images and each concatenated image consists of three different digit images,
this high-dimensional optimization problem is challenging.
The size of a fixed-size pool is 80,000.
As shown in~\figref{fig:multi_digit_mnist_results},
our methods show satisfactory performance compared to other baseline methods.

Missing details of these experiments are shown in~\secref{sec:exp_details}.

\section{Discussion}
\label{sec:discussion}

We discuss interesting topics on our methods and the properties of~\ours.
Moreover,
we provide more thorough discussion and limitations in the appendices.

\paragraph{Effects of the Number of Points and Sampling Distributions for Unlabeled Points.}

We choose a distribution for unlabeled point sampling as the truncated multivariate normal distribution in order to satisfy the cluster assumption.
To analyze our algorithm thoroughly,
we demonstrate the effects of the number of sampled points and sampling distributions for unlabeled points in~\secref{sec:discussion_sampling},
varying the number of unlabeled points
and utilizing diverse sampling distributions,
i.e., uniform distributions, Halton sequences~\citep{HaltonJH1960nm}, and Sobol' sequences~\citep{SobolIM1967russian}.

\paragraph{Effects of Pool Sampling.}

Because the computational complexity of label propagation and label spreading depends on a pool size,
we need to reduce a pool size appropriately in order to speed up the algorithms.
Therefore,
we uniformly sample a subset of the fixed-size set, which is used as unlabeled points.
\figref{fig:discussion_pool_sampling_time} reports elapsed times over subset sizes for pool sampling.
Larger subsets make the framework slower as expected.
More detailed analysis on the effects of pool sampling is demonstrated in~\secref{sec:discussion_pool_sampling}.

\paragraph{Effects of Threshold Ratios.}

We study the effects of a threshold ratio $\zeta$
in order to understand how we can choose $\zeta$.
As shown in~\figref{fig:discussion_zeta},
both small $\zeta$ and large $\zeta$ generally represent worse performance;
see the results with $\zeta = 0.01$ and $\zeta = 0.8$.
While it has to depend on optimization problems,
$\zeta = 0.33$ and $\zeta = 0.5$ are generally appropriate choices according to this analysis.
The details of this study can be found in~\secref{sec:discussion_threshold_ratios} and \figref{fig:discussion_zeta}.

\paragraph{Flat Landscape of Class Probabilities over Inputs.}

Regardless of the use of either supervised or semi-supervised classifier,
a flat landscape of class probabilities can occur in the framework of DRE-based Bayesian optimization.
To overcome the issue of optimizing a flat landscape,
we add a simple heuristic
of selecting a random query point from points with identical highest class probabilities if the landscape is flat,
as described in~\secref{sec:proposed_method}.
Since off-the-shelf local optimization methods
struggle to optimize a flat landscape,
this simple heuristic is effective.

\paragraph{General Framework of~\ours.}

As a future direction,
we expect that a general framework of~\ours~can be defined,
which is similar to a likelihood-free approach by~\citet{SongJ2022icml}.
However, it is difficult to define an utility function involved in both labeled and unlabeled data.
For example, if we assume that
an utility function is $u(y; y^\dagger) = \max(y^\dagger - y, 0)$,
$y$ for an unlabeled data point is unknown.
Although it depends on the form of utility function,
we need to define $y$ of an unlabeled data point
by utilizing the information we have
if the utility function is related to $y$.

\section{Conclusion}
\label{sec:conclusion}

In this paper we have proposed a DRE-based Bayesian optimization
framework with semi-supervised learning, named~\ours.
Unlike the existing work such as BORE and LFBO,
our methods make use of semi-supervised classifiers
where unlabeled data points are sampled or given.
The superior results by our methods and the thorough analyses on the characteristics of \ours~exhibit the validity of our proposed algorithms.

\section*{Acknowledgements}

This research was supported in part by the University of Pittsburgh Center for Research Computing through the resources provided. Specifically, this work used the H2P cluster, which is supported by National Science Foundation award number OAC-2117681.

\section*{Impact Statement}

Our work does not have any direct negative societal impacts
because it proposes a new Bayesian optimization method to optimize a black-box function.
However, in the sense of optimization,
this line of research can be used in optimizing any objectives including unethical optimization tasks.

\bibliography{kjt}

\begin{thebibliography}{73}
\providecommand{\natexlab}[1]{#1}
\providecommand{\url}[1]{\texttt{#1}}
\expandafter\ifx\csname urlstyle\endcsname\relax
  \providecommand{\doi}[1]{doi: #1}\else
  \providecommand{\doi}{doi: \begingroup \urlstyle{rm}\Url}\fi

\bibitem[Ament et~al.(2023)Ament, Daulton, Eriksson, Balandat, and
  Bakshy]{AmentS2023neurips}
Ament, S., Daulton, S., Eriksson, D., Balandat, M., and Bakshy, E.
\newblock Unexpected improvements to expected improvement for {Bayesian}
  optimization.
\newblock In \emph{Advances in Neural Information Processing Systems
  (NeurIPS)}, volume~36, pp.\  20577--20612, New Orleans, Louisiana, USA, 2023.

\bibitem[Attia et~al.(2020)Attia, Grover, Jin, Severson, Markov, Liao, Chen,
  Cheong, Perkins, Yang, Herring, Aykol, Harris, Braatz, Ermon, and
  Chueh]{AttiaPM2020nature}
Attia, P.~M., Grover, A., Jin, N., Severson, K.~A., Markov, T.~M., Liao, Y.-H.,
  Chen, M.~H., Cheong, B., Perkins, N., Yang, Z., Herring, P.~K., Aykol, M.,
  Harris, S.~J., Braatz, R.~D., Ermon, S., and Chueh, W.~C.
\newblock Closed-loop optimization of fast-charging protocols for batteries
  with machine learning.
\newblock \emph{Nature}, 578\penalty0 (7795):\penalty0 397--402, 2020.

\bibitem[Balandat et~al.(2020)Balandat, Karrer, Jiang, Daulton, Letham, Wilson,
  and Bakshy]{BalandatM2020neurips}
Balandat, M., Karrer, B., Jiang, D.~R., Daulton, S., Letham, B., Wilson, A.~G.,
  and Bakshy, E.
\newblock {BoTorch}: A framework for efficient {Monte-Carlo} {Bayesian}
  optimization.
\newblock In \emph{Advances in Neural Information Processing Systems
  (NeurIPS)}, volume~33, pp.\  21524--21538, Virtual, 2020.

\bibitem[Belkin \& Niyogi(2002)Belkin and Niyogi]{BelkinM2002neurips}
Belkin, M. and Niyogi, P.
\newblock Using manifold stucture for partially labeled classification.
\newblock In \emph{Advances in Neural Information Processing Systems
  (NeurIPS)}, volume~15, Vancouver, British Columbia, Canada, 2002.

\bibitem[Ben-David et~al.(2008)Ben-David, Lu, and P{\'a}l]{BenDavidS2008colt}
Ben-David, S., Lu, T., and P{\'a}l, D.
\newblock Does unlabeled data provably help? worst-case analysis of the sample
  complexity of semi-supervised learning.
\newblock In \emph{Proceedings of the Annual Conference on Learning Theory
  (COLT)}, pp.\  33--44, Helsinki, Finland, 2008.

\bibitem[Bengio et~al.(2006)Bengio, Delalleau, and {Le
  Roux}]{BengioY2006sslbook}
Bengio, Y., Delalleau, O., and {Le Roux}, N.
\newblock Label propagation and quadratic criterion.
\newblock In Chapelle, O., Sch{\"o}lkopf, B., and Zien, A. (eds.),
  \emph{Semi-Supervised Learning}, pp.\  193--216. MIT Press, 2006.

\bibitem[Bergstra et~al.(2011)Bergstra, Bardenet, Bengio, and
  {K\'egl}]{BergstraJ2011neurips}
Bergstra, J., Bardenet, R., Bengio, Y., and {K\'egl}, B.
\newblock Algorithms for hyper-parameter optimization.
\newblock In \emph{Advances in Neural Information Processing Systems
  (NeurIPS)}, volume~24, pp.\  2546--2554, Granada, Spain, 2011.

\bibitem[Bishop(2006)]{BishopCM2006book}
Bishop, C.~M.
\newblock \emph{Pattern Recognition and Machine Learning}.
\newblock Springer, 2006.

\bibitem[Botev(2017)]{BotevZI2017jrssb}
Botev, Z.~I.
\newblock The normal law under linear restrictions: simulation and estimation
  via minimax tilting.
\newblock \emph{Journal of the Royal Statistical Society B}, 79\penalty0
  (1):\penalty0 125--148, 2017.

\bibitem[Breiman(2001)]{BreimanL2001ml}
Breiman, L.
\newblock Random forests.
\newblock \emph{Machine Learning}, 45\penalty0 (1):\penalty0 5--32, 2001.

\bibitem[Brochu et~al.(2010)Brochu, Cora, and {de Freitas}]{BrochuE2010arxiv}
Brochu, E., Cora, V.~M., and {de Freitas}, N.
\newblock A tutorial on {Bayesian} optimization of expensive cost functions,
  with application to active user modeling and hierarchical reinforcement
  learning.
\newblock \emph{{arXiv} preprint {arXiv}:1012.2599}, 2010.

\bibitem[Byrd et~al.(1995)Byrd, Lu, Nocedal, and Zhu]{ByrdRH1995siamjsc}
Byrd, R.~H., Lu, P., Nocedal, J., and Zhu, C.
\newblock A limited memory algorithm for bound constrained optimization.
\newblock \emph{SIAM Journal on Scientific Computing}, 16\penalty0
  (5):\penalty0 1190--1208, 1995.

\bibitem[Carmon et~al.(2019)Carmon, Raghunathan, Schmidt, Duchi, and
  Liang]{CarmonY2019neurips}
Carmon, Y., Raghunathan, A., Schmidt, L., Duchi, J.~C., and Liang, P.~S.
\newblock Unlabeled data improves adversarial robustness.
\newblock In \emph{Advances in Neural Information Processing Systems
  (NeurIPS)}, volume~32, pp.\  11192--11203, Vancouver, British Columbia,
  Canada, 2019.

\bibitem[Chapelle et~al.(2006)Chapelle, Sch{\"o}lkopf, and
  A.Zien]{ChapelleO2006book}
Chapelle, O., Sch{\"o}lkopf, B., and A.Zien.
\newblock \emph{Semi-Supervised Learning}.
\newblock MIT Press, 2006.

\bibitem[Chen \& Guestrin(2016)Chen and Guestrin]{ChenT2016kdd}
Chen, T. and Guestrin, C.
\newblock {XGBoost}: A scalable tree boosting system.
\newblock In \emph{Proceedings of the ACM SIGKDD Conference on Knowledge
  Discovery and Data Mining (KDD)}, pp.\  785--794, San Francisco, California,
  USA, 2016.

\bibitem[Cowen-Rivers et~al.(2022)Cowen-Rivers, Lyu, Tutunov, Wang, Grosnit,
  Griffiths, Maraval, Jianye, Wang, Peters, and
  Bou-Ammar]{CowenRiversAI2022jair}
Cowen-Rivers, A.~I., Lyu, W., Tutunov, R., Wang, Z., Grosnit, A., Griffiths,
  R.-R., Maraval, A.~M., Jianye, H., Wang, J., Peters, J., and Bou-Ammar, H.
\newblock {HEBO}: Pushing the limits of sample-efficient hyper-parameter
  optimisation.
\newblock \emph{Journal of Artificial Intelligence Research}, 74:\penalty0
  1269--1349, 2022.

\bibitem[Dong \& Yang(2019)Dong and Yang]{DongX2019iclr}
Dong, X. and Yang, Y.
\newblock {NAS-Bench-201}: Extending the scope of reproducible neural
  architecture search.
\newblock In \emph{Proceedings of the International Conference on Learning
  Representations (ICLR)}, New Orleans, Louisiana, USA, 2019.

\bibitem[Dong et~al.(2021)Dong, Liu, Musial, and Gabrys]{DongX2021ieeetpami}
Dong, X., Liu, L., Musial, K., and Gabrys, B.
\newblock {NATS-Bench}: Benchmarking {NAS} algorithms for architecture topology
  and size.
\newblock \emph{IEEE Transactions on Pattern Analysis and Machine
  Intelligence}, 44\penalty0 (7):\penalty0 3634--3646, 2021.

\bibitem[Eriksson et~al.(2019)Eriksson, Pearce, Gardner, Turner, and
  Poloczek]{ErikssonD2019neurips}
Eriksson, D., Pearce, M., Gardner, J.~R., Turner, R., and Poloczek, M.
\newblock Scalable global optimization via local {Bayesian} optimization.
\newblock In \emph{Advances in Neural Information Processing Systems
  (NeurIPS)}, pp.\  5496--5507, Vancouver, British Columbia, Canada, 2019.

\bibitem[Friedman(2001)]{FriedmanJH2001aos}
Friedman, J.~H.
\newblock Greedy function approximation: a gradient boosting machine.
\newblock \emph{The Annals of Statistics}, 29:\penalty0 1189--1232, 2001.

\bibitem[Gammerman et~al.(1998)Gammerman, Vovk, and Vapnik]{GammermanA1998uai}
Gammerman, A., Vovk, V., and Vapnik, V.
\newblock Learning by transduction.
\newblock In \emph{Proceedings of the Annual Conference on Uncertainty in
  Artificial Intelligence (UAI)}, pp.\  148--155, Madison, Wisconsin, USA,
  1998.

\bibitem[Garnett(2023)]{GarnettR2023book}
Garnett, R.
\newblock \emph{{Bayesian Optimization}}.
\newblock Cambridge University Press, 2023.

\bibitem[Garrido-Merch{\'a}n \& Hern{\'a}ndez-Lobato(2020)Garrido-Merch{\'a}n
  and Hern{\'a}ndez-Lobato]{GarridoEC2020neucom}
Garrido-Merch{\'a}n, E.~C. and Hern{\'a}ndez-Lobato, D.
\newblock Dealing with categorical and integer-valued variables in {Bayesian}
  optimization with {Gaussian} processes.
\newblock \emph{Neurocomputing}, 380:\penalty0 20--35, 2020.

\bibitem[Genz(1992)]{GenzA1992jcgs}
Genz, A.
\newblock Numerical computation of multivariate normal probabilities.
\newblock \emph{Journal of Computational and Graphical Statistics}, 1\penalty0
  (2):\penalty0 141--149, 1992.

\bibitem[Guo et~al.(2017)Guo, Pleiss, Sun, and Weinberger]{GuoC2017icml}
Guo, C., Pleiss, G., Sun, Y., and Weinberger, K.~Q.
\newblock On calibration of modern neural networks.
\newblock In \emph{Proceedings of the International Conference on Machine
  Learning (ICML)}, pp.\  1321--1330, Sydney, Australia, 2017.

\bibitem[Halton(1960)]{HaltonJH1960nm}
Halton, J.~H.
\newblock On the efficiency of certain quasi-random sequences of points in
  evaluating multi-dimensional integrals.
\newblock \emph{Numerische Mathematik}, 2:\penalty0 84--90, 1960.

\bibitem[Harris et~al.(2020)Harris, Millman, {van der Walt}, Gommers, Virtanen,
  Cournapeau, Wieser, Taylor, Berg, Smith, Kern, Picus, Hoyer, {van Kerkwijk},
  Brett, Haldane, {del R\'{i}o}, Wiebe, Peterson, G\'{e}rard-Marchant,
  Sheppard, Reddy, Weckesser, Abbasi, Gohlke, and Oliphant]{HarrisCR2020nature}
Harris, C.~R., Millman, K.~J., {van der Walt}, S.~J., Gommers, R., Virtanen,
  P., Cournapeau, D., Wieser, E., Taylor, J., Berg, S., Smith, N.~J., Kern, R.,
  Picus, M., Hoyer, S., {van Kerkwijk}, M.~H., Brett, M., Haldane, A., {del
  R\'{i}o}, J.~F., Wiebe, M., Peterson, P., G\'{e}rard-Marchant, P., Sheppard,
  K., Reddy, T., Weckesser, W., Abbasi, H., Gohlke, C., and Oliphant, T.~E.
\newblock Array programming with {NumPy}.
\newblock \emph{Nature}, 585:\penalty0 357--362, 2020.

\bibitem[Hutter et~al.(2011)Hutter, Hoos, and {Leyton-Brown}]{HutterF2011lion}
Hutter, F., Hoos, H.~H., and {Leyton-Brown}, K.
\newblock Sequential model-based optimization for general algorithm
  configuration.
\newblock In \emph{Proceedings of the International Conference on Learning and
  Intelligent Optimization (LION)}, pp.\  507--523, Rome, Italy, 2011.

\bibitem[Hvarfner et~al.(2024)Hvarfner, Hellsten, and Nardi]{HvarfnerC2024icml}
Hvarfner, C., Hellsten, E.~O., and Nardi, L.
\newblock Vanilla {Bayesian} optimization performs great in high dimensions.
\newblock In \emph{Proceedings of the International Conference on Machine
  Learning (ICML)}, pp.\  20793--20817, Vienna, Austria, 2024.

\bibitem[Jang et~al.(2024)Jang, Lee, Kim, and Lee]{JangC2024neurips}
Jang, C., Lee, H., Kim, J., and Lee, J.
\newblock Model fusion through {Bayesian} optimization in language model
  fine-tuning.
\newblock In \emph{Advances in Neural Information Processing Systems
  (NeurIPS)}, pp.\  29878--29912, Vancouver, British Columbia, Canada, 2024.

\bibitem[Joachims(2003)]{JoachimsT2003icml}
Joachims, T.
\newblock Transductive learning via spectral graph partitioning.
\newblock In \emph{Proceedings of the International Conference on Machine
  Learning (ICML)}, pp.\  290--297, Washington, District of Columbia, USA,
  2003.

\bibitem[Jones et~al.(1998)Jones, Schonlau, and Welch]{JonesDR1998jgo}
Jones, D.~R., Schonlau, M., and Welch, W.~J.
\newblock Efficient global optimization of expensive black-box functions.
\newblock \emph{Journal of Global Optimization}, 13:\penalty0 455--492, 1998.

\bibitem[Kim \& Choi(2022)Kim and Choi]{KimJ2022aistats}
Kim, J. and Choi, S.
\newblock On uncertainty estimation by tree-based surrogate models in
  sequential model-based optimization.
\newblock In \emph{Proceedings of the International Conference on Artificial
  Intelligence and Statistics (AISTATS)}, pp.\  4359--4375, Virtual, 2022.

\bibitem[Kim \& Choi(2023)Kim and Choi]{KimJ2023joss}
Kim, J. and Choi, S.
\newblock {BayesO}: A {Bayesian} optimization framework in {Python}.
\newblock \emph{Journal of Open Source Software}, 8\penalty0 (90):\penalty0
  5320, 2023.

\bibitem[Kim et~al.(2024)Kim, Li, Li, G\'{o}mez, Hinder, and Leu]{KimJ2024dd}
Kim, J., Li, M., Li, Y., G\'{o}mez, A., Hinder, O., and Leu, P.~W.
\newblock {Multi-BOWS}: Multi-fidelity multi-objective {Bayesian} optimization
  with warm starts for nanophotonic structure design.
\newblock \emph{Digital Discovery}, 3\penalty0 (2):\penalty0 381--391, 2024.

\bibitem[Kingma \& Ba(2015)Kingma and Ba]{KingmaDP2015iclr}
Kingma, D.~P. and Ba, J.~L.
\newblock {Adam}: A method for stochastic optimization.
\newblock In \emph{Proceedings of the International Conference on Learning
  Representations (ICLR)}, San Diego, California, USA, 2015.

\bibitem[Klein \& Hutter(2019)Klein and Hutter]{KleinA2019arxiv}
Klein, A. and Hutter, F.
\newblock Tabular benchmarks for joint architecture and hyperparameter
  optimization.
\newblock \emph{arXiv preprint arXiv:1905.04970}, 2019.

\bibitem[Kloek \& {van Dijk}(1978)Kloek and {van Dijk}]{KloekT1978econm}
Kloek, T. and {van Dijk}, H.~K.
\newblock {Bayesian} estimates of equation system parameters: an application of
  integration by {Monte Carlo}.
\newblock \emph{Econometrica: Journal of the Econometric Society}, pp.\  1--19,
  1978.

\bibitem[Kushner(1964)]{KushnerHJ1964jbe}
Kushner, H.~J.
\newblock A new method of locating the maximum point of an arbitrary multipeak
  curve in the presence of noise.
\newblock \emph{Journal of Basic Engineering}, 86\penalty0 (1):\penalty0
  97--106, 1964.

\bibitem[LeCun et~al.(1998)LeCun, Cortes, and Burges]{LeCunY1998mnist}
LeCun, Y., Cortes, C., and Burges, C. J.~C.
\newblock The {MNIST} database of handwritten digits.
\newblock \url{http://yann.lecun.com/exdb/mnist/}, 1998.

\bibitem[{Martinez-Cantin} et~al.(2018){Martinez-Cantin}, Tee, and
  {McCourt}]{MartinezCantinR2018aistats}
{Martinez-Cantin}, R., Tee, K., and {McCourt}, M.
\newblock Practical {Bayesian} optimization in the presence of outliers.
\newblock In \emph{Proceedings of the International Conference on Artificial
  Intelligence and Statistics (AISTATS)}, pp.\  1722--1731, Lanzarote, Canary
  Islands, Spain, 2018.

\bibitem[M{\"u}ller et~al.(2019)M{\"u}ller, Kornblith, and
  Hinton]{MullerR2019neurips}
M{\"u}ller, R., Kornblith, S., and Hinton, G.~E.
\newblock When does label smoothing help?
\newblock In \emph{Advances in Neural Information Processing Systems
  (NeurIPS)}, volume~32, Vancouver, British Columbia, Canada, 2019.

\bibitem[Nadaraya(1964)]{NadarayaEA1964tpia}
Nadaraya, E.~A.
\newblock On estimating regression.
\newblock \emph{Theory of Probability and Its Applications}, 9\penalty0
  (1):\penalty0 141--142, 1964.

\bibitem[Oh et~al.(2018)Oh, Gavves, and Welling]{OhC2018icml}
Oh, C., Gavves, E., and Welling, M.
\newblock {BOCK}: {Bayesian} optimization with cylindrical kernels.
\newblock In \emph{Proceedings of the International Conference on Machine
  Learning (ICML)}, pp.\  3868--3877, Stockholm, Sweden, 2018.

\bibitem[Oliveira et~al.(2022)Oliveira, Tiao, and Ramos]{OliveiraR2022neurips}
Oliveira, R., Tiao, L., and Ramos, F.~T.
\newblock Batch {Bayesian} optimisation via density-ratio estimation with
  guarantees.
\newblock In \emph{Advances in Neural Information Processing Systems
  (NeurIPS)}, volume~35, pp.\  29816--29829, New Orleans, Louisiana, USA, 2022.

\bibitem[Pan et~al.(2024)Pan, Falkner, Berkenkamp, and
  Vanschoren]{PanJ2024icml}
Pan, J., Falkner, S., Berkenkamp, F., and Vanschoren, J.
\newblock {MALIBO}: Meta-learning for likelihood-free {Bayesian} optimization.
\newblock In \emph{Proceedings of the International Conference on Machine
  Learning (ICML)}, pp.\  39102--39134, Vienna, Austria, 2024.

\bibitem[Paszke et~al.(2019)Paszke, Gross, Massa, Lerer, Bradbury, Chanan,
  Killeen, Lin, Gimelshein, Antiga, Desmaison, {K\"{o}pf}, Yang, DeVito,
  Raison, Tejani, Chilamkurthy, Steiner, Fang, Bai, and
  Chintala]{PaszkeA2019neurips}
Paszke, A., Gross, S., Massa, F., Lerer, A., Bradbury, J., Chanan, G., Killeen,
  T., Lin, Z., Gimelshein, N., Antiga, L., Desmaison, A., {K\"{o}pf}, A., Yang,
  E., DeVito, Z., Raison, M., Tejani, A., Chilamkurthy, S., Steiner, B., Fang,
  L., Bai, J., and Chintala, S.
\newblock {PyTorch}: An imperative style, high-performance deep learning
  library.
\newblock In \emph{Advances in Neural Information Processing Systems
  (NeurIPS)}, volume~32, pp.\  8026--8037, Vancouver, British Columbia, Canada,
  2019.

\bibitem[Pedregosa et~al.(2011)Pedregosa, Varoquaux, Gramfort, Michel, Thirion,
  Grisel, Blondel, Prettenhofer, Weiss, Dubourg, Vanderplas, Passos,
  Cournapeau, Brucher, Perrot, and Duchesnay]{PedregosaF2011jmlr}
Pedregosa, F., Varoquaux, G., Gramfort, A., Michel, V., Thirion, B., Grisel,
  O., Blondel, M., Prettenhofer, P., Weiss, R., Dubourg, V., Vanderplas, J.,
  Passos, A., Cournapeau, D., Brucher, M., Perrot, M., and Duchesnay, {\'E}.
\newblock Scikit-learn: Machine learning in {Python}.
\newblock \emph{Journal of Machine Learning Research}, 12:\penalty0 2825--2830,
  2011.

\bibitem[Picheny et~al.(2019)Picheny, Vakili, and Artemev]{PichenyV2019arxiv}
Picheny, V., Vakili, S., and Artemev, A.
\newblock Ordinal {Bayesian} optimisation.
\newblock \emph{arXiv preprint arXiv:1912.02493}, 2019.

\bibitem[Qin(1998)]{QinJ1998biometrika}
Qin, J.
\newblock Inferences for case-control and semiparametric two-sample density
  ratio models.
\newblock \emph{Biometrika}, 85\penalty0 (3):\penalty0 619--630, 1998.

\bibitem[Rasmussen \& Williams(2006)Rasmussen and
  Williams]{RasmussenCE2006book}
Rasmussen, C.~E. and Williams, C. K.~I.
\newblock \emph{Gaussian Processes for Machine Learning}.
\newblock MIT Press, 2006.

\bibitem[Rhodes et~al.(2020)Rhodes, Xu, and Gutmann]{RhodesB2020neurips}
Rhodes, B., Xu, K., and Gutmann, M.~U.
\newblock Telescoping density-ratio estimation.
\newblock In \emph{Advances in Neural Information Processing Systems
  (NeurIPS)}, volume~33, pp.\  4905--4916, Virtual, 2020.

\bibitem[Rigollet(2007)]{RigolletP2007jmlr}
Rigollet, P.
\newblock Generalization error bounds in semi-supervised classification under
  the cluster assumption.
\newblock \emph{Journal of Machine Learning Research}, 8\penalty0 (7):\penalty0
  1369--1392, 2007.

\bibitem[Seeger(2000)]{SeegerM2000tr}
Seeger, M.
\newblock Learning with labeled and unlabeled data.
\newblock Technical Report Technical Report, University of Edinburgh, 2000.

\bibitem[Shields et~al.(2021)Shields, Stevens, Li, Parasram, Damani, Alvarado,
  Janey, Adams, and Doyle]{ShieldsBJ2021nature}
Shields, B.~J., Stevens, J., Li, J., Parasram, M., Damani, F., Alvarado, J.
  I.~M., Janey, J.~M., Adams, R.~P., and Doyle, A.~G.
\newblock {Bayesian} reaction optimization as a tool for chemical synthesis.
\newblock \emph{Nature}, 590\penalty0 (7844):\penalty0 89--96, 2021.

\bibitem[Singh et~al.(2008)Singh, Nowak, and Zhu]{SinghA2008neurips}
Singh, A., Nowak, R.~D., and Zhu, X.
\newblock Unlabeled data: now it helps, now it doesn't.
\newblock In \emph{Advances in Neural Information Processing Systems
  (NeurIPS)}, volume~21, pp.\  1513--1520, Vancouver, British Columbia, Canada,
  2008.

\bibitem[Snoek et~al.(2012)Snoek, Larochelle, and Adams]{SnoekJ2012neurips}
Snoek, J., Larochelle, H., and Adams, R.~P.
\newblock Practical {Bayesian} optimization of machine learning algorithms.
\newblock In \emph{Advances in Neural Information Processing Systems
  (NeurIPS)}, volume~25, pp.\  2951--2959, Lake Tahoe, Nevada, USA, 2012.

\bibitem[Sobol'(1967)]{SobolIM1967russian}
Sobol', I.~M.
\newblock On the distribution of points in a cube and the approximate
  evaluation of integrals.
\newblock \emph{Zhurnal Vychislitel'noi Matematiki i Matematicheskoi Fiziki},
  7\penalty0 (4):\penalty0 784--802, 1967.

\bibitem[Song et~al.(2022)Song, Yu, Neiswanger, and Ermon]{SongJ2022icml}
Song, J., Yu, L., Neiswanger, W., and Ermon, S.
\newblock A general recipe for likelihood-free {Bayesian} optimization.
\newblock In \emph{Proceedings of the International Conference on Machine
  Learning (ICML)}, pp.\  20384--20404, Baltimore, Maryland, USA, 2022.

\bibitem[Springenberg et~al.(2016)Springenberg, Klein, Falkner, and
  Hutter]{SpringenbergJT2016neurips}
Springenberg, J.~T., Klein, A., Falkner, S., and Hutter, F.
\newblock {Bayesian} optimization with robust {Bayesian} neural networks.
\newblock In \emph{Advances in Neural Information Processing Systems
  (NeurIPS)}, volume~29, pp.\  4134--4142, Barcelona, Spain, 2016.

\bibitem[Srinivas et~al.(2010)Srinivas, Krause, Kakade, and
  Seeger]{SrinivasN2010icml}
Srinivas, N., Krause, A., Kakade, S., and Seeger, M.
\newblock {Gaussian} process optimization in the bandit setting: No regret and
  experimental design.
\newblock In \emph{Proceedings of the International Conference on Machine
  Learning (ICML)}, pp.\  1015--1022, Haifa, Israel, 2010.

\bibitem[Srivastava et~al.(2023)Srivastava, Han, Xu, Rhodes, and
  Gutmann]{SrivastavaA2023tmlr}
Srivastava, A., Han, S., Xu, K., Rhodes, B., and Gutmann, M.~U.
\newblock Estimating the density ratio between distributions with high
  discrepancy using multinomial logistic regression.
\newblock \emph{Transactions on Machine Learning Research}, 2023.

\bibitem[Sugiyama et~al.(2012)Sugiyama, Suzuki, and
  Kanamori]{SugiyamaM2012book}
Sugiyama, M., Suzuki, T., and Kanamori, T.
\newblock \emph{Density ratio estimation in machine learning}.
\newblock Cambridge University Press, 2012.

\bibitem[Swersky(2017)]{SwerskyK2017thesis}
Swersky, K.
\newblock \emph{Improving {Bayesian} optimization for machine learning using
  expert priors}.
\newblock PhD thesis, University of Toronto, 2017.

\bibitem[Tiao et~al.(2021)Tiao, Klein, Seeger, Bonilla, Archambeau, and
  Ramos]{TiaoLC2021icml}
Tiao, L.~C., Klein, A., Seeger, M., Bonilla, E.~V., Archambeau, C., and Ramos,
  F.
\newblock {BORE}: {Bayesian} optimization by density-ratio estimation.
\newblock In \emph{Proceedings of the International Conference on Machine
  Learning (ICML)}, pp.\  10289--10300, Virtual, 2021.

\bibitem[Virtanen et~al.(2020)Virtanen, Gommers, Oliphant, Haberland, Reddy,
  Cournapeau, Burovski, Peterson, Weckesser, Bright, {van der Walt}, Brett,
  Wilson, Millman, Mayorov, Nelson, Jones, Kern, Larson, Carey, Polat, Feng,
  Moore, VanderPlas, Laxalde, Perktold, Cimrman, Henriksen, Quintero, Harris,
  Archibald, Ribeiro, Pedregosa, {van Mulbregt}, and {SciPy 1.0
  Contributors}]{VirtanenP2020nm}
Virtanen, P., Gommers, R., Oliphant, T.~E., Haberland, M., Reddy, T.,
  Cournapeau, D., Burovski, E., Peterson, P., Weckesser, W., Bright, J., {van
  der Walt}, S.~J., Brett, M., Wilson, J., Millman, K.~J., Mayorov, N., Nelson,
  A. R.~J., Jones, E., Kern, R., Larson, E., Carey, C.~J., Polat, I., Feng, Y.,
  Moore, E.~W., VanderPlas, J., Laxalde, D., Perktold, J., Cimrman, R.,
  Henriksen, I., Quintero, E.~A., Harris, C.~R., Archibald, A.~M., Ribeiro,
  A.~H., Pedregosa, F., {van Mulbregt}, P., and {SciPy 1.0 Contributors}.
\newblock {SciPy} 1.0: fundamental algorithms for scientific computing in
  {Python}.
\newblock \emph{Nature Methods}, 17:\penalty0 261--272, 2020.

\bibitem[Watson(1964)]{WatsonGS1964sijs}
Watson, G.~S.
\newblock Smooth regression analysis.
\newblock \emph{Sankhy{\=a}: The Indian Journal of Statistics, Series A},
  26:\penalty0 359--372, 1964.

\bibitem[Wei et~al.(2020)Wei, Shen, Chen, and Ma]{WeiC2020iclr}
Wei, C., Shen, K., Chen, Y., and Ma, T.
\newblock Theoretical analysis of self-training with deep networks on unlabeled
  data.
\newblock In \emph{Proceedings of the International Conference on Learning
  Representations (ICLR)}, Virtual, 2020.

\bibitem[Yamada et~al.(2011)Yamada, Suzuki, Kanamori, Hachiya, and
  Sugiyama]{YamadaM2011neurips}
Yamada, M., Suzuki, T., Kanamori, T., Hachiya, H., and Sugiyama, M.
\newblock Relative density-ratio estimation for robust distribution comparison.
\newblock In \emph{Advances in Neural Information Processing Systems
  (NeurIPS)}, volume~24, pp.\  594--602, Granada, Spain, 2011.

\bibitem[Zhang et~al.(2022)Zhang, Wang, Liu, Chen, and Xiong]{ZhangS2022iclr}
Zhang, S., Wang, M., Liu, S., Chen, P.-Y., and Xiong, J.
\newblock How does unlabeled data improve generalization in self-training? a
  one-hidden-layer theoretical analysis.
\newblock In \emph{Proceedings of the International Conference on Learning
  Representations (ICLR)}, Virtual, 2022.

\bibitem[Zhou et~al.(2003)Zhou, Bousquet, Lal, Weston, and
  Sch{\"o}lkopf]{ZhouD2003neurips_b}
Zhou, D., Bousquet, O., Lal, T.~N., Weston, J., and Sch{\"o}lkopf, B.
\newblock Learning with local and global consistency.
\newblock In \emph{Advances in Neural Information Processing Systems
  (NeurIPS)}, volume~16, pp.\  321--328, Vancouver, British Columbia, Canada,
  2003.

\bibitem[Zhu(2005)]{ZhuX2005tr}
Zhu, X.
\newblock Semi-supervised learning literature survey.
\newblock Technical Report Computer Sciences TR-1530, University of
  Wisconsin--Madison, 2005.

\bibitem[Zhu \& Ghahramani(2002)Zhu and Ghahramani]{ZhuX2002tr}
Zhu, X. and Ghahramani, Z.
\newblock Learning from labeled and unlabeled data with label propagation.
\newblock Technical Report CMU-CALD-02-107, Carnegie Mellon University, 2002.

\end{thebibliography}
\bibliographystyle{icml2025}

\newpage
\appendix
\onecolumn

\section{Additional Comparisons of BORE and LFBO}

\begin{figure}[ht!]
    \centering
    \subfigure[Random forests, BORE, Iterations 1 to 5]{
        \includegraphics[width=0.19\textwidth]{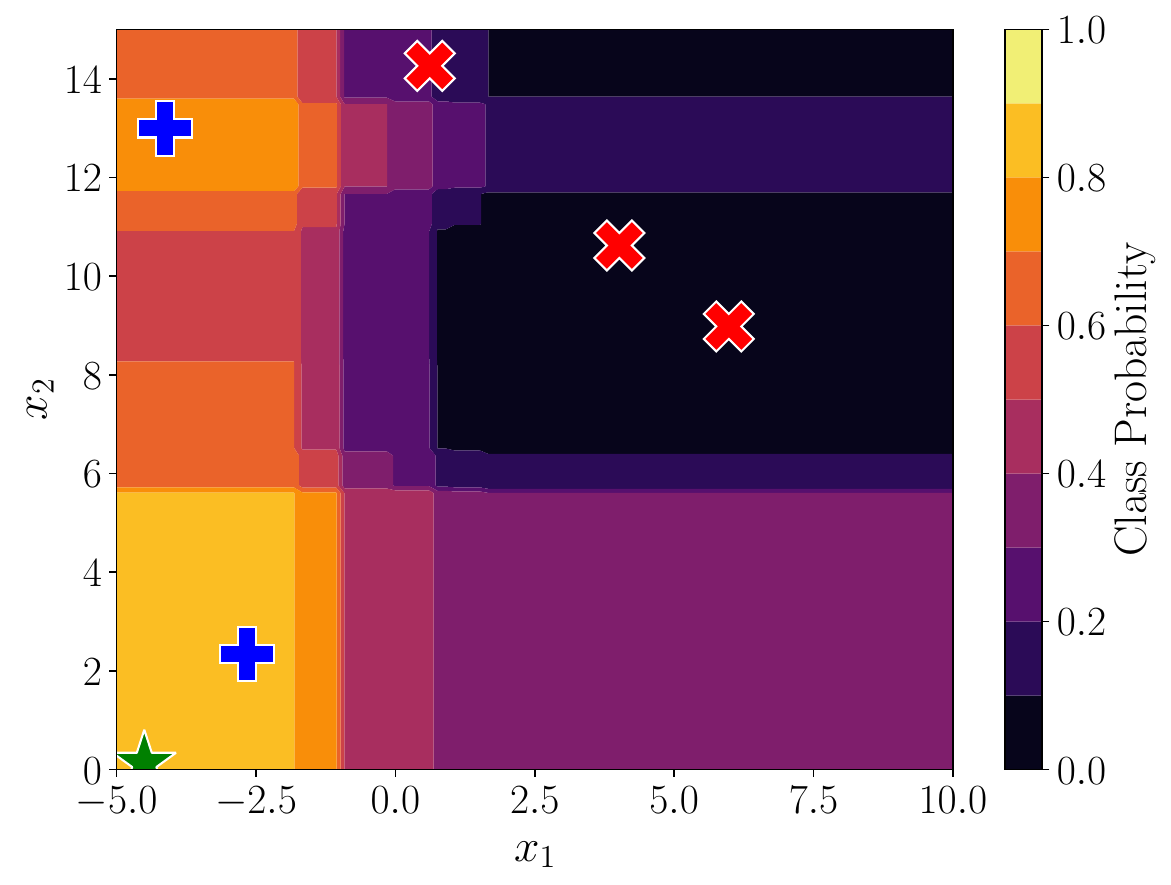}
        \includegraphics[width=0.19\textwidth]{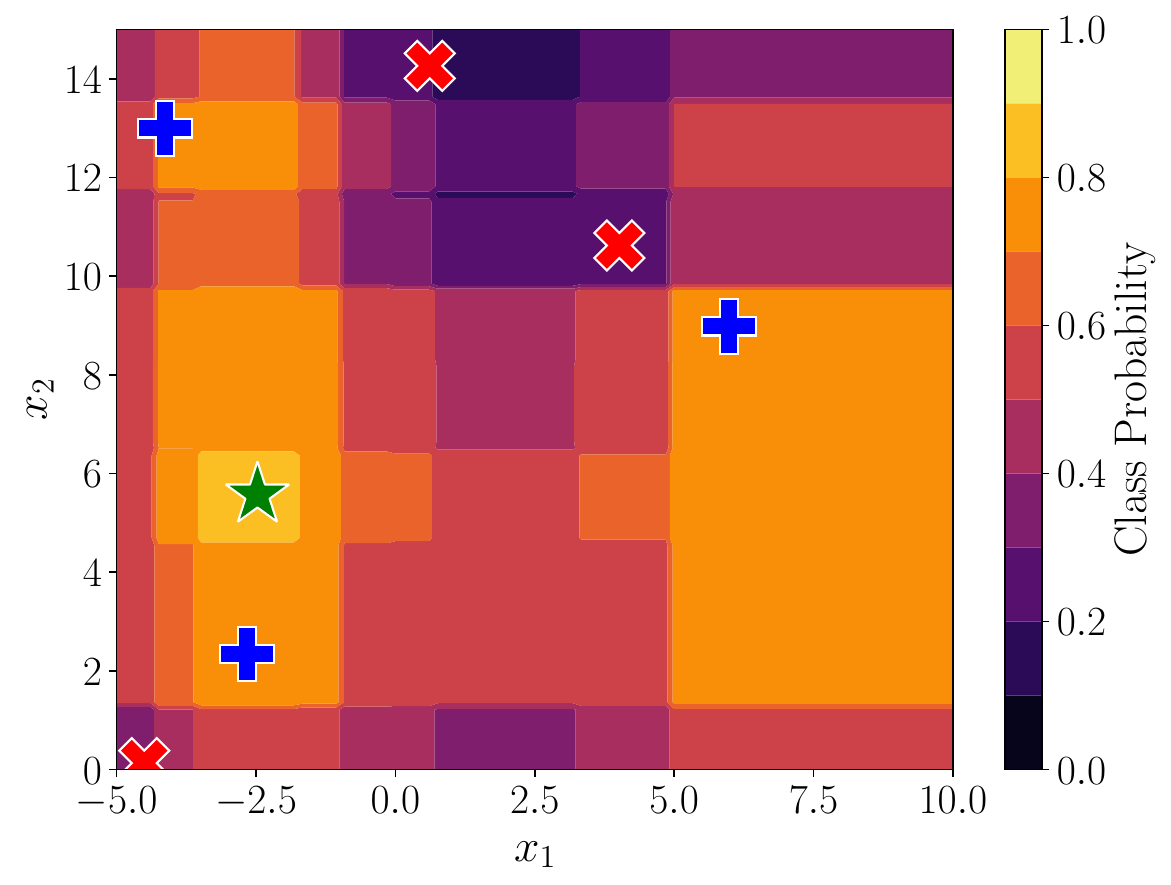}
        \includegraphics[width=0.19\textwidth]{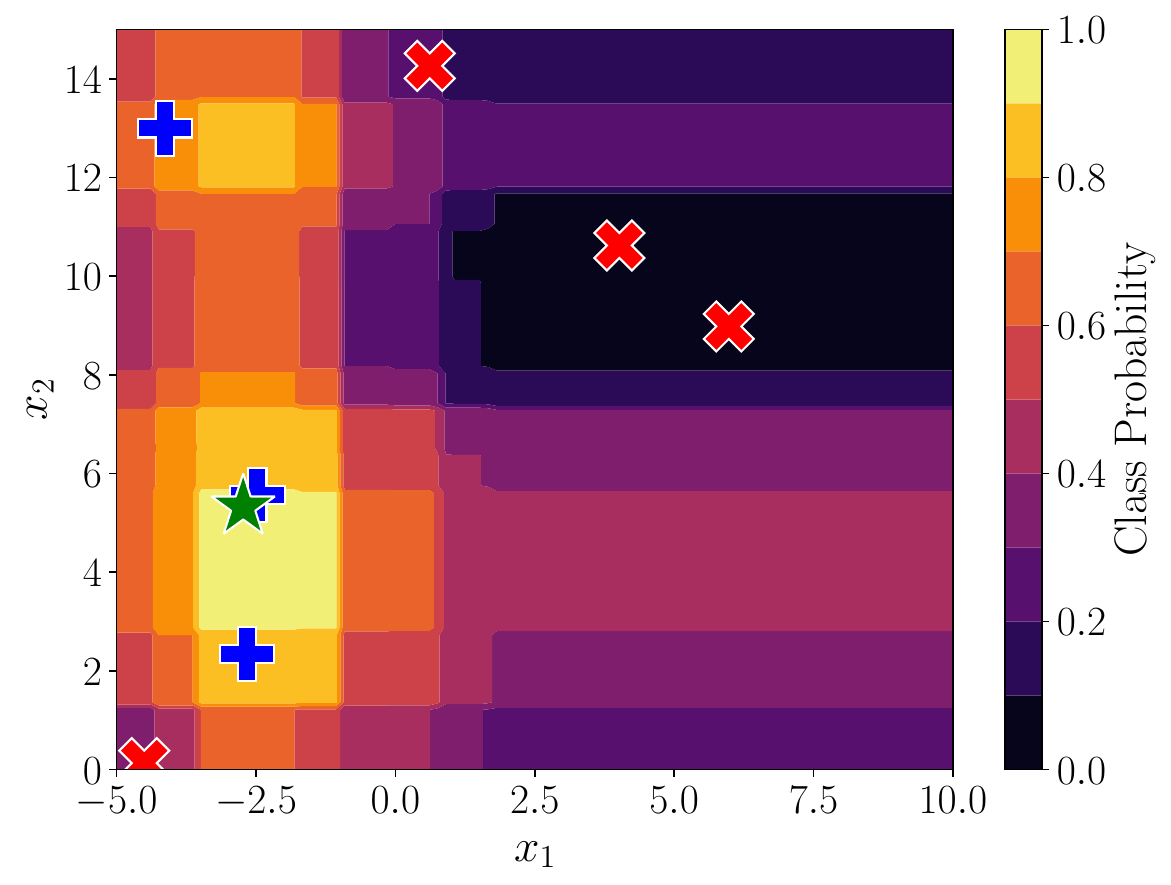}
        \includegraphics[width=0.19\textwidth]{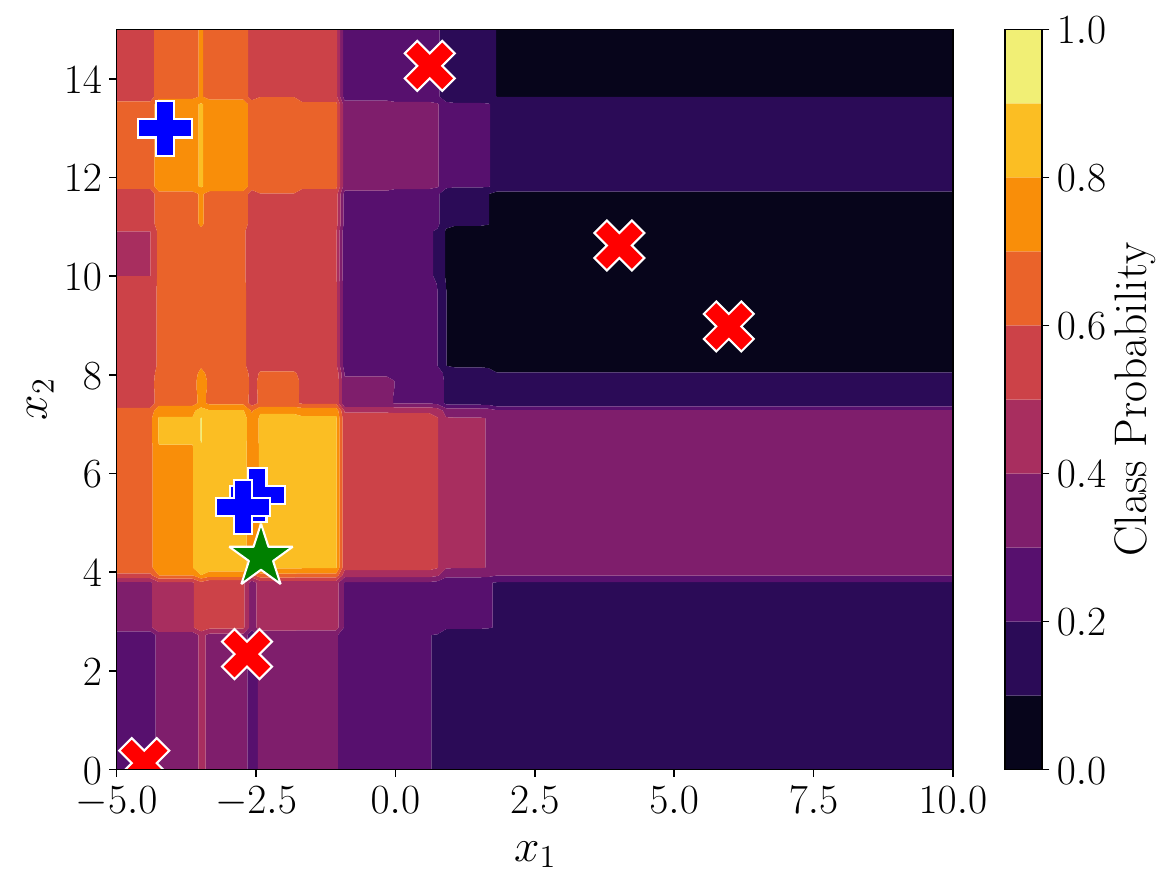}
        \includegraphics[width=0.19\textwidth]{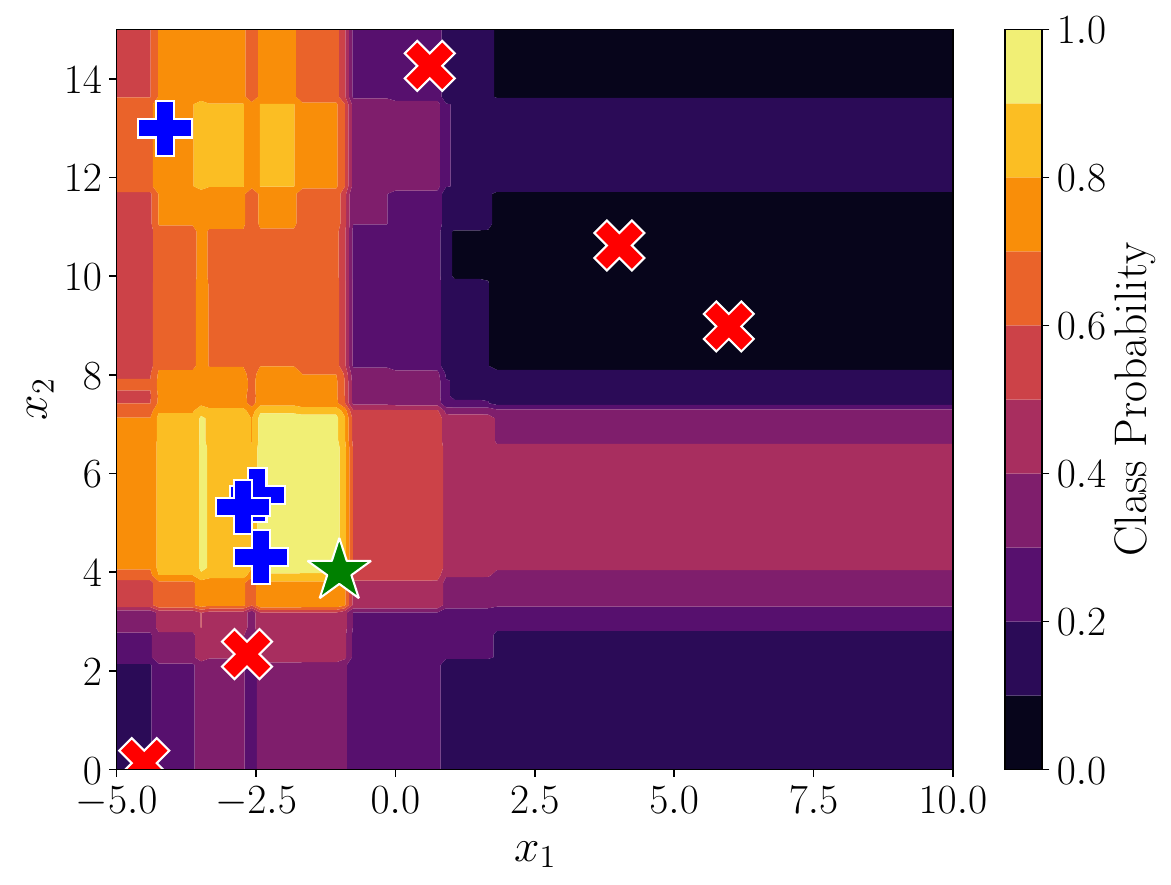}
    }
    \subfigure[Random forests, LFBO, Iterations 1 to 5]{
        \includegraphics[width=0.19\textwidth]{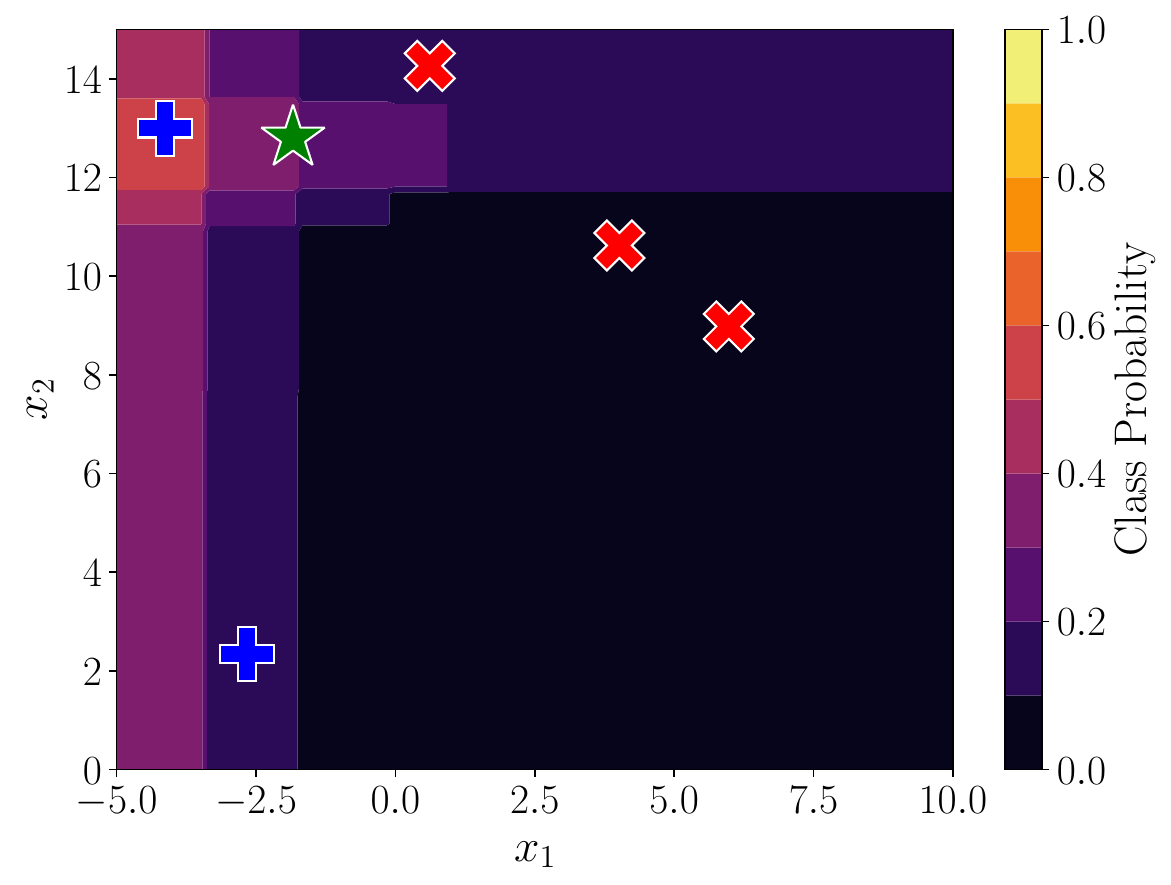}
        \includegraphics[width=0.19\textwidth]{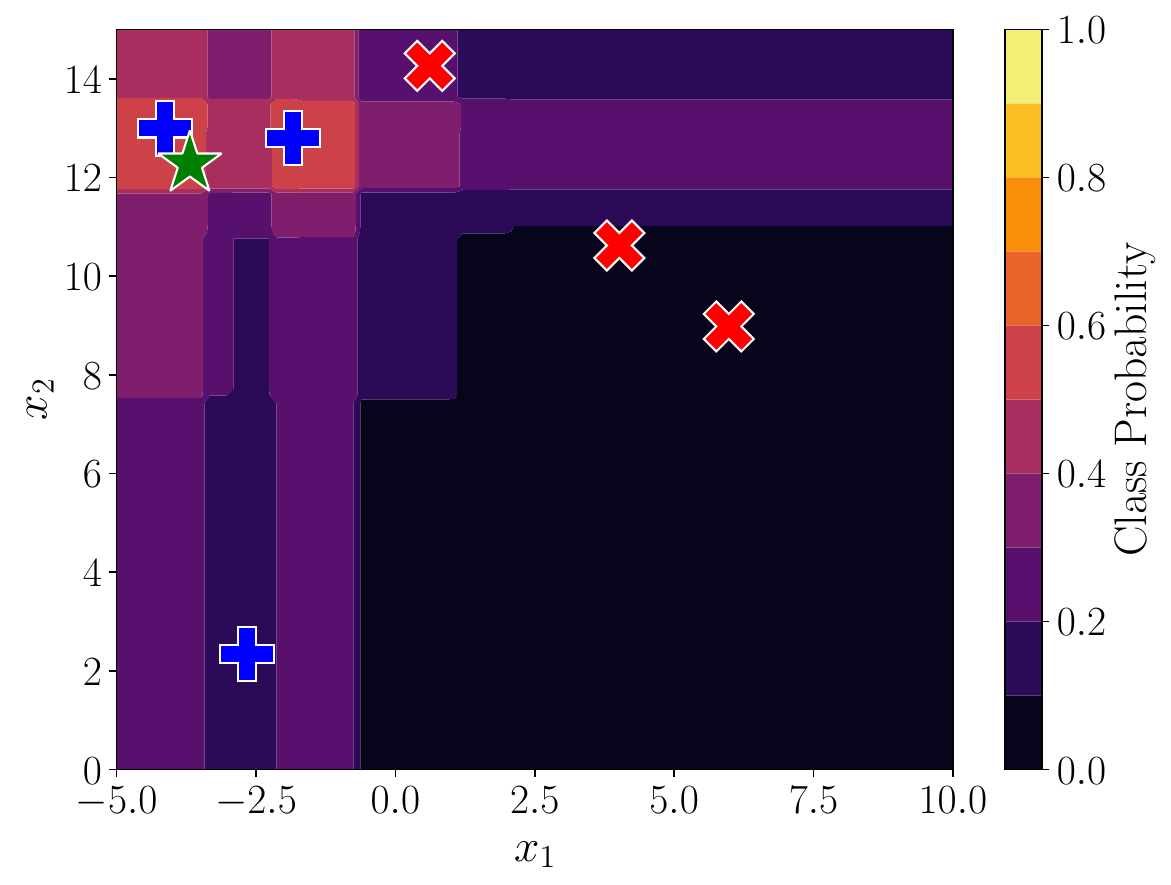}
        \includegraphics[width=0.19\textwidth]{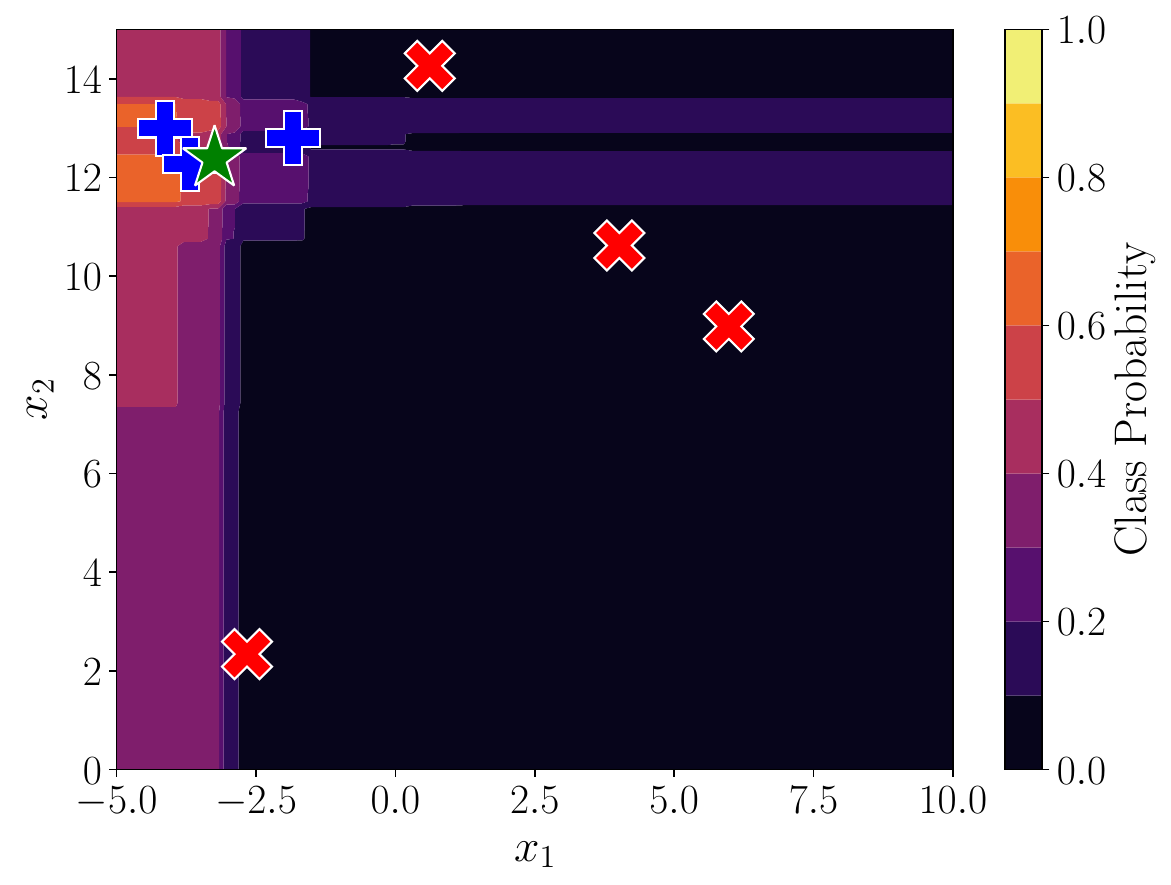}
        \includegraphics[width=0.19\textwidth]{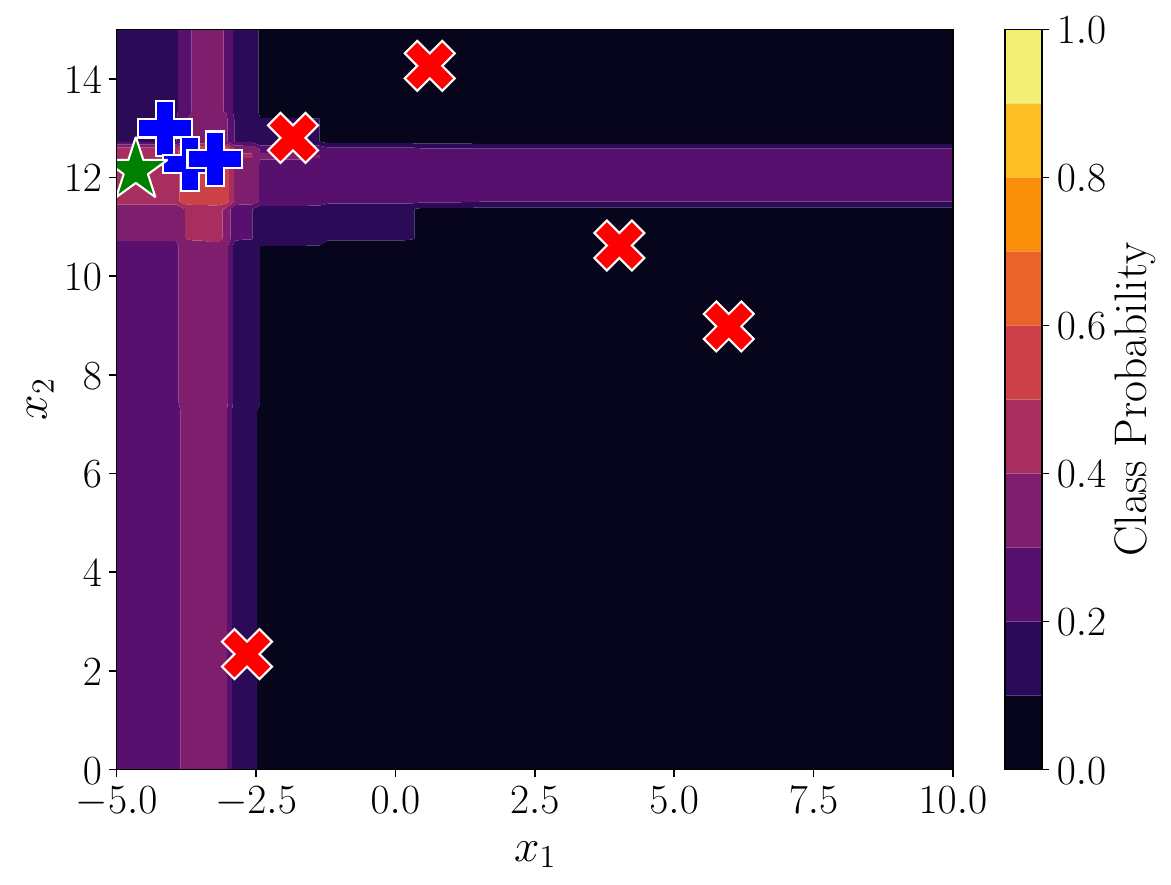}
        \includegraphics[width=0.19\textwidth]{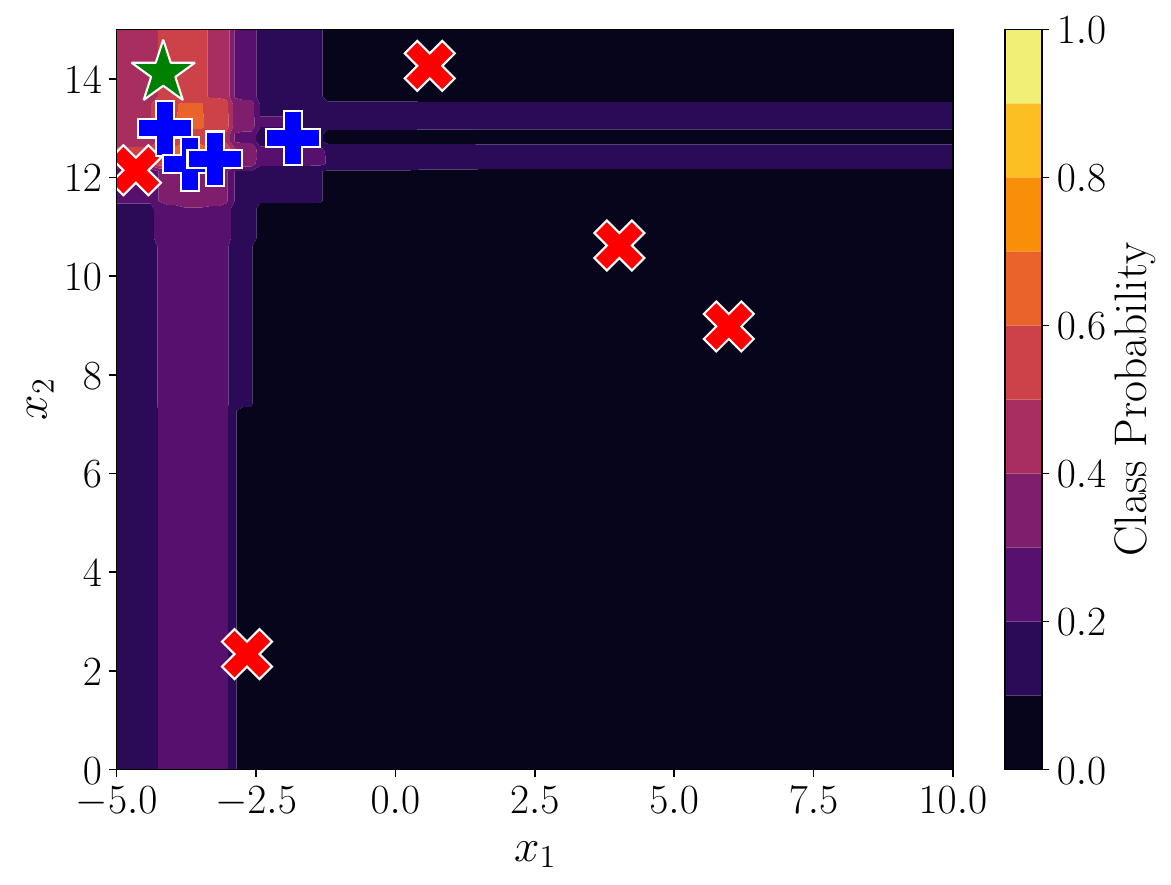}
    }
    \subfigure[Gradient boosting, BORE, Iterations 1 to 5]{
        \includegraphics[width=0.19\textwidth]{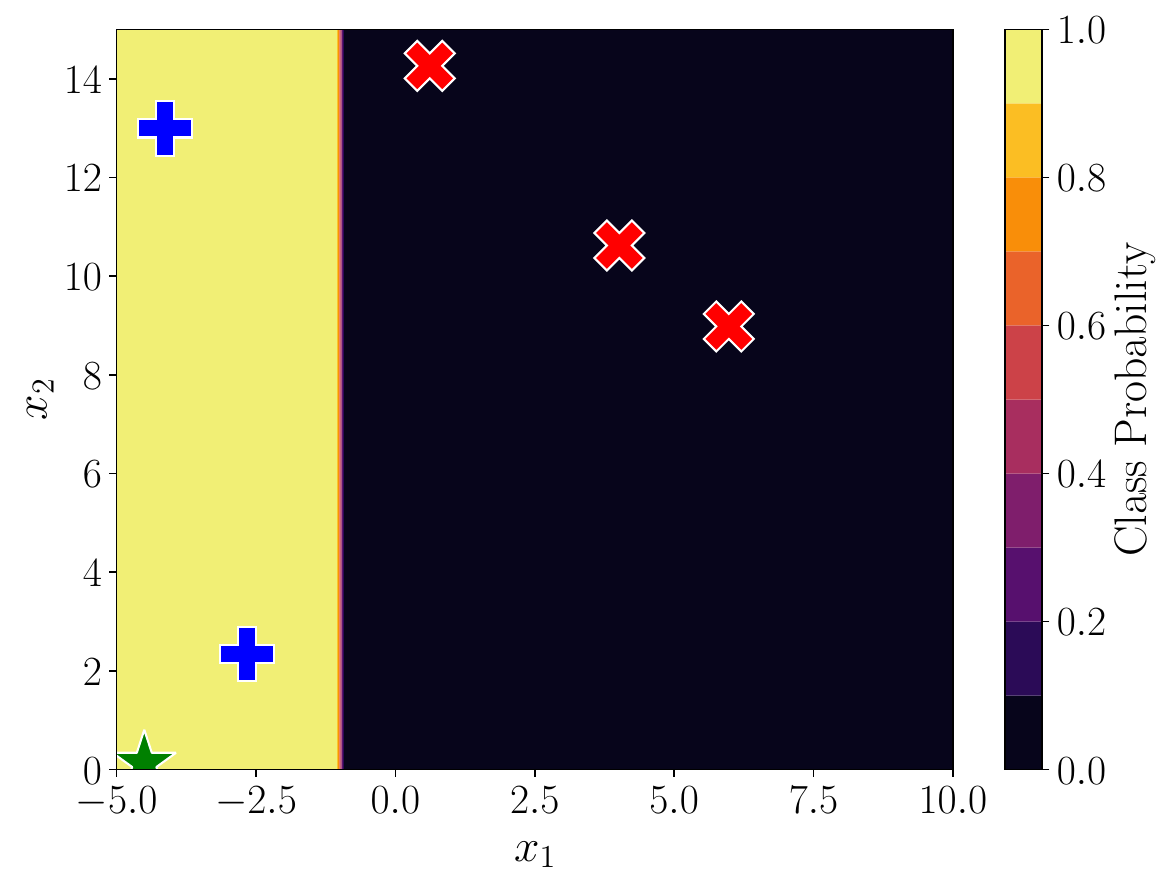}
        \includegraphics[width=0.19\textwidth]{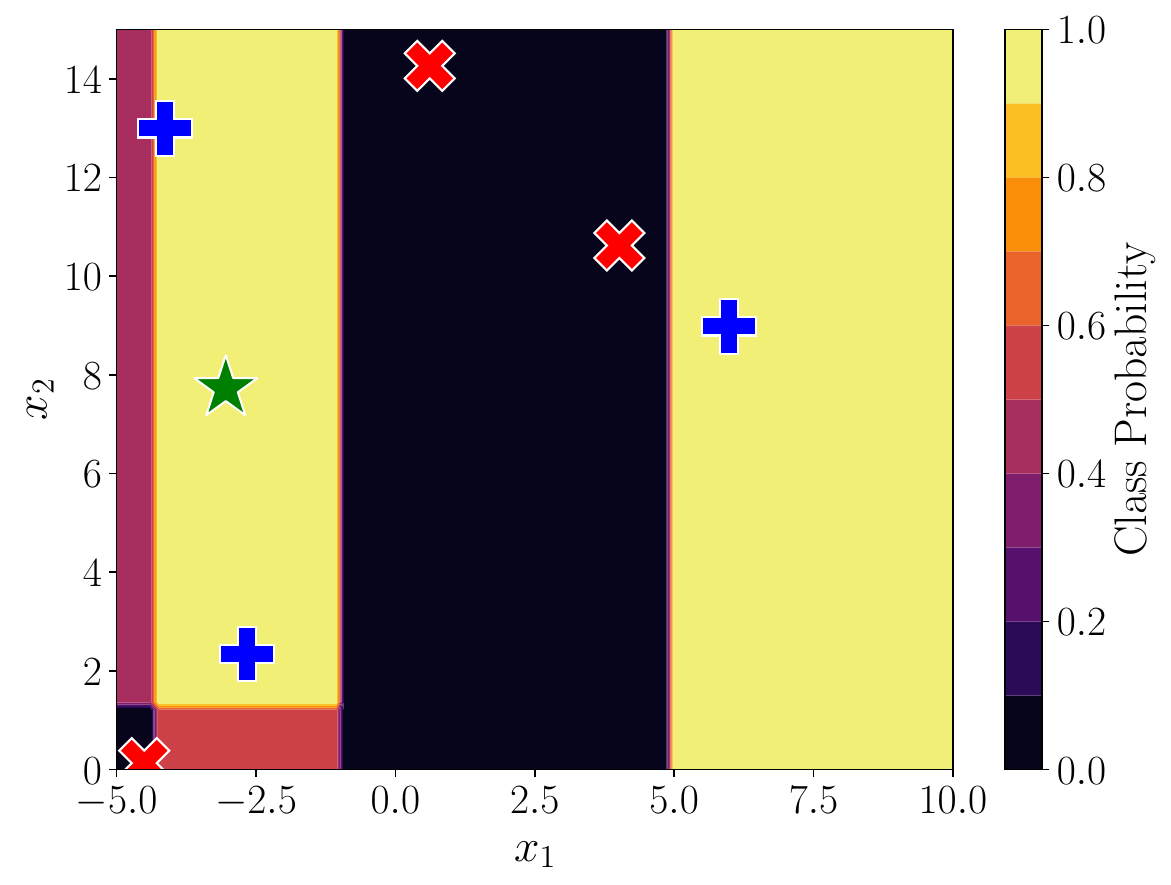}
        \includegraphics[width=0.19\textwidth]{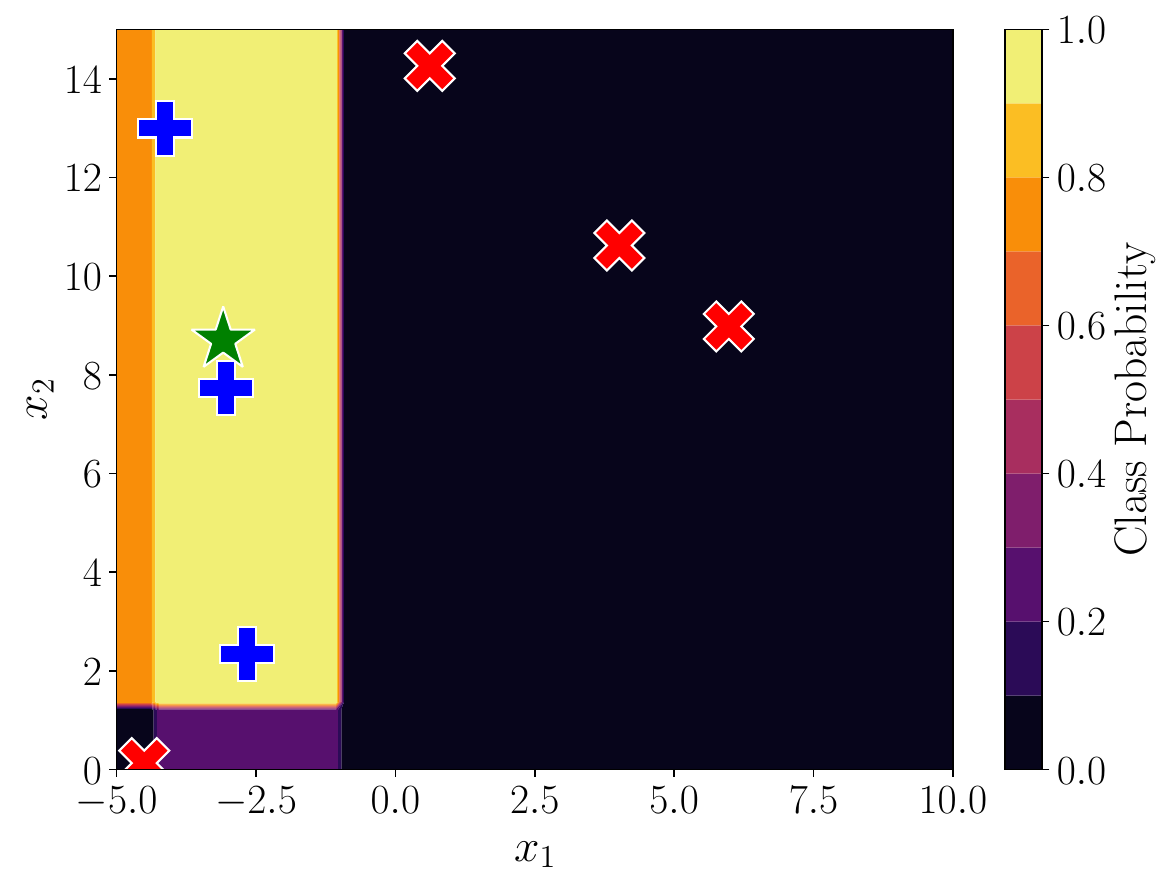}
        \includegraphics[width=0.19\textwidth]{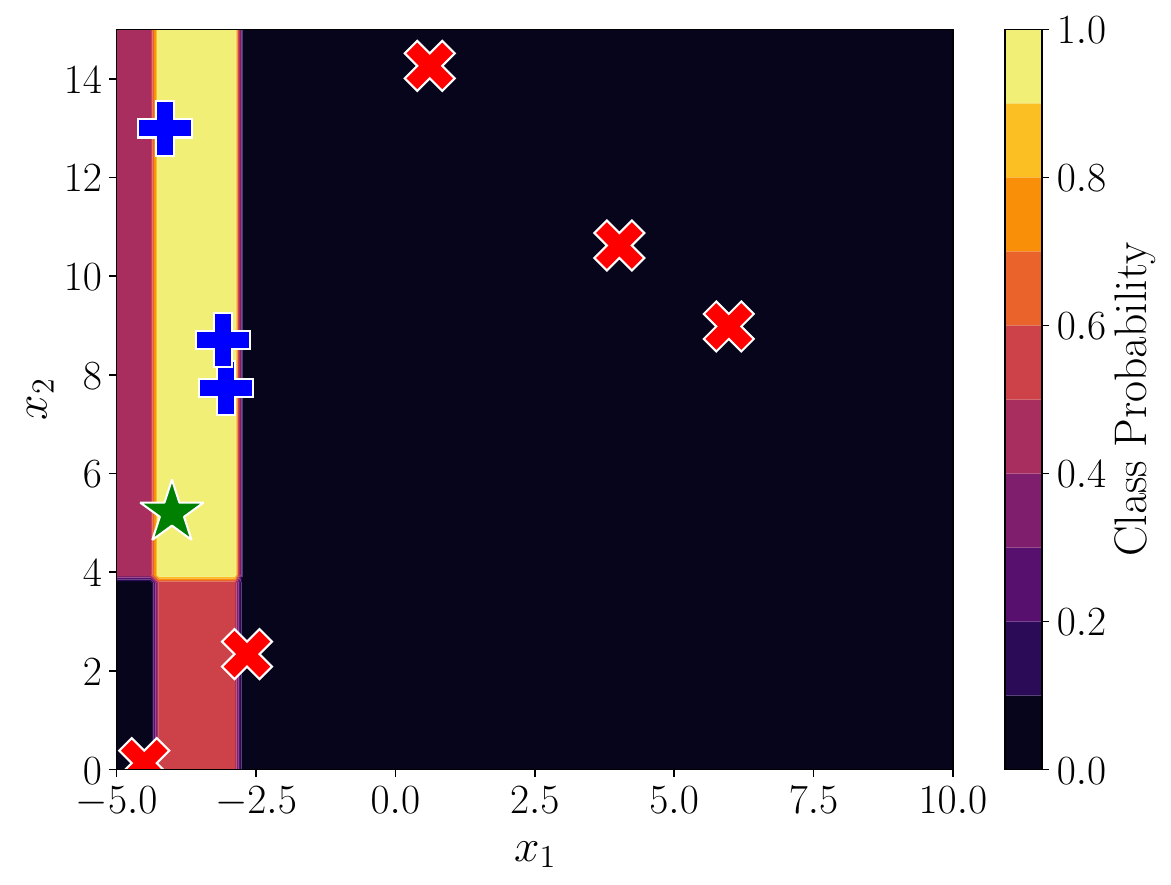}
        \includegraphics[width=0.19\textwidth]{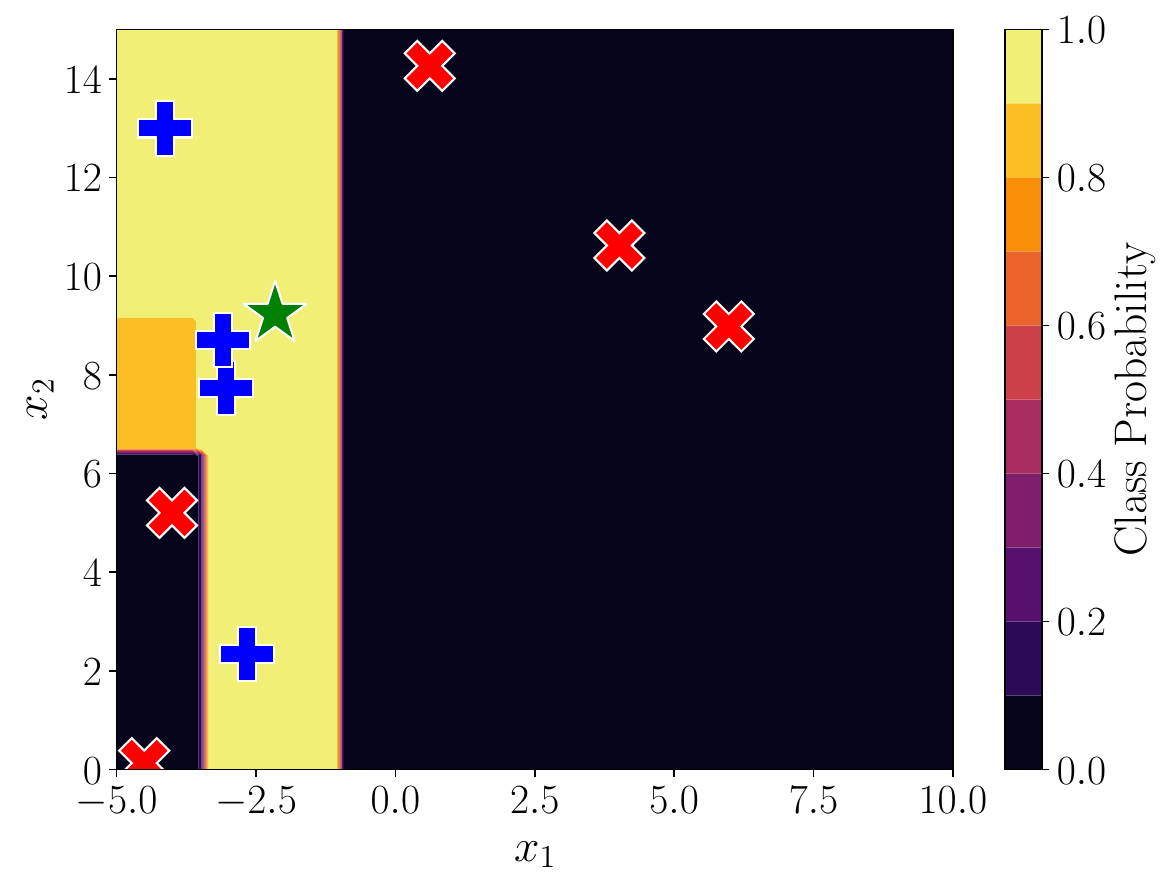}
    }
    \subfigure[Gradient boosting, LFBO, Iterations 1 to 5]{
        \includegraphics[width=0.19\textwidth]{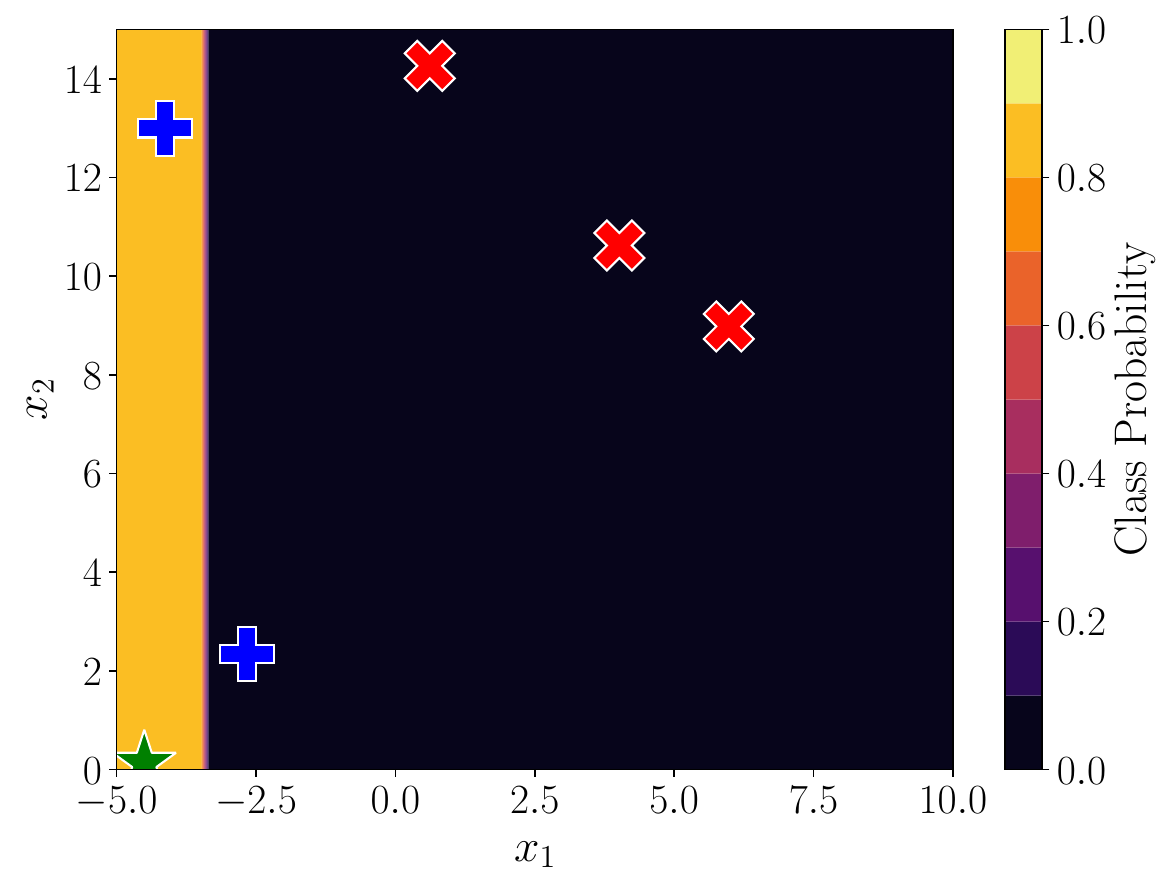}
        \includegraphics[width=0.19\textwidth]{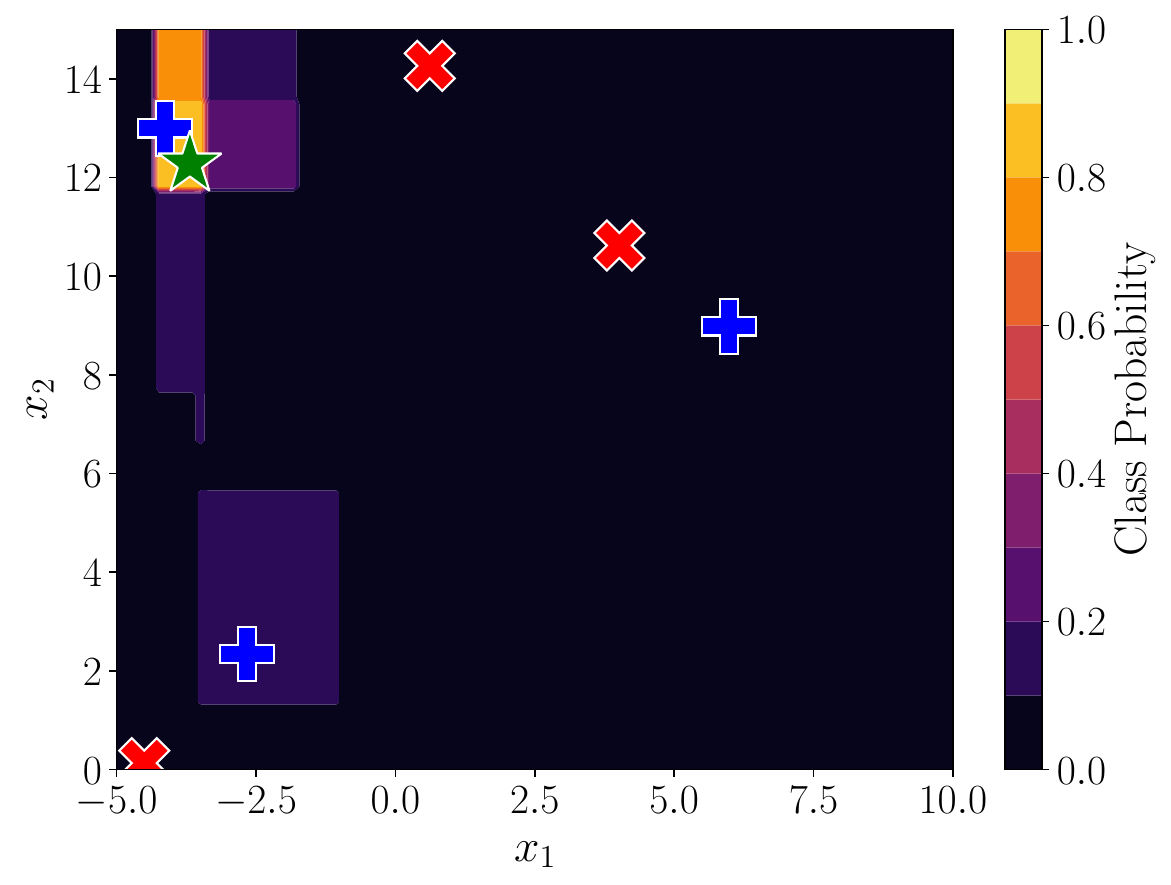}
        \includegraphics[width=0.19\textwidth]{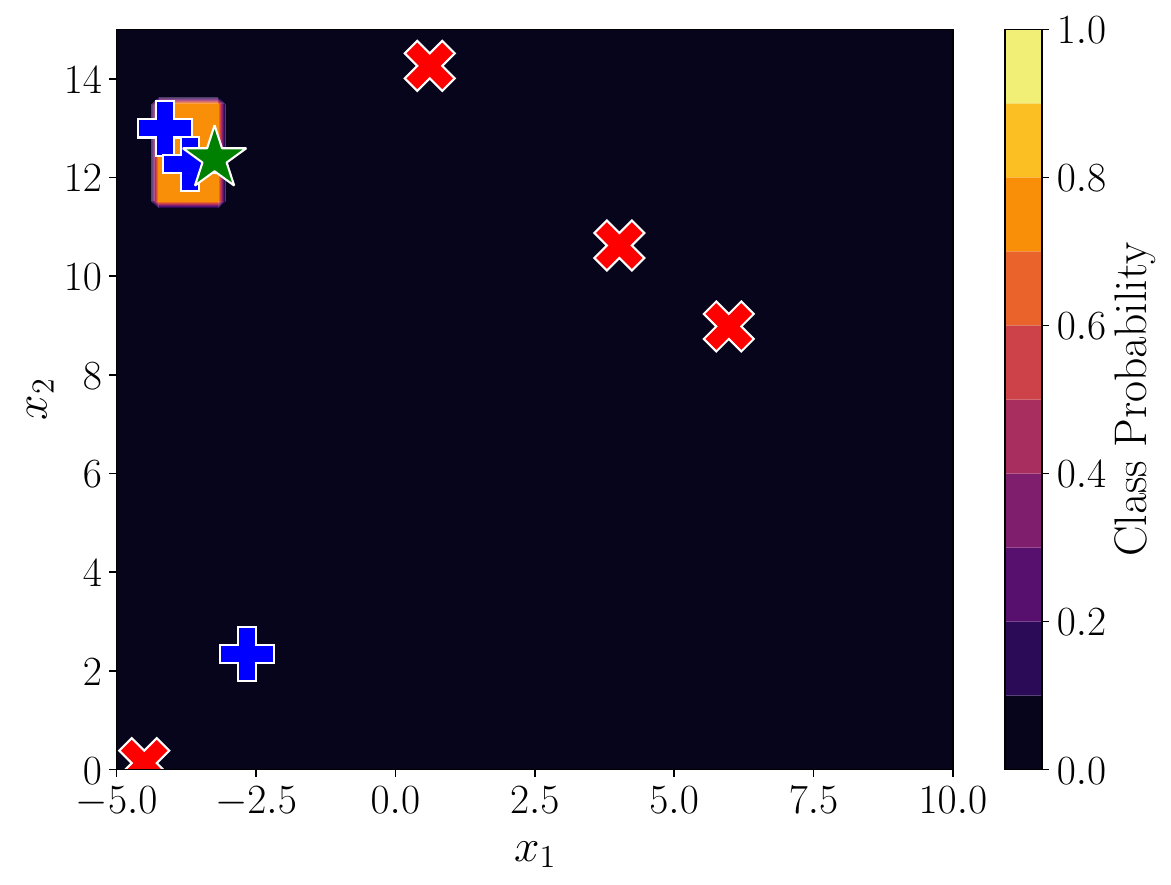}
        \includegraphics[width=0.19\textwidth]{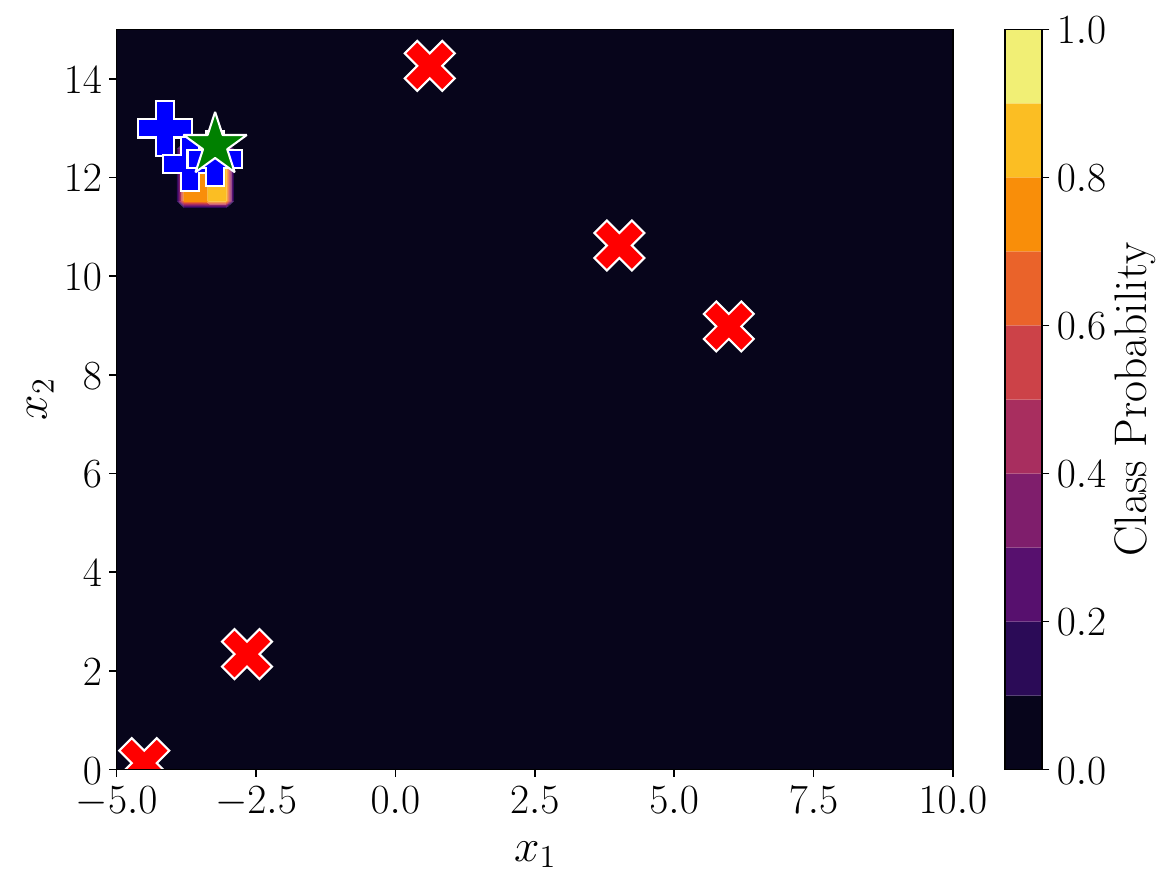}
        \includegraphics[width=0.19\textwidth]{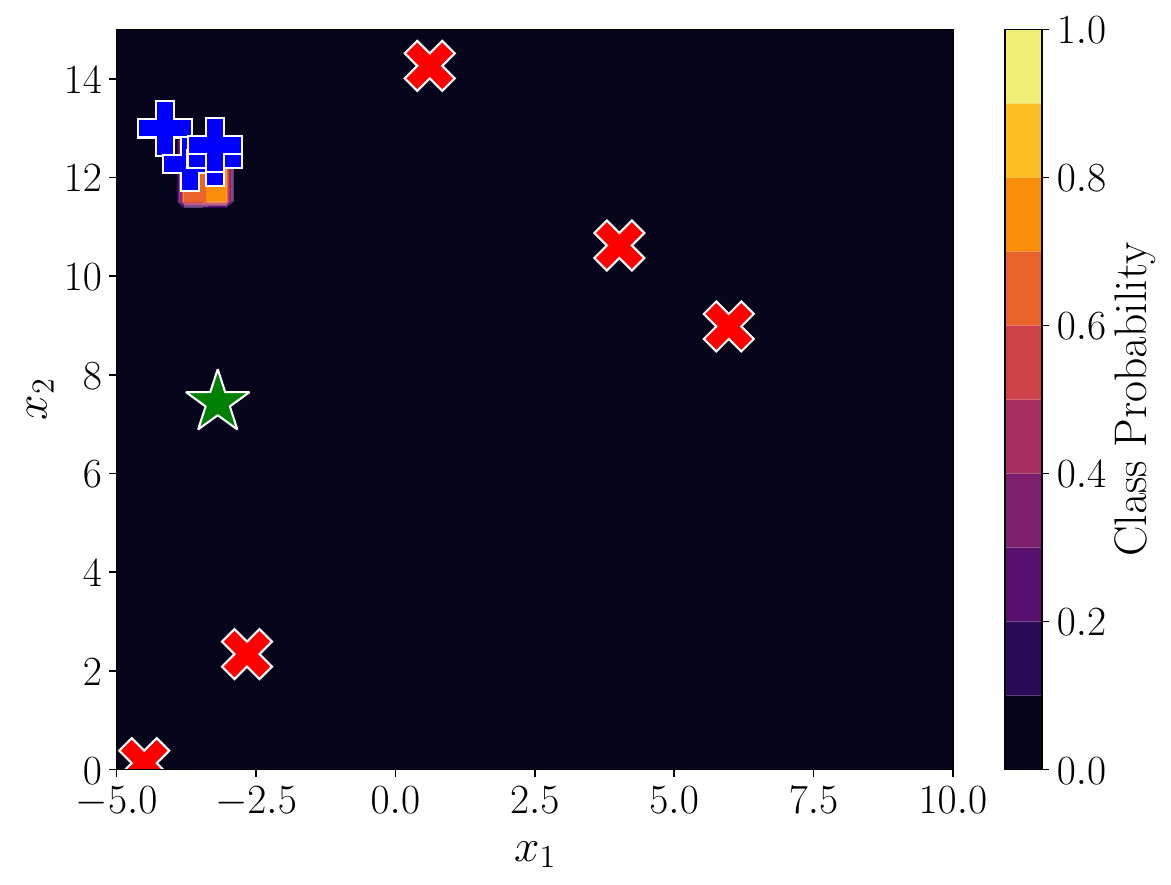}
    }
    \subfigure[XGBoost, BORE, Iterations 1 to 5]{
        \includegraphics[width=0.19\textwidth]{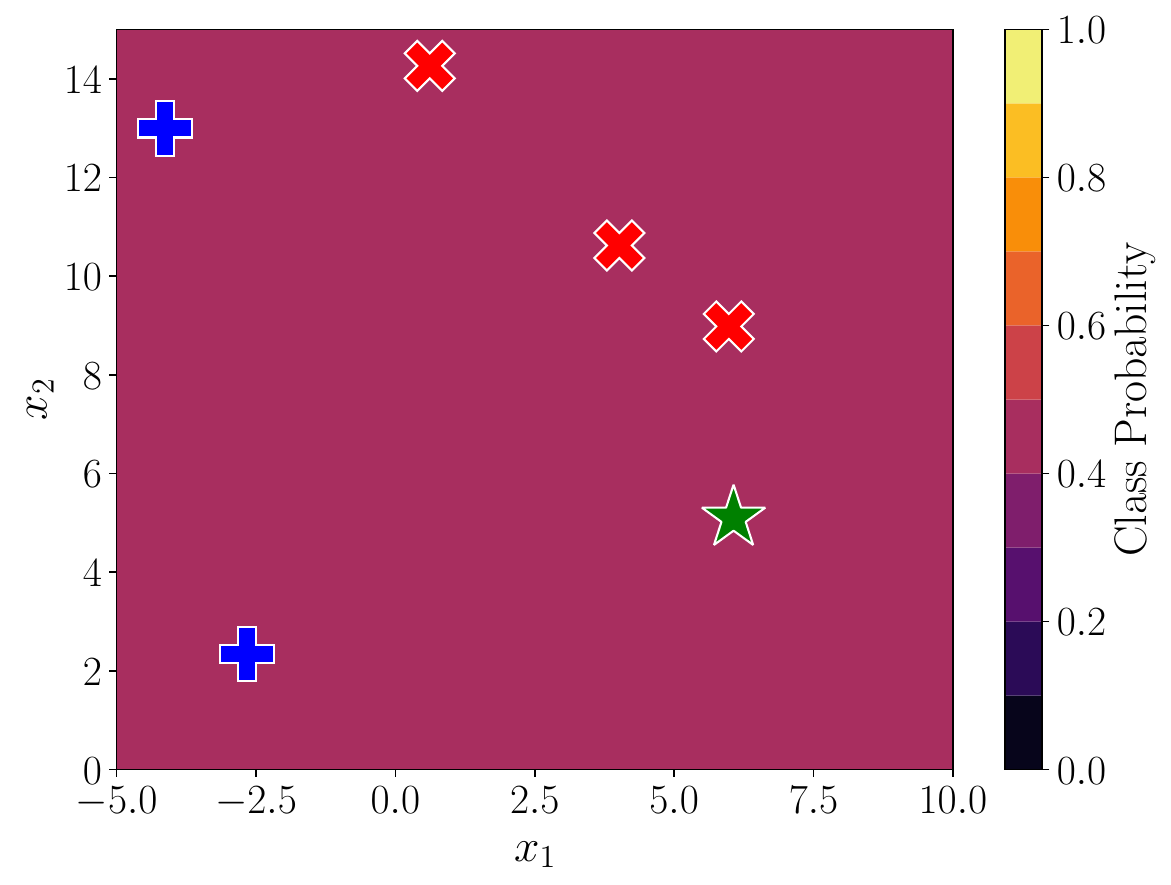}
        \includegraphics[width=0.19\textwidth]{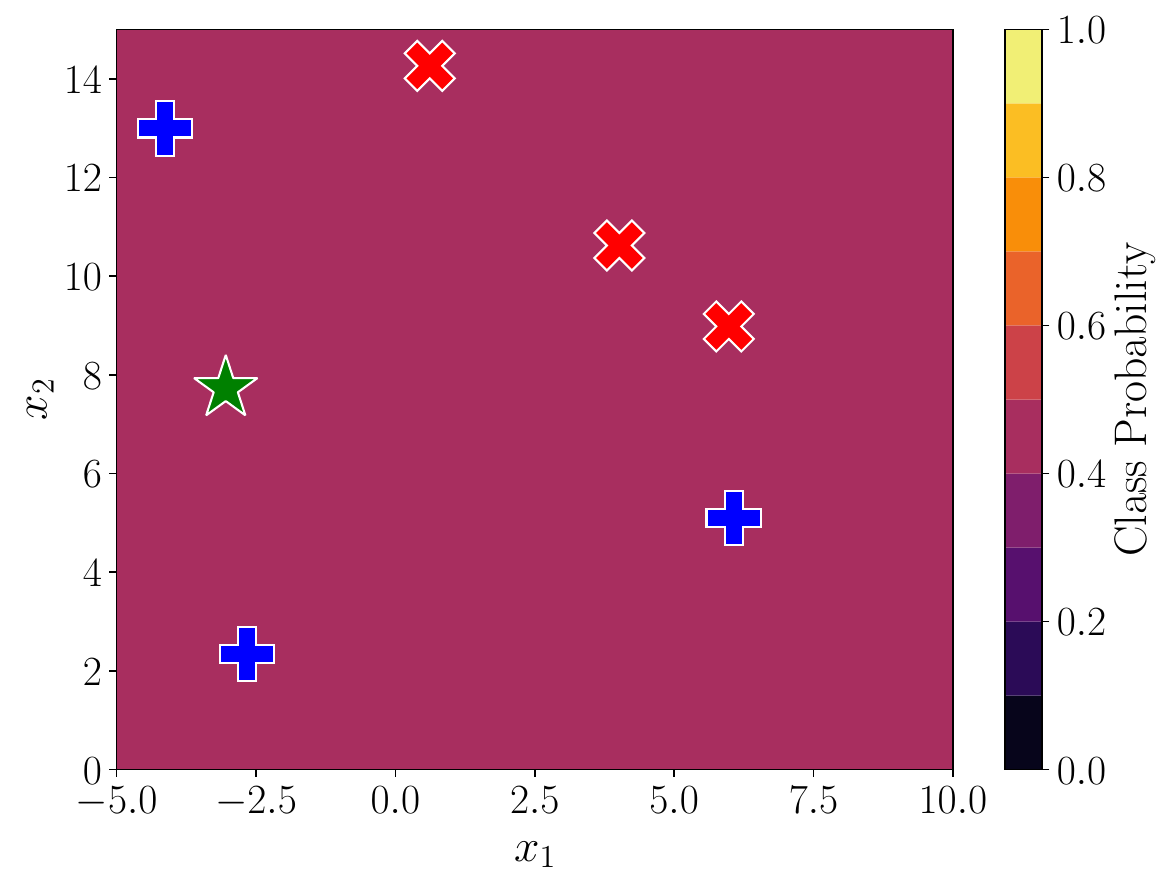}
        \includegraphics[width=0.19\textwidth]{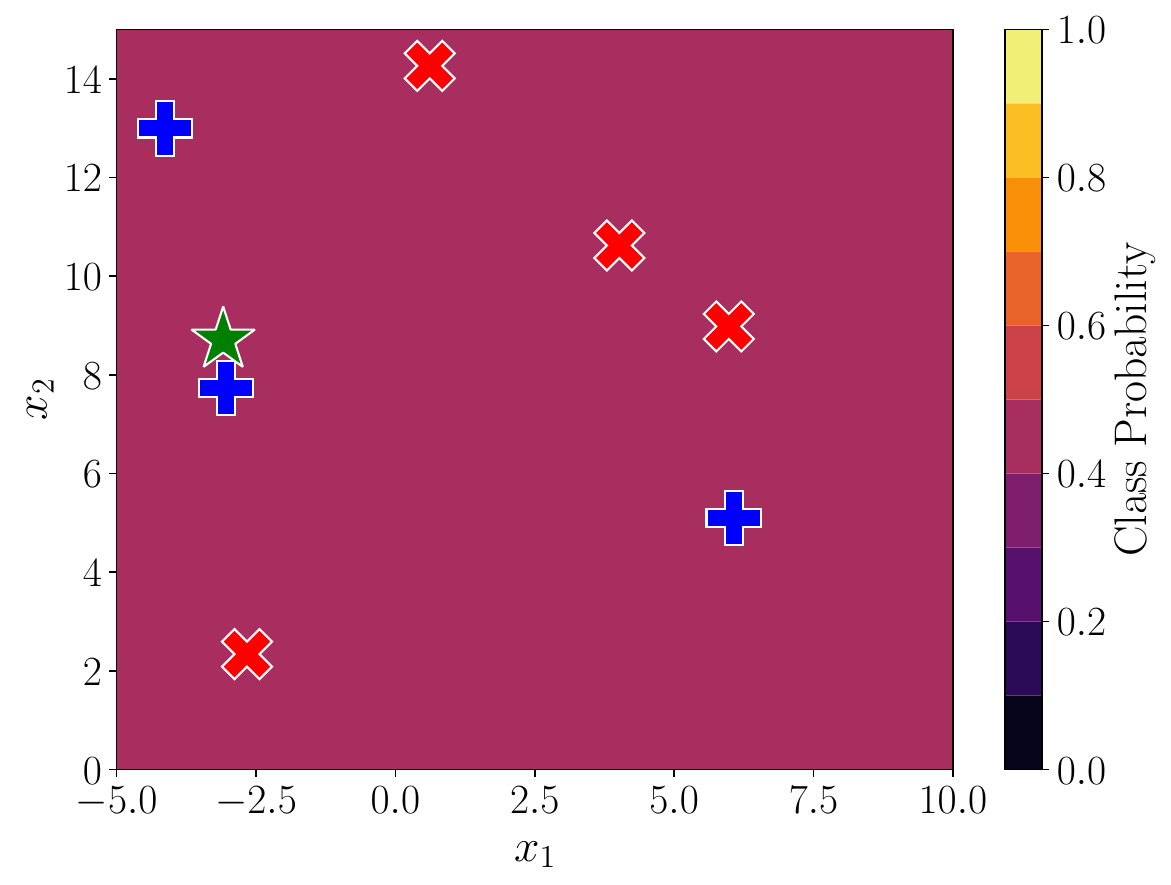}
        \includegraphics[width=0.19\textwidth]{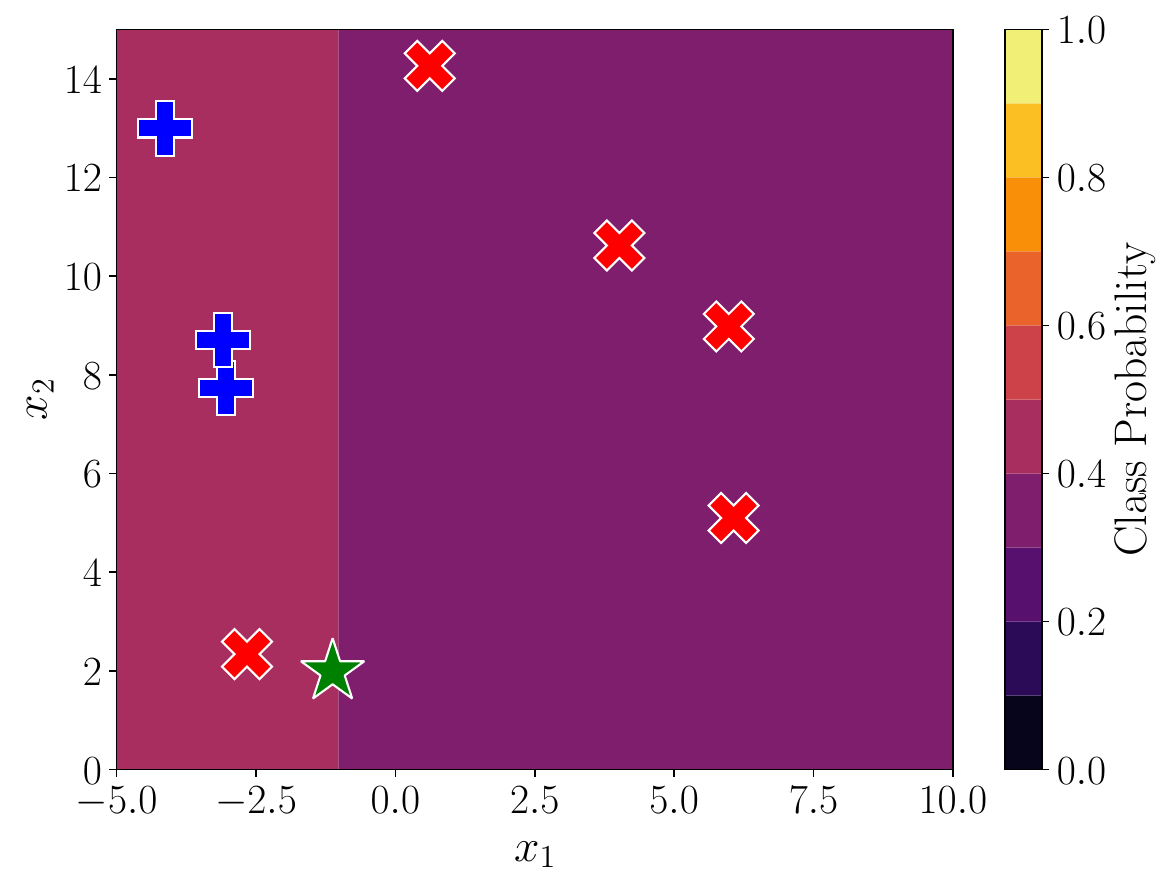}
        \includegraphics[width=0.19\textwidth]{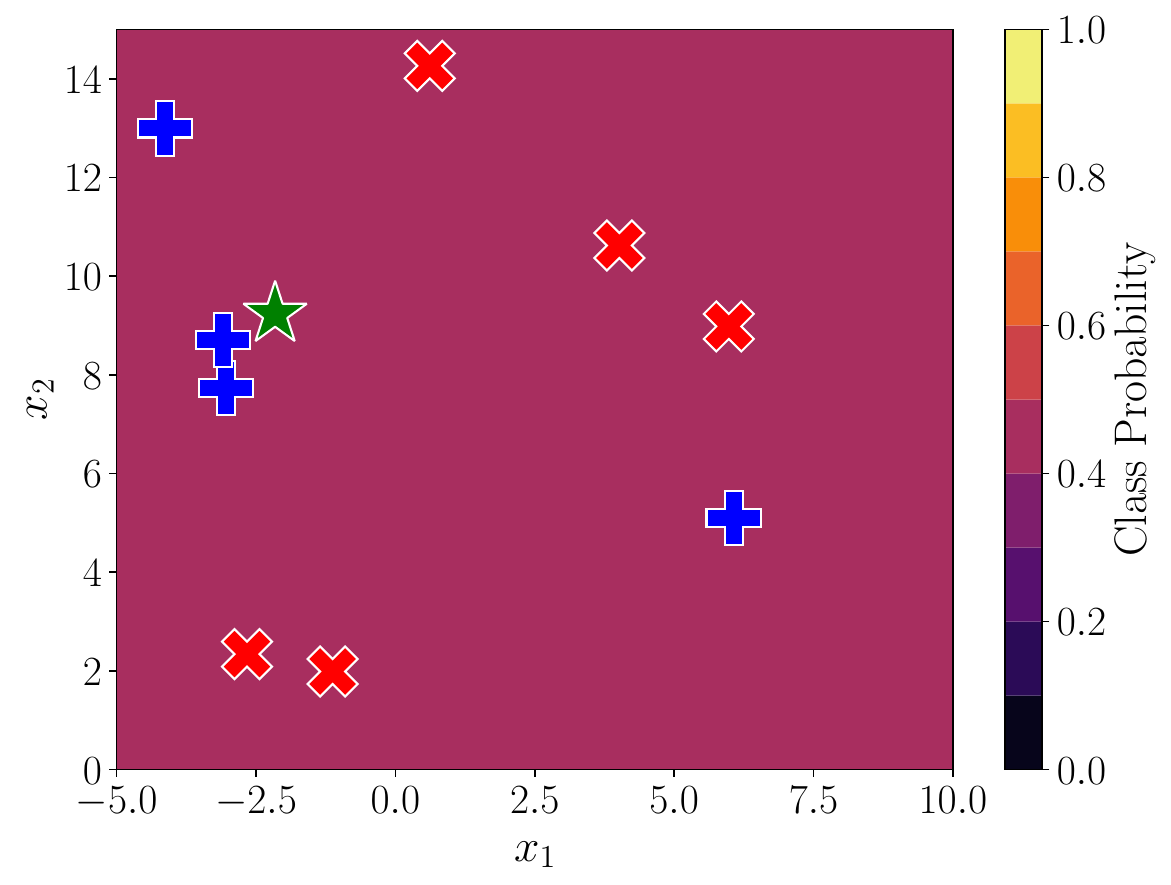}
    }
    \subfigure[XGBoost, LFBO, Iterations 1 to 5]{
        \includegraphics[width=0.19\textwidth]{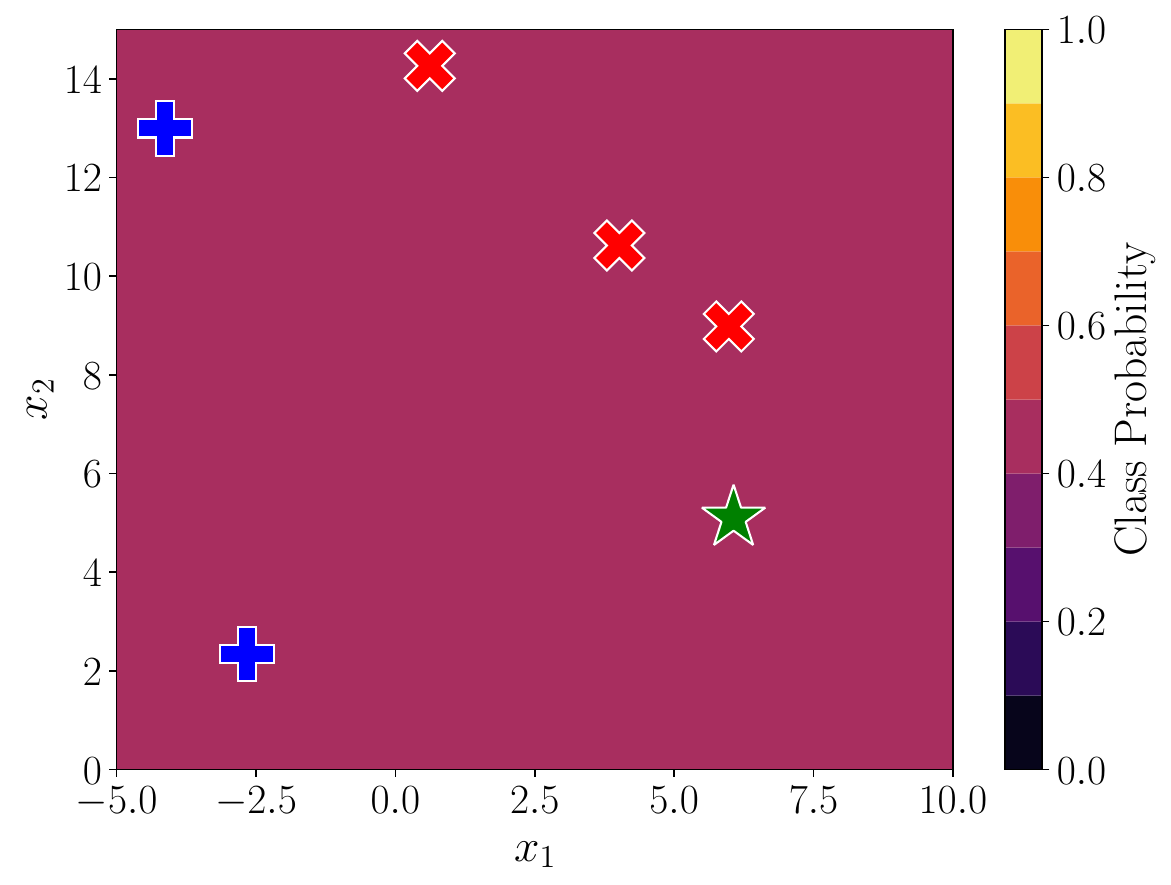}
        \includegraphics[width=0.19\textwidth]{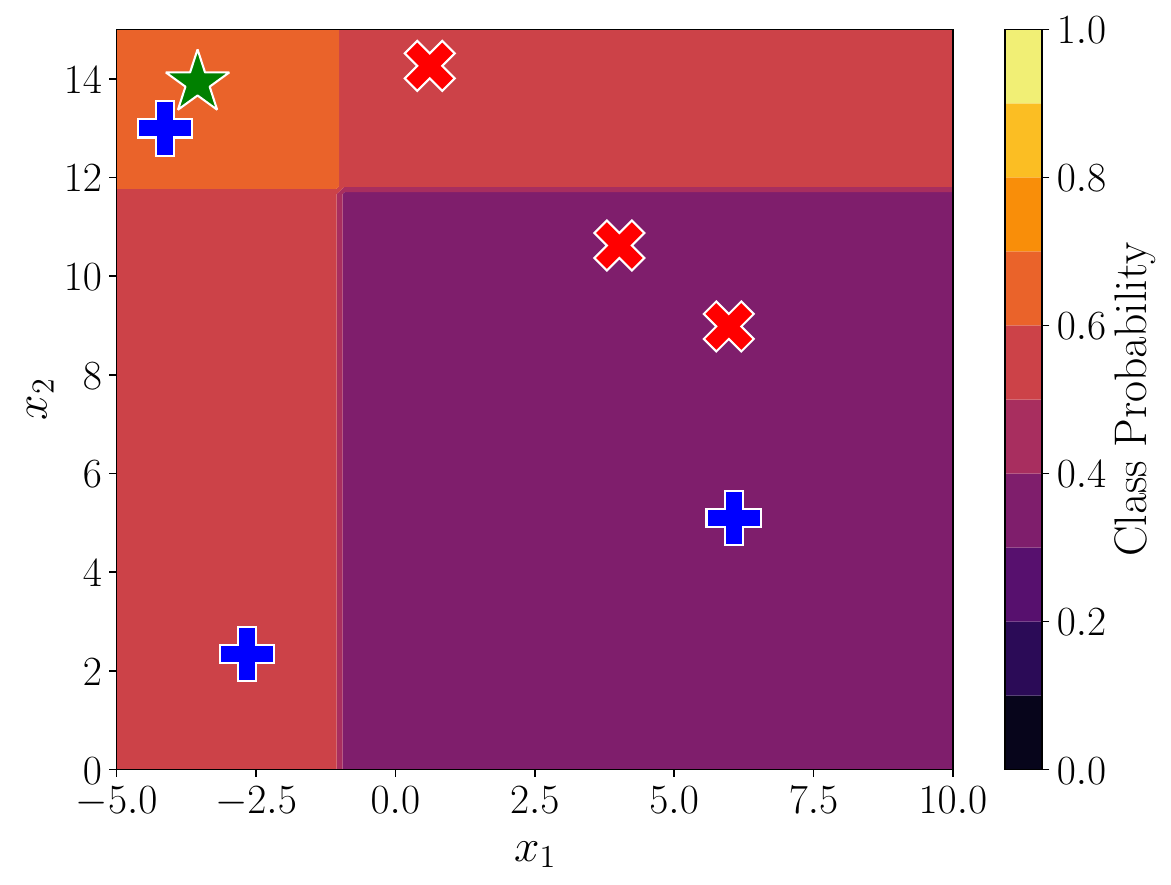}
        \includegraphics[width=0.19\textwidth]{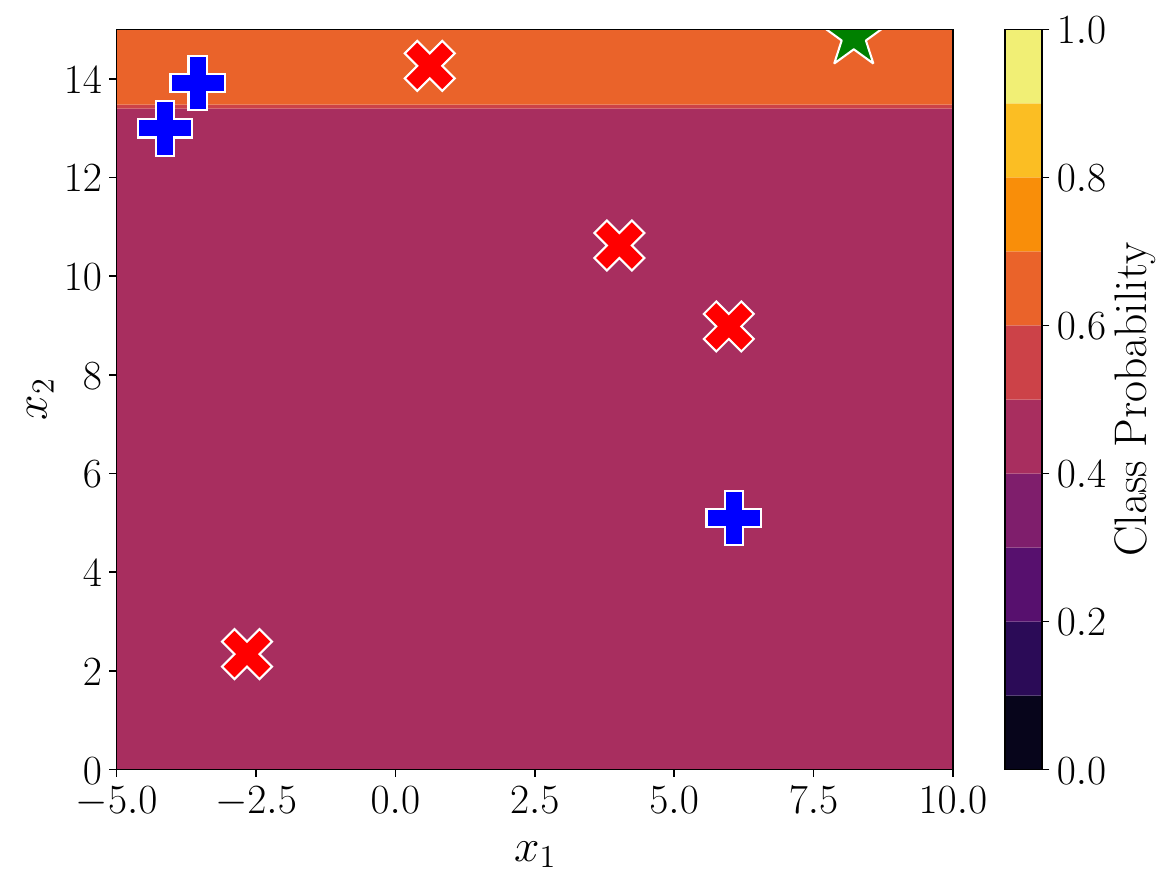}
        \includegraphics[width=0.19\textwidth]{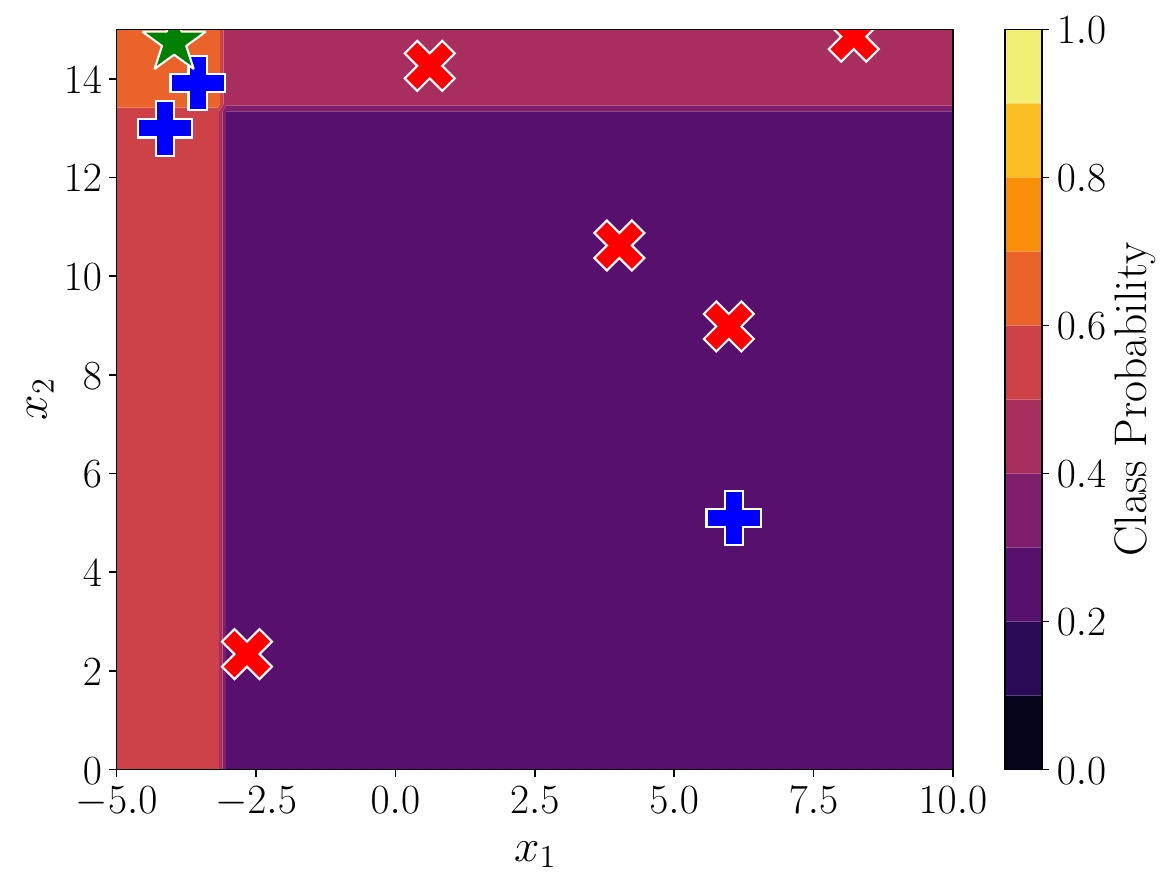}
        \includegraphics[width=0.19\textwidth]{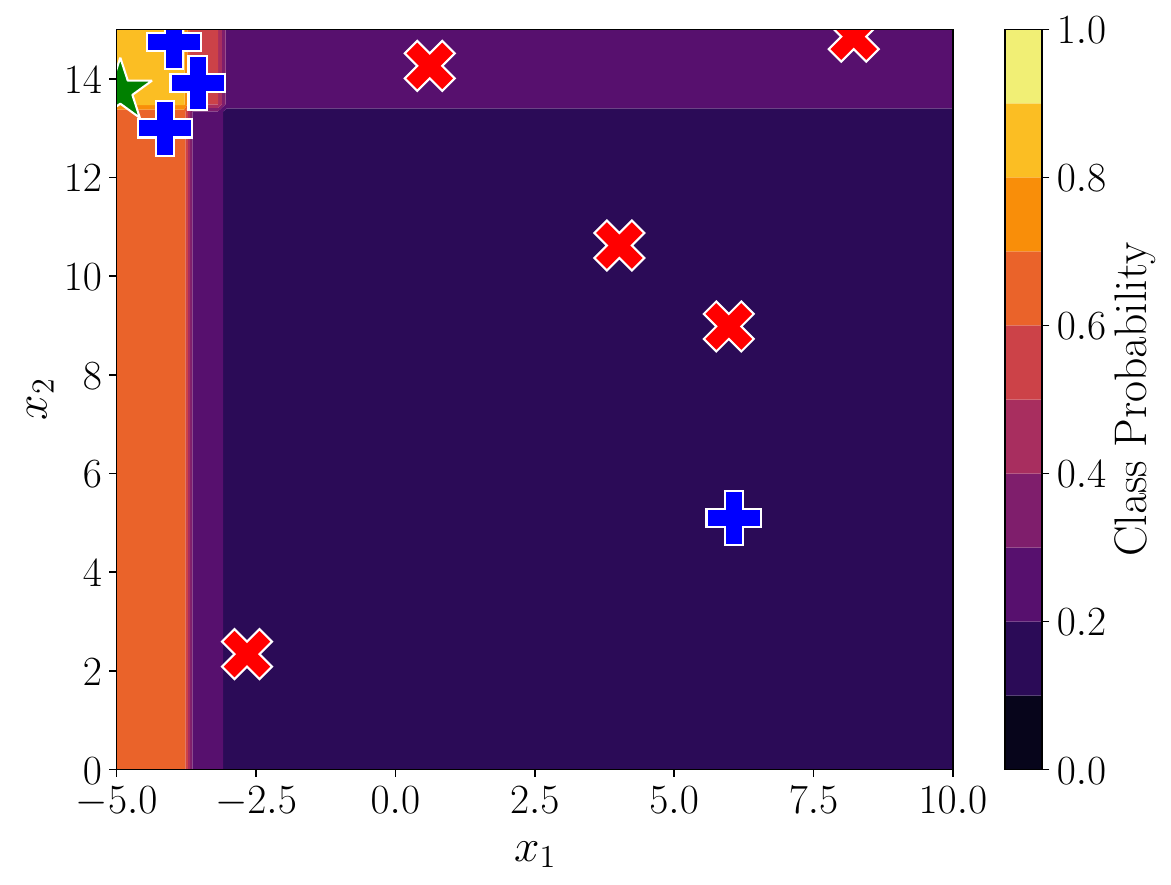}
    }
	\caption{Comparisons of BORE and LFBO by random forests, gradient boosting, and XGBoost for the Branin function. It follows the configurations described in~\figref{fig:prob_spaces}.}
	\label{fig:prob_spaces_random_forest_gradient_boosting_xgboost}
\end{figure}

In addition to \figref{fig:prob_spaces},
we include additional comparisons of BORE and LFBO
by random forests~\citep{BreimanL2001ml},
gradient boosting~\citep{FriedmanJH2001aos},
and XGBoost~\citep{ChenT2016kdd} for the Branin function.
For~\figsref{fig:prob_spaces}{fig:prob_spaces_random_forest_gradient_boosting_xgboost},
we use $\zeta = 0.33$ and $\beta = 0.5$.

\section{Additional Related Work}

Although many recent Bayesian optimization methods using regression models~\citep{ErikssonD2019neurips,BalandatM2020neurips,CowenRiversAI2022jair,AmentS2023neurips} show strong performance on various benchmark functions,
we do not compare our methods to them, as such comparisons are beyond the scope of this work.
The goal of this work is to demonstrate the effectiveness of DRE-based Bayesian optimization with semi-supervised learning in the settings of DRE-based Bayesian optimization.
\citet{PichenyV2019arxiv} propose a novel Bayesian optimization framework that relies on variable ordering in a latent space to handle ill-conditioned or discontinuous objectives.
Regarding the choice of the truncated multivariate normal distribution,
it may be related to the boundary issue of Bayesian optimization;
see the Ph.D. thesis of~\citet{SwerskyK2017thesis}.
Such a similar issue is discussed in the existing work by~\citet{OhC2018icml};
it proposes a Bayesian optimization approach with cylindrical kernels concentrating on a region proximal to the center of the search space.
Moreover,
the prior work by~\citet{HvarfnerC2024icml} also reports the similar consequence in Bayesian optimization.
From the perspective of DRE-based Bayesian optimization,
Gaussian process classification~\citep{RasmussenCE2006book} might be applicable for defining binary classifiers; it is left for future work.

\section{Details of~\ours}

\begin{algorithm}[ht]
	\caption{Labeling Unlabeled Data}
	\label{alg:labeling}
	\begin{algorithmic}[1]
        \REQUIRE Labeled data points $\bX_l$, their labels $\bC_l$, unlabeled data points $\bX_u$, maximum iterations $\tau$, and a tolerance $\varepsilon$. Additionally, a clamping factor $\alpha$ for label spreading.
        \ENSURE Propagated labels $\widehat{\bC}$.
        \STATE Initialize propagated labels $\widehat{\bC}$ of $\bX$.
        \STATE Compute similarities $\bW$ and a degree matrix $\bD$.
        \STATE Compute transition probabilities $\bP$ with $\bW$ and $\bD$.
		\REPEAT
			\STATE Propagate $\widehat{\bC}$ with $\bP$ and the previous $\widehat{\bC}$, and additionally $\alpha$ for label spreading.
            \STATE Normalize $\widehat{\bC}$ row-wise.
		\UNTIL{a change of $\widehat{\bC}$ converging to $\varepsilon$ or reaching $\tau$.}
	\end{algorithmic}
\end{algorithm}

\algref{alg:labeling} describes a procedure to label unlabeled data points;
see the main article for the details of~\ours.

\section{Analysis of~\ours}
\label{sec:analysis}

Under the cluster assumption, i.e.,~\asmref{asm:cluster},
a margin $\gamma$ is defined as a minimum distance between two decision boundaries.

\begin{definition}
    \label{def:margin}
    Let a compact connected decision set be $\calC_i \subseteq \calX$ for $i \in \{0, 1\}$ and a boundary subset, i.e., a set of boundary points, of a compact connected set $\calS$ be $\partial \calS$. A margin $\gamma$ is defined as
    \begin{equation}
        \gamma = (2\bbI(\calC_1 \cap \calC_2 = \emptyset) - 1) \min_{\bx_1 \in \partial \calC_1 \backslash \partial \calX, \bx_2 \in \partial \calC_2 \backslash \partial \calX} \| \bx_1 - \bx_2 \|_2.
    \end{equation}
\end{definition}

Using~\defref{def:margin},
we claim that a semi-supervised classifier
in~\ours~can mitigate the over-exploitation problem presented in~\secref{subsec:overexploitation},
because it can expand a decision set $\calC_1$ for Class 1
by reducing $\gamma$ with unlabeled data.
However, we need to verify if a large decision set
is derived from the characteristics of non-parametric classifiers,
since a semi-supervised classifier we use is converted to the Nadaraya-Watson non-parametric model~\citep{NadarayaEA1964tpia,WatsonGS1964sijs} without unlabeled data.
As shown in~\figref{fig:unlabeled_points},
there is no strong relationship between the performance of
semi-supervised classifiers without unlabeled data,
i.e., the Nadaraya-Watson estimator,
and one of semi-supervised classifiers with unlabeled data.
It implies that configuration selection for a semi-supervised classifier is dependent on a class of objective function,
and the presence of unlabeled data is likely to be effective
for alleviating the over-exploitation problem.

\citet{SinghA2008neurips} provide a sample error bound of supervised and semi-supervised
learners, related to $\gamma$, $n_l$, $n_u$, and $d$.
This work proves that
a semi-supervised learner can be better than any supervised learners,
by assuming that $n_u \gg n_l$ and access to perfect knowledge of decision sets.
However,
these theoretical results cannot be directly applied in our sequential problem
because this work assumes that both labeled and unlabeled data points are independent and identically distributed.
Nevertheless,
these results can hint a theoretical guarantee on better performance of semi-supervised classifiers with unlabeled data points.
Further analysis for the circumstances of Bayesian optimization is left for future work.

\section{Experimental Details}
\label{sec:exp_details}

Here we present the missing details of the experiments shown in the main part.

To carry out the experiments in our work,
we use dozens of commercial Intel and AMD CPUs
such as Intel Xeon Gold 6126 and AMD EPYC 7302.
For the experiments on minimum multi-digit MNIST search,
the NVIDIA GeForce RTX 3090 GPU is used.

To minimize~\eqref{eqn:entropy} for finding an adequate $\beta$ of label propagation and label spreading,
we use L-BFGS-B with a single initialization,
which is implemented in SciPy~\citep{VirtanenP2020nm},
for all the experiments in~Figures~\ref{fig:synthetic_sampling},
\ref{fig:synthetic_pools},
\ref{fig:tabular}, \ref{fig:natsbench},
and~\ref{fig:multi_digit_mnist_results}.
For the other empirical analyses,
we set $\beta$ as a specific fixed value;
see the corresponding sections for the details.

To reduce the computational complexity of \ours~for a scenario with a fixed-size pool,
we randomly sample 2,000 unlabeled points from the predefined pool for all experiments excluding synthetic benchmark functions.
More thorough analysis can be found in~\secref{sec:discussion_pool_sampling}.

To compare baseline methods with our methods,
we assess optimization algorithms using a simple regret:
\begin{equation}
    \textrm{simple regret}(f(\bx_1), \ldots, f(\bx_t), f(\bx^*)) = \min_{i = 1}^t f(\bx_i) - f(\bx^*),
\end{equation}
where $\bx^*$ is a global optimum.

Our proposed methods and baseline methods are implemented with scikit-learn~\citep{PedregosaF2011jmlr},
PyTorch~\citep{PaszkeA2019neurips},
NumPy~\citep{HarrisCR2020nature},
SciPy~\citep{VirtanenP2020nm},
XGBoost~\citep{ChenT2016kdd},
and BayesO~\citep{KimJ2023joss}.
Scikit-learn and SciPy are under the BSD 3-Clause license,
NumPy is under the modified BSD license,
XGBoost is under the Apache 2.0 license,
and BayesO is under the MIT license.
On the other hand,
PyTorch is under its own license;
refer to its repository for the license.

\subsection{Details of Synthetic Benchmarks}

Here we describe the details of synthetic benchmarks.

\paragraph{Beale Function.}

This function is defined as follows:
\begin{equation}
	f(\bx) = (1.5 - x_1 + x_1 x_2)^2 + (2.25 - x_1 + x_1 x_2^2)^2 + (2.625 - x_1 + x_1 x_2^3)^2,
\end{equation}
where $\bx = [x_1, x_2] \in [-4.5, 4.5]^2$.

\paragraph{Branin Function.}

It is defined as follows:
\begin{equation}
	f(\bx) = \left(x_2 - (5.1 / 4 \pi^2) x_1^2 + (5 / \pi) x_1 - 6 \right)^2 + 10 \left(1 - (1 / 8 \pi) \right) \cos(x_1) + 10,
\end{equation}
where $\bx = [x_1, x_2] \in [[-5, 10], [0, 15]]$.

\paragraph{Bukin6 Function.}

This benchmark is given by the following:
\begin{equation}
	f(\bx) = 100 \sqrt{|x_2 - 0.01 x_1^2|} + 0.01 |x_1 + 10|,
\end{equation}
where $\bx = [x_1, x_2] \in [[-15, -5], [-3, 3]]$.

\paragraph{Six-Hump Camel Function.}

It is given by the following:
\begin{equation}
	f(\bx) = \left( 4 - 2.1 x_1^2 + x_1^4 / 3 \right) x_1^2 + x_1 x_2 + (-4 + 4x_2^2) x_2^2,
\end{equation}
where $\bx = [x_1, x_2] \in [[-3, 3], [-2, 2]]$.

\subsection{Details of Tabular Benchmarks}

\begin{table}[ht]
    \centering
    \caption{Search space for Tabular Benchmarks. ``tanh'' and ``relu'' represent hyperbolic tangent and ReLU, respectively.}
    \vspace{5pt}
    \label{tab:tabular}
    \small
    \begin{tabular}{lc}
        \toprule
        \textbf{Hyperparameter} & \textbf{Possible Values} \\
        \midrule
        The number of units for 1st layer & \{16, 32, 64, 128, 256, 512\} \\
        The number of units for 2nd layer & \{16, 32, 64, 128, 256, 512\} \\
        Dropout rate for 1st layer & \{0.0, 0.3, 0.6\} \\
        Dropout rate for 2nd layer & \{0.0, 0.3, 0.6\} \\
        Activation function for 1st layer & \{``tanh'', ``relu''\} \\
        Activation function for 2nd layer & \{``tanh'', ``relu''\} \\
        Initial learning rate & \{$5 \times 10^{-4}$, $1 \times 10^{-3}$, $5 \times 10^{-3}$, $1 \times 10^{-2}$, $5 \times 10^{-2}$, $1 \times 10^{-1}$\} \\
        Learning rate scheduling & \{``cosine'', ``constant''\} \\
        Batch size & \{8, 16, 32, 64\} \\
        \bottomrule
    \end{tabular}
\end{table}

The search space of Tabular Benchmarks~\citep{KleinA2019arxiv}
is described in~\tabref{tab:tabular}.
To handle categorical and discrete variables,
we treat them as integer variables by following the previous literature by~\citet{GarridoEC2020neucom}.
More precisely,
the value of each integer variable corresponds to the index of the original categorical or discrete variable.
When evaluating the variables,
the integer variables are transformed into the original variables.

\subsection{Details of NATS-Bench}

\begin{figure}[ht]
    \centering
    \includegraphics[width=0.95\textwidth]{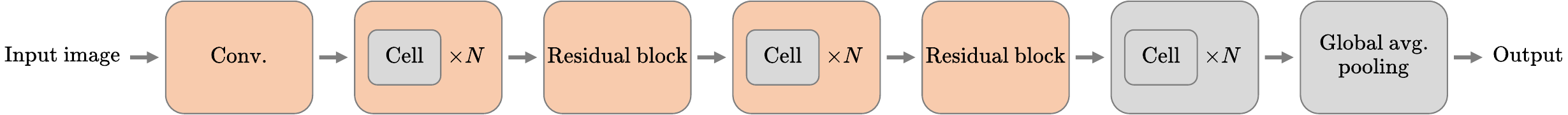}
    \caption{Neural network architecture in NATS-Bench. Orange blocks are optimized.}
    \label{fig:arch_natsbench}
\end{figure}

\begin{table}[ht]
    \centering
    \caption{Search space for NATS-Bench. There exist $8^5 =$ 32,768 models.}
    \vspace{5pt}
    \label{tab:natsbench}
    \small
    \begin{tabular}{lc}
        \toprule
        \textbf{Hyperparameter} & \textbf{Possible Values} \\
        \midrule
        Output channels of 1st convolutional layer & \{8, 16, 24, 32, 40, 48, 56, 64\} \\
        Output channels of 1st cell stage & \{8, 16, 24, 32, 40, 48, 56, 64\} \\
        Output channels of 1st residual block & \{8, 16, 24, 32, 40, 48, 56, 64\} \\
        Output channels of 2nd cell stage & \{8, 16, 24, 32, 40, 48, 56, 64\} \\
        Output channels of 2nd residual block & \{8, 16, 24, 32, 40, 48, 56, 64\} \\
        \bottomrule
    \end{tabular}
\end{table}

We describe the search space for NATS-Bench~\citep{DongX2021ieeetpami}
in~\figref{fig:arch_natsbench}
and~\tabref{tab:natsbench}.

\subsection{Details of Minimum Multi-Digit MNIST Search}

\begin{figure}[t]
    \centering
    \includegraphics[width=0.18\textwidth]{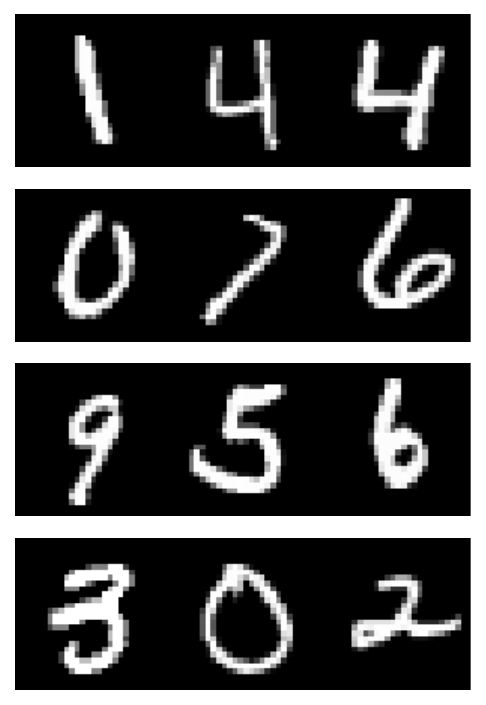}
    \caption{Examples on 64D minimum multi-digit MNIST search.}
    \label{fig:multi_digit_mnist}
\end{figure}

Since multi-digit MNIST, which is composed of images of size $(28, 84)$ and shown in~\figref{fig:multi_digit_mnist},
is high-dimensional,
some of the methods used in this work,
e.g., methods with random forests, gradient boosting, and XGBoost,
struggle to process such data.
Therefore, we embed an original image to a lower-dimensional vector using an auxiliary convolutional neural network.
The convolutional neural network is trained to classify a three-digit image to one of labels from ``000'' to ``999,''
with the following architecture:
\begin{enumerate}
    \item[] First layer: convolutional, input channel $1$, output channel $8$, kernel size $3 \times 3$, padding 1, ReLU, max-pooling $2 \times 2$;
    \item[] Second layer: convolutional, input channel $8$, output channel $16$, kernel size $3 \times 3$, padding 1, ReLU, max-pooling $2 \times 2$;
    \item[] Third layer: convolutional, input channel $16$, output channel $32$, kernel size $3 \times 3$, padding 1, ReLU, max-pooling $2 \times 2$;
    \item[] Fourth layer: fully-connected, input dimensionality $960$, output dimensionality $128$, ReLU;
    \item[] Fifth layer: fully-connected, input dimensionality $128$, output dimensionality $64$, ReLU;
    \item[] Sixth layer: fully-connected, input dimensionality $64$, output dimensionality $1000$, Softmax.
\end{enumerate}
The Adam optimizer~\citep{KingmaDP2015iclr} with learning rate $1 \times 10^{-3}$ is used to train the network for 100 epochs.
To train and test the model fairly,
we create a training dataset of 440,000 three-digit images, a validation dataset of 40,000 three-digit images, and a test dataset of 80,000 three-digit images
using a training dataset of 55,000 single-digit images, a validation dataset of 5,000 single-digit images, and a test dataset of 10,000 single-digit images in the original MNIST dataset~\citep{LeCunY1998mnist}.
For example, supposing that a test dataset has 1,000 single-digit images per class---it is not true for the MNIST dataset, but it is assumed for explanation---and we would like to generate a three-digit image ``753,''
$10^9$ combinations for ``753'' can be created.
We therefore randomly sample a fixed number of three-digit images from a vast number of possible combinations. 
In addition, an early stopping technique is utilized
by comparing the current validation loss to the average of validation losses for the recent five epochs.
Eventually,
our network achieves $99.6\%$ in the training dataset,
$97.0\%$ in the validation dataset,
and $96.9\%$ in the test dataset.

To construct a fixed-size pool,
we use 80,000 embeddings of dimensionality 64,
which are derived from the outputs of the fifth layer without ReLU, by passing the test dataset of three-digit images through the network.

\clearpage

\section{Discussion on a Free Parameter in Label Propagation and Label Spreading}
\label{sec:discussion_free_parameters}

\begin{figure}[ht]
    \centering
    \subfigure[Beale, LP]{
        \centering
        \includegraphics[width=0.23\textwidth]{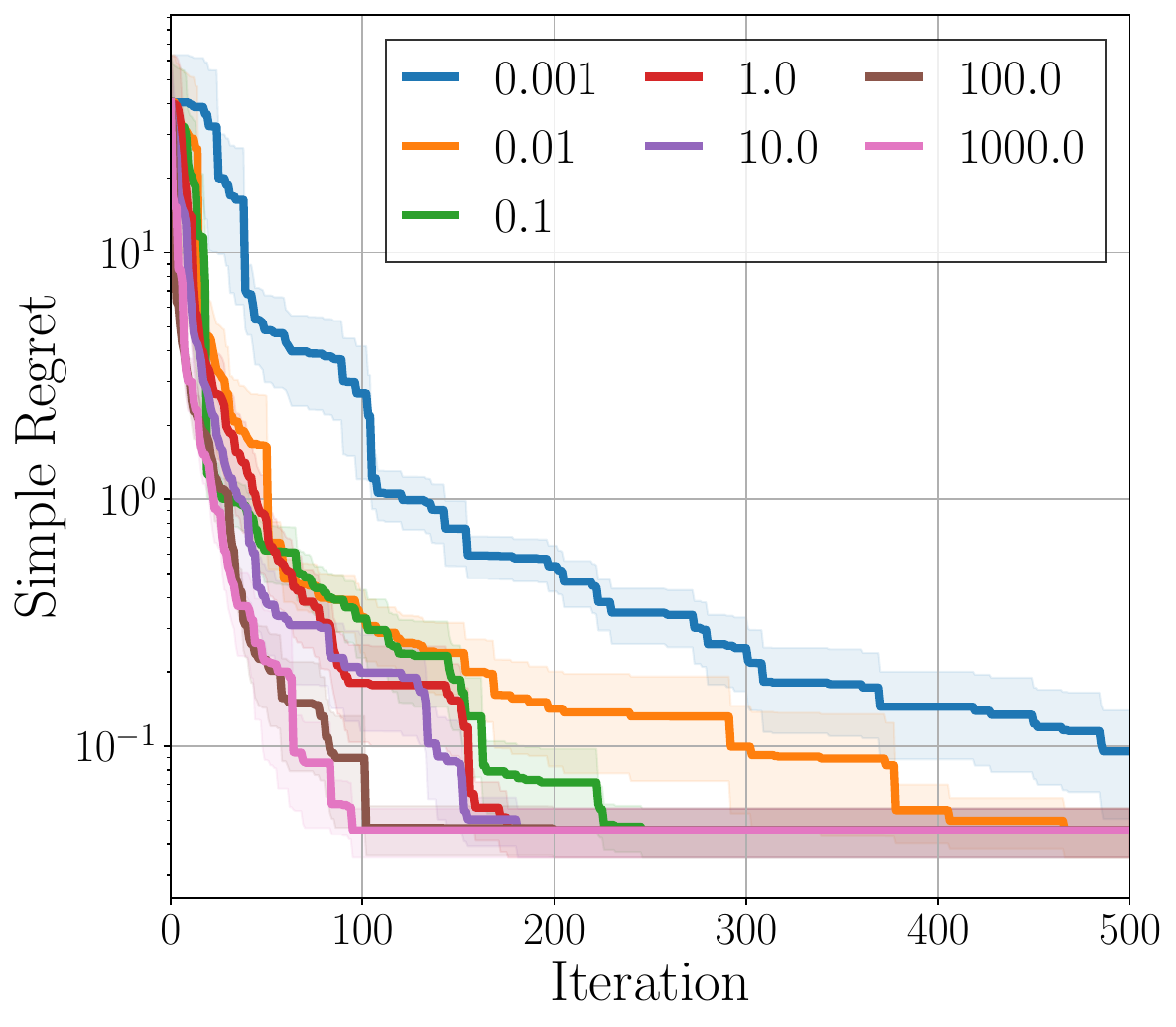}
    }
    \subfigure[Branin, LP]{
        \centering
        \includegraphics[width=0.23\textwidth]{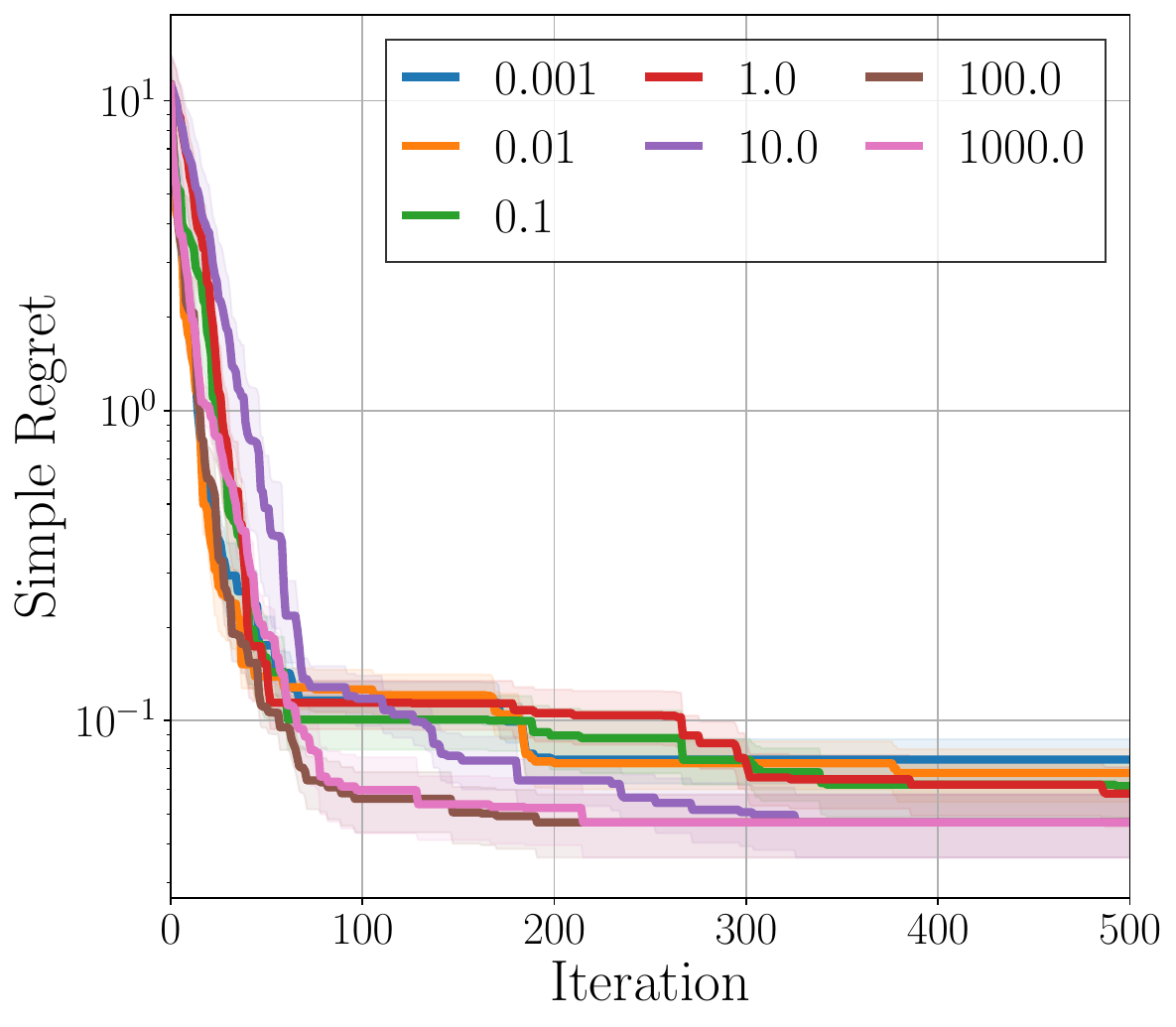}
    }
    \subfigure[Bukin6, LP]{
        \centering
        \includegraphics[width=0.23\textwidth]{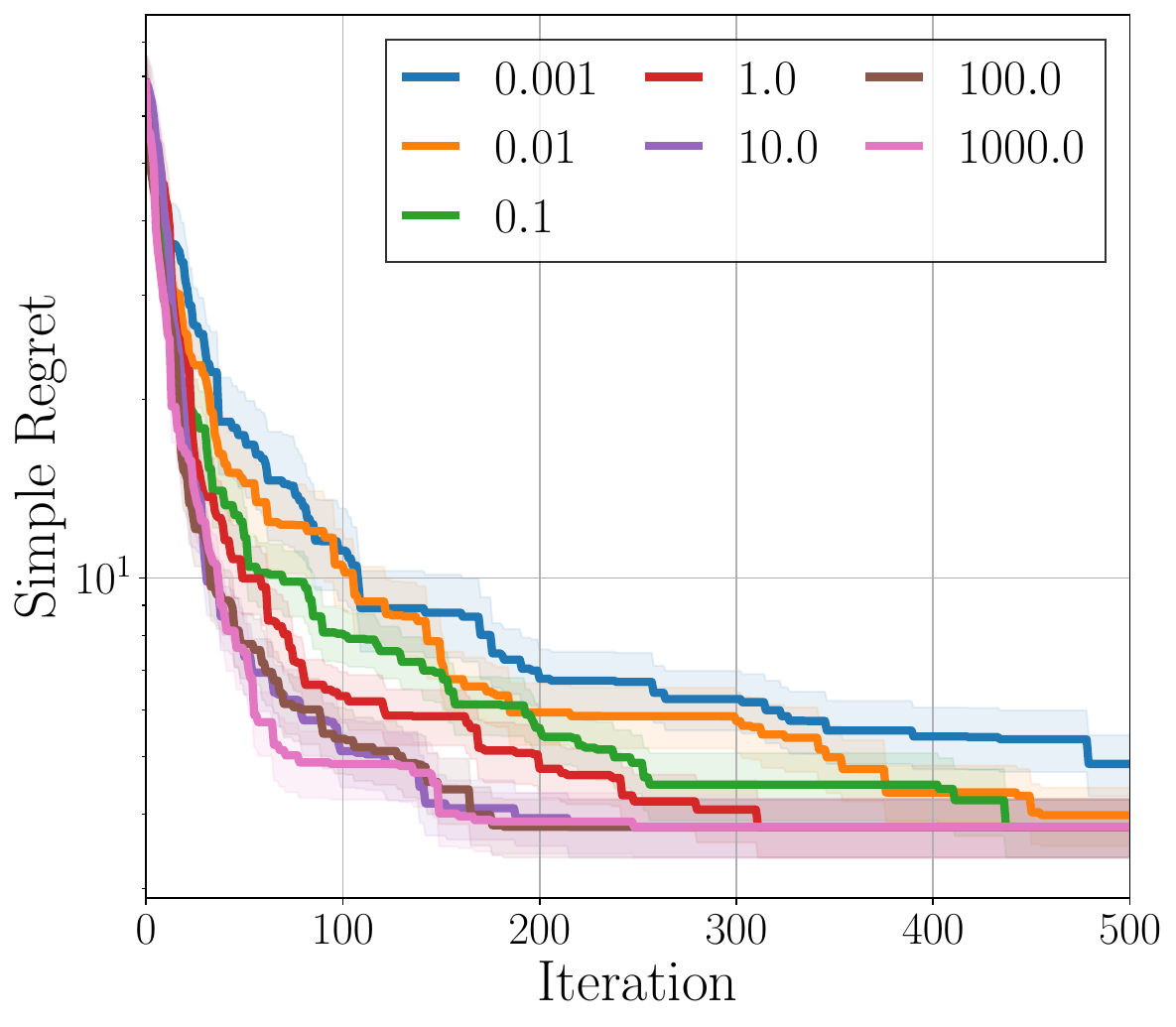}
    }
    \subfigure[Six-hump camel, LP]{
        \centering
        \includegraphics[width=0.23\textwidth]{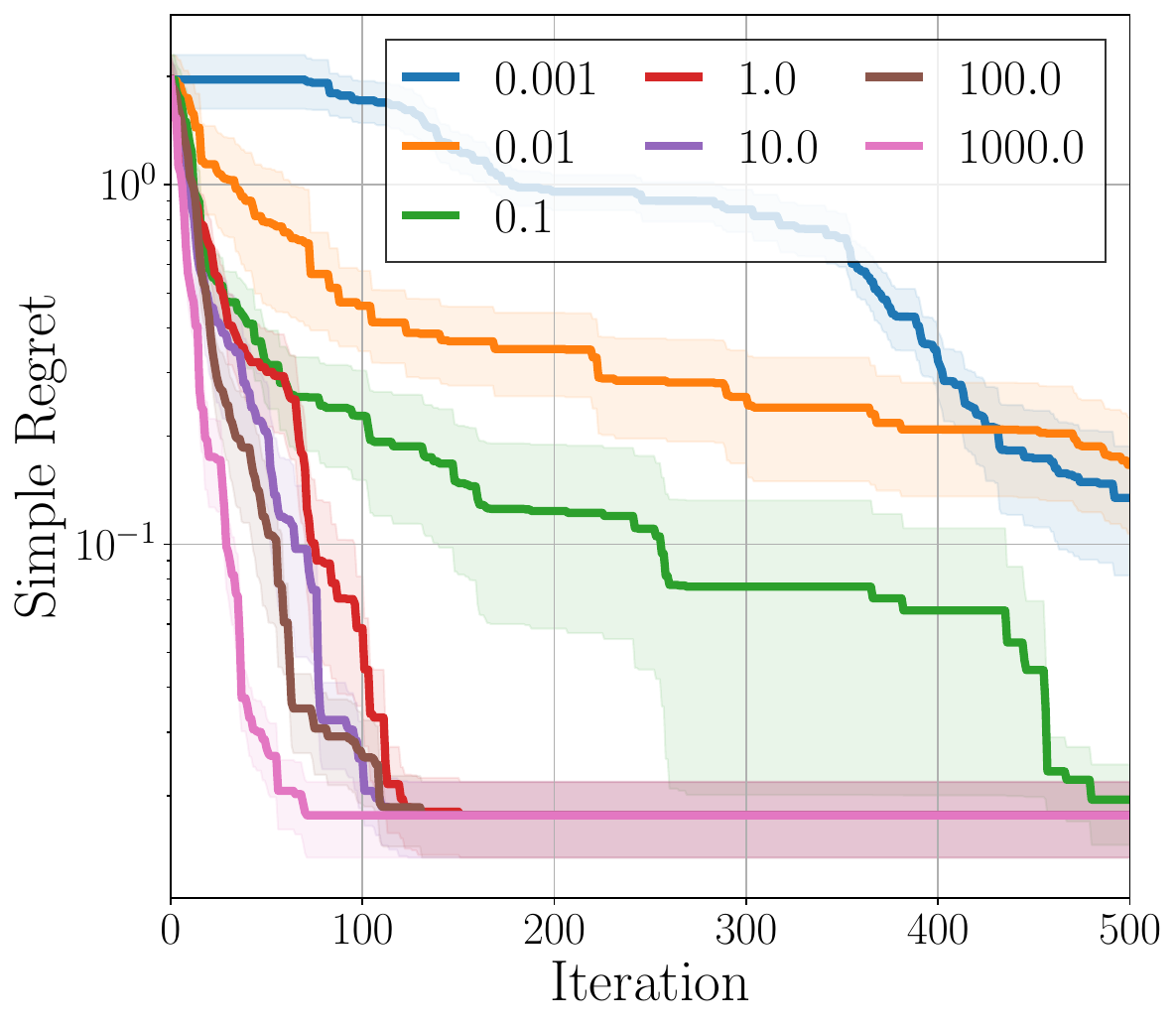}
    }
    \subfigure[Beale, LS]{
        \centering
        \includegraphics[width=0.23\textwidth]{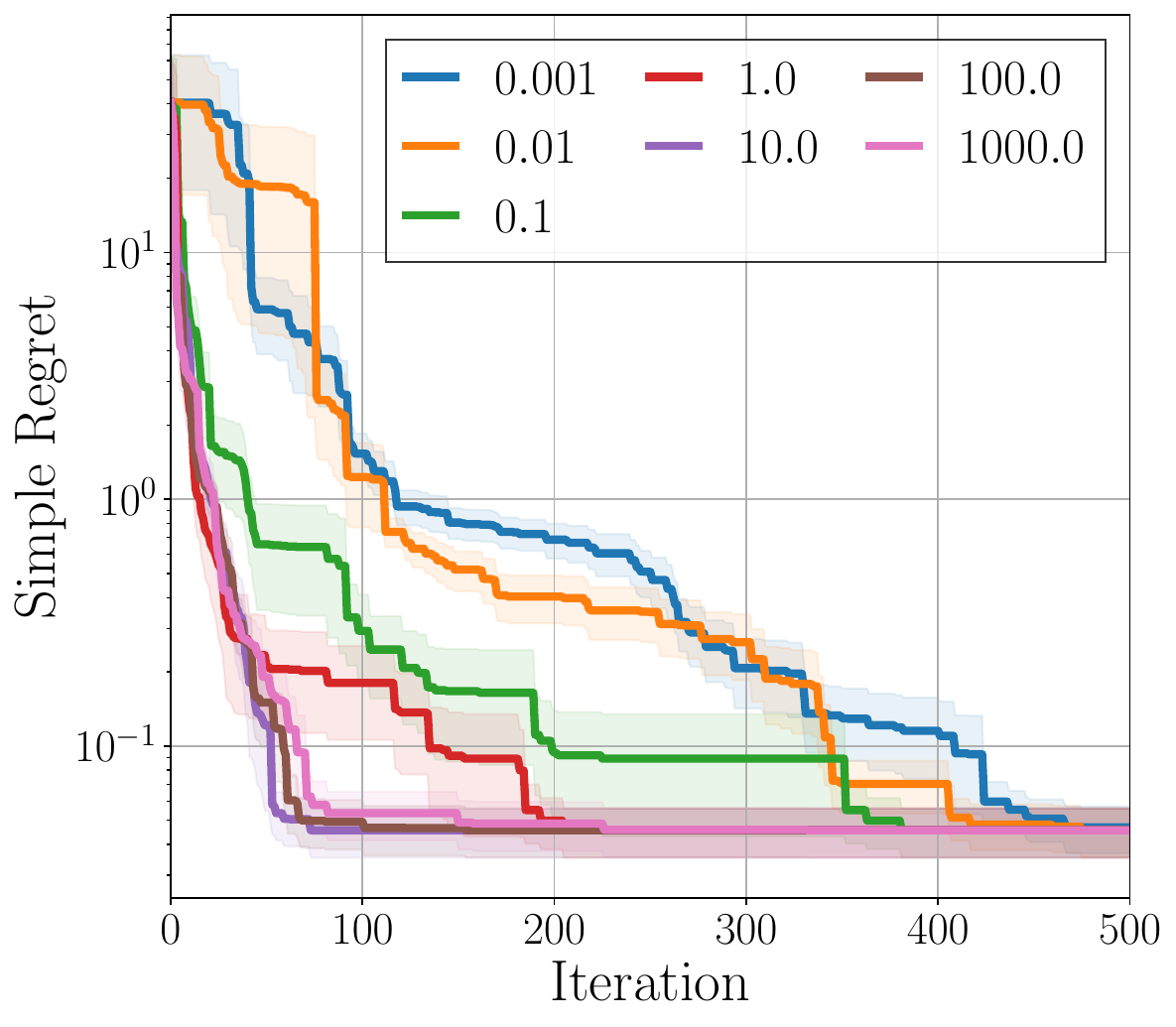}
    }
    \subfigure[Branin, LS]{
        \centering
        \includegraphics[width=0.23\textwidth]{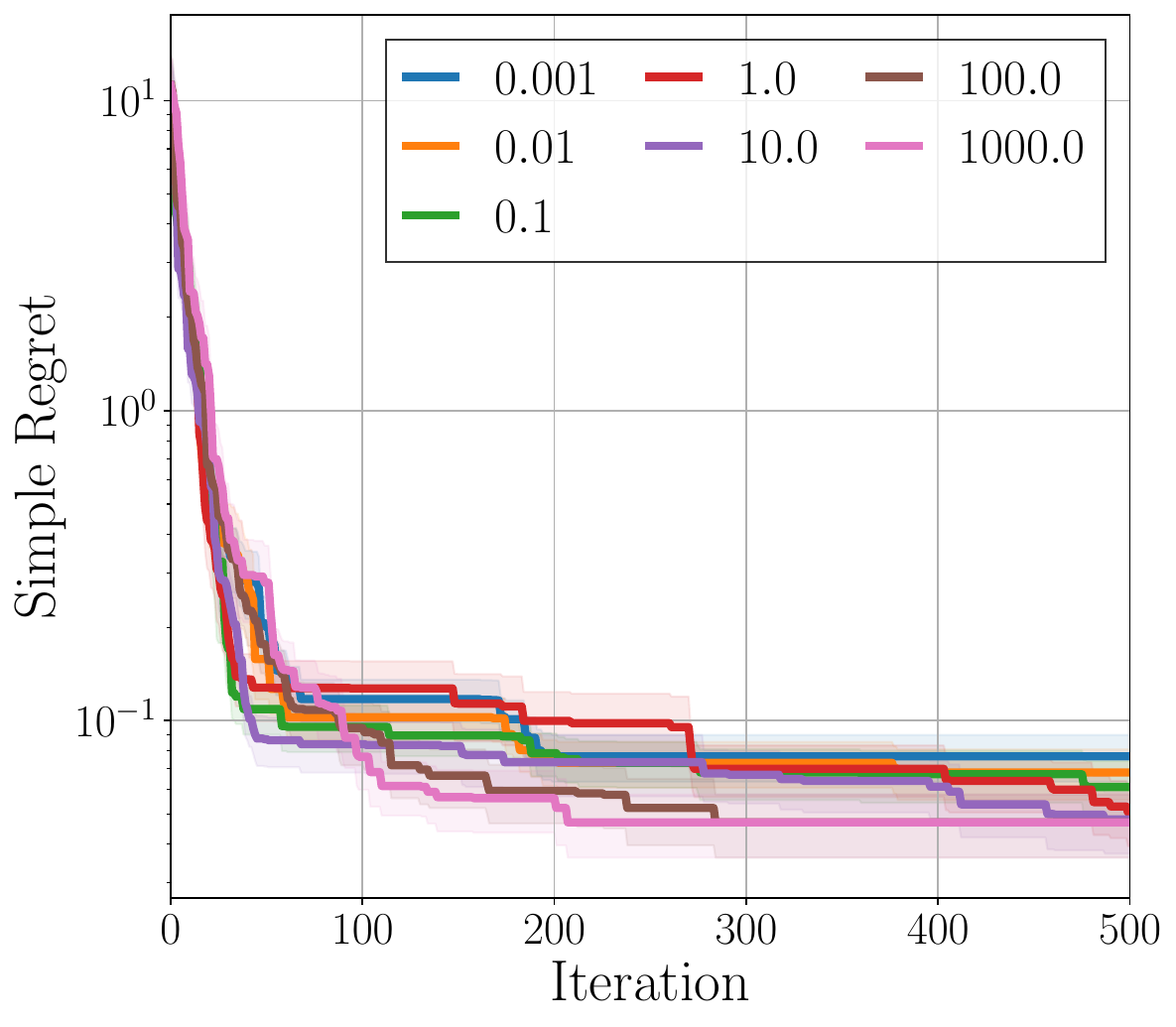}
    }
    \subfigure[Bukin6, LS]{
        \centering
        \includegraphics[width=0.23\textwidth]{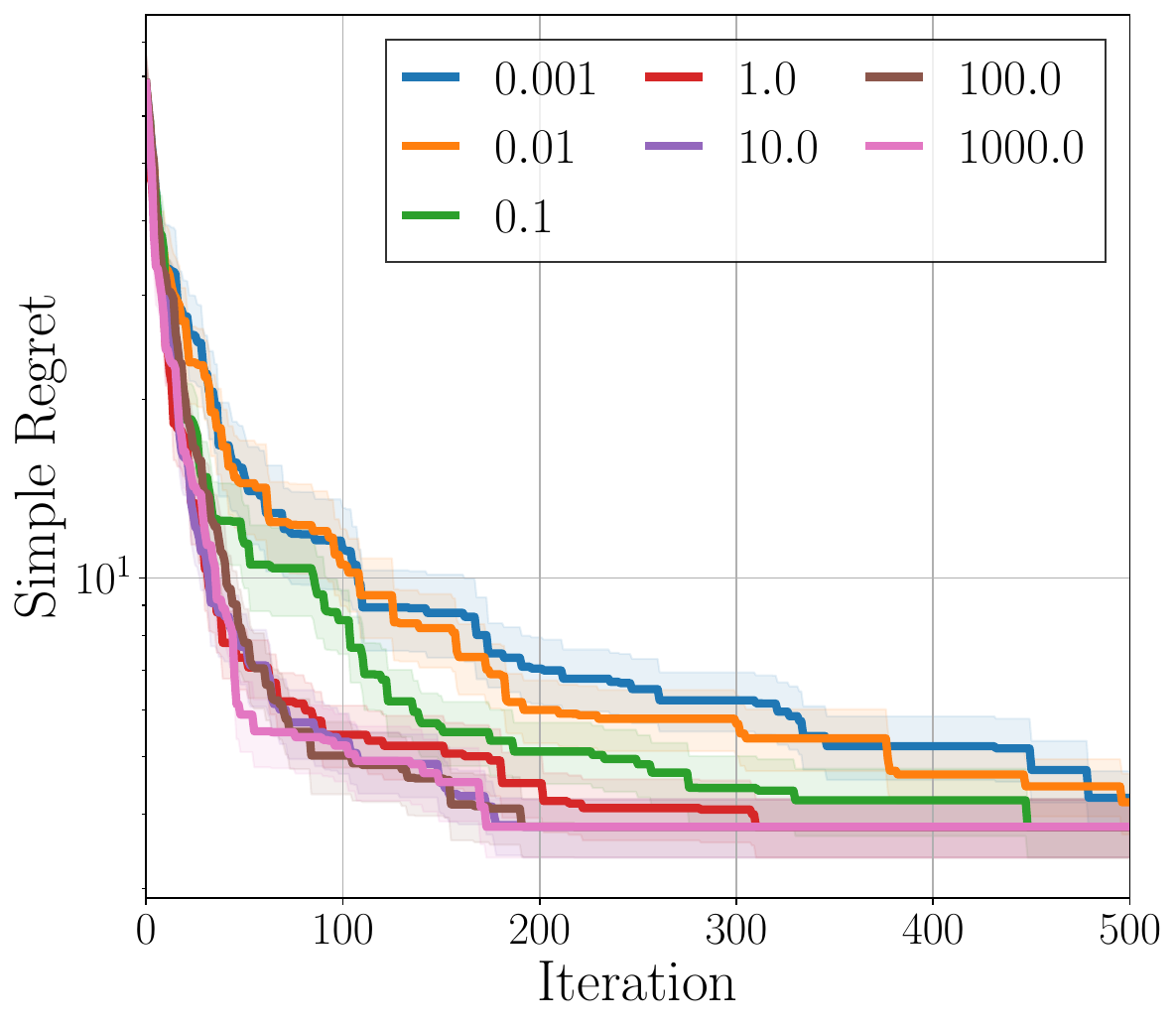}
    }
    \subfigure[Six-hump camel, LS]{
        \centering
        \includegraphics[width=0.23\textwidth]{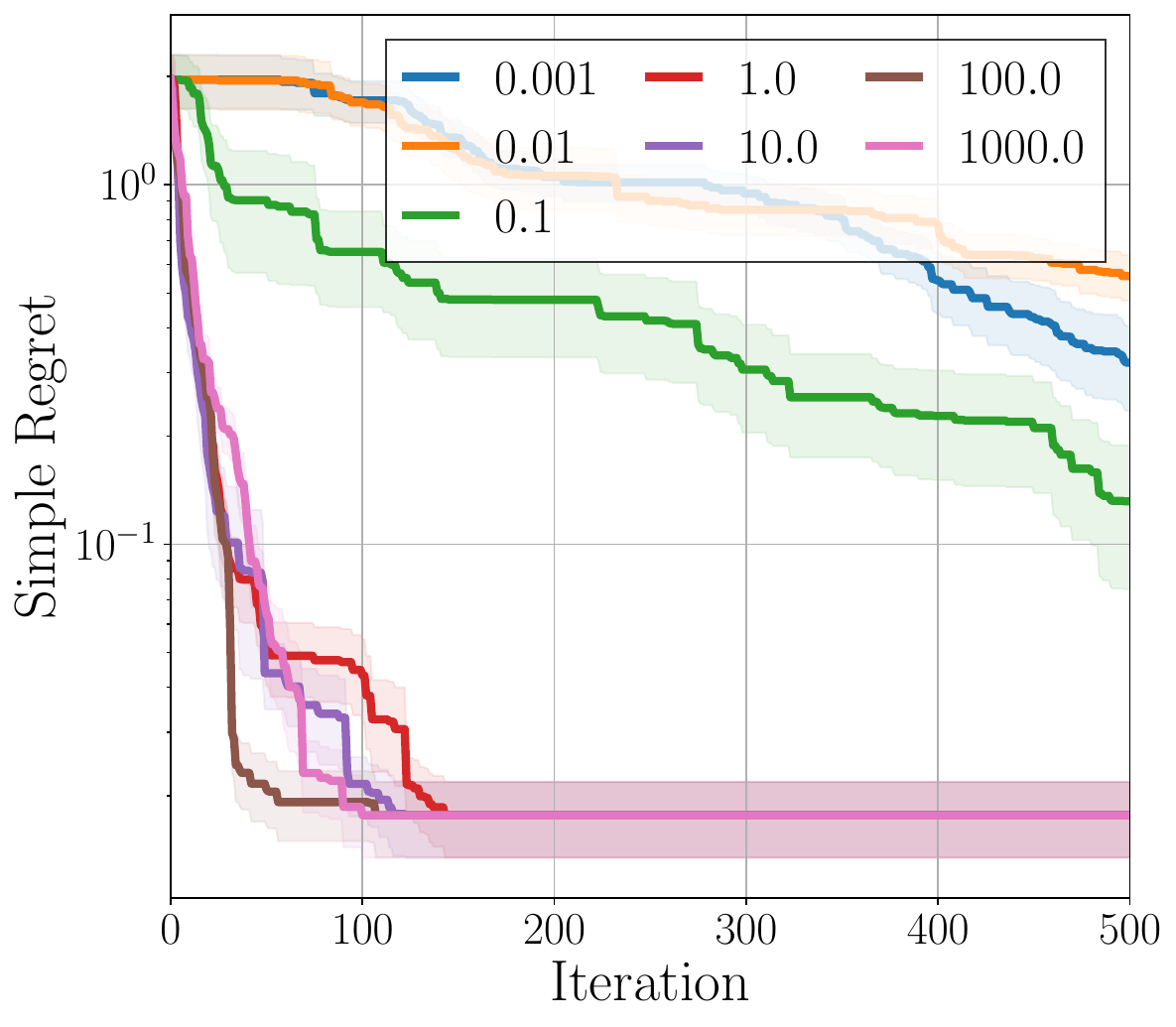}
    }
	\caption{Effects of a free parameter $\beta$ in label propagation, denoted as LP, and label spreading, denoted as LS. All experiments are repeated 20 times.}
	\label{fig:discussion_beta}
\end{figure}

In the Bayesian optimization process of \ours, a free parameter $\beta$ in label propagation and label spreading is learned every iteration by minimizing~\eqref{eqn:entropy};
see~\figref{fig:gammas} for the results on learned $\beta$.
Furthermore,
to show the effects of $\beta$,
we empirically analyze $\beta$
as depicted in~\figref{fig:discussion_beta}.
We sample 1,000 unlabeled points and use all of them as unlabeled points without pool sampling.
For the cases of four benchmark functions,
higher $\beta$ tends to show better performance
than lower $\beta$.
These results considerably correspond with the results in~\figref{fig:gammas}.

\clearpage

\section{Discussion on Unlabeled Point Sampling}
\label{sec:discussion_sampling}

\begin{figure}[ht]
    \centering
    \subfigure[Beale, LP]{
        \centering
        \includegraphics[width=0.23\textwidth]{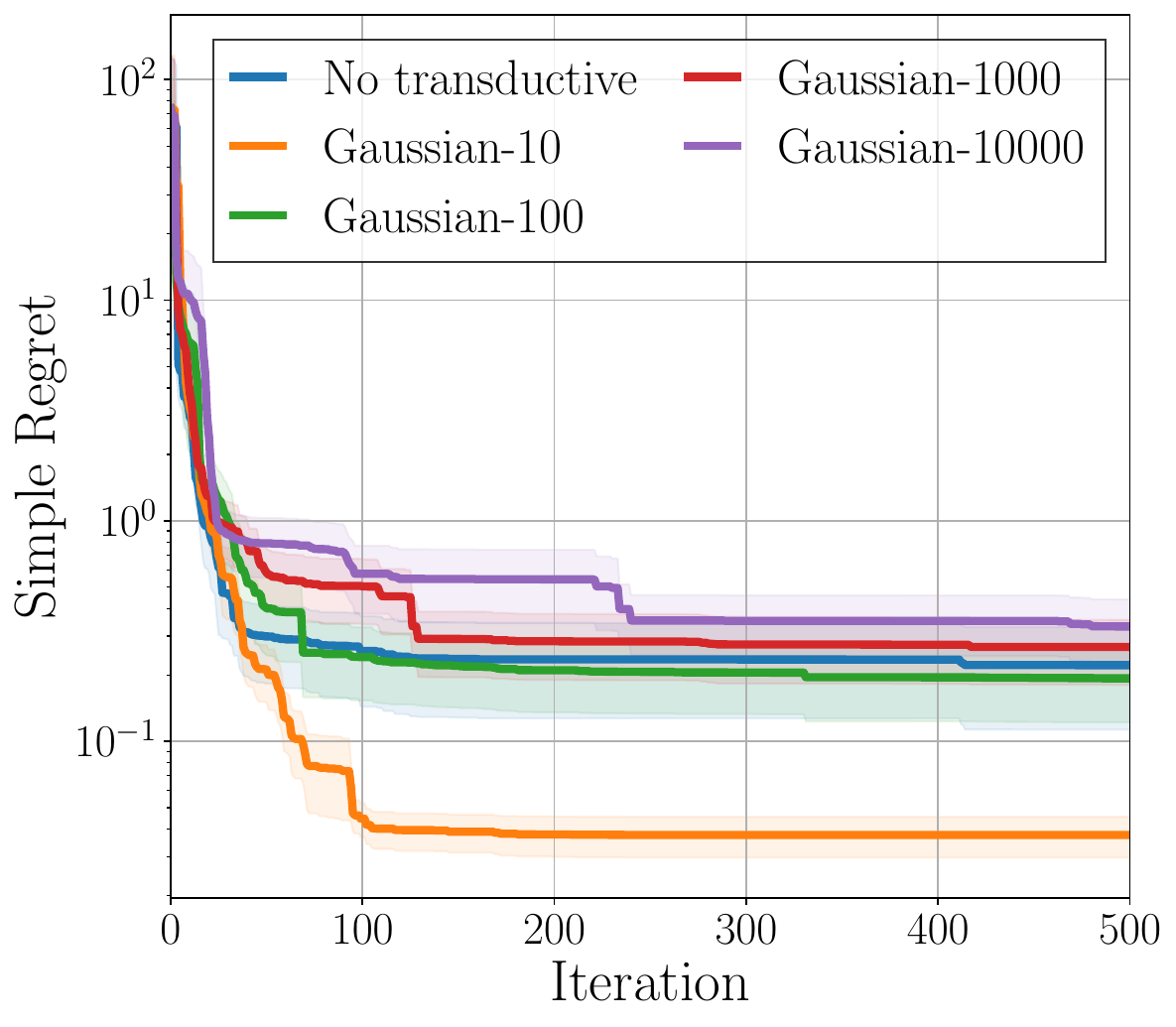}
    }
    \subfigure[Branin, LP]{
        \centering
        \includegraphics[width=0.23\textwidth]{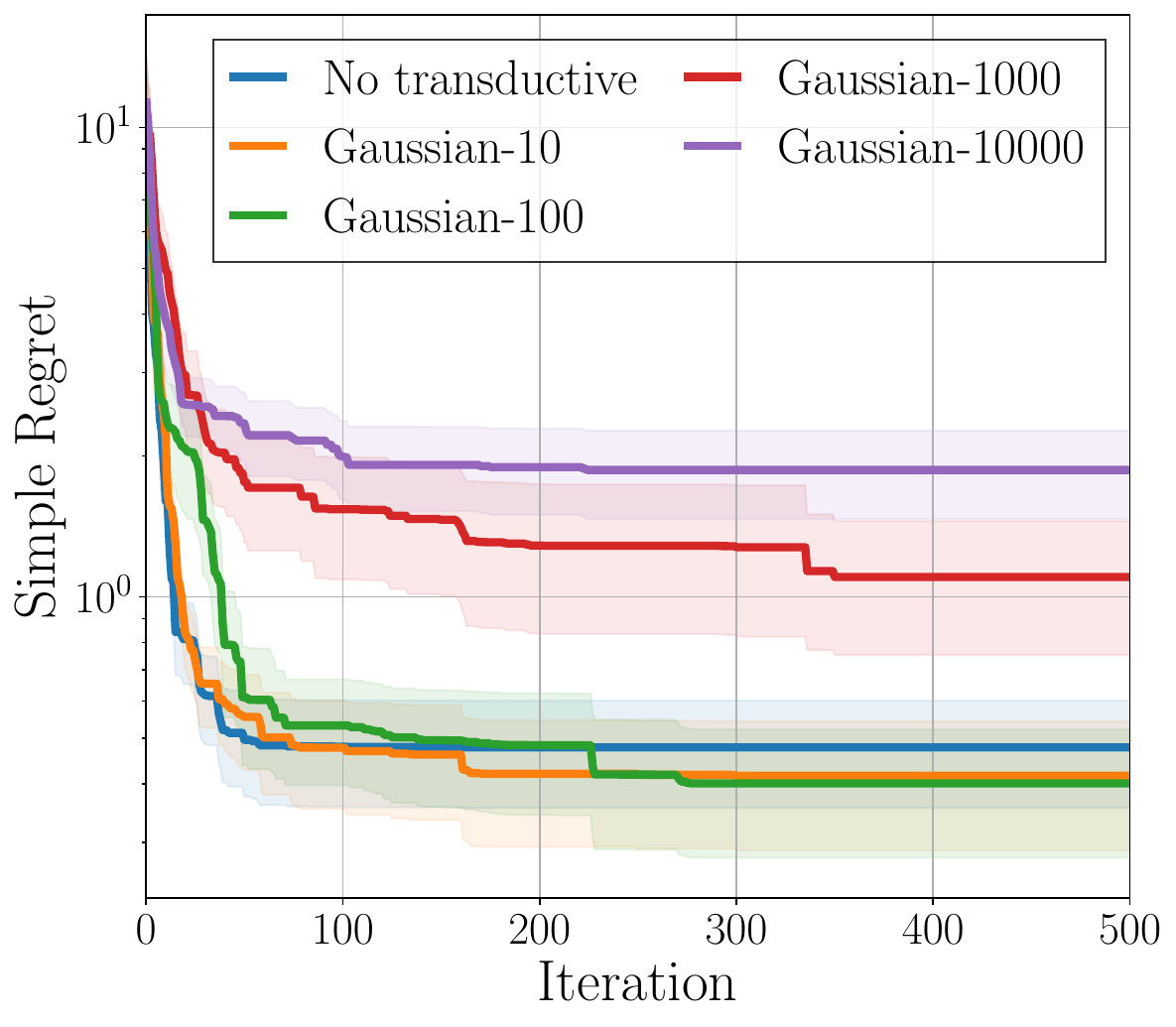}
    }
    \subfigure[Bukin6, LP]{
        \centering
        \includegraphics[width=0.23\textwidth]{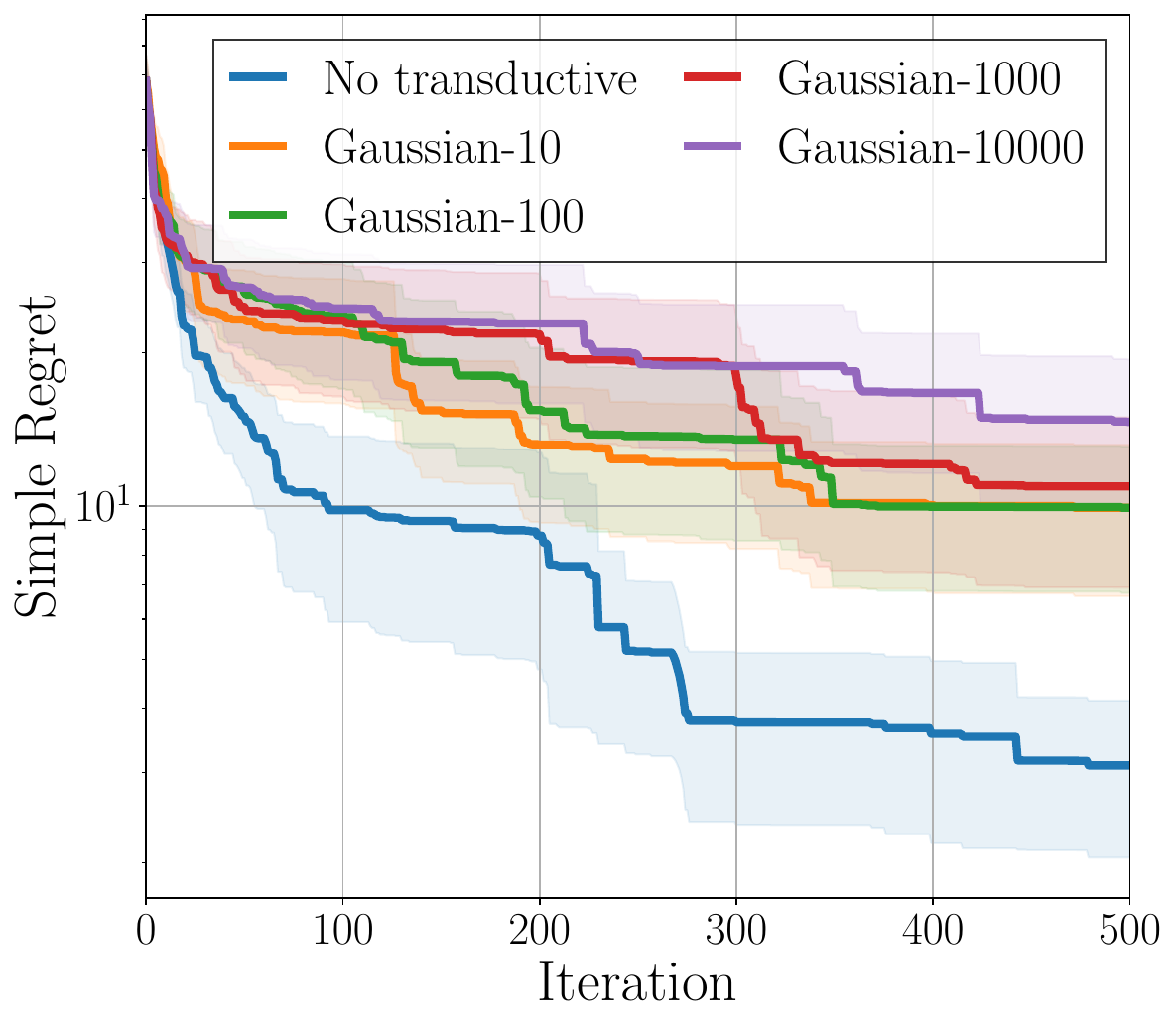}
    }
    \subfigure[Six-hump camel, LP]{
        \centering
        \includegraphics[width=0.23\textwidth]{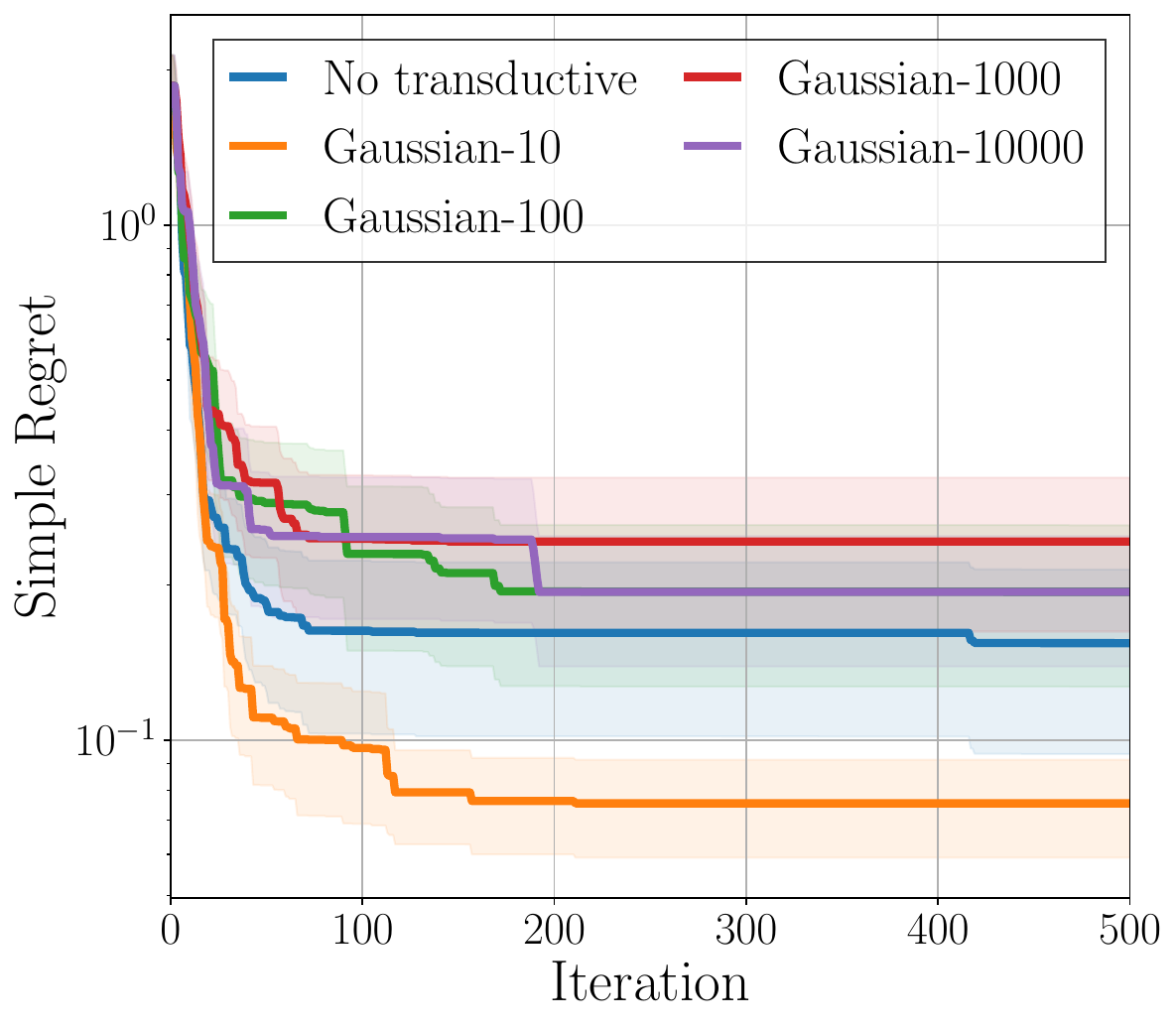}
    }
    \subfigure[Beale, LS]{
        \centering
        \includegraphics[width=0.23\textwidth]{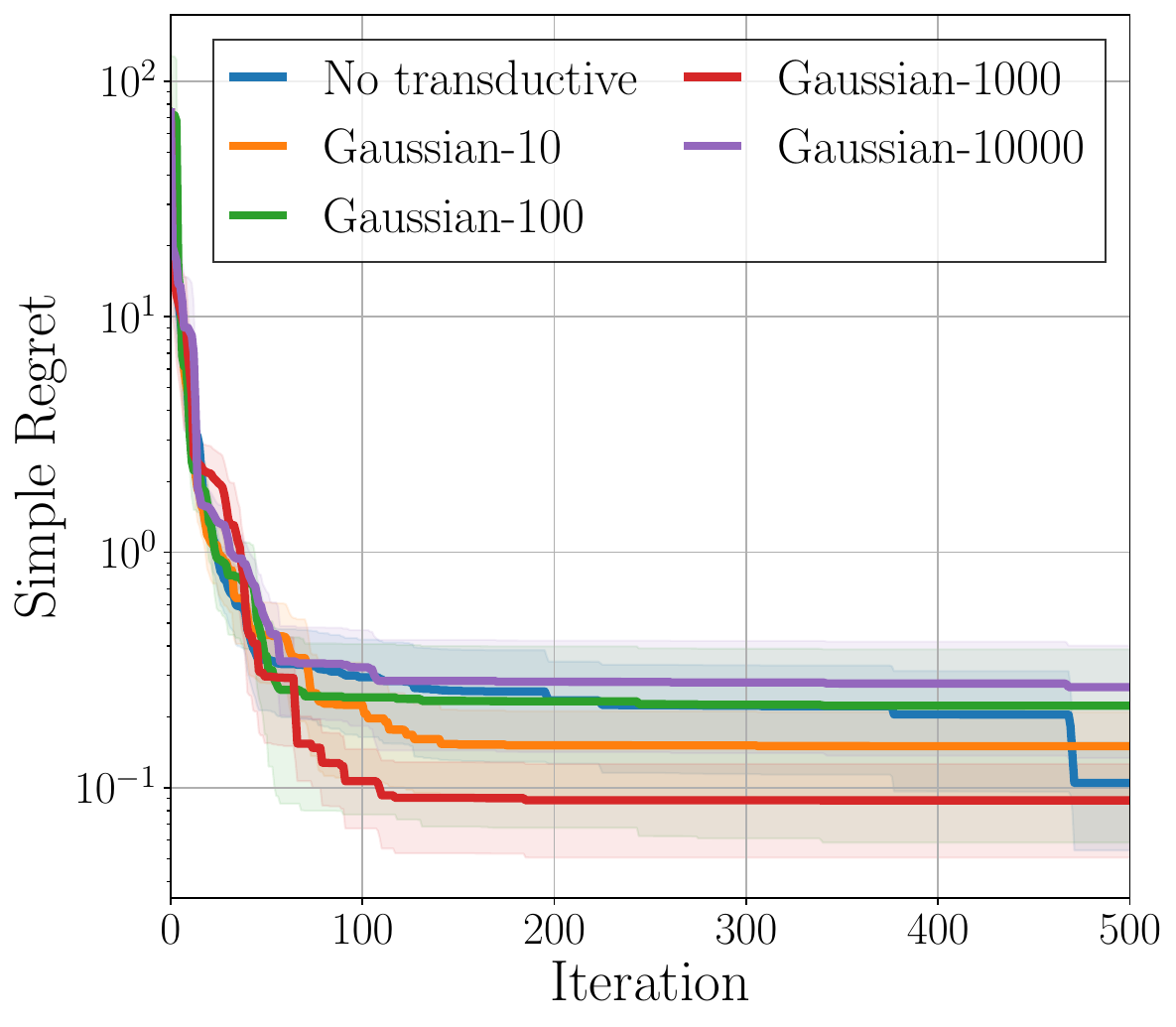}
    }
    \subfigure[Branin, LS]{
        \centering
        \includegraphics[width=0.23\textwidth]{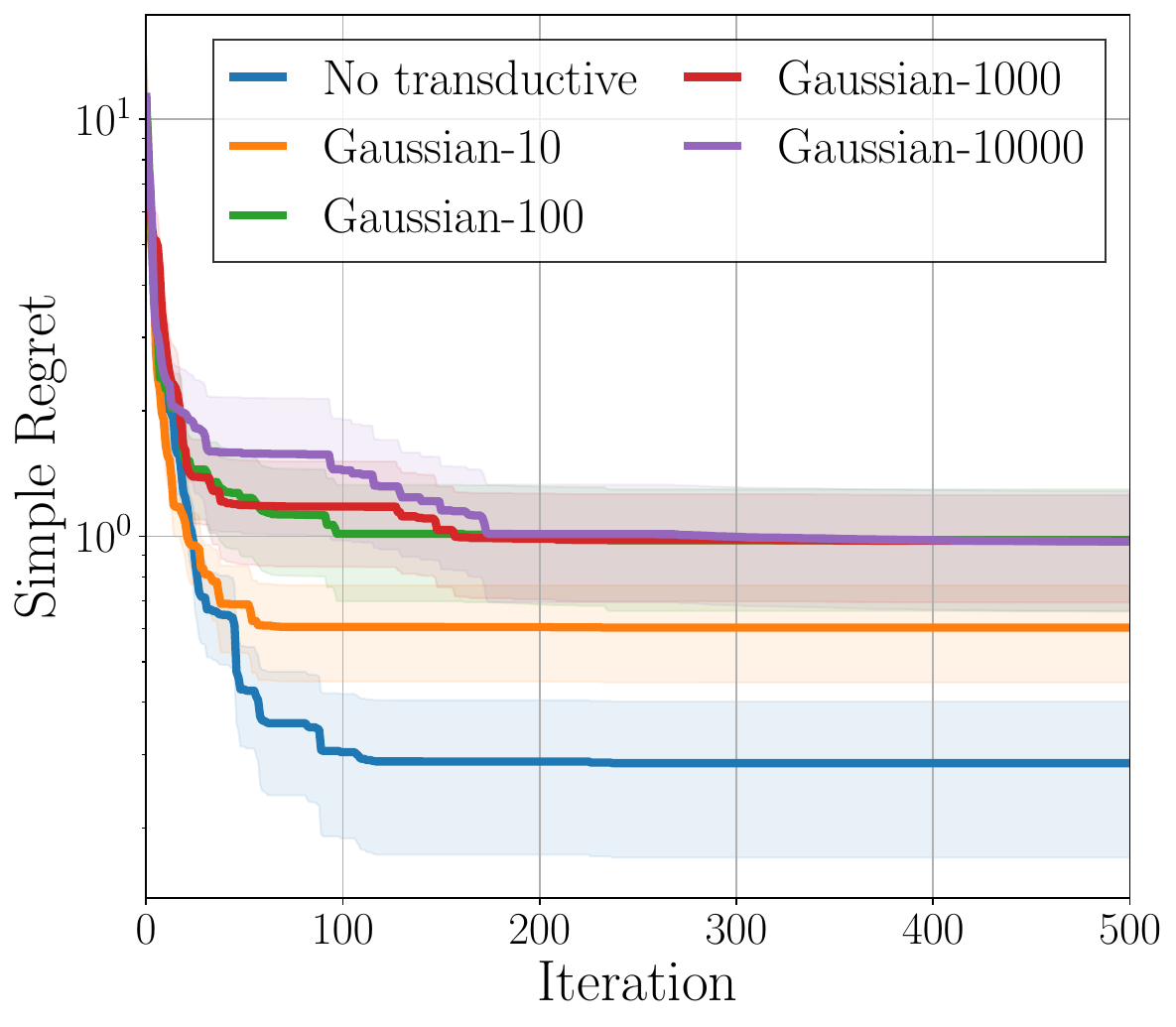}
    }
    \subfigure[Bukin6, LS]{
        \centering
        \includegraphics[width=0.23\textwidth]{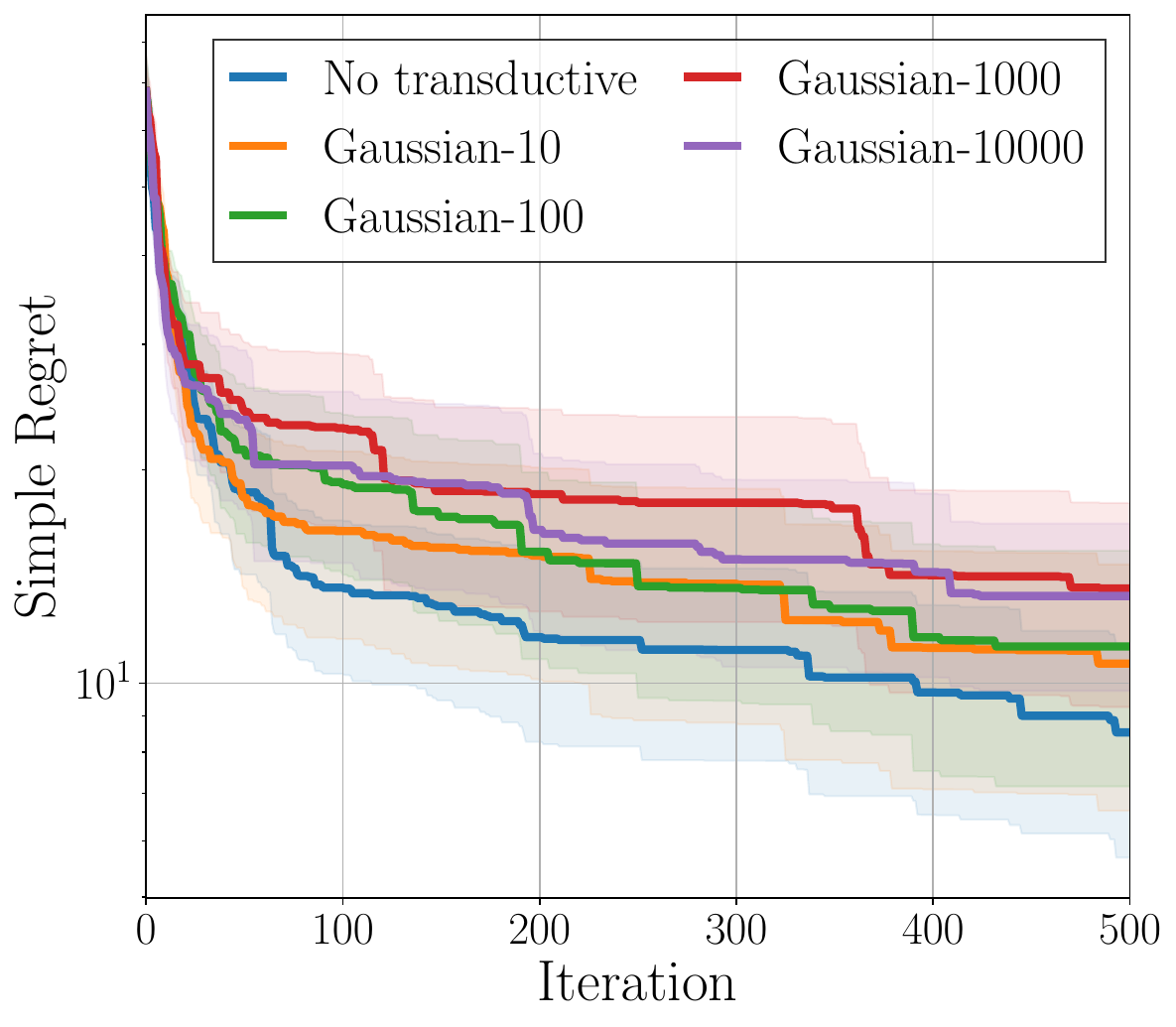}
    }
    \subfigure[Six-hump camel, LS]{
        \centering
        \includegraphics[width=0.23\textwidth]{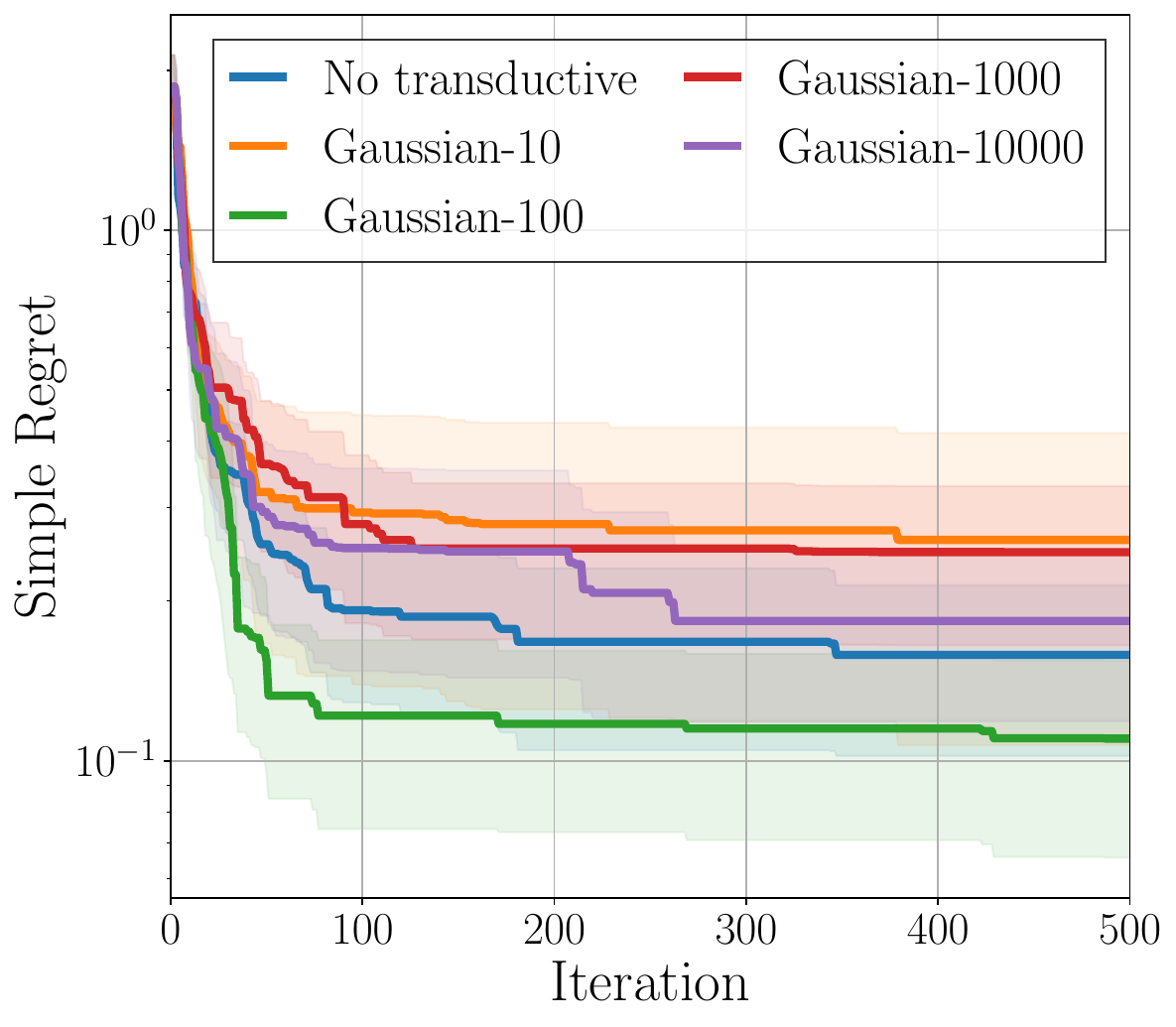}
    }
	\caption{Effects of the number of unlabeled points for unlabeled point sampling. LP and LS stand for label propagation and label spreading, respectively. We repeat all experiments 20 times.}
	\label{fig:unlabeled_points}
\end{figure}

\begin{figure}[ht!]
    \centering
    \subfigure[Beale, LP]{
        \centering
        \includegraphics[width=0.23\textwidth]{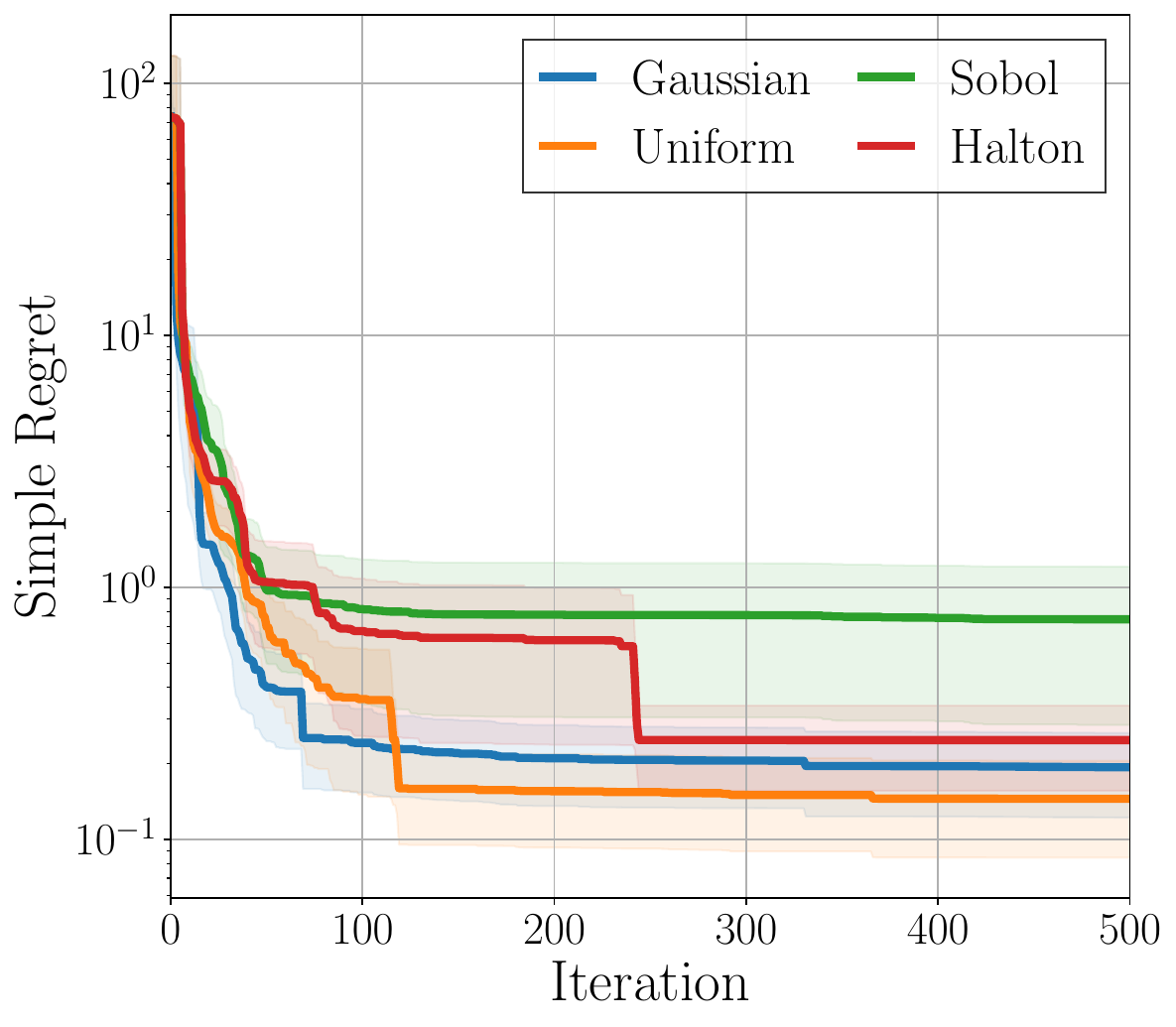}
    }
    \subfigure[Branin, LP]{
        \centering
        \includegraphics[width=0.23\textwidth]{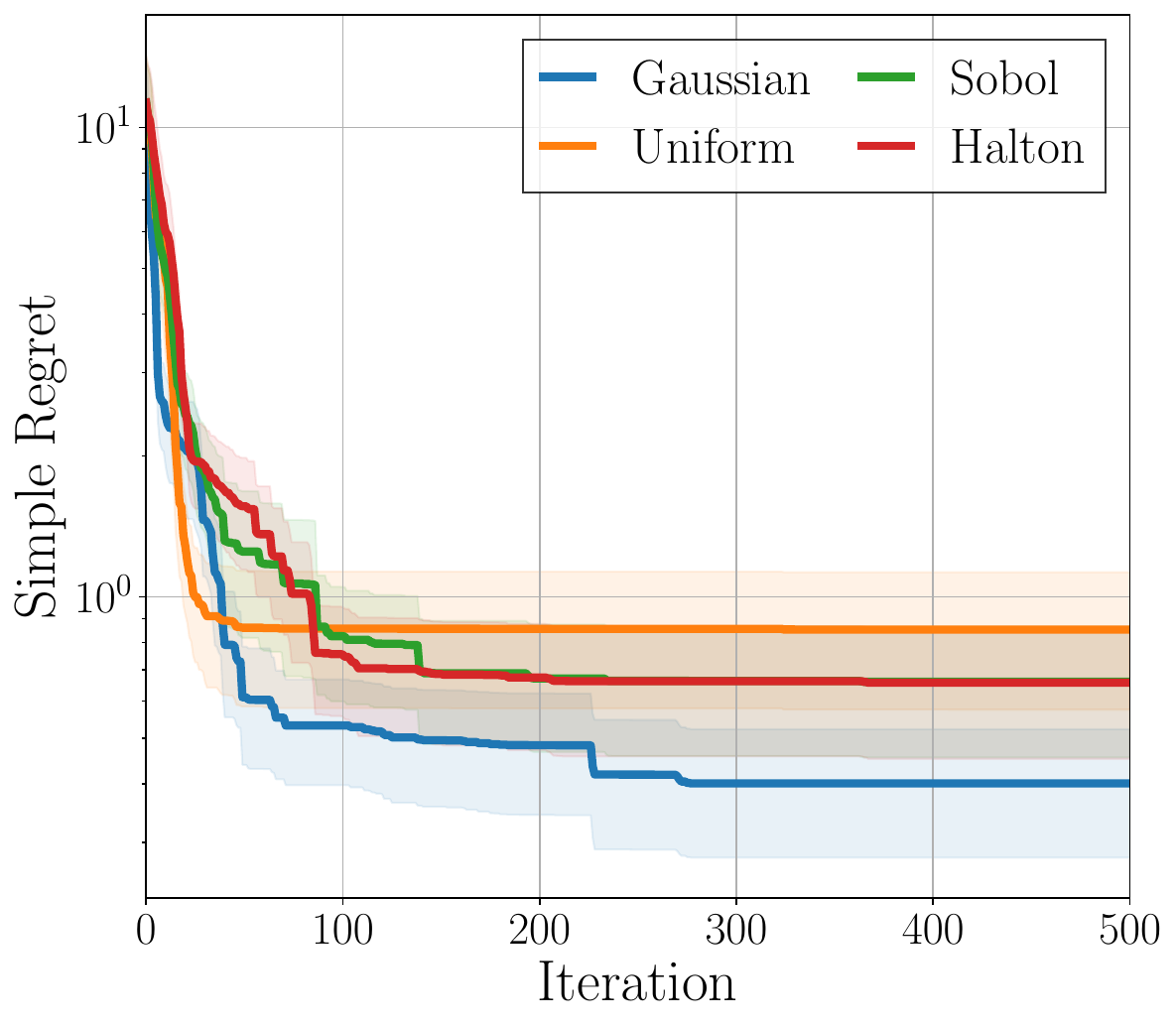}
    }
    \subfigure[Bukin6, LP]{
        \centering
        \includegraphics[width=0.23\textwidth]{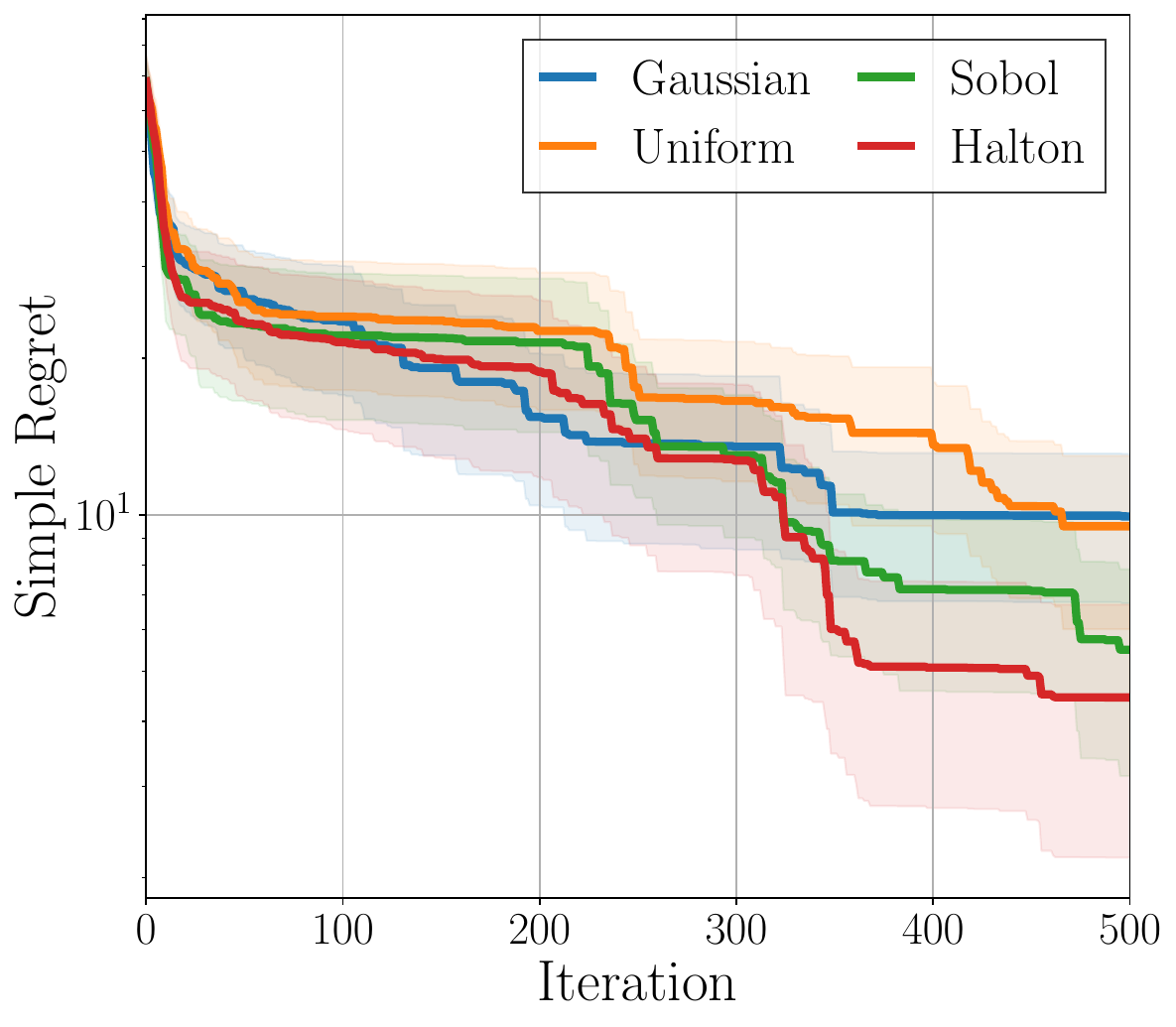}
    }
    \subfigure[Six-hump camel, LP]{
        \centering
        \includegraphics[width=0.23\textwidth]{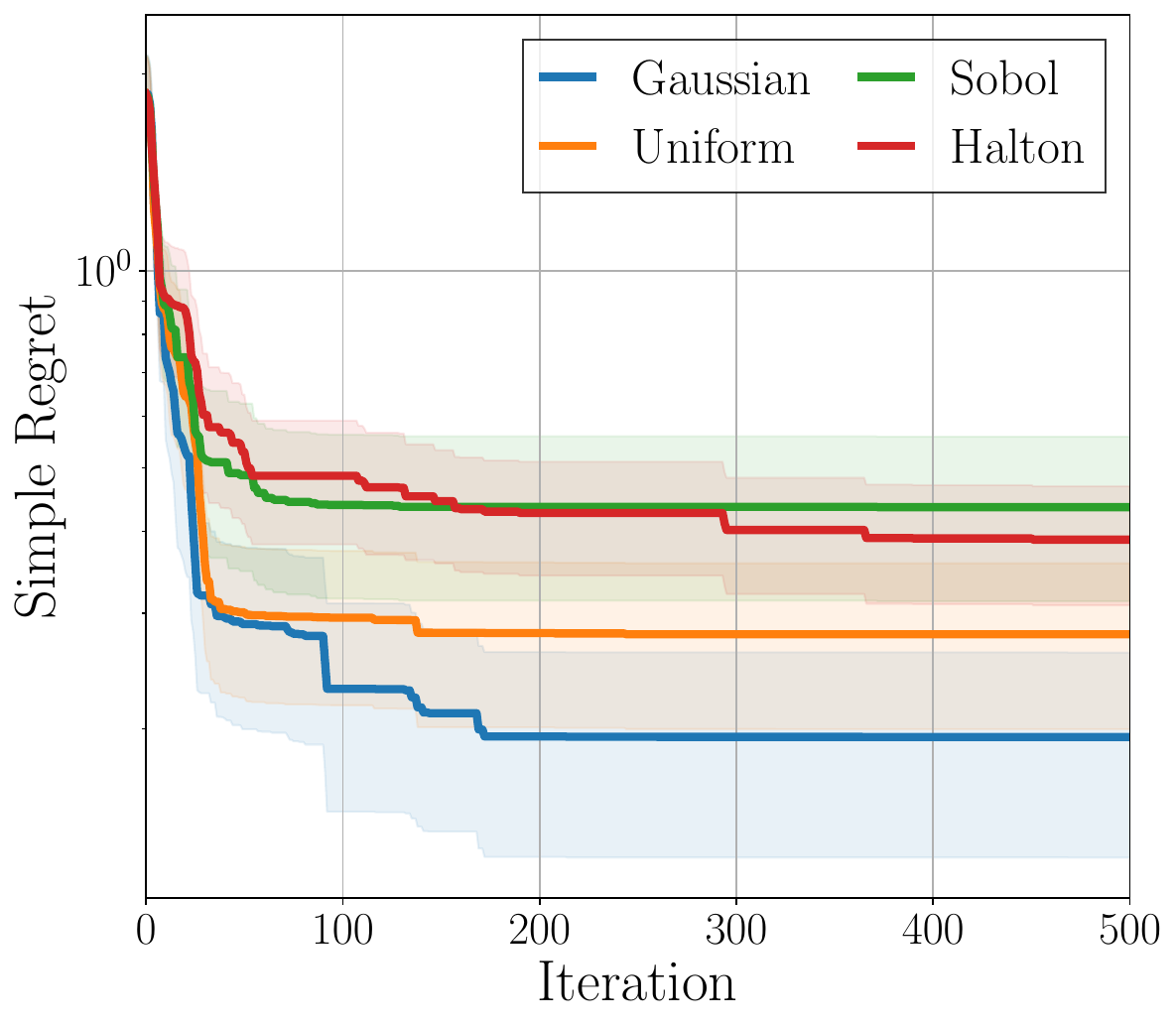}
    }
    \subfigure[Beale, LS]{
        \centering
        \includegraphics[width=0.23\textwidth]{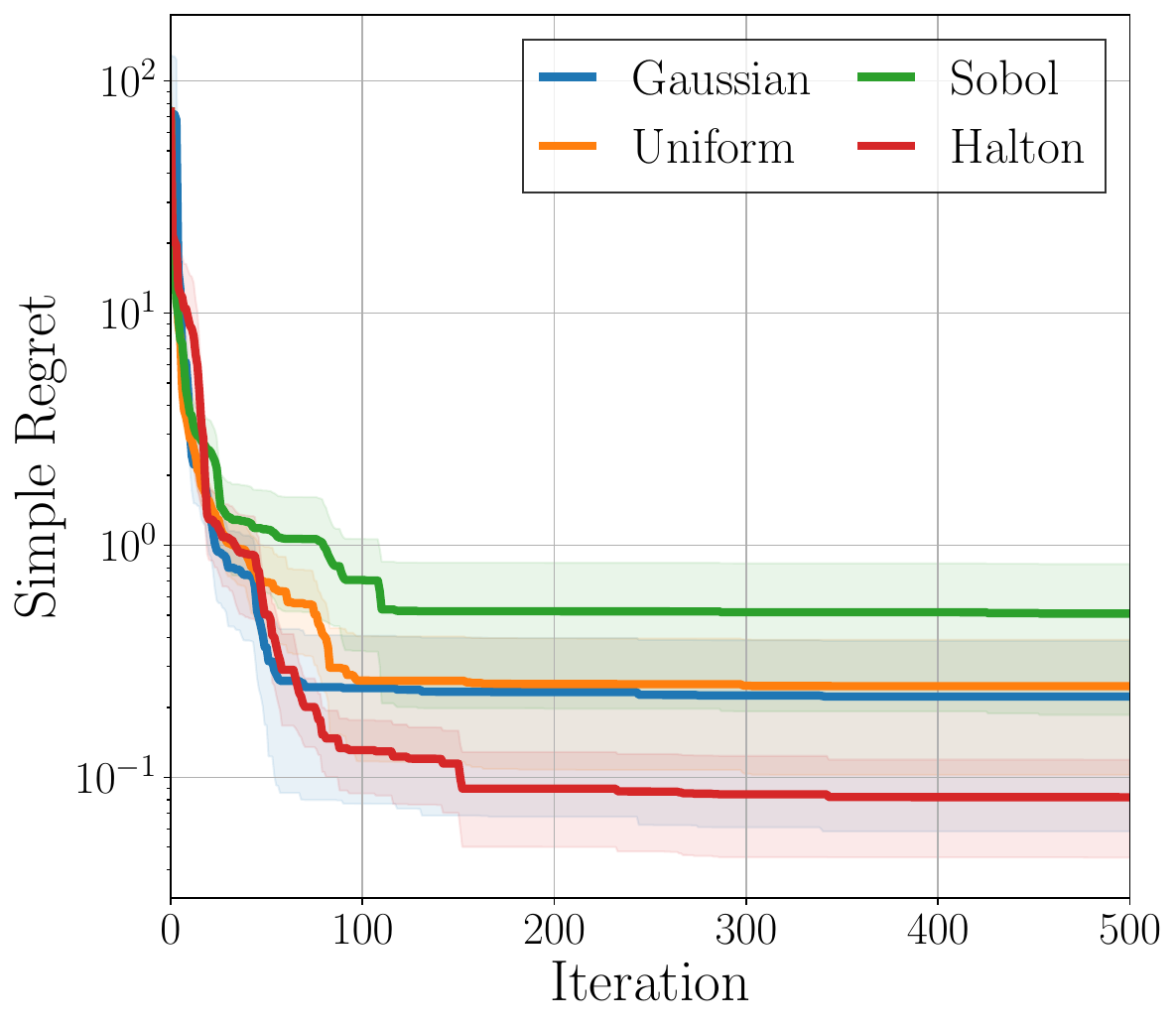}
    }
    \subfigure[Branin, LS]{
        \centering
        \includegraphics[width=0.23\textwidth]{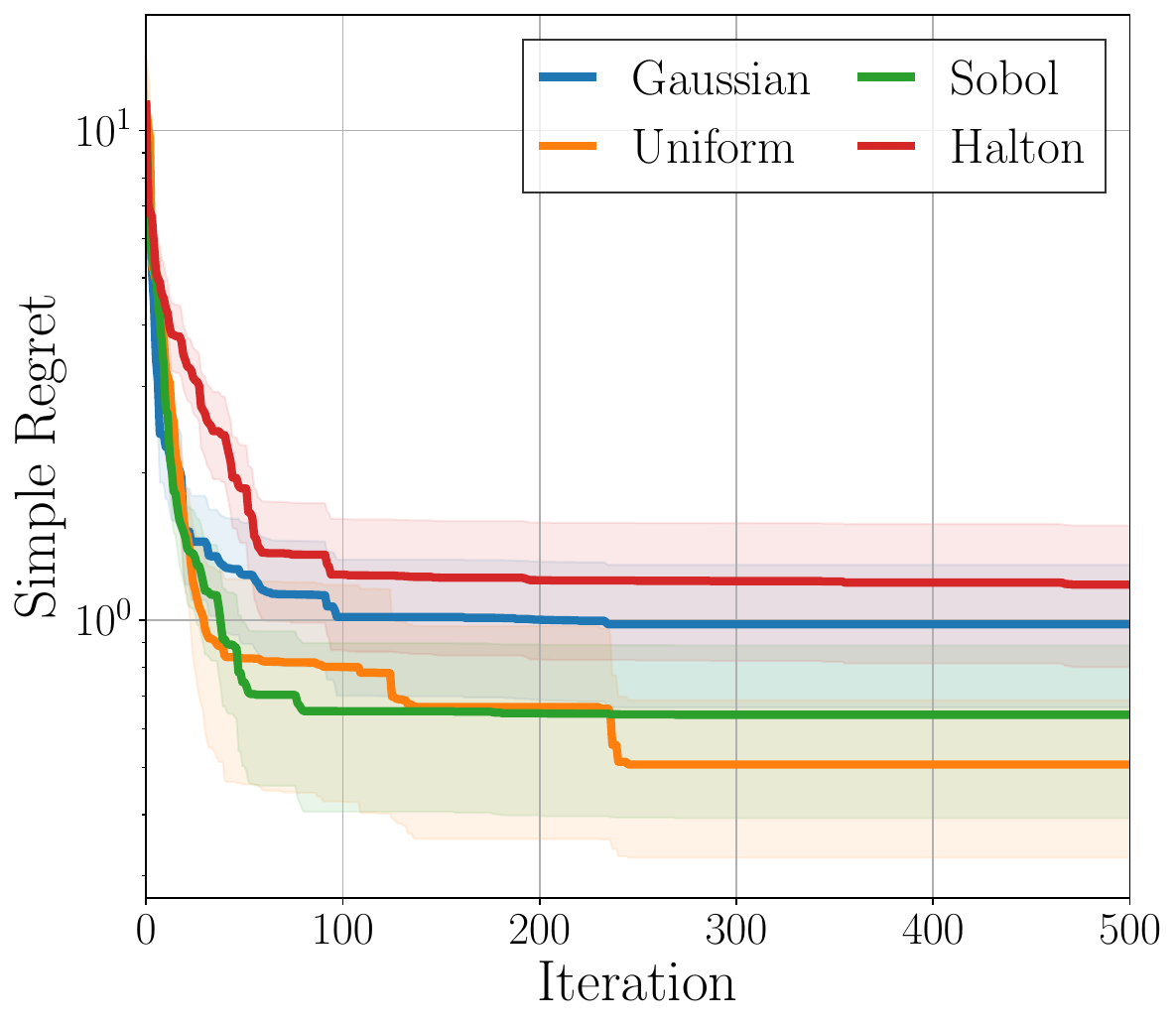}
    }
    \subfigure[Bukin6, LS]{
        \centering
        \includegraphics[width=0.23\textwidth]{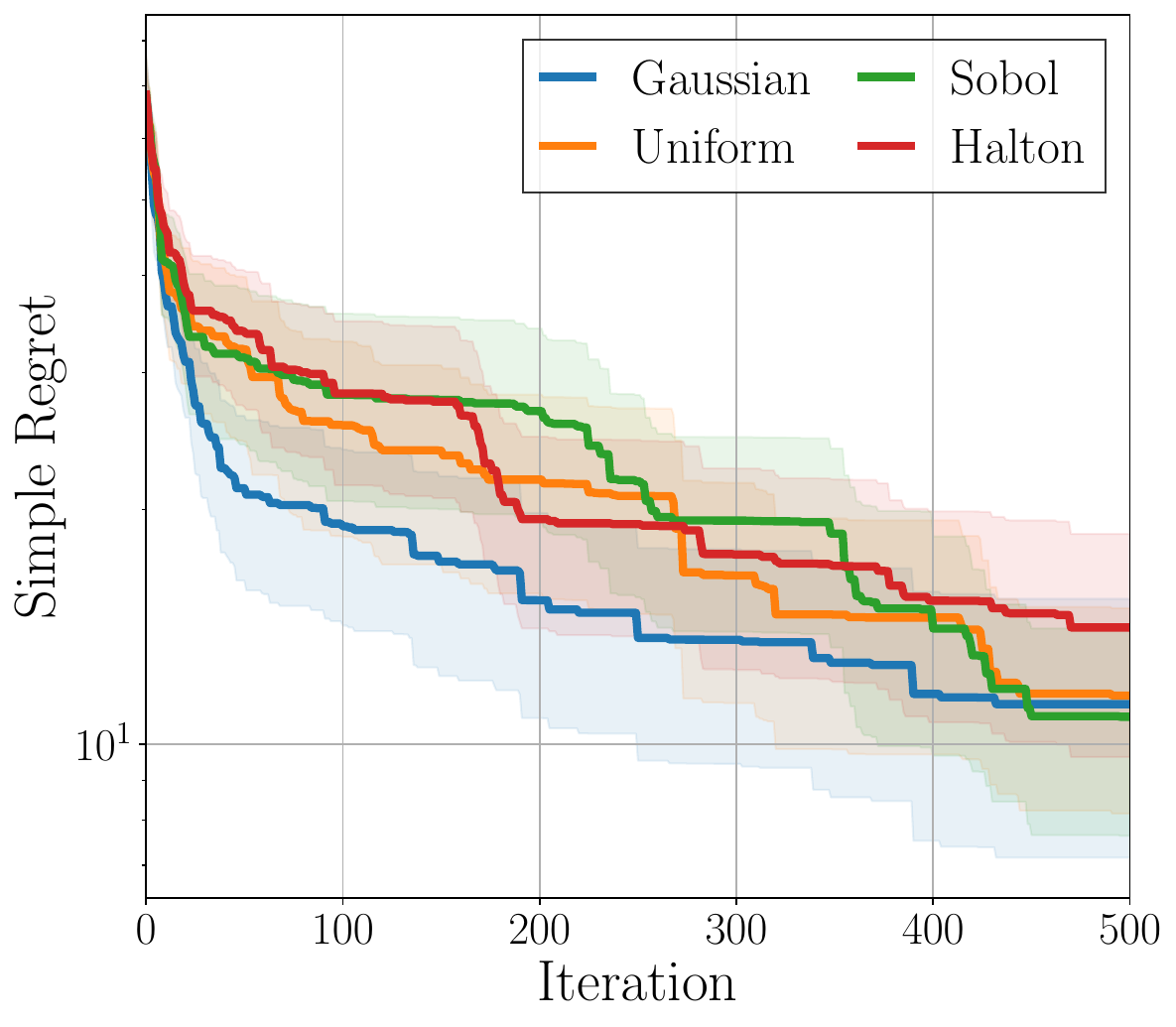}
    }
    \subfigure[Six-hump camel, LS]{
        \centering
        \includegraphics[width=0.23\textwidth]{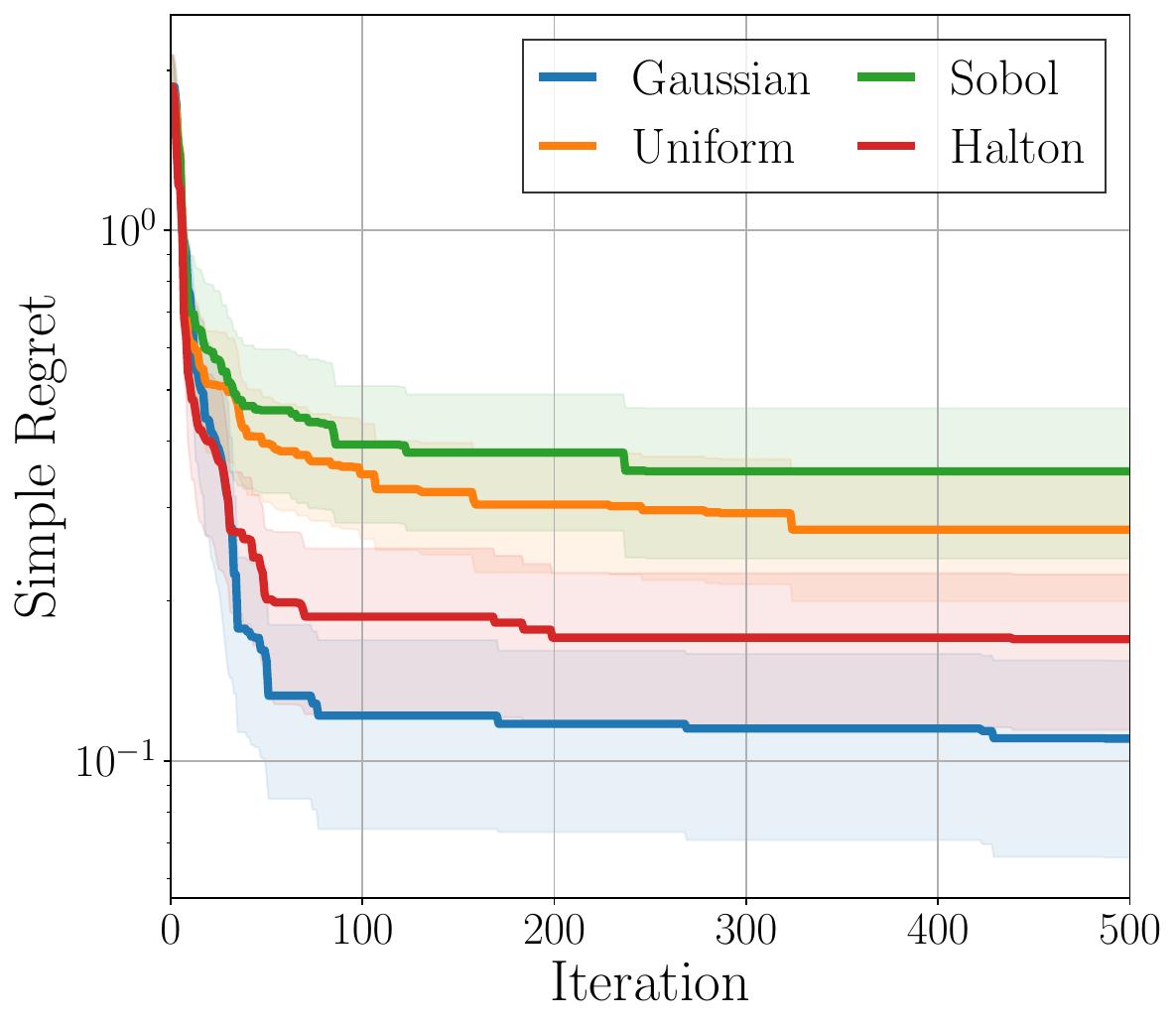}
    }
	\caption{Effects of sampling strategies for unlabeled point sampling. LP and LS stand for label propagation and label spreading, respectively. We repeat all experiments 20 times.}
	\label{fig:unlabeled_sampling}
\end{figure}

We design two studies
to analyze the effects of the number of unlabeled points $n_u$ and sampling strategies in a process of unlabeled point sampling,
where unlabeled points are not provided
and $\beta = 0.5$ is given.

For the first study,
we conduct five settings,
no unlabeled data, which implies that transductive learning is not used,
and $n_u = 10, 100, 1000, 10000$.
Interestingly,
the tendency of the number of unlabeled points
are unclear as presented in~\figref{fig:unlabeled_points}.
It implies that a setting for the number of unlabeled data points depend on the characteristics of benchmark functions,
which is common in Bayesian optimization and black-box optimization.
Besides,
$\gamma$ is different across benchmarks and iterations
and it lets optimization results sensitive to $n_u$.
Therefore, we cannot determine a suitable setting without access to a black-box function of interest.

As another elaborate study,
we test the effects of sampling strategies.
Four strategies,
the truncated multivariate normal distributions,
uniform distributions,
Halton sequences~\citep{HaltonJH1960nm},
and Sobol' sequences~\citep{SobolIM1967russian},
are compared.
As depicted in~\figref{fig:unlabeled_sampling},
the normal distribution is better than
the other sampling methods in four cases
and shows robust performance in most of the cases,
but it is not always the best.
Similar to the previous study on the effects of $n_u$,
we presume that it is also affected by $\gamma$,
which is hard to define in practice.

\clearpage

\section{Discussion on Pool Sampling}
\label{sec:discussion_pool_sampling}

\begin{figure}[ht]
    \centering
    \subfigure[Beale, LP]{
        \centering
        \includegraphics[width=0.23\textwidth]{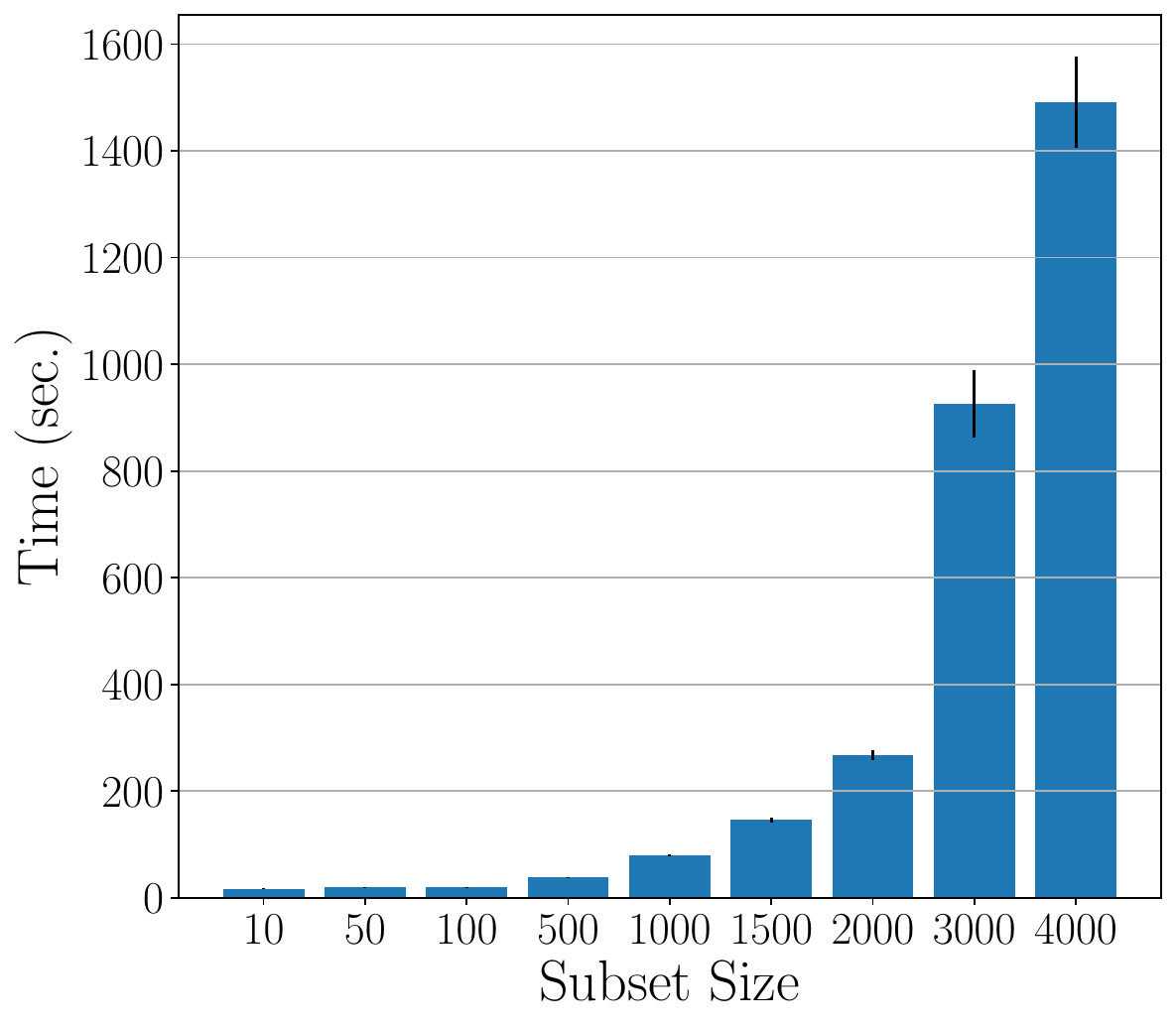}
    }
    \subfigure[Branin, LP]{
        \centering
        \includegraphics[width=0.23\textwidth]{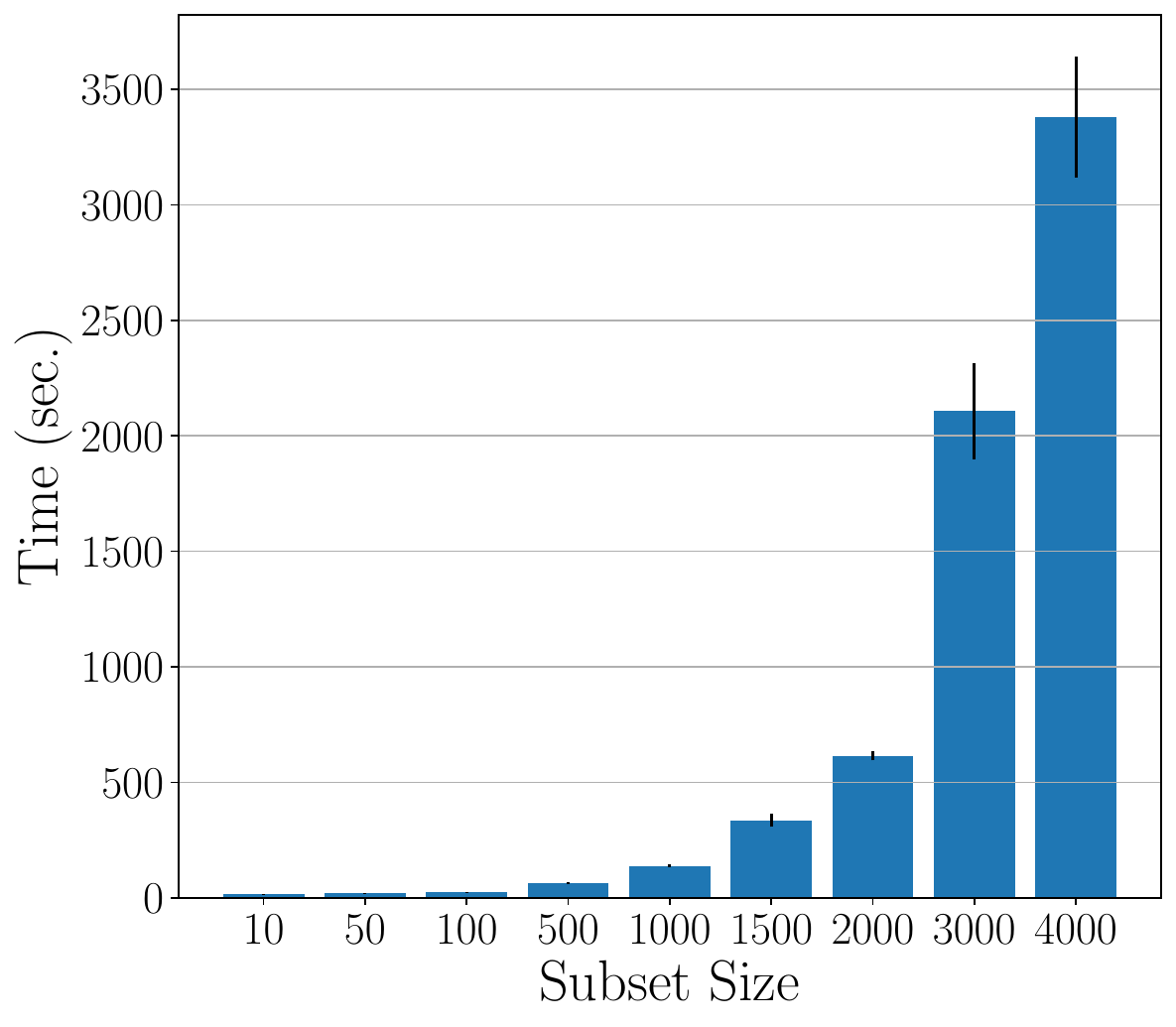}
    }
    \subfigure[Bukin6, LP]{
        \centering
        \includegraphics[width=0.23\textwidth]{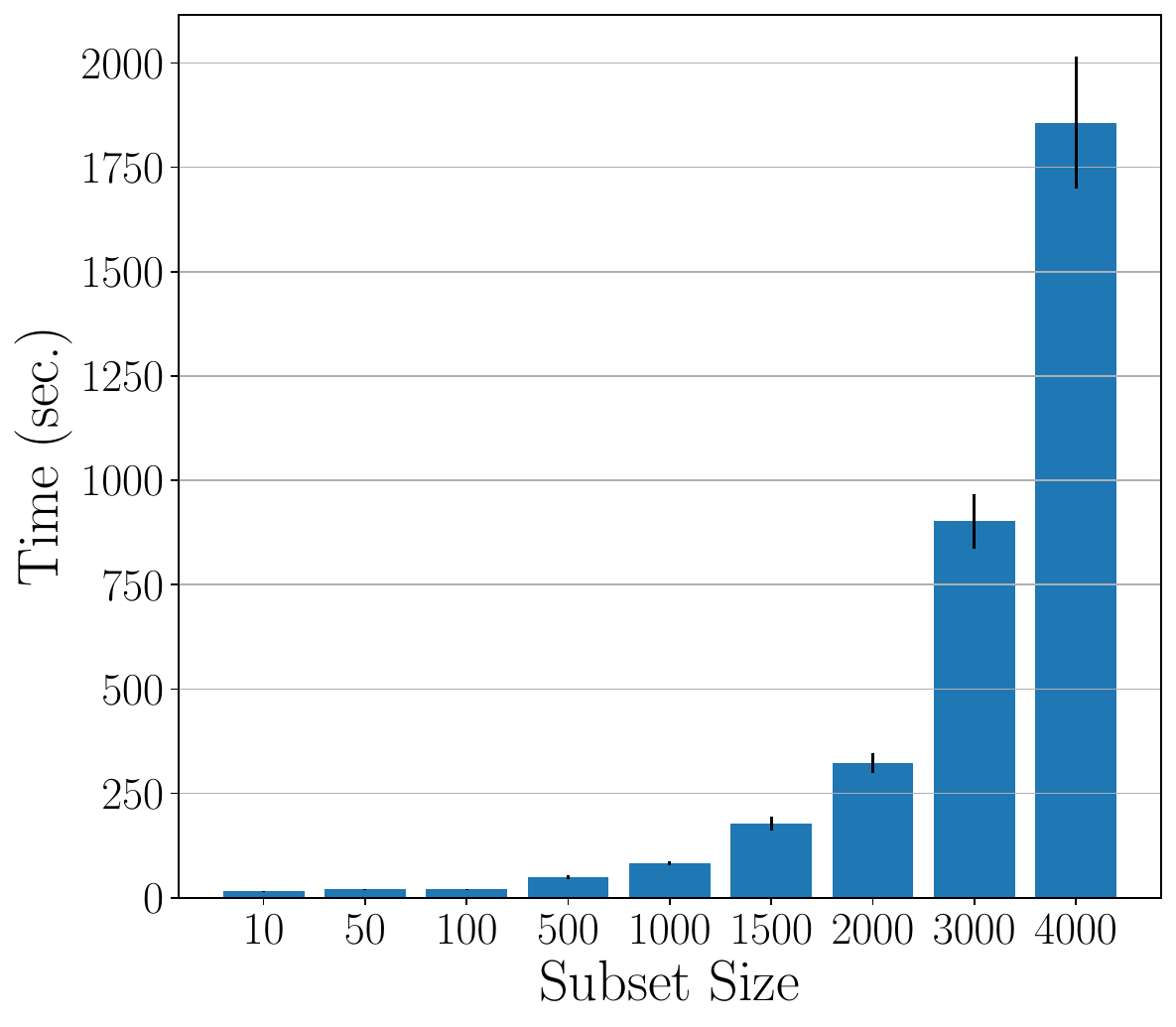}
    }
    \subfigure[Six-hump camel, LP]{
        \centering
        \includegraphics[width=0.23\textwidth]{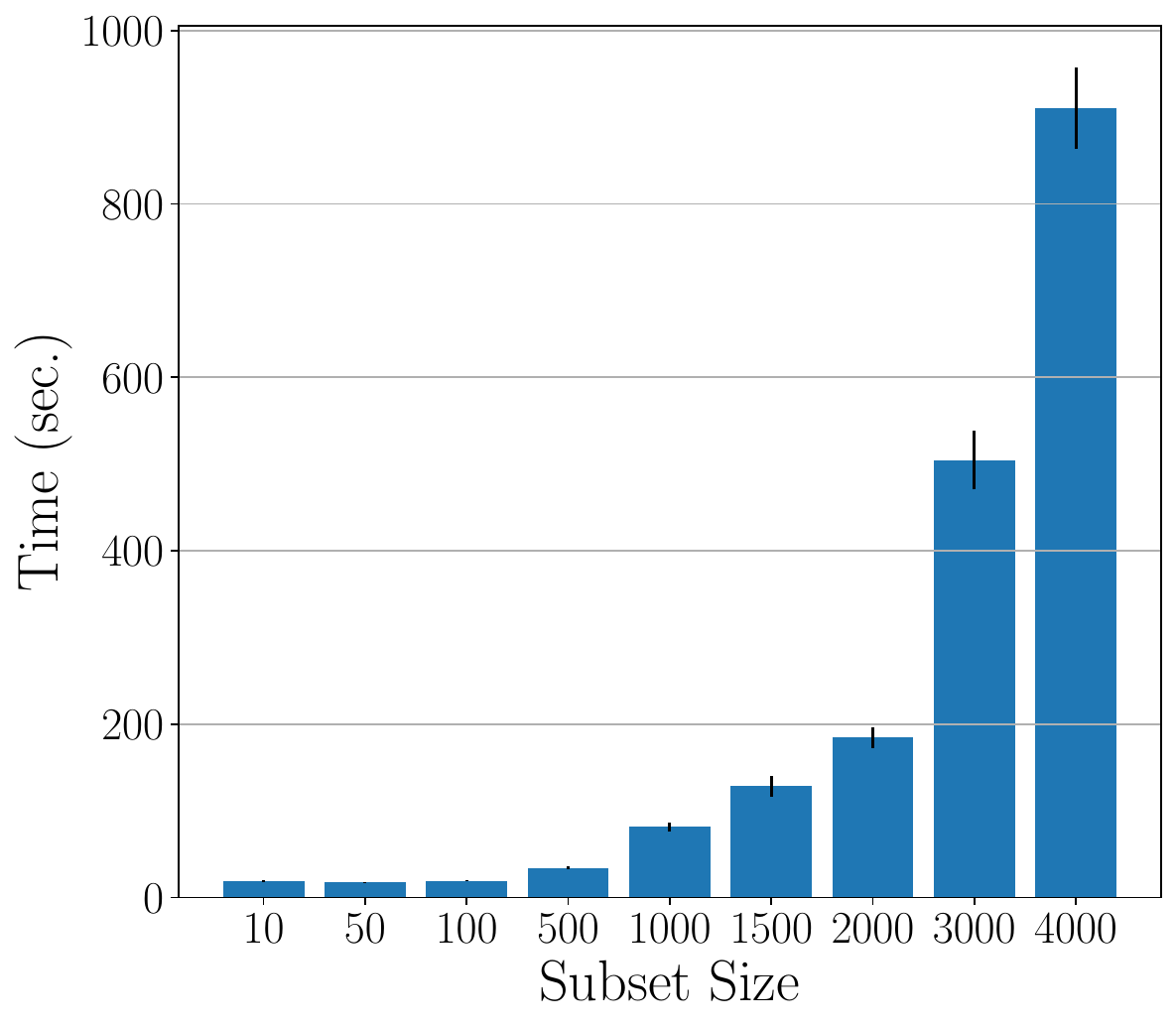}
    }
    \subfigure[Beale, LS]{
        \centering
        \includegraphics[width=0.23\textwidth]{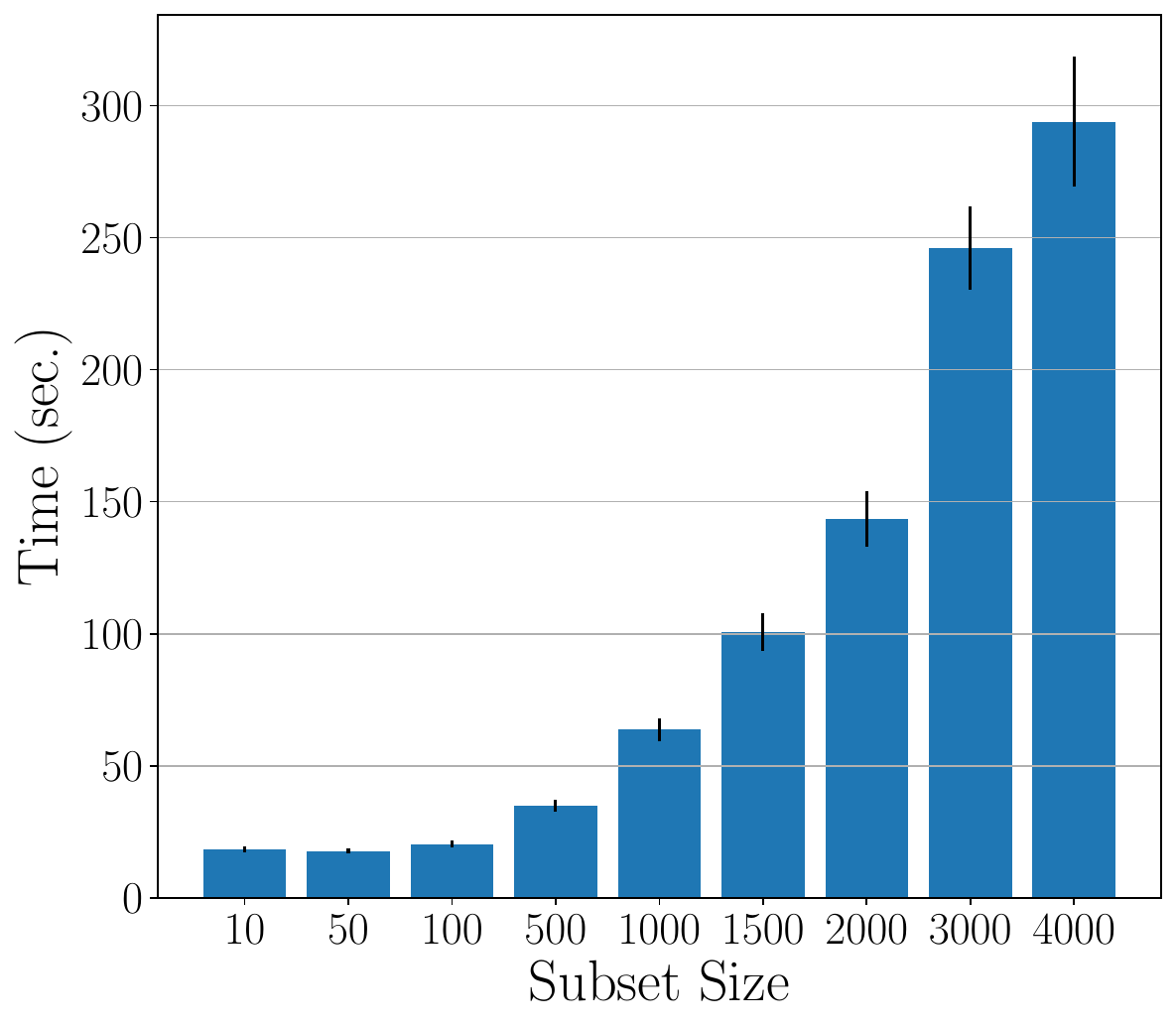}
    }
    \subfigure[Branin, LS]{
        \centering
        \includegraphics[width=0.23\textwidth]{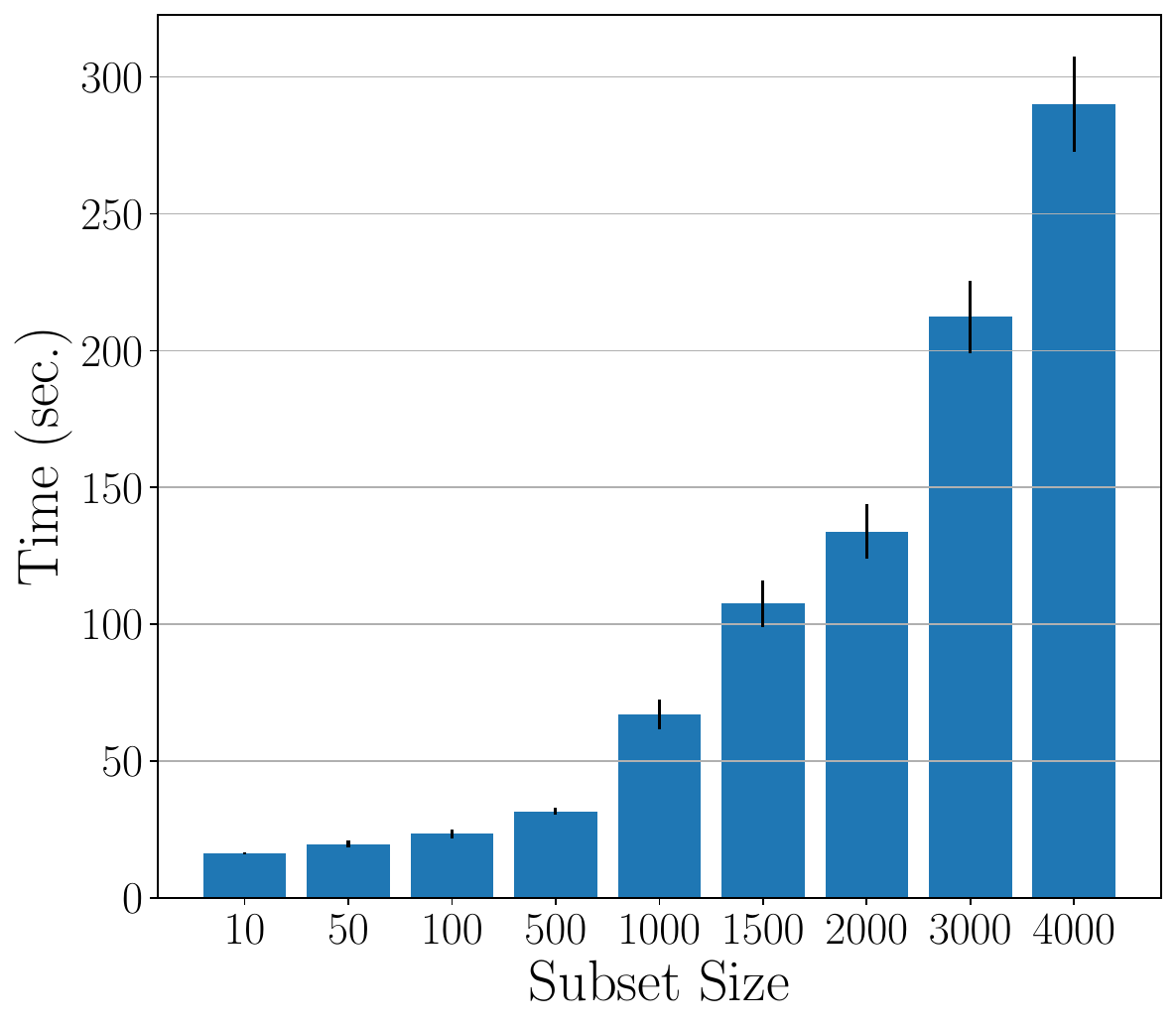}
    }
    \subfigure[Bukin6, LS]{
        \centering
        \includegraphics[width=0.23\textwidth]{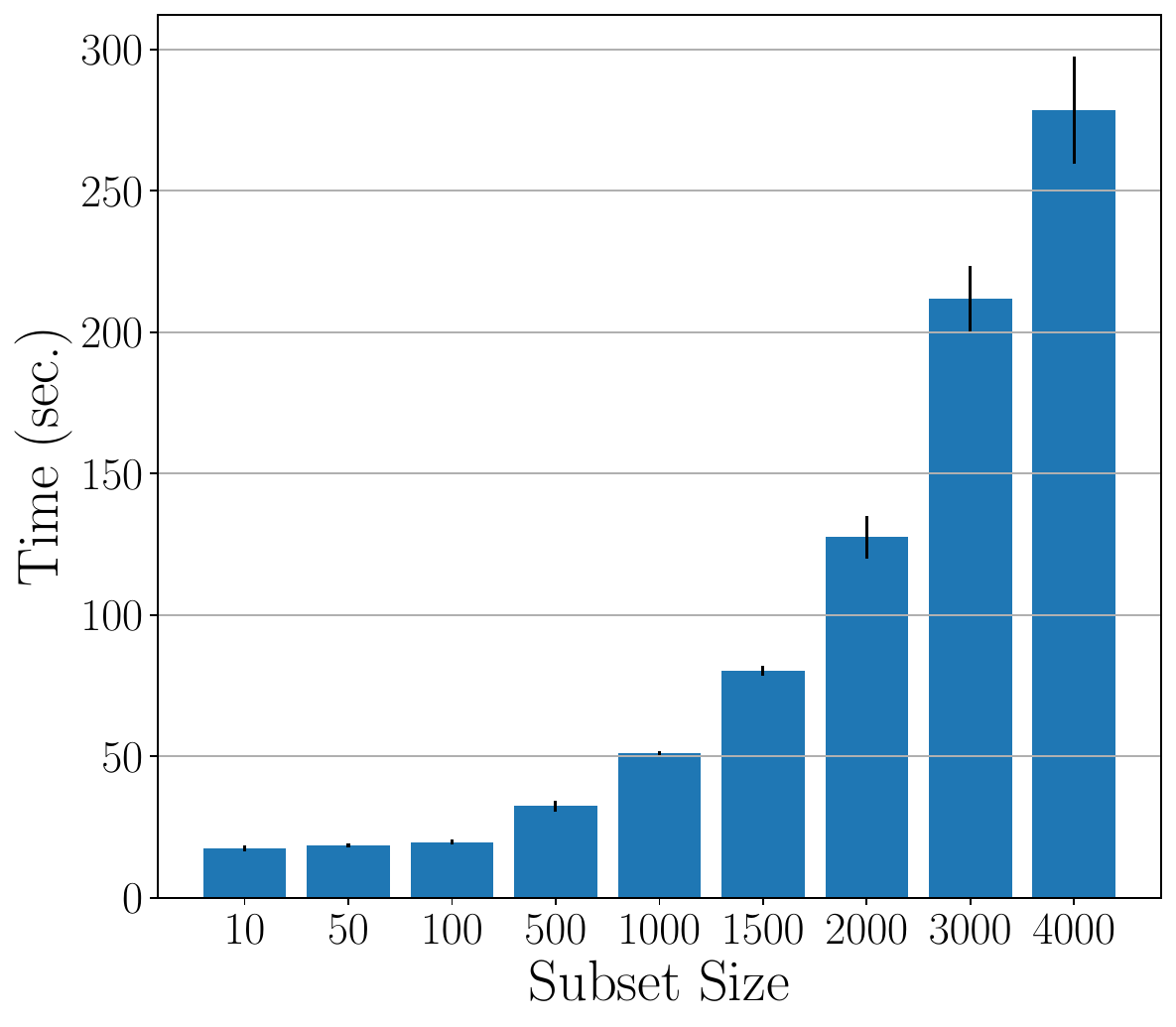}
    }
    \subfigure[Six-hump camel, LS]{
        \centering
        \includegraphics[width=0.23\textwidth]{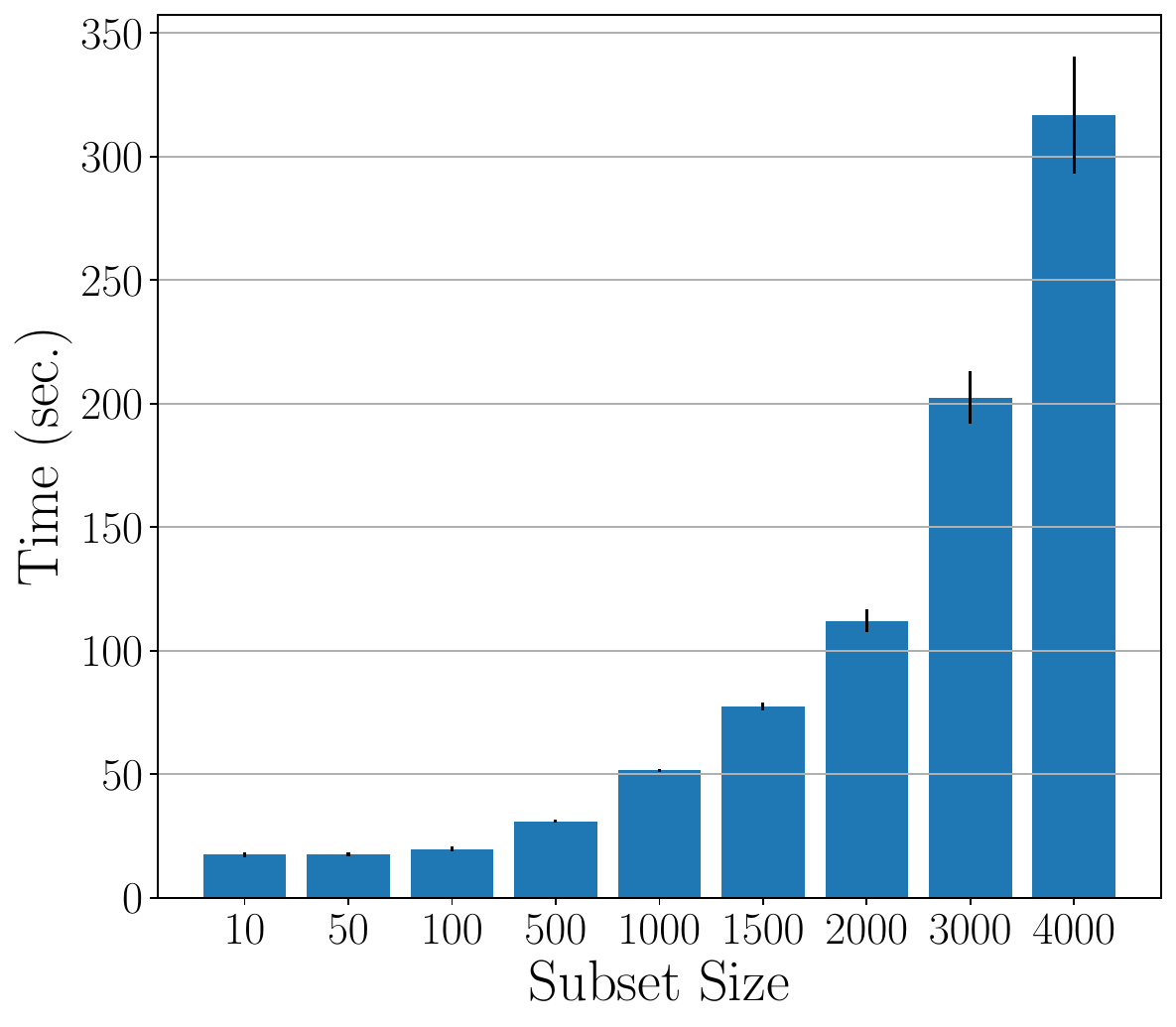}
    }
	\caption{Results with 20 repeated experiments on elapsed times varying subset sizes via pool sampling. LP and LS stand for label propagation and label spreading, respectively.}
	\label{fig:discussion_pool_sampling_time}
\end{figure}

\begin{figure}[ht!]
    \centering
    \subfigure[Beale, LP]{
        \centering
        \includegraphics[width=0.23\textwidth]{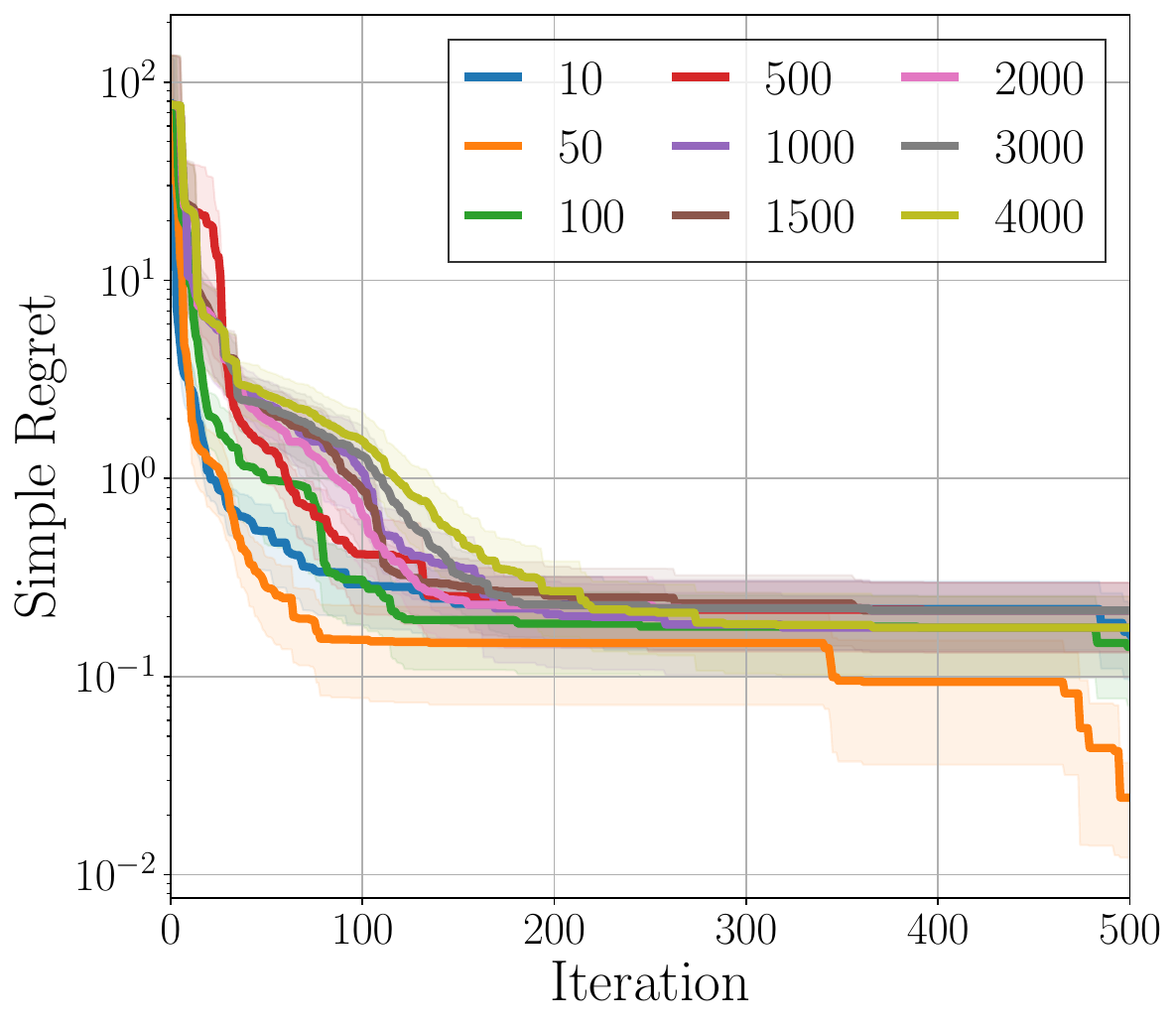}
    }
    \subfigure[Branin, LP]{
        \centering
        \includegraphics[width=0.23\textwidth]{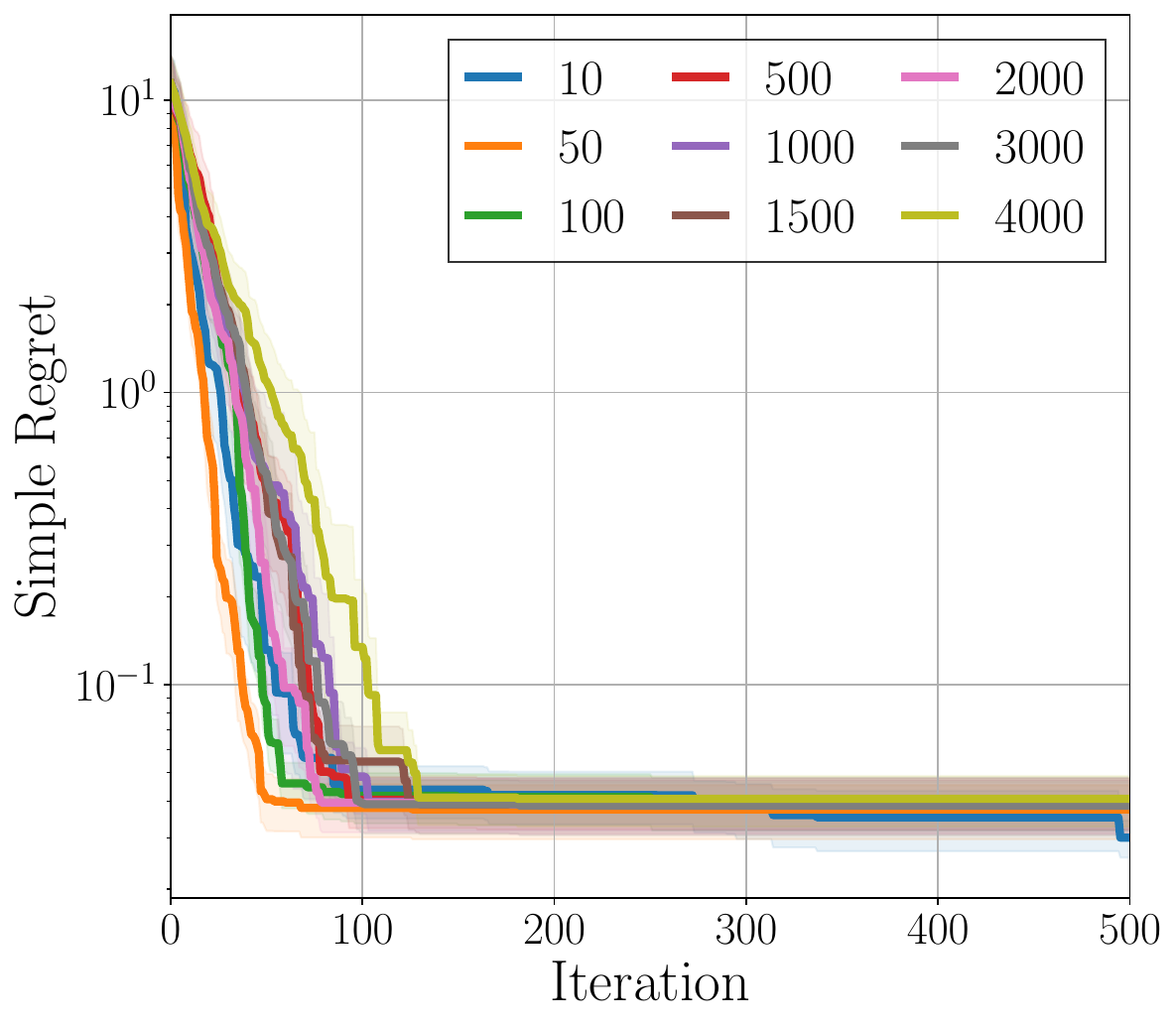}
    }
    \subfigure[Bukin6, LP]{
        \centering
        \includegraphics[width=0.23\textwidth]{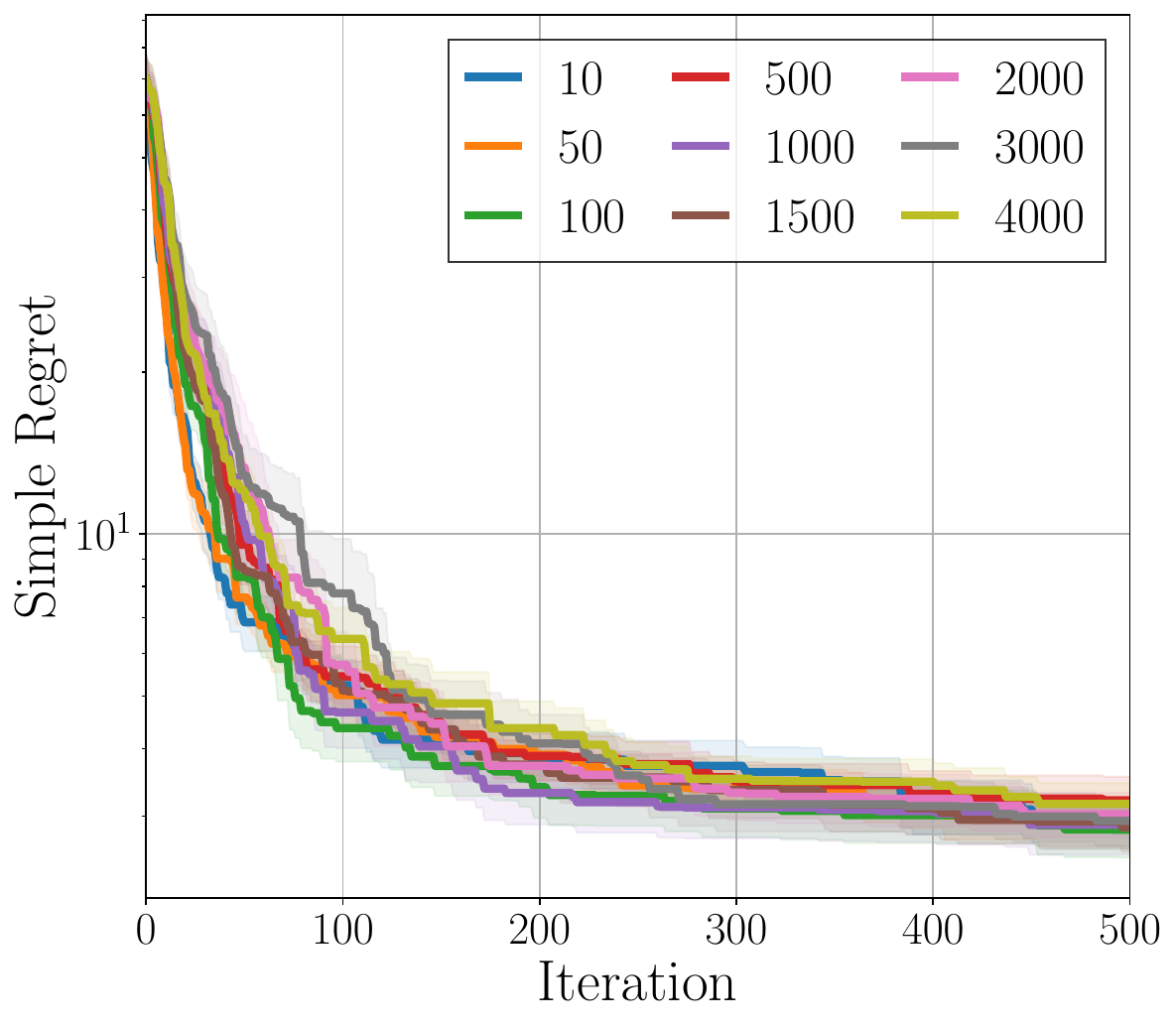}
    }
    \subfigure[Six-hump camel, LP]{
        \centering
        \includegraphics[width=0.23\textwidth]{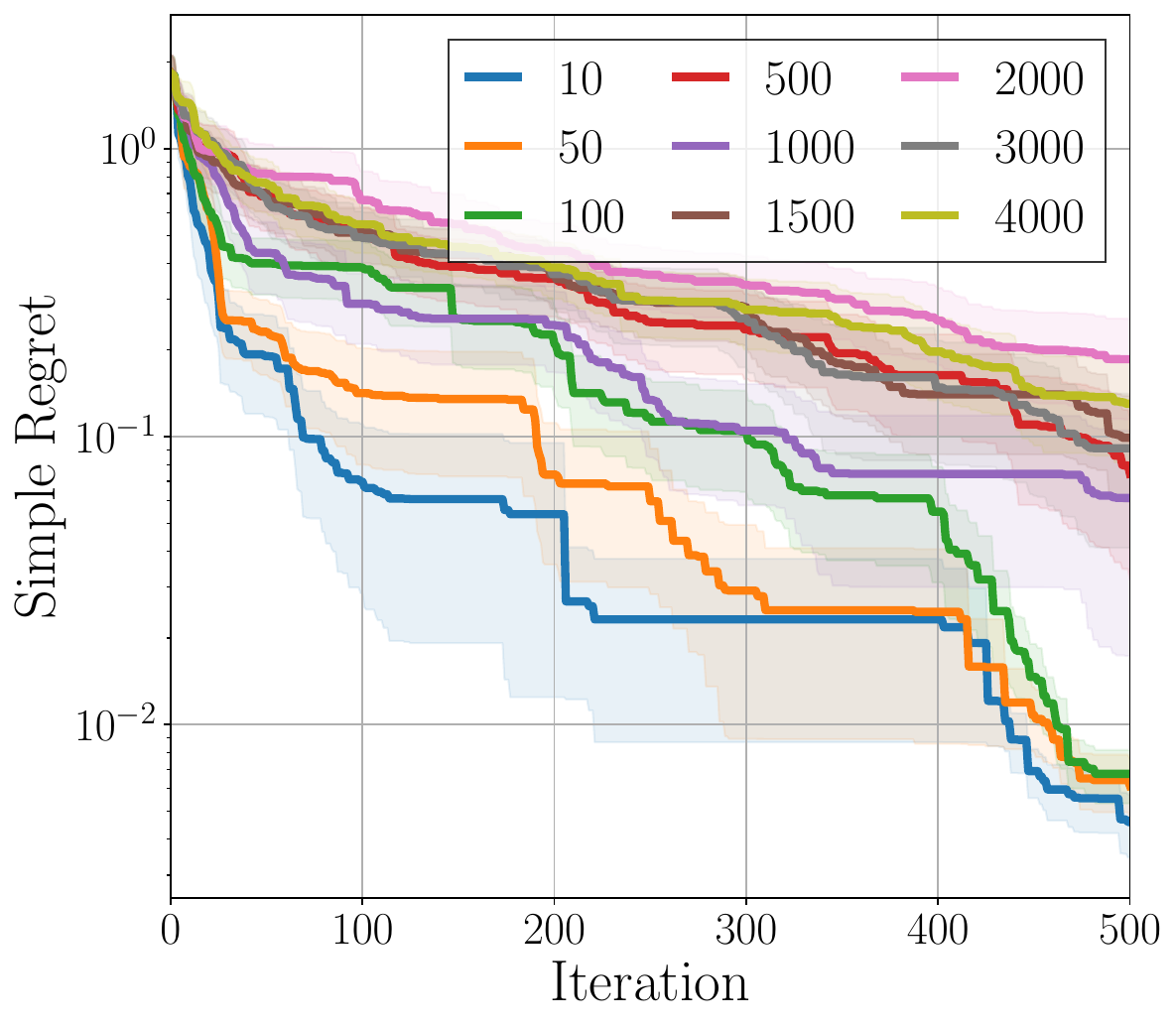}
    }
    \subfigure[Beale, LS]{
        \centering
        \includegraphics[width=0.23\textwidth]{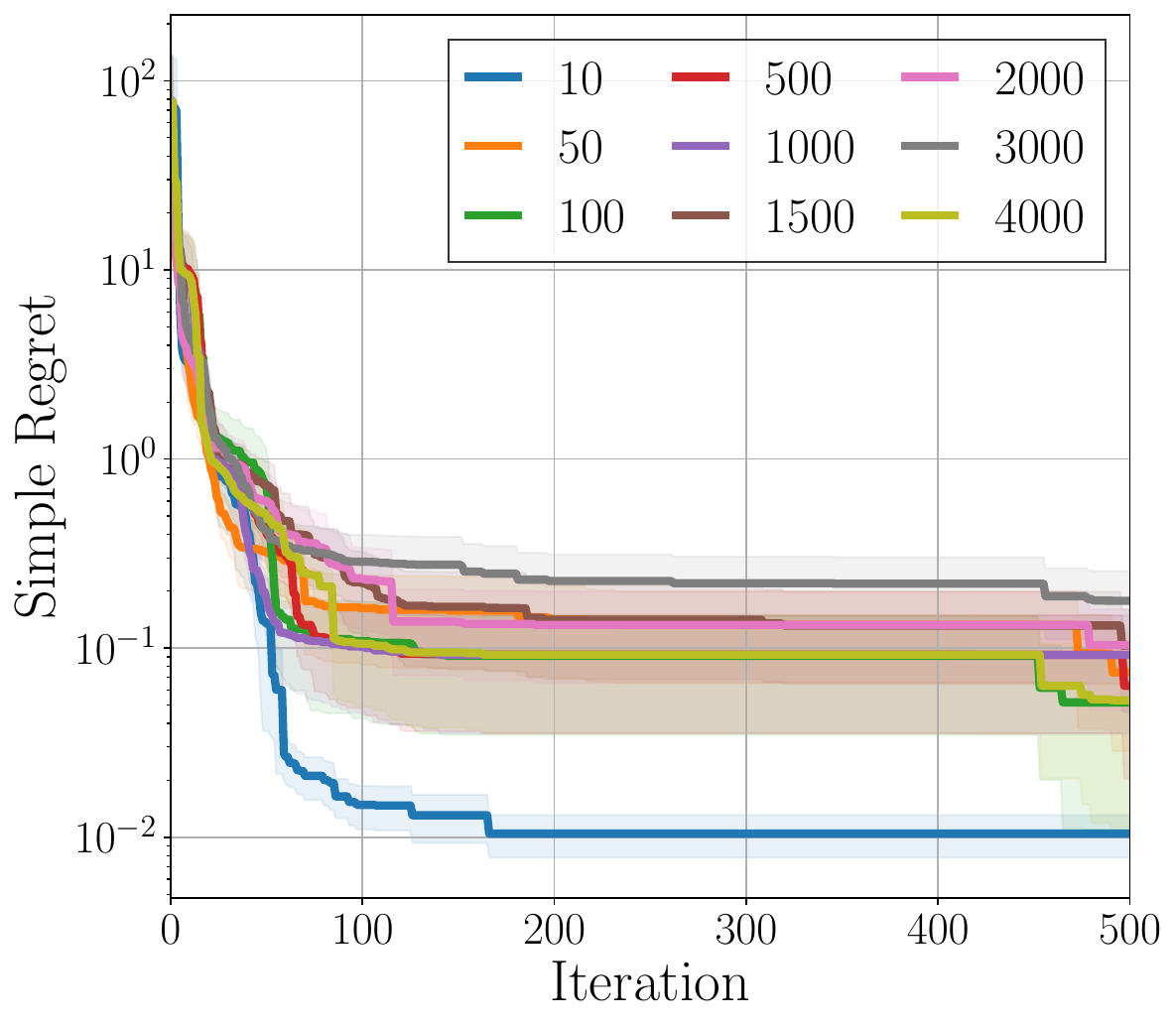}
    }
    \subfigure[Branin, LS]{
        \centering
        \includegraphics[width=0.23\textwidth]{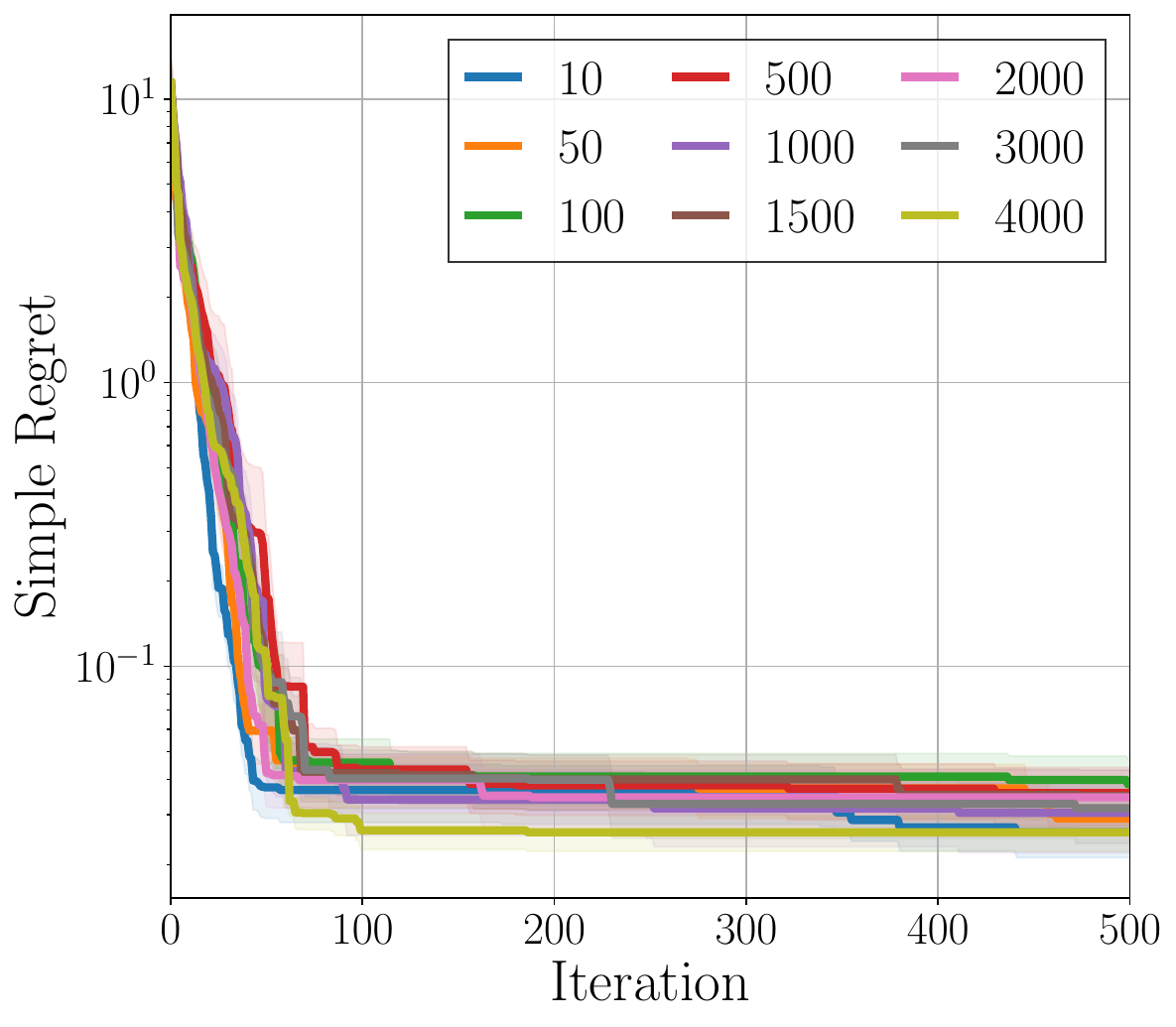}
    }
    \subfigure[Bukin6, LS]{
        \centering
        \includegraphics[width=0.23\textwidth]{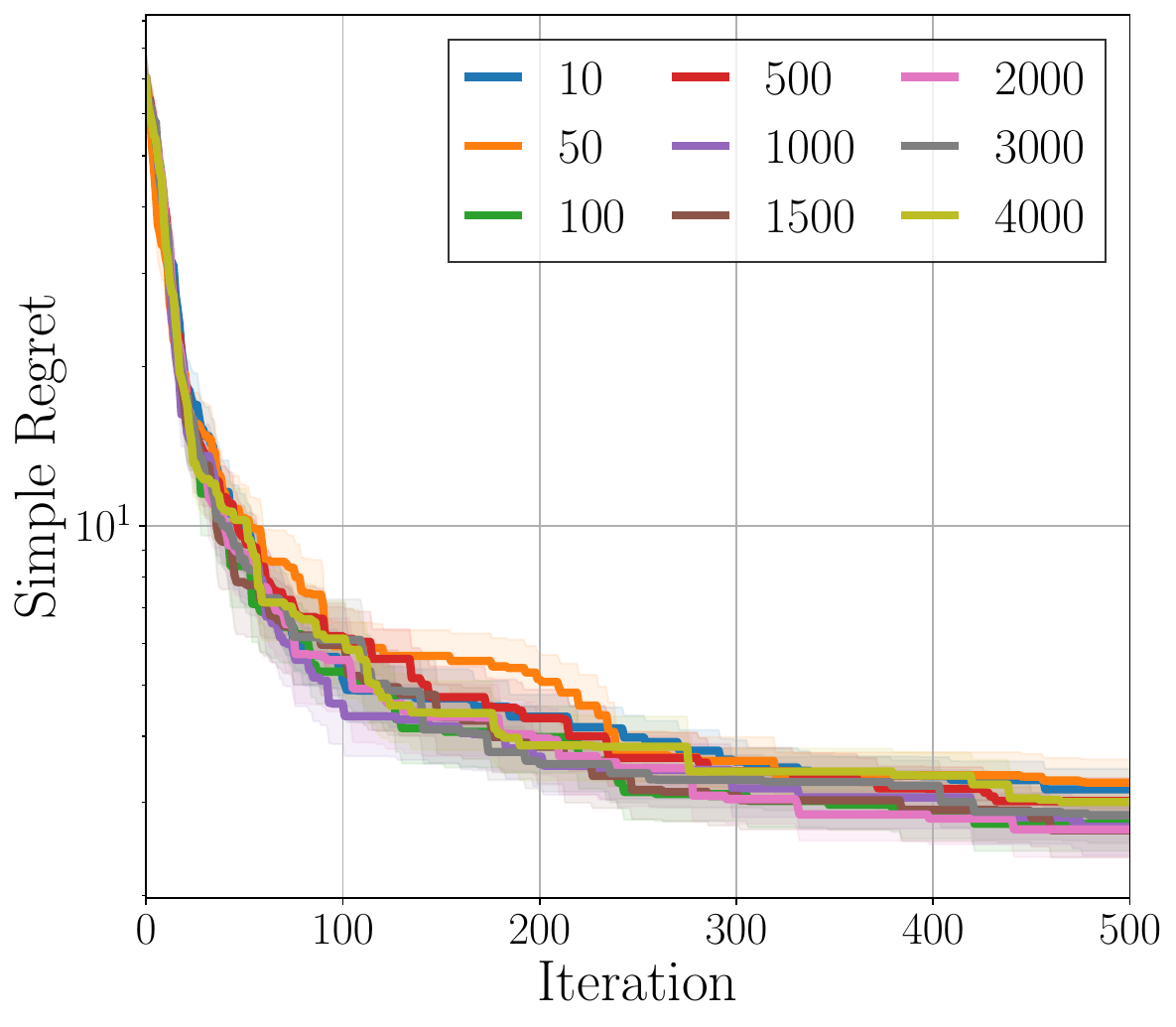}
    }
    \subfigure[Six-hump camel, LS]{
        \centering
        \includegraphics[width=0.23\textwidth]{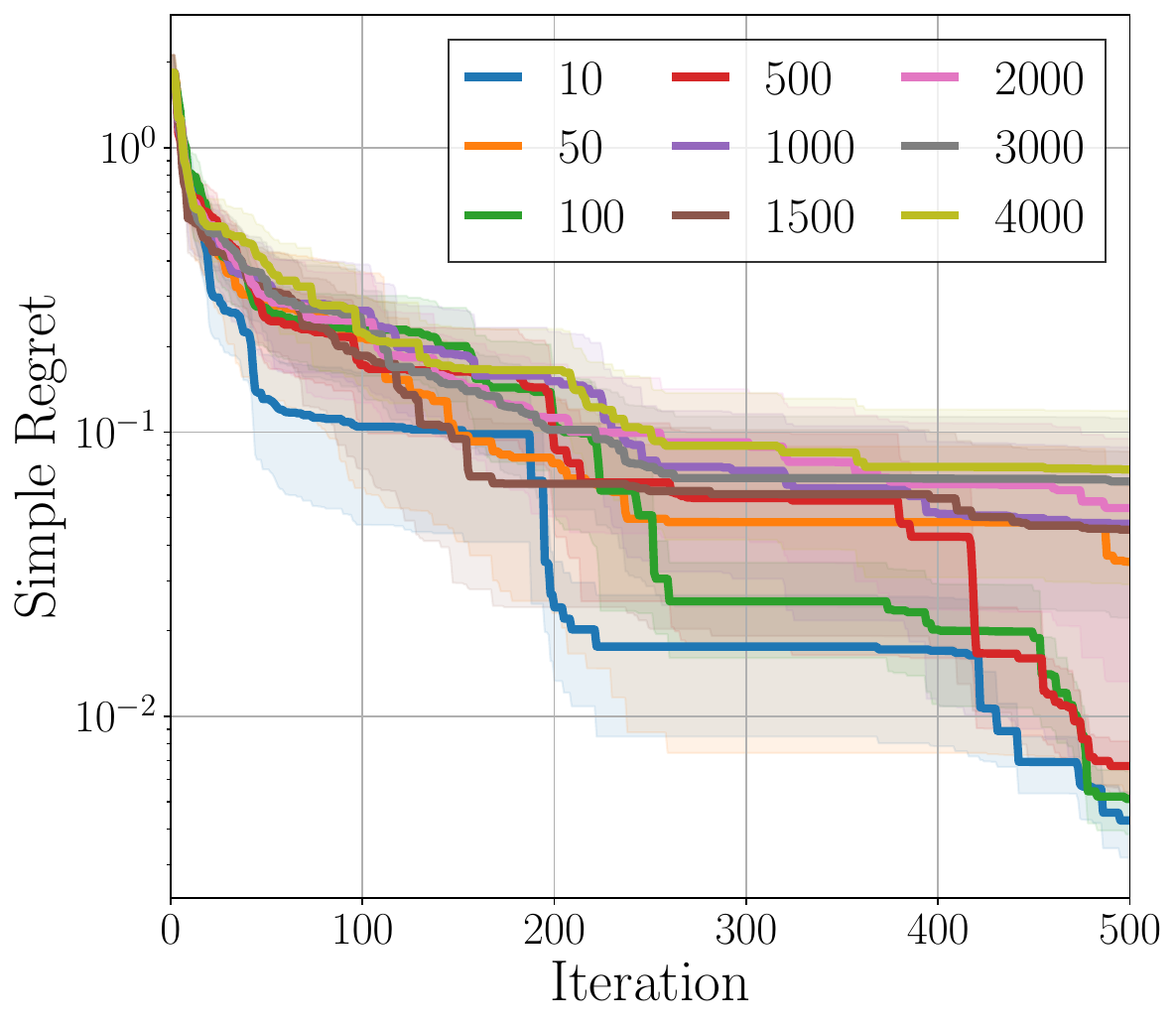}
    }
	\caption{Effects of pool sampling for a case with fixed-size pools. We repeat all experiments 20 times, and LP and LS stand for label propagation and label spreading, respectively.}
	\label{fig:discussion_pool_sampling}
\end{figure}

To see the impact of an additional hyperparameter,
i.e., the size of a subset of the original pool,
which is introduced to speed up semi-supervised learning algorithms,
we demonstrate numerical analysis on pool sampling
where the size of a predefined pool is 4,000
and $\beta = 0.5$ is given.
Based on~\figsref{fig:discussion_pool_sampling_time}{fig:discussion_pool_sampling},
we can accelerate our framework without significant performance loss.

\clearpage

\section{Discussion on a Threshold Ratio}
\label{sec:discussion_threshold_ratios}

\begin{figure}[ht]
    \centering
    \subfigure[Beale, LP]{
        \centering
        \includegraphics[width=0.23\textwidth]{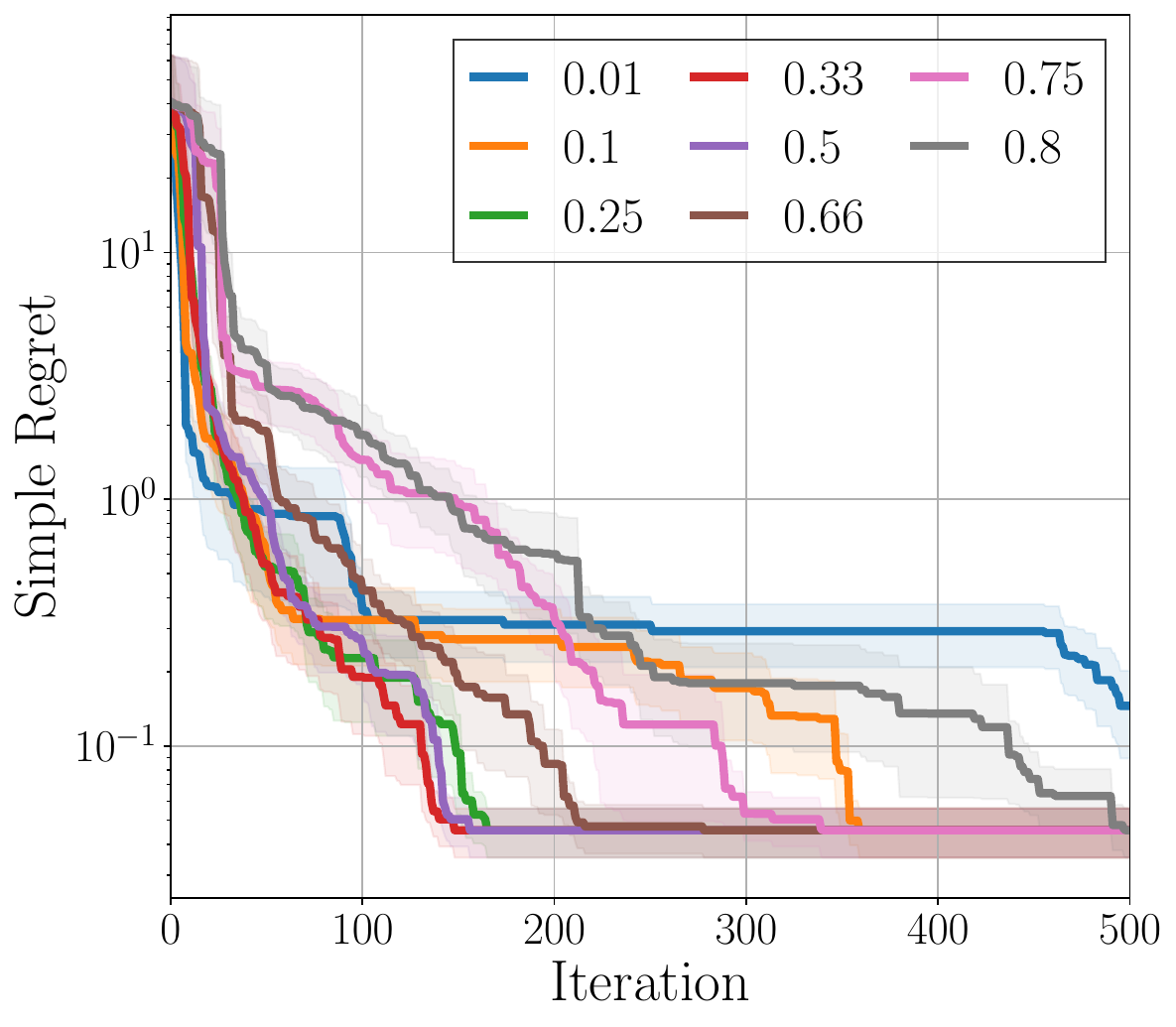}
    }
    \subfigure[Branin, LP]{
        \centering
        \includegraphics[width=0.23\textwidth]{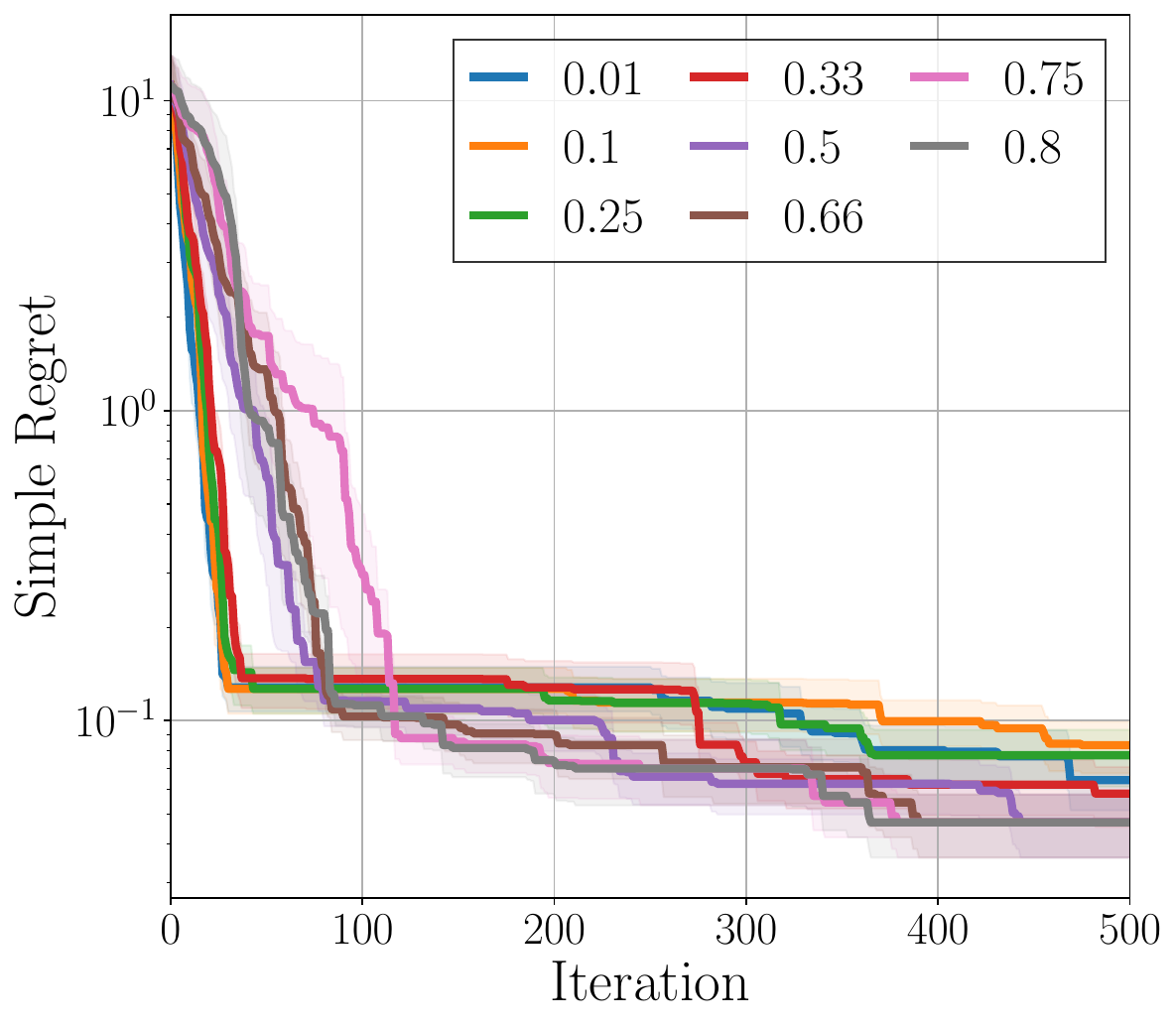}
    }
    \subfigure[Bukin6, LP]{
        \centering
        \includegraphics[width=0.23\textwidth]{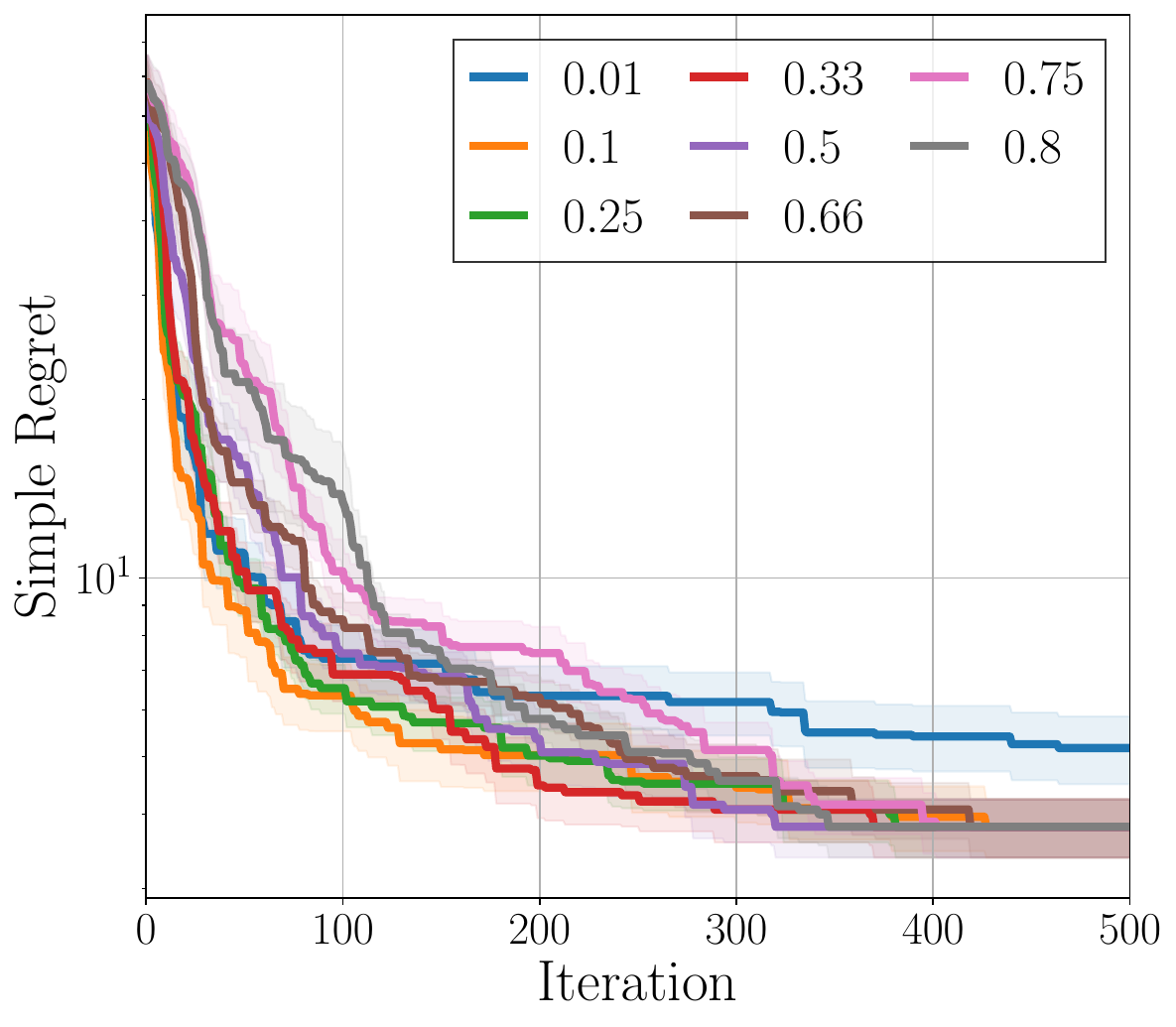}
    }
    \subfigure[Six-hump camel, LP]{
        \centering
        \includegraphics[width=0.23\textwidth]{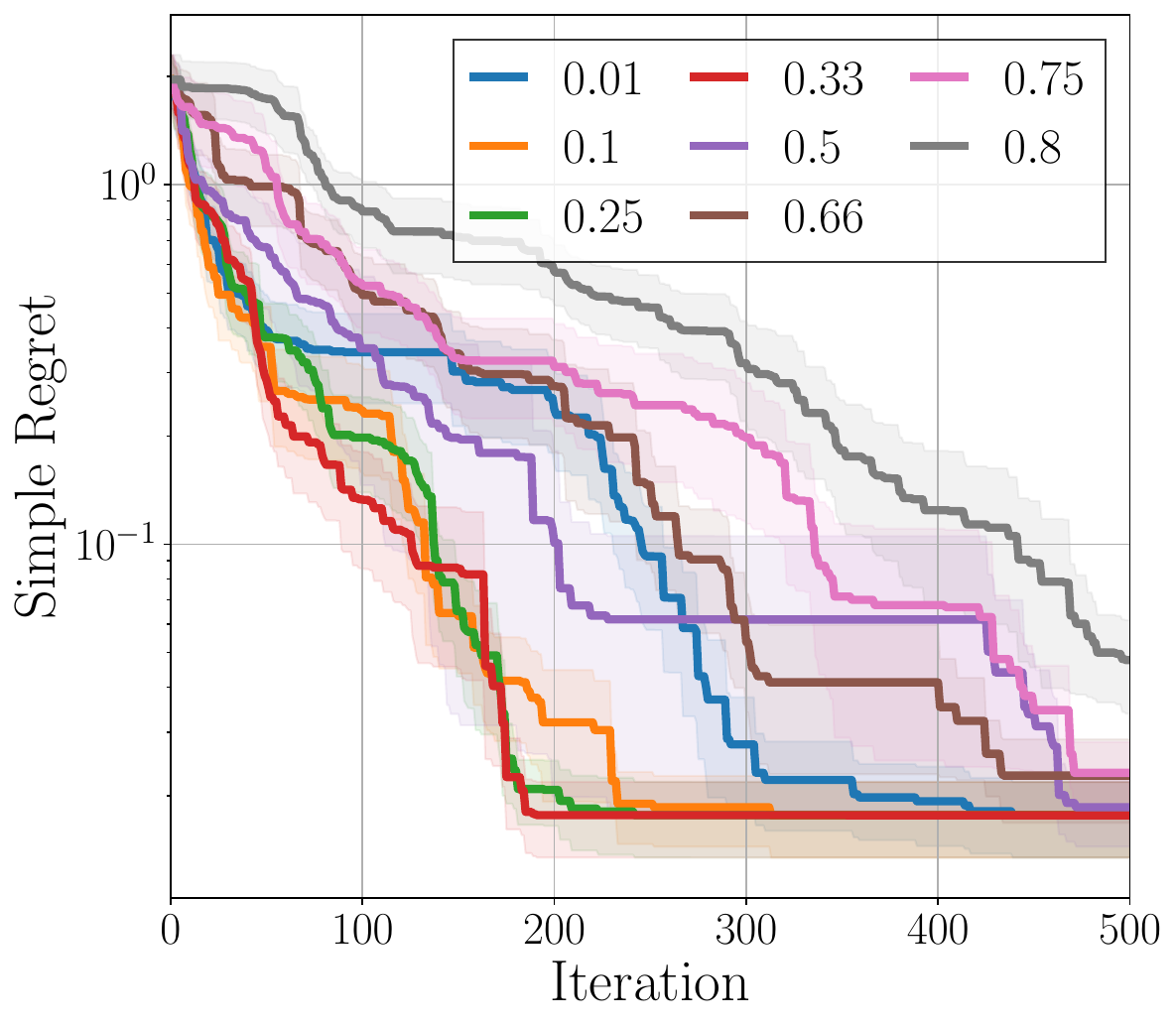}
    }
    \subfigure[Beale, LS]{
        \centering
        \includegraphics[width=0.23\textwidth]{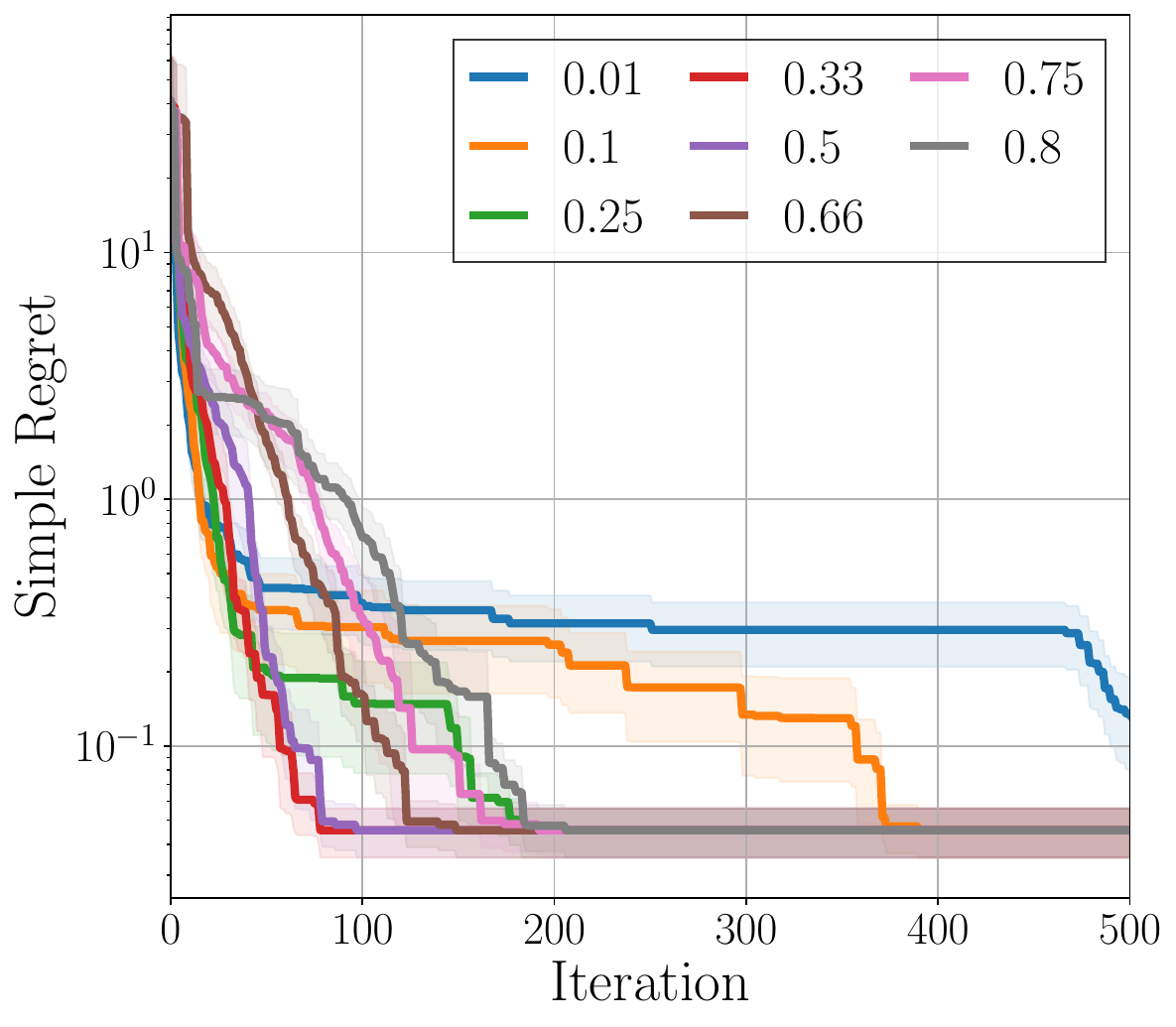}
    }
    \subfigure[Branin, LS]{
        \centering
        \includegraphics[width=0.23\textwidth]{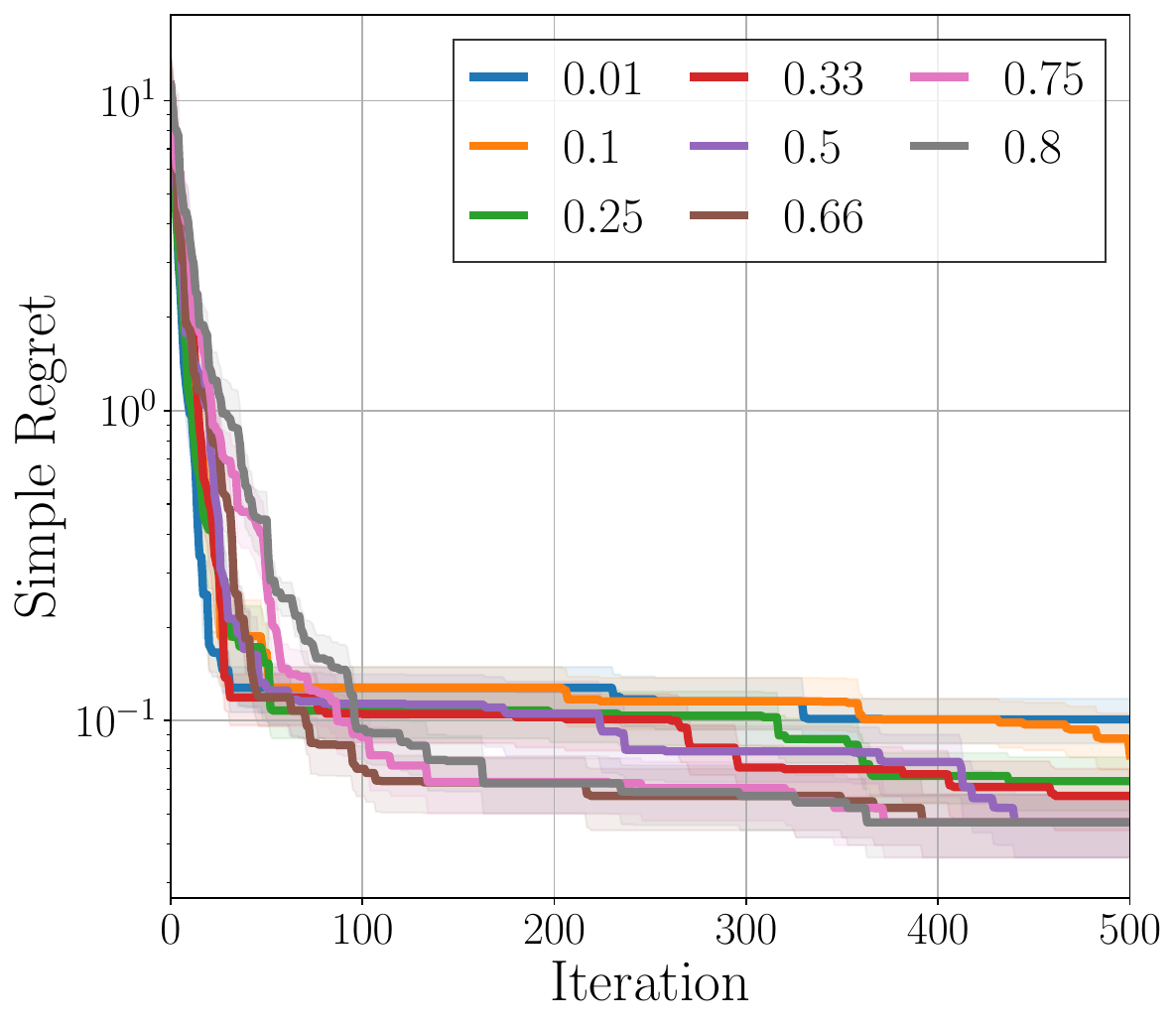}
    }
    \subfigure[Bukin6, LS]{
        \centering
        \includegraphics[width=0.23\textwidth]{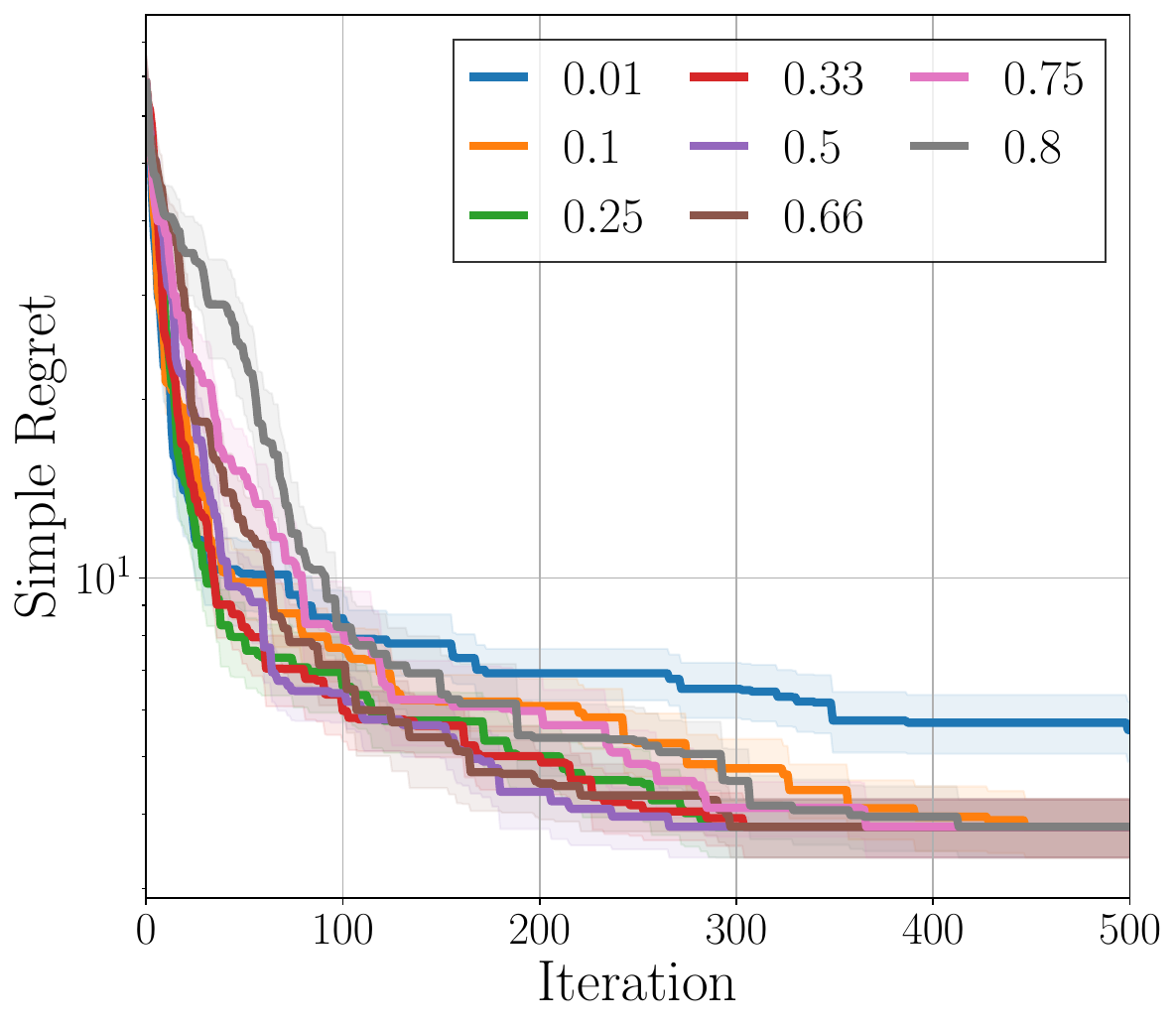}
    }
    \subfigure[Six-hump camel, LS]{
        \centering
        \includegraphics[width=0.23\textwidth]{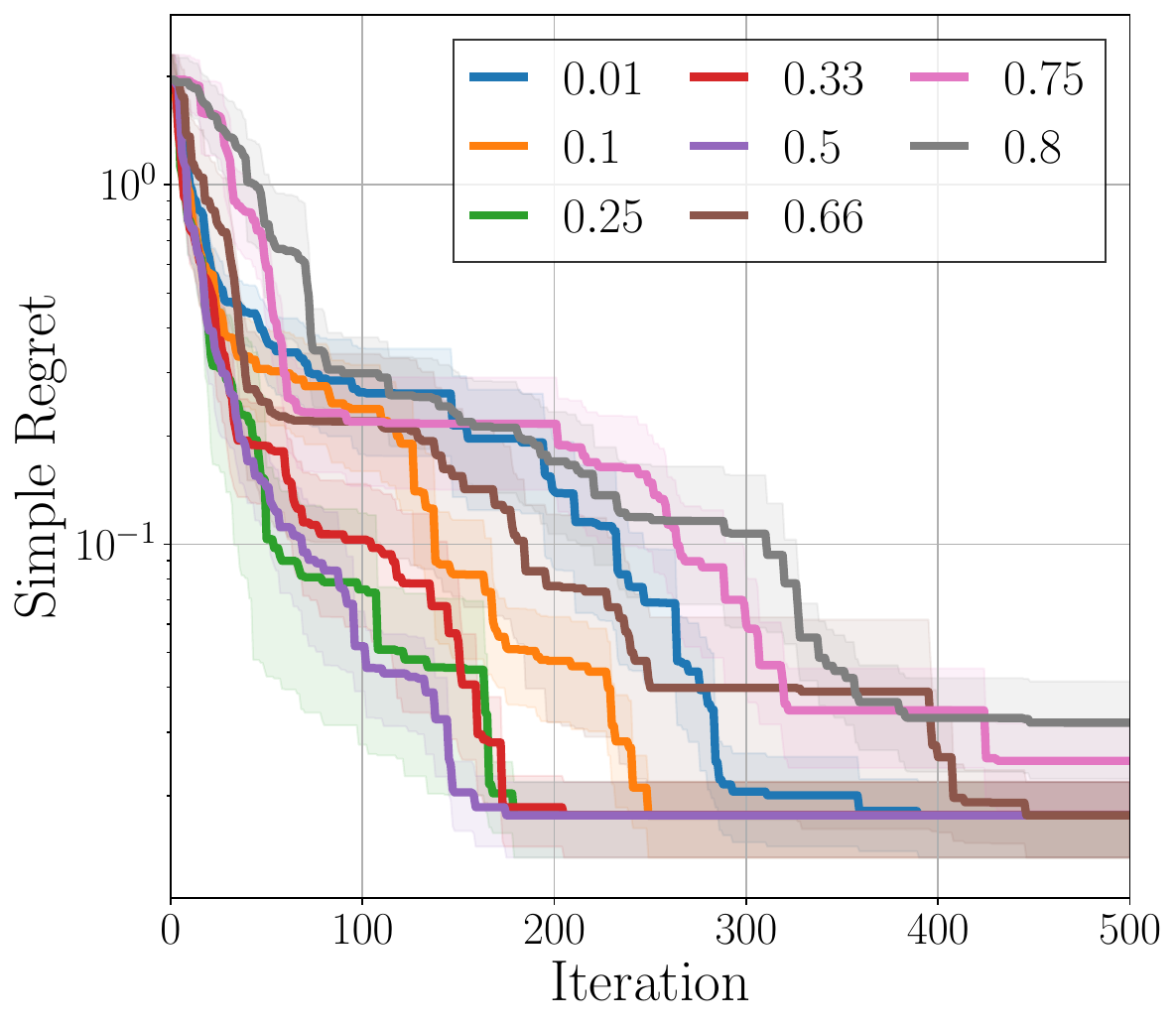}
    }
    \caption{Effects of a threshold ratio $\zeta$ with $\beta = 0.5$. We repeat all experiments 20 times, and LP and LS stand for label propagation and label spreading, respectively.}
    \label{fig:discussion_zeta}
\end{figure}

\figref{fig:discussion_zeta} demonstrates the effects of a threshold ratio $\zeta$,
where we use $\beta = 0.5$.
It follows the setting described in~\secref{sec:discussion_free_parameters}.
As presented in~\figref{fig:discussion_zeta},
a smaller $\zeta$, for example, $\zeta = 0.01$ or $\zeta = 0.1$ tends to show worse performance than a larger $\zeta$,
which implies that the over-exploitation is not due to conservative $y^\dagger$.
Interestingly, the results with $\zeta = 0.8$ also generally under-perform.
We presume that it is basically due to over-exploration.

\section{Limitations}
\label{sec:limitations}

As discussed above, our algorithms slow down if a pool size is significantly large.
As presented in~\figref{fig:discussion_pool_sampling_time},
elapsed times are certainly dependent on subset sizes.
To tackle this issue, we suggest a method to randomly select a subset of the pool,
but a more sophisticated subsampling method can be devised for our framework.
In particular,
we can leverage the impacts of the subset of the pool
by utilizing the geometric information of unlabeled data points.

\end{document}